\documentclass[reprint,floatfix,superscriptaddress]{revtex4-2}

\usepackage{hyperref}
\usepackage{color}
\usepackage[utf8]{inputenc} 
\usepackage[T1]{fontenc}
\usepackage{hyperref} 
\usepackage{url}
\usepackage{booktabs} 
\usepackage{amsfonts} 
\usepackage{nicefrac} 
\usepackage{microtype} 
\usepackage{amsmath}
\usepackage{float}
\usepackage{graphicx}
\usepackage{placeins}

\usepackage{subcaption}

\usepackage{amsthm}
\usepackage[capitalise,noabbrev]{cleveref}
\usepackage{verbatim}
\usepackage{csquotes}
\usepackage{amssymb}
\usepackage{enumerate}
\usepackage{enumitem}

\usepackage{natbib}
\usepackage{bbm}
\usepackage{bigints}
\usepackage{soul}
\usepackage{pgffor}
\usepackage[ruled,vlined,algo2e]{algorithm2e}
\usepackage{algorithm}
\usepackage{algorithmic}
\usepackage{cases}
\usepackage{ifthen}

\theoremstyle{plain}

\usepackage{xr}

\usepackage{appendix} 

\crefname{appendix}{Supplementary Note}{Supplementary Notes}

\renewcommand{\bibsection}{\section*{References}}

\makeatletter

\begin{document}

\title{Deep neural networks have an inbuilt Occam's razor} 

\author{Chris Mingard}\altaffiliation{These authors contributed equally.}
\affiliation{Rudolf Peierls Centre for Theoretical Physics,  University of Oxford; Oxford  OX1 3PU,  UK}
\affiliation{Physical and Theoretical Chemistry Laboratory, University of Oxford; Oxford, OX1 3QZ, UK}
\author{Henry Rees}\altaffiliation{These authors contributed equally.}
\affiliation{Rudolf Peierls Centre for Theoretical Physics,  University of Oxford; Oxford  OX1 3PU,  UK}
\author{Guillermo Valle-P\'erez}
\affiliation{Rudolf Peierls Centre for Theoretical Physics,  University of Oxford; Oxford  OX1 3PU,  UK}
\author{Ard A. Louis} \thanks{email: ard.louis@physics.ox.ac.uk} 
\affiliation{Rudolf Peierls Centre for Theoretical Physics,  University of Oxford; Oxford  OX1 3PU,  UK}

\date{\today}
\begin{abstract}
\vspace{10pt}
\section*{Abstract}
The remarkable performance of overparameterized deep neural networks (DNNs) must arise from an interplay between network architecture, training algorithms, and structure in the data.  To disentangle these three components for supervised learning, we apply a Bayesian picture based on the functions expressed by a DNN.   The prior over functions is determined by the network architecture, which we vary by exploiting a transition between ordered and chaotic regimes. For Boolean function classification, we approximate the
likelihood using the error spectrum of functions on data. Combining this with the prior  yields an accurate prediction for the posterior, measured for DNNs trained with stochastic gradient descent.  This analysis shows that structured data, together with a specific Occam's razor-like inductive bias towards (Kolmogorov) simple functions that exactly counteracts the exponential growth of the number of functions with complexity, is a key to the success of DNNs. 
\end{abstract}

\maketitle

\section{Introduction}

Although deep neural networks (DNNs) have revolutionised modern machine learning~\cite{lecun2015deep,schmidhuber1997discovering}, a fundamental theoretical understanding of why they perform so well remains elusive~\cite{breiman1995reflections,zdeborova2020understanding}.  One of their most surprising features is that they work best in the overparameterized regime, with many more parameters than data points.  As expressed in the famous quip: ``\textit{With four parameters I can fit an elephant, and with five I can make him wiggle his trunk.''} (attributed by Enrico Fermi to John von Neumann~\cite{dyson2004meeting}), it is widely believed that having too many parameters will lead to overfitting: a model will capture noise or other inconsequential aspects of the data, and therefore predict poorly. 

In statistical learning theory~\cite{shalev2014understanding} this intuition is formalised in terms of model capacity. It is not simply the number of parameters, but rather the complexity of the set of hypotheses a model can express that matters.  The search for optimal performance is often expressed in terms of bias-variance trade-off.  Models that are too simple introduce errors due to bias; they can't capture the underlying processes that generate the data. Models that are too complex are over-responsive to random fluctuations in the data, leading to variance in their predictions.

DNNs are famously highly expressive \cite{zhang2016understanding,cybenko1989approximation,hornik1991approximation}, i.e.\  they have extremely high capacity.  Their ability to generalize therefore appears to break basic rules of statistical learning theory.  Exactly how,  without explicit regularisation, DNNs achieve this feat is a fundamental question that has remained open for decades~\cite{breiman1995reflections,zdeborova2020understanding}. Although there has been much recent progress (see \cref{sec:introlit} for a literature overview)
there is no consensus for why DNNs work so well in the overparameterized regime.  

Here we study this conundrum in the context of supervised learning for classification, where inputs $x_i$ are attached to labels $y_i$.  Given a training set $S = \{(x_i,y_i\}^{m}_{i=1}$  of $m$ input-output pairs, sampled i.i.d.\ from a data distribution $\mathcal{D}$,  the task is to train a model on $S$ such that it performs well (has low generalization error) when predicting output labels $\hat{y}_i$ for a test set $T$ of unseen inputs, sampled from $\mathcal{D}$.  
For a DNN $\mathcal{N}(\Theta)$,
with parameters $\Theta \subseteq \mathbb{R}^p$ (typically weights and biases), the accuracy on a training set can be captured by a loss function  $L(\hat{y}_i, y_i)$ that measures how close, for input $x_i$,  the prediction  $\hat{y}_i$ of the DNN is to the true label $y_i$.  Training is typically done via some variant of stochastic gradient descent (SGD) which uses derivatives of $L(\hat{y}_i, y_i)$ to adjust the parameters $\Theta$ in order to minimise the loss on $S$.     Because DNNs are so highly expressive, and because SGD is typically a highly effective optimiser for DNNs, (near) zero training error (all correct labels after thresholding) on $S$ is routinely achieved~\cite{zhang2016understanding}.

\noindent{\bf Functions and inductive bias}
 
\noindent For classification, the question of why overparameterized DNNs don't overfit can conveniently be expressed in terms of functions.
For a given training set $S$ and test set $T$, a function $f$ can be defined on a restricted domain $S+T$. The inputs of $f$ are the $x_i \in S\cup T$, and the outputs include all possible sets of labels $\{\hat{y}_i\}$. Only one function gives the true labels $\{y_i\}$.  For a given set of parameters $\Theta$, the DNN  
then represents a particular function $f$, which can be identified by the labels it outputs on the inputs $x_i \in S\cup T$, after thresholding. Under the assumption that zero training error can be achieved,  functions need only be distinguished by how they behave on the test set $T$.   For $C$ classes there are $N_T = C^{|T|}$ possible functions $f$  with zero error on the training set, this number is typically unimaginably large.   The overwhelming majority of these functions will not generalize well.  Since DNNs are highly expressive, they should be able to represent all (or nearly all)  of these functions.  The fundamental question of overparameterized DNN performance becomes a question of \textit{inductive bias}:
Why, from the unimaginably large set of functions that give zero error on $S$, do DNNs converge on a minuscule subset of functions that generalize well?  
Here, we will argue that a combination of structured data and specific Occam's razor-like inductive bias towards simple functions which cancels out the exponential growth of the number of functions with increasing complexity helps answer this question.

\begin{figure*}[ht]
    \centering
    \begin{subfigure}{0.3\linewidth}
        \includegraphics[width= 5cm, height=3.75cm]{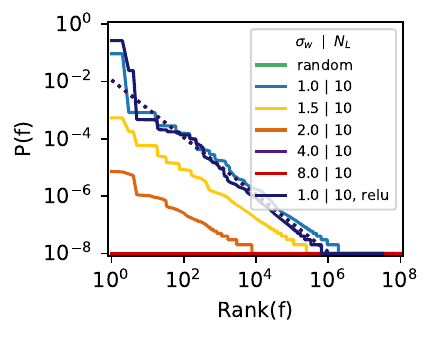}
        \caption{Prior $P(f)$ versus rank.}
        \label{subfig:rank_plot}
    \end{subfigure}
    \begin{subfigure}{0.3\linewidth}
        \includegraphics[width= 5cm, height=3.75cm]{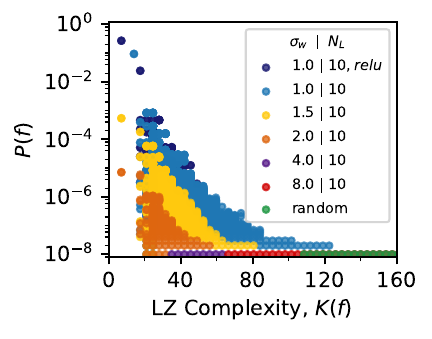}
        \caption{Prior $P(f)$ versus complexity}
        \label{subfig:LZ_order_chaos}
    \end{subfigure}
    \begin{subfigure}{0.3\linewidth}
        \includegraphics[width= 5cm, height=3.75cm]{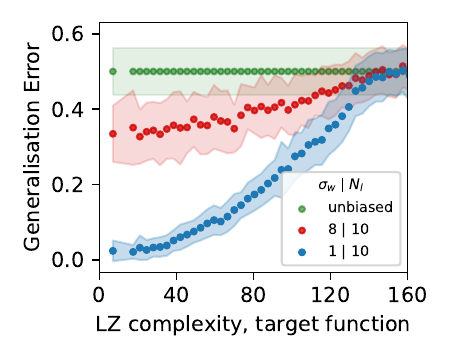}
        \caption{generalization error vs K}
        \label{subfig:trained_DNN_bias}
    \end{subfigure}
    
    \begin{subfigure}{0.3\linewidth}
        \includegraphics[width= 5cm, height=4.25cm]{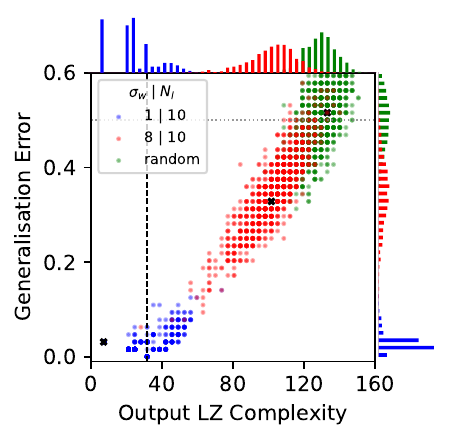}
        \caption{Target function LZ = 31.5}\label{subfig:TF31}
    \end{subfigure}
    \begin{subfigure}{0.3\linewidth}
        \includegraphics[width= 5cm, height=4.25cm]{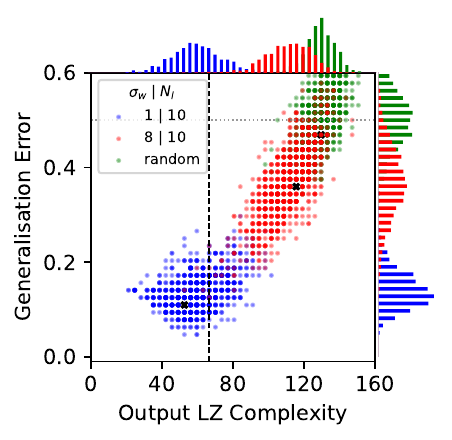}\caption{Target function LZ = 66.5}\label{subfig:TF66}
    \end{subfigure}
    \begin{subfigure}{0.3\linewidth}
        \includegraphics[width= 5cm, height=4.25cm]{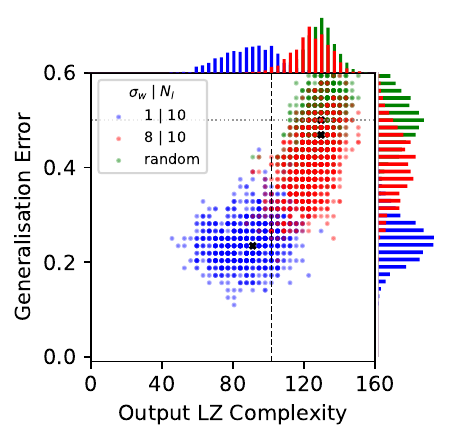}\caption{Target function LZ = 101.5}\label{subfig:TF101}
    \end{subfigure}
    
    \begin{subfigure}{0.3\linewidth}
        \includegraphics[width= 5cm, height=3.75cm]{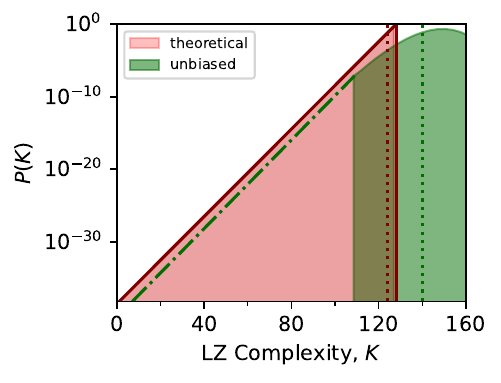}
        \caption{Prior $P(K)$ for  uniform sampling}\label{subfig:P(K|S)_theoretical}
    \end{subfigure}
    \begin{subfigure}{0.3\linewidth}
        \includegraphics[width= 5cm, height=3.75cm]{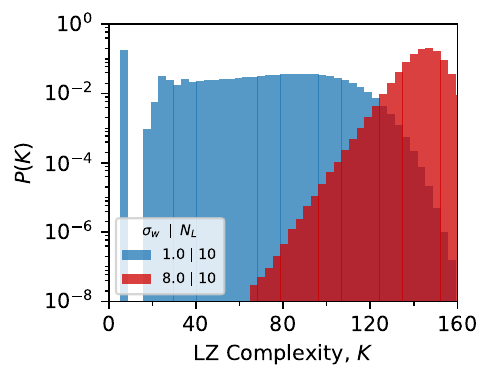}
        \caption{Prior $P(K)$  for $\sigma_w=1, 8$ }
        \label{subfig:order_bar_plot}
    \end{subfigure}
    \begin{subfigure}{0.3\linewidth}
        \includegraphics[width= 5cm, height=3.75cm]{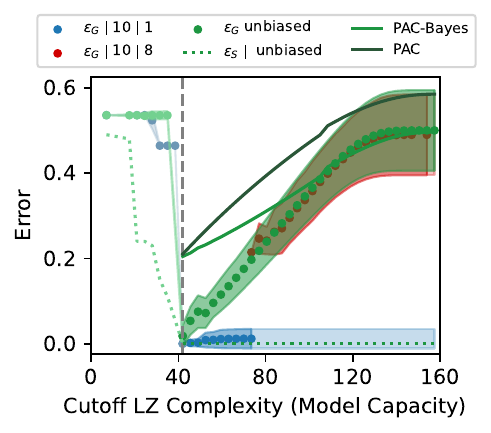}
        \caption{Bias variance}
        \label{subfig:main_bv}
    \end{subfigure}
    \caption{{\bf \small Priors over functions and over complexity}  \small (a) Prior $P(f)$ that a $N_l$-layer FCN with $\tanh$ activations generates $n=7$ Boolean functions $f$, ranked by probability of individual functions, generated from  $10^8$ random samples of parameters $\Theta$ over a Gaussian $P_{\rm par}(\Theta)$ with standard deviations $\sigma_w=1\ldots8$. Also compared is a ReLU-activated DNN.  The dotted blue line denotes a Zipf's law prior~\cite{valle2018deep} $P(f) = 1/((128 \ln 2) Rank(f))$.  (b) $P(f)$ versus LZ complexity $K$ for the networks from (a).  (c) generalization error versus $K$ of the target function for an unbiased learner (green), and $\sigma_w=1,8$ $\tanh$ networks trained to zero error with advSGD~\cite{valle2018deep} on cross-entropy loss with training set $S$ of size $m=64$, for $1000$ random initialisations. The error is calculated on the remaining $|T|=64$ functions. Error bars are one standard deviation (See \cref{fig:app:pb_1c} for PAC-Bayes bounds on this data). (d), (e), (f) Scatterplots of generalization error versus learned function LZ complexity, from $1000$ random initialisations for three target functions from subfigure (c). The dashed vertical line denotes the target function complexity. The black cross represents the mode function. The histograms at the top (side) of the plots show the posterior probability upon training as a function of complexity,$P_{\textrm{SGD}}(K|S)$ (error,$P_{\textrm{SGD}}(\epsilon_G|S)$). 
   (g) The prior probability  $P(K)$  to obtain a function of LZ complexity $K$ for uniform random sampling of  $10^8$, compared to a theoretical perfect compressor. $90\%$ of the probability mass lies to the right of the vertical dotted lines, and the dash-dot line denotes an extrapolation to low $K$. (h) $P(K)$ is relatively uniform on $K$ for the $\sigma_w=1$ system, while it is highly biased towards complex functions for the $\sigma_w=8$ networks.  The large difference in these priors helps explain the significant variation in DNN performance.
  (i) generalization error for the K-learning restriction for the $\sigma_w=1,8$ DNNs and for an unbiased learner, all for $|S|=100$. $\epsilon_S$ is the training error and $\epsilon_G$ is the generalization error on the test set. The vertical dashed line is the complexity $K_t$ of the target. Also compared are the standard realisable PAC and marginal-likelihood PAC-Bayes bounds for the unbiased learner. In $10^4$ samples, no solutions were found with $K \lesssim 70$ for the $\sigma_w=8$ DNN, and with $K\gtrsim70$ for the $\sigma_w=1$ DNN.
   } \label{fig:Cube_plots_1}
\end{figure*}

\noindent{\bf Distinguishing two  questions about generalization}
\noindent The main question we address here is what we call the \\
\noindent \textit{1st-order question of generalization} -- Why do high-capacity learning models (such as DNNs) generalise at all? --
 This question of how inductive bias allows DNNs to break the conventional bias-variance trade-off expectations of classical learning theory has a long history.  For example,  it was famously highlighted in Leo Breiman's  1995 commentary on refereeing for the NeurIPS conference~\cite{breiman1995reflections} (see also Appendix A) who phrased it by asking ``Why don't heavily overparameterised neural networks overfit the data?".       While originally formulated for DNNs, the fact that infinite width limits of DNNs can reduce to neural network Gaussian processes (NNGPs)~\cite{neal1994priors,lee2018deep,matthews2018gaussian}, or kernels such as the  Neural Tangent Kernel (NTK)~\cite{jacot2018neural}, has stimulated a large volume of important theoretical work on GPs and kernels, see e.g.~\cite{pmlr-v80-belkin18a,lee2020finite,arora2019exact,belkin2021fit,bordelon2020spectrum,cohen2021learning,canatar2021spectral,harzli2021double,cui2021generalization,simon2021neural}. In particular, these models recapitulate many properties of finite-width DNNs, including reaching small generalisation errors on standard datasets such as CIFAR10~\cite{lee2020finite,arora2019exact}. While these methods are non-parametric, they have high capacity, and like DNNs~\cite{zhang2016understanding}, can memorize random data~\cite{pmlr-v80-belkin18a}. 
For small capacity (or fewer parameters than data points for DNNs) all these models exhibit classic bias-variance trade-offs, with optimal generalisation performance at intermediate capacity. However, as capacity (or the number of parameters) increases further, the generalization error markedly diminishes. This phenomenon, known as double-descent~\cite{belkin2018reconciling}, is observed in DNNs, kernels, and GPs, illustrating how these high-capacity models deviate from the conventional wisdom of classical statistical learning theory

The relative simplicity of GPs and kernels compared to DNNs has enabled the derivation of analytic estimates of the generalization error in terms of the kernel eigenfunctions and eigenvalues~\cite{bordelon2020spectrum,cohen2021learning,canatar2021spectral,harzli2021double,cui2021generalization,simon2021neural}.  Good generalisation occurs when the kernel eigenfunctions with large eigenvalues align well with the target function being learned. 
Thus, these analyses offer a quantitative measure of precisely \textit{how} the inductive bias of a high-capacity kernel must align with that of the learning task. However, they do not provide a broader explanation of the nature and origin of the inductive bias in these models, nor why it often matches the data they are trained on. That is the big question we will attempt to address here.

We want to disambiguate the broader 1st-order question above from a more specific  \noindent \textit{2nd-order question of generalization} -- Given a high capacity DNN  that generalizes reasonably well (e.g.\ it solves the  1st order overparameterization/large capacity problem), can we understand how to improve its performance further?  ---  
This second question is crucial for deep learning practitioners: variations in architecture, hyperparameter tuning, data augmentation, etc., can significantly enhance performance over basic vanilla DNNs. However, these adjustments and tricks start from a base of a high-capacity model that already confounds expectations from classical learning theory.
Because the two questions are sometimes conflated, we want to emphasise up front that this paper will focus on the 1st-order question, which is relevant to all high-capacity models. A better understanding of this fundamental question should help frame important second-order questions about further improving DNN performance.

\noindent{\bf Learning Boolean functions: a model  system}

\noindent Inspired by calls to study model systems~\cite{zdeborova2020understanding,breiman1995reflections}, we first examine how a fully connected network (FCN)  learns Boolean functions $f: \{0,1\}^n \xrightarrow{} \{0,1\}$, which are key objects of study in computer science.
Just as the Ising model does for magnetism, this simple but versatile model allows us to capture the essence of the overparameterization problem, while remaining highly tractable.   For a system of size $n$, there are $2^n$ inputs, and $2^{2^n}$ Boolean functions.  Given a Boolean target function $f_t$, the DNN is trained on a subset $S$ of $ m < 2^n$ inputs, and then provides a prediction on a test set $T$ which consists of the rest of the inputs. A key advantage of this system is that data complexity can easily be varied by choice of target function $f_t$. 
Moreover, the model's tractability allows us to calculate the prior $P(f)$,  likelihood, $P(S|f)$ and posterior $P(f|S)$ for different functions and targets, and so cast the tripartite schema of architecture, training algorithm, and structured data from~\cite{zdeborova2020understanding} into a textbook Bayesian picture.

\section{Results}

\noindent{\bf Quantifying inductive bias with Bayesian priors}

\noindent The prior over functions, $P(f)$, is the probability that a DNN  $\mathcal{N}(\Theta)$ expresses $f$ upon random sampling of parameters over a parameter initialisation distribution $P_{\textrm{par}}(\Theta)$:  
\begin{equation}
\label{eqn:Pf_vol}
P(f)=\bigintssss \mathbbm{1}[\mathcal{N}(\Theta)==f]P_{\textrm{par}}(\Theta)d\Theta,
\end{equation}
where $\mathbbm{1}$ is an indicator function ($1$ if its argument is true, and $0$ otherwise). Explicitly, this term is $1$ if the neural network $\mathcal{N}(\Theta)$ expresses $f$ with parameters $\Theta$, else 0.
It was shown in \citep{valle2018deep} that, for ReLU activation functions,  $P(f)$ for the Boolean system was insensitive to different choices of $P_{\textrm{par}}(\Theta)$, and that it exhibits an exponential bias of the form $P(f) \lesssim 2^{-a \tilde{K}(f) + b}$ towards simple functions with low descriptional complexity $\tilde{K}(f)$, which is a proxy for the true (but uncomputable) Kolmogorov complexity.   We will, as in~\cite{valle2018deep}, calculate $\tilde{K}(f)$ using $C_{LZ}$, a Lempel-Ziv (LZ)  based complexity measure from~\cite{dingle2018input} on the $2^n$ long bitstring that describes the function, taken on an ordered list of inputs. 
Other complexity measures give similar results~\cite{valle2018deep,bhattamishra2022simplicity}, so there is nothing fundamental about this particular choice. To simplify notation, we will use $K(f)$ instead of $\tilde{K}(f)$.   The exponential drop of $P(f)$ with $K(f)$  in the map from parameters to functions is consistent with an algorithmic information theory (AIT) coding theorem~\cite{li2008introduction} inspired \textit{simplicity bias} bound~\cite{dingle2018input} which works for a much wider set of input-output maps.   It was argued in~\citep{valle2018deep} that if this inductive bias in the priors matches the simplicity of structured data then it would help explain why DNNs generalize so well.     However, the weakness of that work, and related works arguing for such a bias towards simplicity~\cite{arpit2017closer,lin2017does,valle2018deep,de2018random,mingard2019neural,kalimeris2019sgd,huh2021low,farhang2022investigating,bhattamishra2022simplicity,chiangloss}, is that it is typically not possible to significantly change this inductive bias towards simplicity, making it hard to conclusively show that it is not some other property of the network that instead generates the good performance. Here we exploit a particularity of $\tanh$ activation functions that enable us to significantly vary the inductive bias of DNNs.  In particular, for a Gaussian $P_{\textrm{par}}(\Theta)$ with standard deviation $\sigma_w$, it was shown~\cite{poole2016exponential,schoenholz2016deep} that, as $\sigma_w$ increases,  there is a transition to a chaotic regime.  Moreover, it was recently demonstrated that the simplicity bias in $P(f)$ becomes weaker in the chaotic regime~\cite{yang2019fine} (see also~\cref{app:sec:infinite}).   We will exploit this behaviour to systematically vary the inductive bias over functions in the prior.

In \cref{subfig:rank_plot,subfig:LZ_order_chaos} we depict prior probabilities $P(f)$ for functions $f$ defined on all $128$ inputs of a $n=7$ Boolean system upon random sampling of parameters of an FCN with 10 layers and hidden width 40 (which is provably fully expressive for this system~\cite{mingard2019neural}), and $\tanh$ activation functions.  The simplicity bias in $P(f)$  becomes weaker as the width $\sigma_w$ of the Gaussian $P_{\textrm{par}} (\sigma_w)$ increases. By contrast,  for ReLU activations, the bias in $P(f)$ barely changes with $\sigma_w$ (see  \cref{app:fig:sw8relu}(a)). 
The effect of the decrease in simplicity bias on DNN generalization performance is demonstrated in \cref{subfig:trained_DNN_bias} for a DNN trained to zero error on a training set $S$ of size $m=64$ using advSGD (an SGD variant taken from~\cite{valle2018deep}), and tested on the other $64$ inputs $x_i \in T$.   The generalization error (the fraction of incorrect predictions on $T$) varies as a function of the complexity of the target function. Although all these DNNs exhibit simplicity bias, weaker forms of the bias correspond to significantly worse generalization on the simpler targets  (see also   \cref{app:ce_advsgd}). For very complex targets, both networks perform poorly.  For reference, we also show an unbiased learner, where functions $f$ are chosen uniformly at random with the proviso that they exactly fit the training set $S$. Not surprisingly, given the $2^{64} \approx 2 \times 10^{19}$  functions that can fit $S$, the performance of this unbiased learner is no better than random chance. 

The scatter plots of~\cref{fig:Cube_plots_1} (d)--(f) depict a more fine-grained picture of the behaviour of the SGD-trained networks for three different target functions.  For each target,   $1000$ independent initialisations of the SGD optimiser, with initial parameters taken from $P_{\textrm{par}}(\sigma_w)$, are used.
The generalization error and complexity of each function found when the DNN first reaches zero training error are plotted.  Since there are $2^{64}$ possible functions that give zero error on the training set $S$, it is not surprising that the DNN converges to many different functions upon different random initialisations.  For the $\sigma_w=1$ network (where $P(f)$ resembles that of ReLU networks)  the most common function is typically simpler than the target.  By contrast, the less biased network converges on functions that are typically more complex than the target.  As the target itself becomes more complex, the relative difference between the two generalization errors decreases, because the strong inductive bias towards simple functions of the first network becomes less useful.   No free lunch theorems for supervised learning tell us that when averaged over all target functions, the three learners above will perform equally badly~\cite{schaffer1994conservation,wolpert1996lack} (see also  \cref{sec:NFL}). 

\noindent \textbf{Priors over complexity:}
To understand why relatively modest changes in the inductive bias towards simplicity lead to such significant differences in generalization performance,  we need another important ingredient, namely how the \textit{number} of functions vary with complexity.  Basic counting arguments imply that the number of strings of a fixed length that have complexity $K$ scales exponentially as $2^K$~\cite{li2008introduction}. Therefore, the vast majority of functions picked at random will have high complexity.   This exponential growth of the number of functions with complexity can be captured in a more coarse-grained prior, the probability $P(K)$ that the DNN expresses a function of complexity $K$ upon random sampling of parameters over a parameter initialisation function $P_{\rm par}(\Theta)$, which can also be written in terms of functions as $P(K') = \sum_{ f\in \mathcal{H}_{K'}} P(f)$, the weighted sum over the set $\mathcal{H}_{K'}$ of all functions with complexity $\tilde{K}(f)=K'$. In 
 \cref{fig:Cube_plots_1} (g)  $P(K)$ is shown for uniform random sampling of functions for $10^8$ samples using the LZ measure, and also for the theoretical ideal compressor with $P(K) = 2^{K-K_{max}-1}$ over all $2^{128}\approx 3 \times 10^{38}$ functions (see also  \cref{app:priors}).  
 In (h) we display $P(K)$ for functions not sampled at random, but rather from the two networks.  There is a dramatic difference between random sampling functions (as in (g)) and between the network with $\sigma_w =1$, where $P(K)$ is nearly flat. 
 This behaviour follows from the interesting fact that the AIT coding theorem-like scaling~\cite{dingle2018input,valle2018deep} of the prior over functions $P(f)\sim 2^{-\tilde{K}(f)}$  counters the $2^K$ growth in the number of functions.
 
By contrast, even though, relative to the $38$ or so orders of magnitude scale on which  $P(f)$ varies,  the more artefactual  $\sigma_w=8$ system has strong simplicity bias  (we estimate that for the simplest functions, $P(f)$ is about $10^{25}$ times higher than the mean probability $\left<P(f)\right> =2^{-128} \approx 3 \times 10^{-39}$), this is not 
enough to counter the  $2^K$ growth in the number of functions with complexity.  Therefore, this DNN is exponentially more likely to throw up complex functions, an effect that SGD is unable to overcome. 

  More generally, the fact that the number of complex functions grows exponentially with complexity $K$ lies at the heart of the classical explanation of why an insufficiently biased agent suffers from variance: It can too easily find many different functions that all fit the data.  
  The marked differences in the generalization performance between the two networks observed in \cref{fig:Cube_plots_1} (c)--(f) can be therefore traced to differences in the inductive bias of the networks, as measured by the differences in their priors.

\begin{figure*}[h!t!]
    \centering
    \begin{subfigure}[ht]{0.3\linewidth}
        \includegraphics[width=\textwidth]{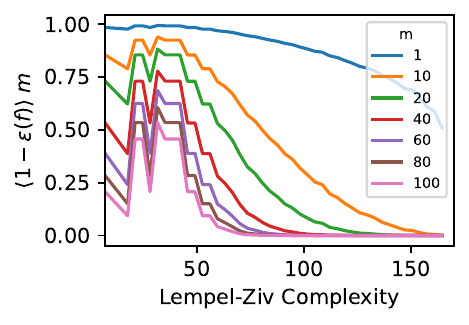}\caption{Target function LZ = 31.5}\label{subfig:eps_KLZ_31}
    \end{subfigure}
    \begin{subfigure}[ht]{0.3\linewidth}
        \includegraphics[width=\textwidth]{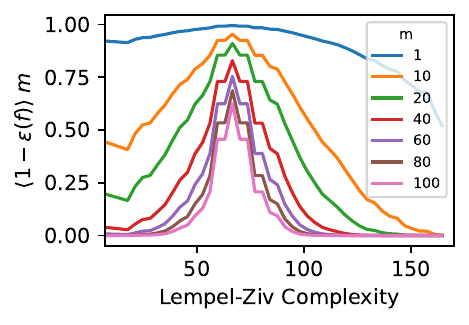}\caption{Target function LZ = 66.5}\label{subfig:eps_KLZ_66}
    \end{subfigure}
    \begin{subfigure}[ht]{0.3\linewidth}
        \includegraphics[width=\textwidth]{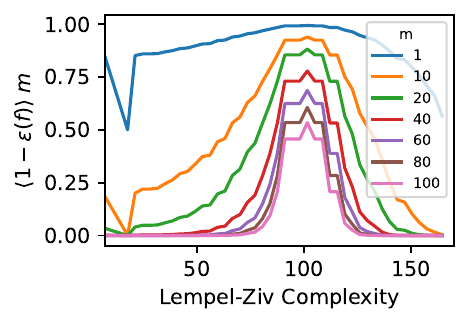}\caption{Target function LZ = 101.5}\label{subfig:eps_KLZ_101}
    \end{subfigure} 

    \begin{subfigure}[ht]{0.3\linewidth} \includegraphics[width=\textwidth]{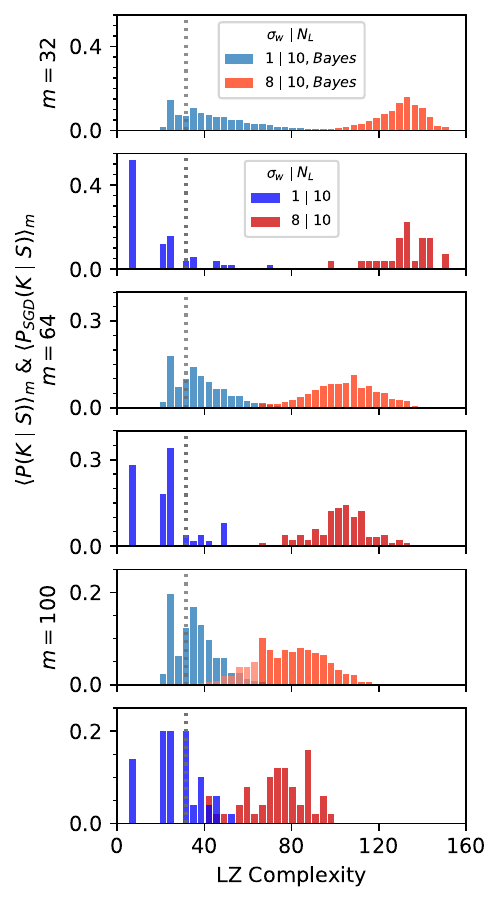}\caption{Target function LZ = 31.5} \label{subfig:Pb_Popt_31}
    \end{subfigure}
    \begin{subfigure}[ht]{0.3\linewidth} \includegraphics[width=\textwidth]{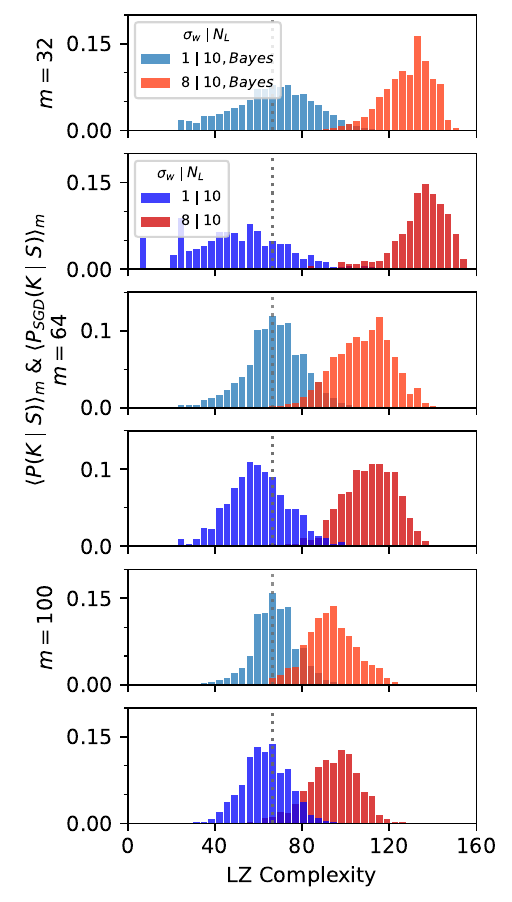}\caption{Target function LZ = 66.5}\label{subfig:Pb_Popt_66}
    \end{subfigure}
    \begin{subfigure}[ht]{0.3\linewidth}
    \includegraphics[width=\textwidth]{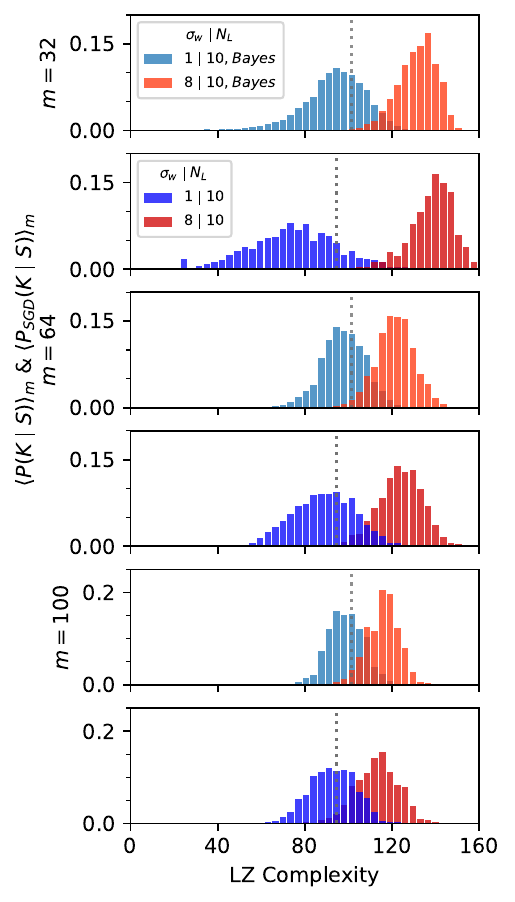}\caption{Target function LZ = 101.5}\label{subfig:Pb_Popt_101}
    \end{subfigure}
    \caption{  \small {\bf \small How training data affects the posteriors:
    } (a), (b) and (c) depict the mean likelihood   $\langle(1 - \epsilon_G(K))^m\rangle_5$ from \cref{eq:PBapprox}, averaged over training sets, and over the 5 lowest error functions at each $K$.  This term depends on data and is independent of the DNN architecture. With increasing $m$ it peaks more sharply around the complexity of the target.    In  (d)--(f) we compare 
the posteriors over complexity, $\langle P_{\rm SGD}(K|S)\rangle_m$, for  SGD (darker blue and red) averaged over training sets of size $m$,  to the prediction of $\langle P(K|S) \rangle_m$ from  \cref{eq:PBapprox} (lighter blue and orange), calculated by multiplying the Bayesian likelihood curves in Figs (a)--(c) by the prior $P(K)$ shown in \cref{fig:Cube_plots_1}(h).  The light (Bayes) and dark (DNN) blue histograms are from the $\sigma_w=1$ system, and the orange (Bayes) and red (DNN) histograms are from the $\sigma_w=8$ system which has less bias towards simple functions.  The Bayesian decoupling approximation (\cref{eq:PBapprox}) captures the dominant trends in the behaviour of the SGD-trained networks as a function of data complexity and training set size. 
Quantitative measures of the similarity between the posteriors can be found in \cref{app:Cube_plots_analysis}.}
    \label{fig:Cube_plots_2}
\end{figure*} 

\noindent \textbf{Artificially restricting model capacity}
To further illustrate the effect of inductive bias we create a K-learner that only allows functions with complexity $\leq K_M$ to be learned and discards all others.  As can be seen in \cref{fig:Cube_plots_1} (i), the learners typically cannot reach zero training error on the training set if  $K_M$ is less than the target function complexity $K_t$.   For $K_M \geq K_t$, zero training error can be reached and not surprisingly, the lowest generalization error occurs when $K_M=K_t$. As the upper limit $K_M$ is increased, all three learning agents are more likely to make errors in predictions due to variance.  The random learner has an error that grows linearly with $K_M$.  This behaviour can be understood with a classic Probably Approximately Correct (PAC) bound~\cite{shalev2014understanding}
where the generalization error  (with confidence $0 \leq (1-\delta) \leq 1$)  scales as $\epsilon_G \leq (\ln|\mathcal{H}_{\leq K_M}| - \ln{\delta})/m$, where  $|\mathcal{H}_{\leq K_M}|~ K\leq K_M$ is the size of the hypothesis class of all functions with $K \leq K_M$;  the bound scales linearly in $K_M$, as the error does (see \cref{app:PAC-Bayes} for further discussion including the more sophisticated PAC-Bayes bound~\cite{mcallester1999some,valle2020generalization}.).  The generalization error for the $\sigma_w=1$ DNN does not change much with $K_M$  for $K_M > K_t$ because the strong inductive bias towards simple solutions means access to higher complexity solutions doesn't significantly change what the DNN converges on.

Finally, we show data for DNNs in the ordered regime with $\sigma_w \ll 1$, and for other optimisers, loss functions, and activation functions in \cref{app:fig:ordered_plots,app:fig:e_vs_lz,app:fig:scatter1,app:fig:single_train,fig:app:1c_extras,fig:app:1d_extras}.   These results broadly exhibit the same behaviour we describe here.

\noindent \textbf{Calculating the Bayesian posterior and likelihood}

\noindent To better understand the generalization behaviour observed in \cref{fig:Cube_plots_1} 
we apply  Bayes' rule, $P(f|S) = P(S|f)P(f)/P(S)$ to calculate the Bayesian posterior $P(f|S)$ from the prior $P(f)$, the  likelihood $P(S|f)$, and the marginal likelihood $P(S)$.  
Since we condition on zero training error, the likelihood takes on a simple form.  $P(S|f) = 1 \textrm{ if } \forall x_i \in S, f(x_i)=y_i$, while $P(S|f)=0$ otherwise.  For a fixed training set, all the variation in $P(f|S)$ for $f \in U(S)$, the set of all functions compatible with $S$, comes from the prior $P(f)$ since $P(S)$ is constant. Therefore, in this Bayesian picture,  the bias in the prior is translated over to the posterior.

The marginal likelihood also takes a relatively simple form for discrete functions, since  $P(S) = \sum_f P(S|f) P(f)=\sum_{f\in U(S)} P(f)$. It is equivalent to the probability that the DNN obtains zero error on the training set $S$ upon random sampling of parameters, and so can be interpreted as a measure of the inductive bias towards the data.    The Marginal-likelihood PAC-Bayes bound~\cite{valle2020generalization} makes a direct link $P(S) \lesssim e^{-m \epsilon_G}$ to the generalization error $\epsilon_G$ which captures the intuition that, for a given $m$,  a better inductive bias towards the data (larger $P(S)$) implies better performance (lower $\epsilon_G$).

One can also define the posterior probability $P_{\textrm{SGD}}(f|S)$, that a network trained with SGD (or another optimiser) on training set $S$, when initialised with $P_{\textrm{par}}(\Theta)$, converges on function $f$.  For simplicity, we take this probability at the epoch where the system first reaches zero training error. Note that in \cref{fig:Cube_plots_1} (d)--(f) it is this SGD-based posterior that we plot in the histograms at the top and sides of the plots, with functions grouped either by complexity, which we will call $P_{\textrm{SGD}}(K|S)$, or by generalization error $\epsilon_G$, which we will call $P_{\textrm{SGD}}(\epsilon_G|S)$.

DNNs are typically trained by some form of SGD, and not by randomly sampling over parameters which is much less efficient. However, a recent study~\cite{mingard2021sgd} which carefully compared the two posteriors has shown that to first order,  $P_{\textrm{B}}(f|S) \approx P_{\textrm{SGD}}(f|S)$, for many different data sets and DNN architectures. We demonstrate this close similarity in \cref{fig:app:gp_nn_2} explicitly for our $n=7$ Boolean system. This evidence suggests that 
Bayesian posteriors calculated by random sampling of parameters, which are much simpler to analyze, can be used to understand the dominant behaviour of an SGD-trained DNN, even if, for example, hyperparameter tuning can lead to 2nd-order deviations between the two methods (see also \cref{sec:introlit}). 

To test the predictive power of our Bayesian picture, we first define the function error $\epsilon(f)$  as the fraction of incorrect labels $f$ produces on the full set of inputs. Next, we average Bayes' rule over all training sets $S$ of size $m$:
\begin{equation}\label{eq:PB}
\langle P(f|S) \rangle_m   =   P(f) \langle \frac{  P(S|f) }{P(S) }   \rangle_{m}  \approx \frac{ P(f) \left(1 - \epsilon(f)\right)^m}{\langle P(S) \rangle_{m}} 
\end{equation}
where the mean likelihood $\langle P(S|f)\rangle_m = (1-\epsilon(f))^m$  is the probability of a function $f$ obtaining zero error on a training set of size $m$. In the second step, we approximate the average of the ratio with the ratio of the averages which should be accurate if $P(S)$ is highly concentrated, as is expected if the training set is not too small. 

\cref{eq:PB} is hard to calculate, so  we  coarse-grain  it by grouping together functions by their complexity:
\begin{equation} \label{eq:PBapprox} 
\langle P(K|S) \rangle_m = \hspace*{-0.5cm} 
  \sum_{C_{LZ}(f)=K} \langle P(f|S) \rangle_m
\propto P(K) \langle \left(1-\epsilon_G(K)\right)^m \rangle_{l},
\end{equation}
and in the second step make a  \textit{decoupling approximation} where we average the  likelihood term  over a small number $l$ of functions with complexity $K$ with lowest generalization error $\epsilon_G(K)$ since the smallest errors in the sum dominate exponentially since $(1-\epsilon_G) \approx e^{-\epsilon_G}$ for $|\epsilon_G| \ll 1$.  We then multiply by $P(K)$, which takes into account the value of the prior and the multiplicity of functions at that $K$, and normalise $\sum_K P(K|S)=1$. For a given target, we make the ansatz that this decoupling approximation provides an estimate that scales as the true (averaged) posterior.

\begin{figure*}[ht]
    \centering
    \begin{subfigure}[ht]{0.3\linewidth}
        \includegraphics[width=\textwidth]{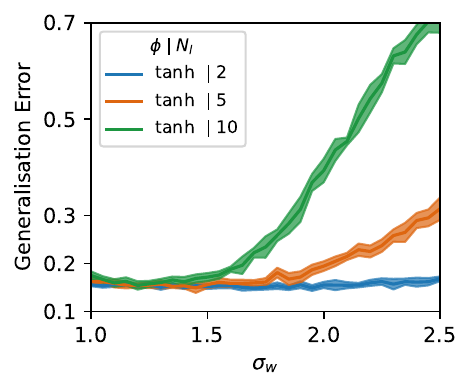}
        \caption{MNIST m=1000}
    \end{subfigure}
    \begin{subfigure}[ht]{0.3\linewidth}
        \includegraphics[width=\textwidth]{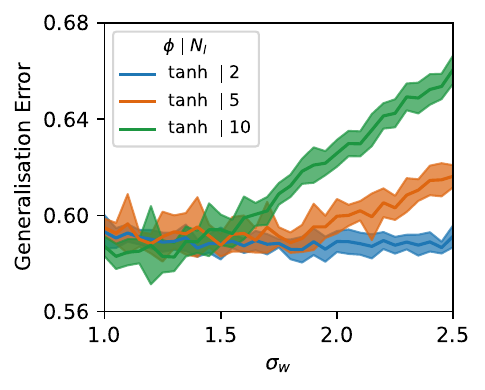}
        \caption{CIFAR10 m=5000}
    \end{subfigure}
    \begin{subfigure}[ht]{0.3\linewidth}
        \includegraphics[width=\textwidth]{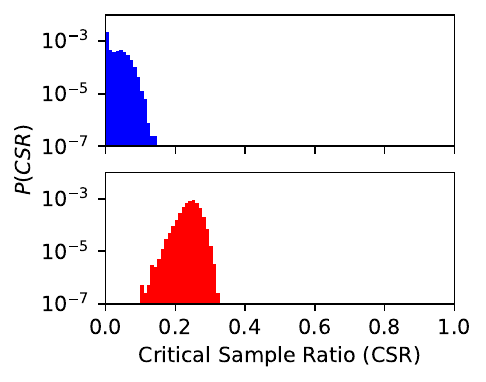}
        \caption{MNIST P(K)}\label{fig:3_c}
    \end{subfigure}
    
    \vspace{2mm}
    \centering
    \begin{subfigure}[ht]{0.3\linewidth}
        \includegraphics[width=\textwidth]{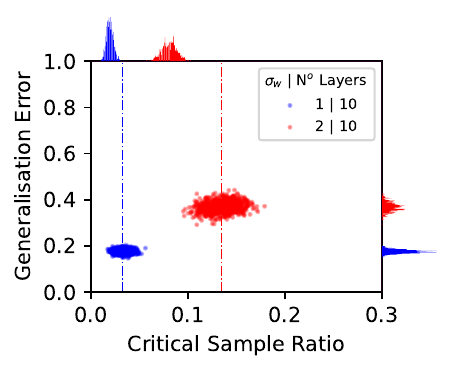}
        \caption{MNIST Uncorrupted, Posterior, m=1000}\label{fig:3_CSR_0}
    \end{subfigure}
    \begin{subfigure}[ht]{0.3\linewidth}
    \includegraphics[width=\textwidth]{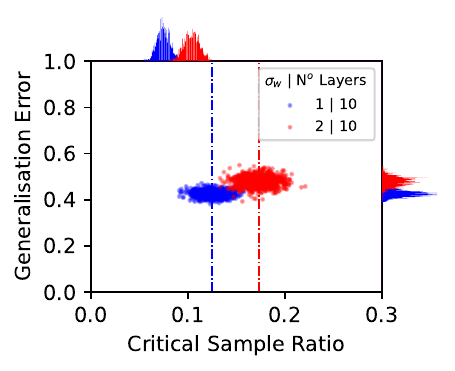}
        \caption{MNIST 25\% corruption, Posterior, m=1000}\label{fig:3_CSR_25}
    \end{subfigure}
    \begin{subfigure}[ht]{0.3\linewidth}
    \includegraphics[width=\textwidth]{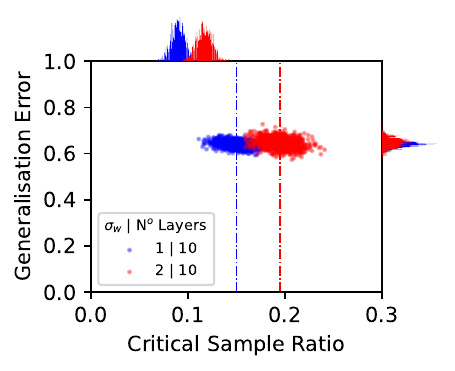}
        \caption{MNIST 50\% corruption, Posterior, m=1000}\label{fig:3_CSR_50}
    \end{subfigure}
    
    \caption{\small {\bf \small MNIST and CIFAR-10 data.} (a) MNIST generalization error for  FCNs on  a $1000$ image training set versus $\sigma_w$ for three depths. (b) CIFAR10 generalization error for  FCNs trained on a $5000$ image training set versus $\sigma_w$ for three depths. The FCNs, made of multiple hidden layers of width $200$,  were trained with SGD with batch size 32 and lr=$10^{-3}$  until $100\%$ accuracy was first achieved on the training set. Error bars are one standard deviation. (c) Complexity prior $P(K)$, for CSR complexity, for $1000$ MNIST images for randomly initialised networks of 10 layers and $\sigma_w =1,2$. Probabilities are estimated from a sample of $2 \times 10^4$ parameters. 
    Figs (d), (e) and (f) are scatterplots of generalization error versus the CSR for $1000$ networks trained to 100\% accuracy on a training set of $1000$ MNIST images and tested on $1000$ different images. In (d) the training labels are uncorrupted, in (e) and (f)  25\% and 50\%  of the training labels are corrupted respectively.  Note the qualitative similarity to the scatter plots in Fig~1 (d)-(f).
   } \label{fig3}
\end{figure*}    

To test our approximations, we first plot, in 
\cref{fig:Cube_plots_2} (a)--(c), the  likelihood term in ~\cref{eq:PBapprox} for three different target functions.  To obtain these curves, we considered a large number of functions (including all functions with up to 5 errors w.r.t.\ the target, with further functions sampled). For each complexity, we average this term over the $l=5$   functions with smallest $\epsilon_G.$      Not surprisingly, functions close to the complexity of the target have the smallest error. These graphs help illustrate how the DNN interacts with data. As the training set size $m$ increases, functions that are most likely to be found upon training to zero training error are increasingly concentrated close to the complexity of the target function.

To test the decoupling approximation from \cref{eq:PBapprox}, we compare in \cref{fig:Cube_plots_2} (d)-(f) the posterior $\langle P(K|S) \rangle_m$, calculated by multiplying the Bayesian likelihood curve from \cref{fig:Cube_plots_2} (a)--(c)  with the two Bayesian priors $P(K)$ from  \cref{fig:Cube_plots_1} (h) and (i), to the posteriors $\langle P_{rm SGD}(K|S) \rangle_m$ calculated by advSGD~\cite{valle2018deep} over a 1000 different parameter initialisations and training sets. It is remarkable to see how well the simple decoupling approximation performs across target functions and training set sizes. In \cref{app:fig:k=1_approx_bayes,app:fig:k=50_approx_bayes}) we demonstrate the robustness of our approach by showing that using $l=1$ or $l=50$ functions does not change the predictions much.  This success suggests that our simple approach captures the essence of the interaction between the data (measured by the likelihood, which is independent of the learning algorithm), and the DNN architecture (which is measured by the prior and is independent of the data). 

We have therefore separated out two of the three parts of the tripartite scheme, which leaves the training algorithm.  In the figures above our Bayesian approximation captures the dominant behaviour of an SGD-trained network. This correspondence is consistent with the results and arguments of \cite{mingard2021sgd}.  We checked this further in \cref{fig:app:gp_nn_2} for a similar set-up using MSE loss, where Bayesian posteriors can be exactly calculated using Gaussian processes (GPs). The direct Bayesian GP calculation closely matches SGD-based results for our much smaller network.  Note that,  in the spirit of model calculations, as called for in~\cite{breiman1995reflections,zdeborova2020understanding}, we mainly used a much smaller DNN. But their agreement with the GP-based posteriors, calculated for the infinite width limit, shows that at the scale of our Bayesian approach to the 1st-order generalisation question we are addressing here, the size of the DNN is not an important factor.  The width of a DNN can, of course, be a factor for 2nd order generalisation questions.

\noindent \textbf{Beyond the Boolean model: MNIST $\bf \&$ CIFAR-10} 
Can the principles worked out for the Boolean system be observed in larger systems that are closer to the standard practice of DNNs?  To this end, we show, in \cref{fig3} (a) and (b) how the generalization error for the popular image datasets MNIST and CIFAR-10 changes as a function of the initial parameter width $\sigma_w$ and the number of layers $N_l$ for a standard FCN, trained with SGD on cross-entropy loss with $\tanh$ activation functions.  Larger $\sigma_w$ and larger $N_l$ push the system deeper into the chaotic regime~\cite{poole2016exponential,schoenholz2016deep} and result in decreasing generalization performance, similar to what we observe for the Boolean system for relatively simple targets. 
In \cref{fig3},  we plot the prior over complexity $P(K)$ for a complexity measure called the critical sample ratio (CSR)~\cite{arpit2017closer}, an estimate of the density of decision boundaries that should be appropriate for this problem.  Again, increasing $\sigma_w$ greatly increases the prior probability that the DNN produces more complex functions upon random sampling of parameters.   Thus the decrease in generalization performance is consistent with the inductive bias of the network becoming less simplicity biased, and therefore less well aligned with structure in the data.  Indeed,  datasets such as MNIST and CIFAR-10 are thought to be relatively simple~\cite{spigler2020asymptotic,goldt2019modelling}.  

These patterns are further illustrated in \cref{fig3} (d)--(f) where we show scatterplots of generalization error v.s. CSR complexity for three target functions that vary in complexity (here obtained by corrupting labels).  The qualitative behaviour is similar to that observed for the Boolean system in \cref{fig:Cube_plots_1}.  The more simplicity-biased networks perform significantly better on the simpler targets, but the difference with the less simplicity-biased network decreases for more complex targets.  While we are unable to directly calculate the likelihoods because these systems are too big, we argue that the strong similarities to our simpler model system suggest that the same basic principles of inductive bias are at work here.  

\section{Discussion}
In order to generalize, high capacity models need a clear inductive bias towards functions that perform well on the data being studied~\cite{mitchell1980need}.
Here we show that DNNs (and also NNGPs and DNN-inspired kernels have a very specific kind of inbuilt Occam's razor (see \cref{Occam} for background on  Occam's razor(s)). 
This inductive bias can be quantified by an AIT coding theorem-like scaling of the prior as $P(f) \propto 2^{-K(f)}$  which can counteract the $2^{K}$ growth of the number of functions with complexity.  If this intrinsic inductive bias is slightly weaker, say $P(f) \propto 2^{- \alpha K(f)}$ with $\alpha < 1$, then it becomes much harder to overcome the $2^{K}$ growth and the learner will likely suffer from strong variance problems.   While we were not able to significantly increase the bias towards simplicity in DNNs,   too strong a bias towards simplicity can mean bias (instead of variance) problems because complex functions become hard to find~\cite{shah2020pitfalls,bhattamishra2022simplicity}.

An interesting direction to explore is the more formal arguments in AIT relating to the optimality of Solomonoff induction~\cite{li2008introduction,rathmanner2011philosophical,grau2024learning} (see also~\cref{app:solomonoff}). In particular, there may be fruitful links between simplicity bias, Solomonoff induction, and compression in deep learning, as recently discussed for example in the context of large language models~\cite{deletang2023language,li2024understanding}.

Another promising direction for exploration is to connect our high-level theory here with the more detailed calculations of generalisation in kernels~\cite{bordelon2020spectrum,cohen2021learning,canatar2021spectral,harzli2021double,cui2021generalization,simon2021neural}.  These works can tell us \textit{how} DNN inspired kernels need to align with data, but not so easily \textit{why} this would be so.  An extra challenge is that our work here has been in the context of discrete functions and classification, whereas the work on kernels is typically for continuous regression scenarios.
However, the classification setting allows us to study the inductive bias of a model over the entire space of input functions and simultaneously make concrete generalisation bounds.
We distinguished the first-order question of why high-capacity DNNs (and related models) generalise at all from the second-order question of how to further improve DNN performance. 
Firstly, the fact that DNNs are strongly biased towards simplicity gives a basis upon which to look for further inductive biases in other directions that may be orthogonal to simplicity.  Our results are derived in a  Bayesian setting, but do not preclude the fact that SGD optimization itself can also introduce useful inductive biases (see also~\cref{sec:introlit}).   In particular, SGD likely enables feature learning. Our understanding of how this works, even in the infinite width kernel or GP limit~\cite{seroussi2023separation,pacelli2023statistical,yang2023spectral}, remains in early stages, but feature-learning is likely to play an important role in explaining why DNNs outperform other methods on large datasets.

Much of the exciting recent progress in foundation models such as GPT-4 likely arises from optimiser-induced inductive biases beyond the simplicity bias at initialisation that may also play an important role there~\cite{bhattamishra2022simplicity}. Direct inspiration from biology can also provide inspiration for alternate inductive biases, as recently shown for image recognition in CNNs~\cite{evans2022biological}. Systematically studying and understanding how these inductive biases interact with data remains one of the key challenges in modern machine learning.

Finally, our observations about inductive bias can be inverted to identify characteristics -- such as limits on the complexity --  of the data that DNNs can successfully learn.  In particular,  the remarkable success of DNNs on 
a broad range of scientific problems~\cite{ching2018opportunities,mehta2019high,carleo2019machine,karniadakis2021physics}
suggests that their inductive biases must recapitulate something deep about the structure of the natural world~\cite{lin2017does}. By understanding why DNNs select specific solutions, we may, in turn, gain profound insights into the structure of nature itself. 

\acknowledgements
We thank Satwik Bhattamishra, Kamal Dingle, Nayara Fonseca, and  Yoonsoo Nam  for helpful discussions. C.M. acknowledges funding from an EPSRC iCASE grant with IBM (grant number EP/S513842/1)

\section*{Data Availability}
All code to generate the data used in this study is publicly available in the following zenodo repository \url{https://doi.org/10.5281/zenodo.13997032}

\section*{Code Availability}
All code to generate the figures in this study is publicly available in the following zenodo repository \url{https://doi.org/10.5281/zenodo.13997032}

\section*{Author Contributions}
CM and HR designed, conducted, and analyzed the experiments, and contributed to writing the manuscript. GVP conducted and analyzed experiments. AAL designed the project, analyzed the experiments, and contributed to writing the manuscript. 

\section*{Competing Interests}
The Authors declare no competing interests

\bibliography{MLrefs}
\bibliographystyle{naturemag}

\onecolumngrid

\appendix

\newpage

\setcounter{figure}{0}
\renewcommand{\figurename}{Fig.}
\renewcommand{\thefigure}{S\arabic{figure}}
\renewcommand{\thesection}{\arabic{section}}
\section{Background literature on generalization in DNNs}\label{sec:introlit}

In an opinionated 
 1995 commentary entitled ``Reflections after refereeing for NIPS [NeurIPS]''\citet{breiman1995reflections}, the influential late statistician Leo Breiman  wrote: 

\begin{quote} \textit{
There are many important questions regarding neural networks
which are largely unanswered. There seem to be conflicting stories regarding the
following issues:
\begin{enumerate}
    \item Why don't heavily parameterized neural networks overfit the data?
    \item What is the effective number of parameters?
    \item Why doesn't backpropagation head for a poor local minima?
    \item When should one stop the backpropagation and use the current parameters?
\end{enumerate}}
\end{quote}
In spite of enormous technical advances in the use of neural networks over the ensuing 28 years since Breiman's paper was written, these basic theoretical questions remain surprisingly relevant. 
Issues 1 and 3 were famously highlighted in 2016 by Zhang et~al.~\citep{zhang2016understanding} who demonstrated that AlexNet, a convolutional neural network (CNN) architecture, can obtain 100\% accuracy on CIFAR10 for both randomised and non-randomised labels.    Given that DNNs were known to have  extremely high expressivity/large capacity~\cite{zhang2016understanding,cybenko1989approximation,hornik1991approximation,hanin2019universal}, it is perhaps not surprising that they can in principle fit randomised labels on images, which is captured by a measure of capacity called  Rademacher complexity.     What was perhaps more surprising than their high Rademacher complexity was how efficient SGD is at finding these   
 complex solutions for randomized data.
The main impact of the Zhang et~al.~\cite{zhang2016understanding} paper, however,  was not its illustration of  the power of SGD, but rather the evocative way it framed and popularised, for a new generation of researchers, Breiman's big question (1) of \textbf{why}
 DNNs generalise despite their high expressivity/capacity.  Their work can be interpreted as demonstrating that  
   DNNs can easily ``memorize'' randomised data. For such solutions, generalization performance should be no better than random chance. This then begs the question: 
 given that DNNs can easily find these memorized solutions,  why do they instead normally converge  to solutions that do generalize well?    The authors demonstrated that explicit regularisation techniques (e.g.\ weight decay, dropout) were not necessary for DNNs to achieve these feats, further confounding the common intuition that regularization is needed to avoid overfitting. For our purposes, a key takeaway from their work is the demonstration that the good inductive bias observed in DNNs is something quite basic to how they operate.  

Another study that captured the broader imagination was by Belkin et~al.~\citep{belkin2018reconciling} who showed that
DNNs (and related kernel methods) exhibit a `double-descent'' phenomenon:  In the under-parameterised regime the generalization error exhibits 
the classical ``U-shaped'' bias-variance tradeoff that peaks when the number of parameters is equal to the number of data points. In the overparameterized regime, the generalization error decreases (the second or ``double'' descent)  with increasing ratio of parameters to data points down to some minimal risk (dependent on the dataset and architecture).   Again, while the phenomenon of double descent (and even multiple descent) has a much longer history in machine learning, see e.g.~\cite{loog2020brief} for some historical comments, the Belkin et~al.~paper~\cite{belkin2018reconciling} played an important role in popularising and framing the question of why DNNs behave so differently from the classical expectations of statistical learning theory. Furthermore, more subtle double descent effects with respect to the quantity of data, training epochs and model size have been observed in e.g. \cite{nakkiran2021deep}. These effects suggest that the bias induced by the architecture can be significantly affected in critical regimes that differ for each problem.

Belkin et al. \cite{belkin2018reconciling} also stimulated a number of follow-on works, including a series of important papers that used random matrix theory (RMT) techniques to analytically calculate the generalization error through the whole double-descent regime, albeit for typically simpler architectures, e.g.\  a two-layer neural network with random first-layer weights, and a simple data generating process~\cite{randomfeature,surprises,advani2020high}.  The calculations use similar techniques to earlier pioneering calculations by Seung and Sompolinsky~\cite{mcallester1998some}.  Such RMT methods have also been extended to kernel methods, where they can be used to analytically calculate the generalization error of neural network-inspired  processes (NNGPs)~\cite{harzli2021double}.  The same double-descent phenomena can be observed not only in NNGPs, but also in several other key machine-learning techniques, which all exhibit surprisingly good generalization in the overparameterized regime~\cite{pmlr-v80-belkin18a}.  What this similarity suggests is that the answer to Breiman's first question is most likely  broader than the direct context of DNNs trained by SGD and extends to other methods as well. 

A series of important recent papers, building on the pioneering work of Neal~\cite{neal1994priors} have shown that, in the limit of infinite width, DNN architectures can be described by neural network Gaussian processes NNGPs~\citep{lee2018deep,matthews2018gaussian,novak2018bayesian,garriga-alonso2018deep,yang2019tensor,yang2020feature} or by kernel methods such as the Neural Tangent Kernel (NTK)\cite{jacot2018neural}.   Importantly, these methods often display remarkably similar generalization behaviour to SGD trained DNNs of finite width~\cite{lee2020finite}.  It is not just the values of the generalization error that are similar, but also other trends, such as learning curves and differences between architectures (e.g.\ CNN-GPs perform better than FCN-GPs on classic image datasets).   These kernel methods appear to  capture essential aspects of how DNNs interact with the data, even if they may not capture everything (for example, feature learning \cite{yang2020feature}).  These methods are typically non-parametric, so counting parameters doesn't work. However,  they are very high capacity and so beg the same question 1 that Breiman raised.

It is well known, of course, that these infinite width limits differ from finite width optimiser-trained (and finite width Bayesian) neural networks in that they cannot perform feature learning, see e.g.~\citep{geiger2020disentangling,yang2020feature,roberts2022principles}. One way to define feature learning is in terms of correlations between $l$'th layer outputs $\mathbb{E}_{\theta}[f_l(x_i)f_l(x_j)]$. If they  remain fixed during training, as is the case for an infinite-width network, then no feature learning is said to occur.  For finite-width neural networks, these correlations do change, which  allows for transfer learning. A greater degree of representation learning has been linked to simpler functions \citep{oymak2019generalization,lou2021equilibrium} and better generalization \citep{baratin2021implicit}. 
While these differences with the kernel limit can be important in practice,   given that 
the kernel limits already are able to generalise despite high capacity e.g.\  they could also find ``memorizing'' solutions, it is unlikely that the first-order answer to Breiman's question will hinge on the differences between a full DNN and their kernel limits.

Empirical practice shows that there are many other factors that affect generalization for finite-width neural networks.    For example, Keskar et~al.~\citep{keskar2016large} showed that DNNs trained with small-batch SGD generalise better (by up to $5\%$) than the same models trained with large-batch SGD. They also provided empirical evidence for a correlation between smaller batch sizes and minima that are less ``sharp'' (inspired by flatness arguments from \citep{schmidhuber1997discovering}). 
A series of papers that examined this result further \citep{goyal2017accurate,hoffer2017train,smith2017don} suggest that better generalization with small batch SGD may be due to the increased number of optimisation steps taken per epoch. For example, \citep{hoffer2017train} showed that overtraining not only does not negatively impact generalization, but improves the performance of large-batch SGD to around the performance of small-batch SGD, see also \cite{geiping2021stochastic} for a comparison of GD to small-batch SGD.   The effect of batch size and learning rate on generalization performance may depend  on details such as the type of data and other factors. Even if the full story is complex,  it remains the case that tuning SGD hyperparameters to optimise generalization plays an important role in the practical use of DNNs.  Improvements can also come from Bayesian averaging procedures~\cite{izmailov2018averaging,wilson2020bayesian}. 
Differences of $5 \%$ can be enormously important in practice, and so it is not surprising that the literature on ways to improve DNN performance by tuning hyperparameters, or by data augmentation, is huge.  But since  DNNs already can perform reasonably well without hyperparameter tuning,  these further improvements in generalization cannot be the solution to Breiman's question 1.

In the main text, we therefore distinguish Breiman's first-order  question above from a second question:
   
\medskip
 \noindent \textbf{The 2nd order question of fine-tuned generalization:} Given a DNN that generalises reasonably well (e.g.\ it solves the overparameterisation/large capacity problem), can we understand how to improve its performance further?
\medskip

It is the first-order question of why DNNs generalise in the first place that we are mainly trying to address in this paper.   However, experience tells us that these two questions are often conflated and confused.    Improvements in generalization, driven by many different tricks such as hyperparameter tuning or data augmentation, are what most practitioners have experience with.  But as important as these tricks are for applications, they only improve on what is already a remarkable performance in the first place. 
Another reason these questions may be conflated links to the power of SGD to efficiently find low error solutions for DNNs trained on training sets.   But  this success does not  mean that it is a special property of SGD itself that is the solution to the first-order question. For that, one would have to show (perhaps with a great deal of patience and compute power) that other optimisation methods somehow converge onto hypotheses that generalise significantly less well. 

Having said that, it is important to recognize that there is a large and sophisticated  literature that locates the answer to Breiman's question 1 in specific properties of SGD.  These arguments go beyond the obvious point that if an SGD-trained DNN generalises well, then SGD played a part.  Rather, the kinds of  special properties of SGD that are considered include,  for example, that it can be proven for linear models that SGD on  linearly separable data converges to the maximum-margin linear classifier~\cite{soudry2018implicit}, which could help explain the implicit regularisation of DNNs.  
Full batch training is technically challenging, but appears to produce generalization accuracy that is quite close to that of smaller batch SGD~\cite{geiping2021stochastic}, so it may be that one should primarily investigate the properties of simple gradient descent to search for hints as to why DNNs generalise well. 
Moreover, kernel methods such as NNGPs are not trained by SGD, and yet generalise (nearly as) well as DNN counterparts do.
While it may be the case that the solutions of NNGP  kernels are related to gradient descent somehow, it is clearly the case that their good generalization does not depend on hyperparameter tuning of SGD.  

In this context, we also mention a study 
\citep{mingard2021sgd} that investigated how close the posterior distributions $P_{\textrm{SGD}}(f|S)$ of DNNs trained by stochastic gradient descent are to their fully Bayesian counterparts, which use an NNGP (or NTK)   kernel to calculate the posterior  probability $P_{\textrm{B}}(f|S)$ that the DNN expresses $f$ upon random sampling of its parameters, conditioned on $S$. 
They found that  $P_{\textrm{SGD}}(f|S) \approx P_{\textrm{B}}(f|S)$ for DNNs  -- including FCNs, CNNs, LSTMs on MNIST, Fashion-MNIST and the IMDb movie review dataset--  initalized with typical hyperparameters and typical learning rates. These results suggest that (to 1st order) SGD selects functions with a probability close to that of a Bayesian sampler that knows nothing about SGD.  Along a similar vein of logic, \citep{valle2020generalization} found that 
the marginal-likelihood PAC-Bayes bound 
captured the learning curves for finite-width SGD-trained neural networks, including the effects of architecture and data complexity, even though the marginal likelihood $P(S)$ itself was calculated with NNGPs, so that the bound calculation knows nothing of SGD.   Similarly, it was shown in~\citep{bernstein2021kernel} that some of these bounds  could be made tighter by using the NTK posterior mean.   The success of these bounds at capturing key aspects of generalization behaviour suggests again that a different property of DNNs beyond just the SGD optimiser alone is responsible for their solution to Breiman's question 1.

As mentioned above, the  fundamental reason why overparameterized DNNs don't fall prey to over-fitting, as predicted from the classical bias-variance trade-off arguments, may be shared by kernel methods (including NTK and NNGPs, which are close in performance to DNNs), and other related techniques.  All this evidence points to the importance of deriving a new and general statistical learning theory in the overparameterized regime.  One promising direction has been to  study simpler systems, such as linear regression, where the concept of ``benign overfitting'', an effect where the noise is fit without seriously affecting the simpler underlying model prediction,  can be more easily understood~\cite{bartlett2020benign}. These concepts have also been observed and partially explained for kernel regression~\cite{belkin2021fit,mallinar2022benign}  which is even closer to DNNs.   Another important series of  recent papers  used the replica trick to analytically calculate generalization errors for  kernel regression methods closely linked to DNNs~\citep{bordelon2020spectrum,spigler2020asymptotic,canatar2021spectral,cui2021generalization,simon2021neural}.  One key outcome of that work is that the overlap between the learner and the data (sometimes called model-task alignment~\cite{canatar2021spectral}) helps explain why these methods generalise well on the typical data set.   Kernel methods can be expanded into eigenfunctions, and if only a small number are needed to also describe the data (which happens when there is good model-task alignment), then the learner will have a low generalization error.  What this very interesting stream of work also helps illustrate is the importance of understanding structure in the data.  We are hopeful that such methods may lead to a new unified theory of generalization in the overparameterized regime. 

It is a widely held belief that DNNs are biased towards simple functions 
\citep{arpit2017closer,de2018random,mingard2019neural,valle2018deep,farhang2022investigating,bhattamishra2022simplicity,chiangloss}. Many independent bodies of work in the literature point towards the idea that the data that DNNs work well on is in one way or another simple. For example, it is well known that  
the effective number of parameters needed to describe a trained network is typically much less than the total number of parameters in typical neural networks -- aggressive pruning \citep{li2016pruning,frankle2018lottery} can result in networks performing equally well even with less than 10\% of the original number of parameters.  (Such works also suggest new insights into Breiman's question 2 above).  The success of pruning suggests simpler functions which need fewer parameters to express, even if a full DNN may be easier to train than a reduced one.
 
Another line of argument is exemplified by 
Lin et~al.~\citep{lin2017does}, who argue that deep learning works well because the laws of physics typically select for function classes that are ``mathematically simple''  and so easy to learn. 

The complexity of data is an important ingredient in understanding DNN performance.
 Several studies have attempted to directly calculate measures of  the complexity of commonly used image datasets. For example,  \citep{goldt2019modelling} studied the way in which neural networks learn low dimensional data embedded in a manifold of much higher dimension. 
\citep{spigler2020asymptotic} calculated the effective dimension $d_{eff}\approx15$ for MNIST (much lower than the $28^2=784$ dimensional manifold in which the data is embedded) and $d_{eff}\approx35<<3072$ for CIFAR10. For MNIST, individual numbers can have effective dimensions that are even lower, ranging from 7 to 13~\citep{hein2005intrinsic}.  These results suggest that the data lives on a much reduced manifold compared to what could be possible in the full dimensional space.  In other words, it is relatively simple compared to what it could be.

Yet another series of ideas that point in the same direction come from studies that investigate the order in which a DNN learns aspects of data
\cite{rahaman2019spectral,kalimeris2019sgd,refinetti2022neural,xu2022overview}.  By different measures of complexity, all these studies agree that, as it is trained, SGD first learns simpler representations of the data, and then more complex ones with more epochs.  This temporal behaviour is again consistent with data that is best described by simpler functions. 

Finally, we note a recent study by Shah et~al.~\cite{shah2020pitfalls} who argue that DNNs may be too simplicity biased, and thus miss some subtler and complex aspects of the data. This may indeed be an issue (see also Domingos on Occam's razor~\cite{domingos1999role}), but these arguments are mainly about 2nd order effects, not about the 1st order problem we are trying to address in this paper.  Another famous problem where a bias towards simplicity can be detrimental is in shortcut learning~\cite{geirhos2020shortcut}, where a DNN learns a simpler shortcut that works on sample data but which leads to poor performance in more general settings beyond the training model.

\section{Simplicity bias for DNNs on Boolean functions}

In this section, we review previous work on simplicity bias for supervised learning on Boolean functions. 
It was shown in~\cite{valle2018deep,mingard2019neural} that, upon random sampling, DNNs with ReLU activations are exponentially biased towards functions with lower Lempel-Ziv (LZ) complexity.   This bias holds, to first order, for a wide range of different parameter initialisations.  The observed bias was linked to the coding theorem from algorithmic information theory (AIT)~\cite{levin1974laws,li2008introduction}.  More specifically, it was argued that these (non-chaotic) DNNs follow a simpler simplicity bias bound~\cite{dingle2018input,dingle2020generic}, inspired by the AIT coding theorem, which takes the form:
\begin{align}\label{eqn:levin}
 P(f)\leq  2^{-a \tilde{K}(f) + b}
\end{align}
where  $\tilde{K}(f)$ is a suitable approximation for the Kolmogorov complexity (for example, LZ complexity $K_{LZ}(f))$ and $a$ and $b$ are parameters that depend on the choice of $\tilde{K}(f)$ and the map, but not on $f$.
Indeed, this bound (as well as a statistical lower bound~\cite{dingle2020generic}) is consistent with the observed data upon random sampling of parameters.   For further details of when it holds, and other systems that it can be applied to, see e.g.~\cite{dingle2018input,dingle2020generic,camargo2020Boolean,johnston2022symmetry}.

One of the advantages of classification of Boolean functions is that the approximations to the complexity of Boolean functions can relatively easily be defined and measured.  In the work above we represent each function $f$ as a binary string of length $2^n$  by ordering inputs ascending by their binary value.  The complexity of the function is then that of the string, and   the main complexity measure used is  based on Lempel-Ziv-76~\cite{lempel1976complexity}, which is somewhat less commonly used than other LZ methods, but is thought to correlate better with Kolmogorov complexity (see~\cite{dingle2018input} and \cref{app:complexitymeasures} for a longer discussion). 
 One might worry that the simplicity bias is somehow linked to LZ itself.  
Importantly,  however, different complexity measures have been shown to correlate well with one another for this problem.   In~\cite{valle2018deep}, it was shown that this measure correlates well with measures such as  the Boolean complexity measure, defined by the number of operations in the resulting Boolean expression,  and  the critical sample ratio (CSR)~\cite{arpit2017closer}, an estimate of the density of decision boundaries, and direct measures of Boolean complexity.  Similarly, in Bhattamishra et~al.~\cite{bhattamishra2022simplicity} sensitivity was used as a complexity measure, and this was also shown to correlate well with other complexity measures for LSTMs and Transformers.  In related work~\citep{de2018random} it was shown using Gaussian processes that upon randomly sampling parameters of an FCN acting on Boolean inputs, the functions obtained were on average much less sensitive to inputs than if functions were randomly sampled, similarly to the findings of~\cite{bhattamishra2022simplicity}. Functions with low input sensitivity are expected to be simple, thus proving another manifestation of simplicity bias present in these systems.
This robustness to the complexity measure is particularly interesting in light of Hao Huang's recent proof of the sensitivity conjecture~\cite{huang2019induced} since sensitivity (or mutational robustness~\cite{mohanty2023robustness}) is correlated with  $\log P(f)$.    Taken together, these results suggest that the qualitative features  of results we study in this paper should not be too sensitive to the exact complexity measure (see also \cref{app:fig:K_measures})  That being said, such comparisons of complexity measures have not yet, to our knowledge, been made for popular image datasets such as MNIST and CIFAR-10.

Another way to study simplicity bias is by direct calculations on simplified networks. For example, it was proven  
in \citep{mingard2019neural} that upon randomly sampling the parameters (with a distribution satisfying certain weak assumptions),  a perceptron with no bias term will produce any value of class-imbalance (``entropy'') with equal likelihood for Boolean functions. Because there are many more functions with high ``entropy'' than low ``entropy'', this implies an exponential  bias towards low ``entropy'' functions compared to a uniform distribution over those functions on the Boolean dataset that are expressible by a perceptron.
Because low $K(f)$ is implied by low entropy (this was shown exactly for the Boolean circuit complexity measure), these results imply a bias in the prior $P(f)$ of perceptrons towards simple functions.
In this work, it was also proven that for infinite-width ReLU DNNs, this bias becomes monotonically stronger as the number of layers grows. 
Furthermore, it was shown that within classes of functions of any given entropy, there was further bias towards functions with low LZ complexity (for example, in the maximum entropy class, the function $0101\dots$ was much more common than the mean).  Taken together, all these approaches point to the same result, namely an intrinsic bias towards simplicity in DNNs on Boolean data.

Finally, Yang et~al.~\citep{yang2019fine} showed that the simplicity bias starts to disappear when DNNs with erf or tanh activations enter the ``chaotic regime'' -- which happens for weight variances above a certain threshold, as the depth grows \citep{poole2016exponential} (note that ReLU networks don't exhibit a chaotic regime in the same way).
While these hyperparameters and activation functions are not typically used for training DNNs in practice, they do demonstrate that there exist regimes where simplicity bias breaks down. The explanation for the disappearance of the simplicity bias in the prior can be found in the next section which proves that in the fully chaotic limit, the prior over functions is fully uniform.

\section{Infinite depth chaotic tanh-activated DNNs have a uniform prior}\label{app:sec:infinite} 

In this section, we discuss the prior of fully chaotic tanh-activated DNNs.
As defined in \cref{app:Boolean:defs},
weights in our neural networks are initialised from Gaussian distributions with mean $0$ and variance $\sigma_w^2/N_l$, where $N_l$ is the width of the $l$'th layer.
\citep{hayou2019impact} showed that for tanh activated neural networks with bias initialisation $\sigma_b=0$, the border between the ordered and chaotic regimes occurs at $\sigma_w=1$.

Signal propagation in deep chaotic networks was studied extensively in~\citep{poole2016exponential,schoenholz2016deep}. The most relevant result for our purposes concerns the correlations between activations after $l$ layers, for neural networks at initialization, taking the mean field approximation (where preactivations $z^l_i$ are replaced by a Gaussian whose first two moments match those of $z^l_i$). This mean field approximation is exact for infinitely wide neural networks.

Specifically, it is shown in \citep{poole2016exponential}  that for neural networks with tanh activations and combination of $\sigma_w$ and $\sigma_b$ in the ``chaotic regime'' (for example, $\sigma_w=2.5$, $\sigma_b=0.0005$), as the number of layers tends to infinity
\begin{equation}\label{app:eqn:mftanh}
    \mathbb{E}[z^l_i(x_{a})z^l_j(x_{b})]\xrightarrow{l\rightarrow\infty}q^*\delta_{ij}\delta_{ab}+q^*c_{ij}(\sigma_w,\sigma_b)(1-\delta_{ij})
\end{equation}
where $x_a$ and $x_b$ are inputs to the network, and $z^l_i$ is the $i$'th preactivation in the $l$'th layer. $c_{ij}(\sigma_w,\sigma_b)$ is a correlation term between inputs $i$ and $j$, and it is further shown that $c_{ij}(\sigma_w,\sigma_b)\xrightarrow{\sigma_w\rightarrow\infty}0$ for any fixed $\sigma_b$.
By contrast, neural networks in the ordered regime (characterised by a small $\sigma_w$ and large $\sigma_b$) have an attractive fixed point at $c_{ij}=1$, that is all inputs become perfectly correlated.

For a neural network with $l\rightarrow \infty$, $\sigma_w\rightarrow \infty$, infinitely wide hidden layers and a single output neuron, \cref{app:eqn:mftanh} can be used for a single output node (by setting $i=j=1$). With $f(x_a)$ the final layer's output for input $x_a$,
\begin{equation*}
    K(x_a,x_b)=\mathbb{E}[f(x_a)f(x_b)]\xrightarrow{\sigma_w\rightarrow\infty,l\rightarrow\infty}q^*\delta_{ab},
\end{equation*}
which defines the kernel used in neural network Gaussian processes (NNGPs). The prior probability of an output $y$, $P(y)$ is given by
\begin{equation*}
    P(y)\propto \exp{-y^T K^{-1} y},
\end{equation*}
where $K$ is the kernel matrix for $m$ inputs, and $y$ the output vector (pre-thresholded) for the $m$ outputs.
As $K^{-1}$ is a diagonal matrix, all predictions are uncorrelated and have equal chance of being 0 or 1, so the probability of any thresholded function $f$ must be uniform:
\begin{equation*}\label{eq:uniform}
    P(f) = 2^{-m}.
\end{equation*}

A similar result was proved in \citep{hayou2019mean} but for the neural tangent kernel NTK. Indeed, NTK in the limit of infinite depth converges to an equicorrelated matrix with correlation $c=c(\sigma_w,\sigma_b)$, with the same limiting behaviour as the NNGPs.

Therefore DNNs are predicted to have no inductive bias at all on all functions in the limit of the full chaotic regime. In this light, our observation that the simplicity bias decays the deeper one enters into the chaotic regime  is not surprising.   A still open question is how to perform a quantitative calculation of the rate at which the bias disappears as a function of how far one moves into the chaotic regime. 

\section{Occam's razors}
\label{Occam} 

\subsection{Historical and philosophical background to Occam's razor} 

The assumption that simplicity is a virtue is widespread in science and philosophy~\cite{baker2004simplicity,sober2015ockham}, with historical antecedents going back at least to Aristotle  (384–322 BCE), who writes in \textit{Posterior Analytics}:
\begin{quote} \textit{
    We may assume the superiority, other things being equal, of the demonstration which derives from fewer postulates or hypotheses.}
    \end{quote} 
    Principles of this type are often called  Occam's razor, named after the English Franciscan friar and scholastic philosopher William of Occam (or Ockham) (1287–1347).  The most common form of the razor: 
    \begin{quote}\textit{
        Entities are not to be multiplied without necessity}
 \end{quote}
comes from another Franciscan philosopher, John Punch, who attributed this phrase to Ockham in his  1639 commentary on the works of Ockham's contemporary and teacher, John Duns Scotus. Although these exact words have not been found in Ockham's works, similar statements were commonplace in the scholarly discourse of the day, and their applications were mainly to metaphysics and theology, and not to science.  

With the advent of modern science, we can find many great thinkers invoking Occam's razor.  Isaac Newton's Principia Mathematica  has as the first of his   ``Rules of Reasoning in Philosophy'':
\begin{quote} \textit{
Rule I. No more causes of natural things should be admitted than are both true and sufficient to explain their phenomena. As the philosophers say: Nature does nothing in vain [Aristotle], and more causes are in vain when fewer suffice [Duns Scotus and Ockham]. }
\end{quote}
which echoes Aristotle, Duns Scotus~\footnote{Newton is probably echoing Duns Scotus's who wrote: \textit{
we should always posit fewer things when the appearances can be saved thereby . . . therefore in positing more things we should always indicate the manifest necessity on account of which so many things are posited.} from \textit{ Questions on the Books of Metaphysics of Aristotle, Book VIII, Q.1, n.22} } and Ockham. 

Albert Einstein presented his own version -- sometimes called Einstein's razor --  in his 1933 Herbert Spencer lecture entitled ``On the Method of Theoretical Physics''~\cite{einstein1934method}:
\begin{quote}\textit{It can scarcely be denied that the supreme goal of all theory is to make the irreducible basic elements as simple and as few as possible without having to surrender the adequate representation of a single datum of experience.}
\end{quote} 
This statement is often simplified to the slogan
\begin{quote}
\textit{``Everything should be kept as simple as possible, but not simpler''}
\end{quote}
which is pithier (and fits better on a t-shirt), but there is no evidence that Einstein used these words.

Notwithstanding the enduring popularity of Occam's razor across many disciplines,  serious questions remain about its \textbf{definition}, its \textbf{usage} and its \textbf{justification}.  In part, this state of affairs is caused by the widely  diverging contexts where it is employed, often in subtly different ways. A key distinction is often made between the ontological razor which minimises the number  of things being postulated,  and the syntactic razor, which minimises the complexity of  the hypotheses.  The former is often called parsimony, and may be closer to the formulation attributed to Occam, and the latter is sometimes called elegance, and may be closer to the most common use of the razor in sciences such as physics.   However, these delineations are not tight.  One may care more about the number of kinds of things being postulated rather than the total number of entities, for example.  

Another important set of difficulties arises from attempts to define exactly what simplicity means.  A description may be shorter in one representation than in another, which can introduce another layer of ambiguity. Algorithmic information theory (AIT) may have a formal answer to this question in terms of the Kolmogorov complexity, but, for philosophers, this solution opens up as many questions as it answers.

There are differences in how the razor is used.  For example, 
in his classic textbook~\cite{sober2015ockham}, Elliott Sober makes an important distinction between the razor of silence, which recommends agnosticism about unneeded causes, and the razor of denial, which argues that superfluous causes don't exist.  

Not surprisingly, there are also serious questions about the ultimate justification of Occam's razor.  Earlier thinkers typically used theological arguments to ground the razor in ontology.  More recent attempts have often focused on epistemological arguments.  Popper famously argued that simpler theories are superior because they are easier to falsify~\cite{popper1989logik}.  There have been many attempts to ground the principle in properties of Bayes theorem,  a famous example is David Mackay's argument about Bayes factors that penalise more complex models, see e.g.\ chapter 28 of \cite{mackay2003information}.  Kevin Kelly has argued that always choosing the simplest theory compatible with the evidence leads to an important efficiency gain in scientific reasoning~\cite{kelly2007new} and claims that   this may be the only non-circular justification of the razor.   

In the context of the natural sciences, one can of course take a different tack and argue that the success of applications of Occam's razor in the past provides an \textit{a-posteriori}  justification for the razor.  But this doesn't explain \textit{a-priori} why it works, and so only begs the question.  Moreover, there are many  subtle variations in the manner in which the razor is used in science.  The \textit{a-posteriori} justification therefore needs to be tempered by carefully addressing which version of the razor is doing the work in a scientific argument.

Clearly,  serious difficulties in providing non-circular justifications for Occam's razor remain.  In this context, it  is notable that one of the most extended and famous defences of Occam's razor by the philosopher Richard Swinburne simply declares that: 
\begin{quote} \textit{ .. it is an ultimate a priori epistemic principle that simplicity is evidence of truth.}
\end{quote}
In other words, Swinburne believes that Occam's razor in some sense basic, just as the law of non-contradiction is.  In fairness,  Swinburne  does provide many arguments in favour of Occam's razor in his book, but if he could have found an obvious rational justification he would have used it.  For further background on this  philosophical literature, we point to standard works~\cite{baker2004simplicity,sober2015ockham}.  

\subsection{Occam's razors in machine learning} 

Not surprisingly, there have been many invocations of Occam's razor in machine learning, see e.g.\ these reviews~\cite{domingos1999role,bargagli2022simple}.    Given the complex intellectual history briefly sketched above, it is not surprising that upon  careful reading one often encounters the same  problems of application, definition, and justification in this literature.  
On the other hand, because problems in machine learning are often simpler to define and control than typical questions from science or philosophy,  there is a growing literature connecting statistical learning theory to broader philosophical questions in epistemology, see e.g.~\cite{harman2007reliable}.

One way to interrogate the use of Occam's razor in machine learning is through the paradigm of supervised learning. We consider here, as in the main text, discrete problems such as classification.  Given a training set $S = \{(x_i,y_i\}^{m}_{i=1}$  of $m$ input-output pairs, sampled i.i.d.\ from a data distribution $\mathcal{D}$,  the machine learning task is to train  a model on $\mathcal{S}$ such that it performs well (has low generalization error) at predicting output labels $\hat{y}_i$ for a test set $\mathcal{T}$ of unseen inputs, sampled  i.i.d.\ from $\mathcal{D}$.  The model (often called a concept or a hypothesis $h$) can be chosen from a hypothesis class $\mathcal{H}$. For a hypothesis $h \in\mathcal{H}$ that minimises the error on the training set $S$, then  a standard result from statistical learning theory is that the  smaller the size of the hypothesis class, $|\mathcal{H}|$, the lower the expected generalization error of $h$ on the test set $T$~\cite{shalev2014understanding}.  

In a famous paper entitled ``Occam's Razor'', Blumer et~al.~\cite{blumer1987occam} made what at first sight appears to be a link between the razor and basic concepts in statistical learning theory.  
They  argue that using simple models can lead to guarantees (closely related to PAC-learning)  that a learning algorithm which performs well on the training data will also perform well on the fresh test data. Note that finding the minimum consistent hypothesis for many classes of Boolean functions can be NP-hard, so that they advocate losing the restriction of minimal simplicity in order to obtain polynomial learnability. 
They define model simplicity by assigning bit-strings to encode each hypothesis.  Then, roughly speaking, a hypothesis class of size $|\mathcal{H}| = 2^n$ can be described by codes of at most length $n$.  So a smaller hypothesis class is related to simpler hypotheses on this measure.  However, (and this is sometimes confused in the literature), there is no reason that the individual hypotheses need to be simple.  In other words, if one can exactly fit the training data with a hypothesis class that consists of a small number of very complex hypotheses, then these bounds also predict a relatively low expected generalization error on a test set. Despite their title, and some of the introductory text,  the authors of~\cite{blumer1987occam}  were self-consciously building on closely related and similar earlier results by Pearl~\cite{pearl1978connection}, who explicitly warns readers not to make the connection to Occam's razor (see also ~\cite{schaffer1993overfitting} for a classic discussion of these points). Their main purpose was not to prove Occam's razor, but rather to connect algorithms that can learn a certain near minimum complexity hypothesis in polynomial time with the classic notion of PAC learning.   In other words, while these famous results are theoretically sound, the authors do not, and  likely did not mean to  tell us what we might expect from a classic Occam's razor argument: Namely,  that in the situation where a set of complex hypotheses fits the training data as well as a set of simple ones, we should prefer simpler  versus more complex hypotheses.   In other words, despite the title and some of the introductory text, this famous paper turns out not to really be about Occam's razor at all.    

In his mini-review~\cite{domingos1999role}, Domingos forcefully argues that if two models have the same error on a training set, then there is no \textit{a-priori} reason to believe that the simpler one will perform better on unseen test data than  the complex one will.  He gives numerous examples where a complex model may in fact perform better.  By contrast, he advocates for a  different version of  Occam's razor:  If two models have identical \textit{generalization} performance, then there may be other reasons, such as interpretability or computational efficiency, that could favour choosing the simpler model over the more complex one. 

There may indeed be many reasons to prefer simpler functions.  For example,  random noise  will make the data appear more complex than the underlying generating signal. If one expects this to be the case, then  an inductive bias towards simplicity could help to avoid high-variance overfitting to the noise.  Such bias  may be advantageous  regardless of whether or not such bias would aid or hamper generalization performance on the underlying (de-noised) data.  In other words, one can imagine a scenario where the presence of noise implies that an inductive bias towards simpler models than the true underlying generating (de-noised) model may still lead to better performance on average.
 Although these are arguments to prefer simpler models, one wouldn't necessarily think of such reasons to prefer simpler theories as a classical Occam's razor argument.

\subsection{No Free Lunch theorems for supervised learning and Occam's razor } \label{sec:NFL}

A set of arguments that are frequently wielded against naive applications of Occam's razor come from the famous no-free-lunch (NFL) theorems for supervised learning~\cite{schaffer1994conservation,wolpert1996lack}, which are related to, and predate, no-free-lunch theorems for optimization~\cite{wolpert1997no}.  A  precursor to these papers was published in 1980 by Mitchell~\cite{mitchell1980need}, who emphasised that learning could not occur without some kind of inductive bias that reflected knowledge about the data.  The NFL theorems tell us (roughly) that two supervised learning algorithms will have the same performance when averaged over all possible input-output pairs \textit{and} all data distributions $\mathcal{D}$. Since one could simply use a completely random guessing algorithm, this implies that no algorithm will, on average over all $\mathcal{D}$ and all target functions, perform better than random chance.  Therefore, the NFL  implies that having a bias towards simplicity will, on average, not bring any performance gain, and so it formally negates this form of Occam's razor.

To make the  link between the NFL supervised learning theorem and Occam's razor  more concrete, consider the $n=7$ Boolean function problem we discuss in the main text. The NFL tells us that, when averaged over all $N_f =2^{128} \approx 3 \times 10^{38}$ possible Boolean target functions (and all data distributions), no learning algorithm will perform better than any other one. Therefore,  on average the generalization error will simply be that of random chance = 0.5.   As can be seen in Fig.~1.i, both a DNN and  a simple Occam-like K-learner that only allows hypothesis complexities $\leq K_M$,  generalise better than random chance on simpler target functions.  For the simple K-learner, the performance can be easily rationalised in terms of the reduction of the size of the hypothesis class $|\mathcal{H}|$, for example using arguments from~\cite{blumer1987occam}. 
For a DNN, similar arguments hold because the effective size of the hypothesis class is small when one includes the relative weighting of each hypothesis.  For an unknown target,  one could imagine a K-learning algorithm where $K_M$ is slowly increased until zero training error can be achieved.  The NFL tells us that on average over all targets, this algorithm will not perform better than random chance.  Since it clearly does better than random chance if  the target function complexities are not too large, it must do slightly worse on target functions with larger complexities, in order to compensate.  The same compensation  should hold for a DNN, which does significantly better than random chance on target functions of lower complexity. For an example of this compensation, see the DNN performance on the parity function in \cref{app:fig:bv_n5}, where the DNN generalization can be worse than random chance.  

It should be kept in mind that the number of functions grows as $2^K$ (see the green curve in Main text Fig.~1(g)), For this kind of ideal scaling, half of all functions will have the maximum complexity, so that the better than random performance on the lower $K_t$ targets can be easily offset even if performance on this  relatively large number of complex functions is only marginally worse than random chance. 

What this example also illustrates is that while algorithms with a bias towards simplicity cannot perform better on average than any other when averaged over all target functions, they may still do much better than random chance if one has some expectation that the target function complexity is less than the maximum complexity $K_{\rm max}$. 
The vast majority of functions are complex, so  having some upper bound on the complexity that is below  $K_{\rm max}$ immediately creates a strong restriction on the size of the hypothesis class. For the simple K-learning algorithm, for example, every bit lower than $K_{\rm max}$ implies a hypothesis class that is smaller by a factor of $2$, and therefore brings with it the  expectation of better generalization the more $K_M$ can be restricted.   Perhaps the key takeaway from this discussion is that  restricting the complexity of the hypothesis class is an extremely efficient and rather natural way to achieve a  smaller hypothesis class with concomitant improvement in generalization performance.   

As discussed above, arguments such as those of~\cite{blumer1987occam} are really about restricting the size of the hypothesis class, and not about complexity \textit{per se}.  For a slightly artefactual example, take $n=7$ Boolean functions, with the ordering explained in the main text, but restricted  to only those with $32$ $1$'s (a fixed entropy). Then the hypothesis class would be hugely reduced (by a factor of about a factor $10^9$) compared to the set of all possible functions for $n=7$.  If one achieved zero error on a training set with this restriction, then the classical arguments above suggest a generalization error less than random chance on a test set.   However, some functions in this class may be complex, e.g.\ those with random placement of the $1$'s, whereas others could be relatively simple, e.g.\ a sequence with 32 $1$'s followed by 96 $0$'s.   In other words, if we know something about our data, e.g.\ that it has a fixed class imbalance,  then that information provides an inductive bias that can be used to achieve better than random predictions.   What the NFL tells us is that if we genuinely know nothing about our target and the data distribution, then no inductive bias will a-priori work better than any other one.  The problem with this argument of course is that having absolutely no knowledge about the data one is trying to learn is an extremely odd state of affairs.     In other words, what the NFL theorem suggests is that a justification of Occam's razor  must be predicated on some prior knowledge that can be used to restrict the potential hypothesis class.    As mentioned above, a restriction on the maximum complexity of hypotheses, either as a hard cutoff, or else as a bias, is a very natural way to achieve such a restriction.

\subsection{Solomonoff induction and Occam's razor}\label{app:solomonoff}

Another important body of literature that links  machine-learning and Occam's razor was pioneered by Ray Solomonoff~\cite{solomonoff1960preliminary,solomonoff1964formal}, who was heavily influenced by the  programme of inductive logic of the philosopher Rudolf Carnap.   To first order, Solomonoff asked a simple question:  How likely is a universal Turing machine (UTM) to produce a particular outcome $x$ (for simplicity,  a bitstring), if one inputs randomly chosen programmes $q$?  The most likely programme to be chosen is the shortest one that can describe $x$.  This definition was independently discovered by Kolmogorov in 1965 \citep{kolmogorov1968three}, and by Chaitin in 1968 \citep{chaitin1969simplicity}, and forms the heart of the field of algorithmic information theory (AIT).  For more detail on AIT we refer to standard texts such as~\cite{li2008introduction}; here we provide a quick introduction to the main concepts.

The first concept we discuss is the \textit{Kolmogorov complexity} of a string $x$, which  is defined as the length $l$ of the shortest computer program $q$ run on a (suitably chosen) UTM $U$ that produces $x$:
\begin{equation}
C_U(x) = \min_q\{l(q):U(q)=x\}.
\end{equation}
This concept is formally uncomputable because of the halting problem of UTMs used to define it. While Kolmogorov complexity is defined w.r.t.\ a given UTM, the  invariance theorem states that for two UTMs $U$ and $V$,  \begin{equation}
    |C_U(x)-C_V(x)|\leq c
\end{equation}
for some $c(U,V)$ that is independent from $x$. This suggests that the Kolmogorov complexity is an intrinsic property of a string. 

The Kolmogorov complexity of most strings $x$ is extremely close to their length. Intuitively, this occurs because, for the $2^n$ strings of length $n$, there are $2^n-1$ strings of length less than $n$.  Therefore, half the strings in $\{0,1\}^n$ can at best be compressed  to a string in $\{0,1\}^{n-1}$; one quarter into strings in $\{0,1\}^{n-2}$ and so on.  

The next key concept, and the one pioneered by Solomonoff,  is the
\textit{algorithmic probability} \citep{solomonoff1964formal}, of a string $x$, which is defined for a prefix UTM $U$ (a prefix UTM is a UTM where no program is the prefix of any other program; required so Kraft's theorem can be applied to guarantee that $\sum_x P_U(x)\leq 1$ \footnote{A prefix Turing machine can be constructed as a Turing machine with one unidirectional (where the head can only move from left to right) input tape, one unidirectional output tape, and a bidirectional (the head can only move in either direction) work tape. Input tapes are read-only, and output tapes are write-only. All tapes are binary (no blank symbol), and work tapes are initially filled with zeros. For a program to be valid, the prefix UTM must halt with the input head at some point on the input tape. As the input head can only move left to right, it can never access further symbols on the input tape, and thus no valid program can be the prefix of any other valid program. \citep{li2008introduction}}). Prefix Kolmogorov complexity $K(x)$ is related to plain Kolmogorov complexity $C(x)$ via $K(x)=C(x)+C(C(x))+\mathcal{O}(C(C(C(x))))$ and $C(x)=K(x)-K(K(x))+\mathcal{O}(K(K(K(x))))$ \citep{bauwens2016relating}; given this close similarity we will rather loosely refer to prefix Kolmogorov Complexity as Kolmogorov complexity. Using a prefix Turing Machine, the  algorithmic probability is a discrete lower-computable semi-measure, defined as
\begin{equation}\label{app:eqn:uprior}
    P_U(x) = \sum_{q':U(q')=x} 2^{-l(q')},
\end{equation}
where $q'$ is any program that halts. This probability is what Solomonoff was essentially discussing in his famous 1964 paper~\cite{solomonoff1964formal}. It captures the probability of outputs $x$ when programs $q'$ are generated randomly (i.e.\ appending 0 or 1 to the end of the program with equal probability until the program runs). It is guaranteed to be $\leq 1$ by the Kraft inequality.  Also called the universal distribution, the algorithmic probability is uncomputable (for similar reasons as Kolmogorov complexity) and possesses many intriguing and powerful mathematical properties~\cite{kirchherr1997miraculous}.

Bounds on the algorithmic probability were established in 1974 by Levin \citep{levin1974laws}
\begin{equation}\label{app:eqn:levin}
    2^{-K(x)}\leq P_U(x)\leq 2^{-K(x)+\mathcal{O}(1)}
\end{equation}
where the $\mathcal{O}(1)$ is independent of $x$ but dependent on the UTM $U$.  This bound tells us that strings $x$ that have a low $K(x)$ are exponentially more likely to appear upon random sampling of programmes than strings that have larger $K(x)$.  Intriguingly, the proof of Levin's upper bound relies on the uncomputability of the algorithmic probability $P_U(x)$; it doesn't work for computable probability measures. 

One important way this concept has been used in machine learning is through the concept of 
\textit{Solomonoff induction}, which uses Bayes' rule with a 0-1 likelihood and the algorithmic probability from \cref{app:eqn:uprior} as the prior to predict the next characters in a binary string. Explicitly, given the a string $x$, the distribution over continuations $y$ (of length $l$ of $x$ is given by 
\begin{equation}
    P(y\mid x) = \frac{P_U(xy)}{P_U(x)},
\end{equation}
where the $xy$ denotes the concatenation of string $x$ with string $y$.
The posterior distribution minimises expected surprise if the source of the strings is algorithmic in nature \citep{solomonoff1964formal}, and thus the string with the shortest description length is most likely to generalise well.

Solomonoff induction is uncomputable, but it has formed the  backbone of other interesting theoretical learning agents. For example, Marcus Hutter's AIXI \citep{hutter2000theory} is a theoretical reinforcement learning agent which combines Solomonoff induction with sequential decision theory. It is Pareto-optimal  (there is no other agent that performs at least as well as AIXI in all environments while performing strictly better in at least one environment), and self-optimising (the performance of the policy $p$ will approach the theoretical maximum for the environment $\mu$ when the length of the agent's lifetime goes to infinity, when such a policy exists) \citep{hutter2004universal}. AIXI algorithms are also uncomputable, although there are computable approximations like AIXItl, which relies on cutting down the space of allowed programs by setting maximum run-times (avoiding the halting problem).

In the machine learning literature, another influential computable approximation is \textit{Schmidhuber's speed prior}~\citep{schmidhuber2002speed}, which includes not just programme length, but also runtime (speed).  Schmidhuber pointed out that some programs that are simple may be hard to compute (have a long runtime) which does not match entirely with our intuitions of simplicity. The complexity measure associated with this speed prior $Kt(x_n)=\min_q \{l(q) + \log t(q,x_n)\}$ where $x_n$ is a length-$n$ bit string, and a prefix program $q$ computed $x_n$ in $t$ timesteps.
The advantage of the speed prior is that it is computable, even if it is not as optimal as the true universal distribution.

Another set of influential ideas that provide computable approximations to concepts inspired  by Solomonoff induction come under the rubric of Minimum Description Length (MDL) principle, originally proposed by Rissanen~\cite{rissanen1978modeling}. The basic idea is to find the hypothesis $H$ that can be used to compress the data $D$ such that the length of the compressed data plus that of the hypothesis is minimised. 
In other words, a very complex hypothesis may lead to a very compact description of the data, but the size of the hypothesis gets penalized in MDL.  Similarly, a simple hypothesis may not compress the data as well, and so also be penalized. 
Thus the MDL is somewhat similar in spirit to an Occam's razor in that short descriptions are preferred, but what it minimizes is more than just the hypothesis, as one might do in the simplest versions of Occam's razor.      For more on this  rich field, see e.g.\ the classic book by Gr\"unwald~\cite{grunwald2007minimum}. 

Rathmaner and Hutter~\cite{rathmanner2011philosophical} provide a long discussion of potential philosophical implications of the universal distribution, including the claim that it can ground Occam's razor (see also~\cite{kirchherr1997miraculous}).      These authors argue that one of the advantages of this formulation of the razor is that on the one hand, it keeps all possible hypotheses  that are consistent with the data (Epicurus's  principle), while at the same time giving a larger weight to simpler ones (Occam's razor).  Moreover, it does so in a way that can be proven to be optimal (albeit with some key caveats, including that it is uncomputable).  In a more technical paper,  Lattimore and Hutter \citep{lattimore2013no} argued a ``free lunch'' exists when using the universal prior, for all strings of interest (i.e.\ non-random strings). 

Do these profound arguments from AIT provide the long-sought-after justification for Occam's razor?  This question has recently been explored at length in an important study by Sterkenburg \citep{sterkenburg2016solomonoff,sterkenburg2018universal}. Results relying on Solomonoff induction suffer from the issue that any string can be made to be arbitrarily simple by some apt choice of Turing Machine. Sterkenburg argues that although these choices become irrelevant in the limit of infinite data streams, arguments such as those in \citep{lattimore2013no} reduce to the assumption of effectiveness (broadly speaking, computability), which he claims undermines the justification of Occam's razor.  In a similar vein,  \citep{neth2022dilemma} shows that the convergence of $K_U$ to $K_V$ in the limit of infinite string length does not hold for computable approximations for $K_U$ and $K_V$.  As Solomonoff induction is uncomputable, these approximations would be key for any practical implementation, and the author argues that because these computable approximations would be dependent on the computer used,  they weaken the justification Solomonoff induction makes for Occam's razor.  

Without doubt, the last philosophical word has not yet been written on this fascinating connection between AIT and Occam's razor.  Although DNNs are computable, and therefore cannot reproduce a universal distribution, since the latter  depends critically on uncomputability,  there is an interesting qualitative similarity between Solomonoff induction and learning  with simplicity-biased DNNs.  In both cases, all hypotheses are included in principle (Epicurus) while simpler hypotheses are exponentially preferred (Occam).

\section{Definitions for Boolean functions and neural networks}

\subsection{The Boolean system}\label{app:Boolean:defs}

In this section, we will carefully and explicitly show how a Boolean function $f$ can be represented in a form for representation by a neural network, and how it can be represented as a bitstring for which the LZ complexity estimator~\cref{eq:CLZ} can measure the LZ complexity $K$.

Consider a function
\begin{equation*}
    f:\{0,1\}^n\rightarrow \{0,1\}.
\end{equation*}
There are $2^{2^n}$ different functions of this form. For a neural network, we represent $f$ as a dataset
\begin{equation*}
    \mathcal{B}(f)_n=\{(x_i, f(x_i)), \dots, (x_{2^n}, f(x_{2^n}))\}
\end{equation*}
where $x_i\in \{0,1\}^n$ are binary vectors of dimension $n$, and $f(x_i)\in \{0, 1\}$. $f$ can be represented as a binary string of length $2^n$, by ordering outputs $f(x_0)f(x_1)\dots f(x_{2^n})$. Inputs are ordered by ascending binary value (for example, for $n=2$: $00$, $01$, $10$, $11$).
The LZ complexity of $f$ is defined as the complexity of its string representation, using the LZ estimator~\cref{eq:CLZ}. This representation has the ordering property built in -- the string can be decoded into the $(x_i,y_i)$ pairs using $\mathcal{B}(f)_n=\{(bin(i),f[i])\}_{i=0}^{i=2^n}$ where $f[i]$ denotes the i'th digit in $f$, and $bin(i)$ is the binary representation of $i$.
Using a much more complex ordering scheme would change the values of LZ complexity measured (and lead to very different distributions over $P(K)$); but this would not capture the true complexity of the dataset well, as one would need to specify the ordering with more bits.

\vspace{10pt}
For example, consider $n=2$, and the binary representation of a particular $f$,
\begin{equation*}
    f=0110
\end{equation*}
for which we could calculate its LZ complexity, $K_{LZ}(0110)$.
The DNN, which requires inputs in $\mathbb{R}^2$ and outputs in $\mathbb{R}$ will receive the following input-target pairs: 
\begin{equation*}
    \mathcal{B}(f)_2=\{([0,0],[0]),([0,1],[1]),([1,0],[1]),([1,1],[0])\}.
\end{equation*}

A training set $S$ of size $m$ is a subset of $\mathcal{B}_n(f)$ with $m$ elements. It has a corresponding test set $T = \mathcal{B}_n(f) \setminus S$. Explicitly, we could write $S_m(f)$ to denote the fact that $S$ is of size $m$ and comes from the function $f$, however, being this explicit is not usually required.
There are $\binom{2^n}{m}$
unique training sets of size $m$ for each function $f$.

Because Boolean functions are  foundational in computer science, there is a long history of studying them with DNNs.  For example, Minsky and Papert~\cite{minsky1969perceptrons} famously showed that a 1-layer Perceptron could not represent XOR, which is said to have contributed to the first AI winter.   Since then a lot of work studying what functions DNNs can represent has worked with Boolean functions, see e.g.~\cite{mozeika2020space} for a recent study of function space.
\subsection{Deep Neural Networks}\label{app:dnns:defs}

Throughout we use standard fully connected neural networks (FCNs). 

\textbf{For Boolean functions}
We restrict the functions expressed by the DNN $\mathcal{N}(x;\Theta)$ to the domain $x\in\{0,1\}^n$ and codomain $\{0,1\}$ by defining the output functions $f(x)$
$$
f(x)=\mathbbm{1}(\mathcal{N}(x;\theta)>0).
$$
When training with MSE loss, $-1/1$ targets are used on the raw (un-thresholded)  output of the DNN,  $\mathcal{N}(x;\theta)$.

For an n-dimensional Boolean system, a neural network with input dimension $n$ and output dimension $1$ has the correct dimensions to model $\mathcal{B}_n$. Throughout the main text, we use $n=7$, with fully connected DNNs with $1\leq N_L\leq 10$ hidden layers with width $40$. 

The weights of the $l$'th layer are initialized i.i.d.\ from Gaussian distributions with mean $0$ and variance $\sigma_w^2/N_l$, where $N_l$ is the width of the $l$'th layer. In the text, we drop the $N_l$ factor, and show $1\leq\sigma_w\leq 8$ (with additional experiments with $\sigma_w<1$ in e.g.\ \cref{app:fig:ordered_plots}).
Bias terms were initialized i.i.d.\ from Gaussians with mean $0$ and variance $\sigma_b^2 = 0.1\sigma_w^2/N_l$.
We experimented with different values of $\sigma_b$, and found no induced substantial differences in network behaviour for the properties we studied, particularly after training. The DNNs use tanh activations, unless otherwise specified.

\textbf{For MNIST and CIFAR-10}
We use the full 10-class datasets, and as such use DNNs with 10 output neurons and softmax functions for training.  We used FCNs of 2 to 10 hidden layers of width $200$. They  were trained with SGD with batch size 32 and lr=$10^{-3}$  until $100\%$ accuracy was first achieved on the training set; we tested that our results are robust to changes in the learning rate, batch size and layer width.   While these FCNs are far from SOTA architectures, they have the advantage that they are easy to interpret, and are still highly expressive, and so solve the 1st order question of generalization that we are addressing in this paper. 

\subsection{Transitions to chaos}

FCNs are in the ordered or chaotic regimes (or the edge of chaos, the boundary between the two) depending on the values of $\sigma_w$ and $\sigma_b$. In the ordered regime, $\sigma_b$ dominates, and as inputs share common bias vectors all inputs approach the random bias. In the chaotic regime, $\sigma_w$ dominates, and inputs are randomly projected \cite{lee2018deep}. As depth increases, networks approach their ordered and chaotic limits exponentially quickly, except at the edge of chaos, where the decay towards the fixed point is much slower (non-exponential).

Lee et al. \cite{lee2018deep} further show a theoretical phase diagram for FCNs with tanh activations (Figure 4a in their paper). Our experiments use $\sigma_b=0.1\sigma_w$. In \cref{fig:Cube_plots_1}(a,b), the tanh-activated FCN with $\sigma_w$=1 is slightly in the ordered regime, and all other values of $\sigma_w$ are in the chaotic regime. The ReLU-activated DNN is in the ordered regime. Experiments in \cref{app:fig:ordered_plots} use $\sigma_b=0$, but are all in the ordered regime as $\sigma_w<1$ in all cases (which is a sufficient condition \cite{lee2018deep}).

The trained networks are initialised without bias terms, so the FCNs with $\sigma_w=1$ in \cref{fig:Cube_plots_1}(c-f) are all initialised on the edge of chaos. The other networks are initialised in the chaotic regime.

\section{Bias-variance tradeoff for learning Boolean functions with a K-learning cutoff.}\label{app:bias_variance_main}

\subsection{Background on bias-variance tradeoff} 
The bias-variance tradeoff is traditionally formulated   for regression with mean square error (MSE) loss. Given some training data $S=\{(x_1,y_1),\dots,(x_m,y_m)\}$ drawn from a joint distribution $P(X,Y)$, and an approximation $f_{\mathcal{S}}$ that minimises the MSE loss on $\mathcal{S}$, assuming a unique solution for each $\mathcal{S}$, the  expected error can be \citep{desa2018}
\begin{equation*}
\underbrace{\mathbb{E}_{x, y, S} \left[\left(f_S(x) - y(x)\right)^{2}\right]}_\mathrm{Expected\;Test\;Error} = \underbrace{\mathbb{E}_{x, S}\left[\left(f_S(x) - \bar{f}(x)\right)^{2}\right]}_\mathrm{Variance} + \underbrace{\mathbb{E}_{x, y}\left[\left(\bar{y}(x) - y(x)\right)^{2}\right]}_\mathrm{Noise} + \underbrace{\mathbb{E}_{\mathbf{x}}\left[\left(\bar{f}(x) - \bar{y}(x)\right)^{2}\right]}_\mathrm{Bias^2}.
\end{equation*}
$f_S$ is the model's unique prediction for $y(x)$ -- the true labels -- given training data $S$, $\bar{f}=\mathbb{E}_S f_S$ is the mean prediction averaging over datasets, and $\bar{y}(x)= \int y P(y|x) dy$ is the mean label for datapoint $x$.

These terms are traditionally called the variance, noise and bias terms. The variance term measures the expected square difference between the function fitted to $S$ and the average function fitted to $S$)
However, this decomposition does not demonstrate in full generality the bias-variance tradeoff. For a start, it assumes that there is a 1-1 correspondence between $S$ and $f_S$.  

More importantly for this paper, for classification problems and 0-1 loss, there is no neat decomposition into bias and variance~\citep{kohavi1996bias,domingos2000unified,geurts2009bias}. Here is  an intuitive argument for why this must be so: the performance of a maximally incorrectly biased learning agent (one which always arrives at the wrong answers) will be improved by the addition of some variance.
Nevertheless, there have been attempts to capture the essence of bias-variance tradeoff for classification\citep{kohavi1996bias,domingos2000unified,geurts2009bias}. These are:
\begin{itemize}
    \item Noise is the unavoidable component of the error (independent of the learning algorithm)
    \item Bias is error incurred by the main (mean or mode depending on loss) prediction relative to the optimal prediction. This is either due to (1) the model is not sufficiently expressive enough to express all functions or (2) the model's inductive bias is worse than random.
    \item 
    In the underparameterised regime, variance is traditionally equated with overfitting, where too complex a model fits random noise in addition to the simpler underlying signal.   For overparameterized models there are further sources of variance. Because of the high capacity, there are typically multiple hypotheses that the model can produce that can fit the complete data. Without a regularization procedure, this becomes a source of variability.  Furthermore,  the sparse nature of the training set w.r.t.\ the model's capacity can lead to different hypotheses being chosen upon different i.i.d.\ instantiations of a training set from the full data distribution.   These sources of variance are present in both regression and classification problems.
\end{itemize}

To study aspects of the bias-variance tradeoff we used a K-learner in the main text which cuts off all hypotheses with complexity greater than a maximum $K_M$. (See \cref{subfig:main_bv}). In this appendix, we perform further experiments for $n=5$ and $n=7$ Boolean functions, (see \cref{app:fig:bv_n7,app:fig:bv_n5}).   The  hypothesis class of all functions with LZ complexity less  than or equal to $K_M$ is defined as:
$$
\mathcal{H}_{K_M} = \{f:K_{LZ}(f)\leq K_M\},
$$
We used this limited hypothesis class for both a uniform learner and for DNNs, and describe these methods in more detail below.

\subsection{The unbiased learner}
The basic algorithm for the unbiased learner $U_{\leq K_M}$ is quite simple.   We fix a target function $f_t$, a training set size $m$, and a cutoff maximum complexity $K_M$.  The training set sizes are chosen such that we can exactly enumerate over all functions that exactly fit the data, which avoids worries about a uniform sampling over functions.   
For each function $f$, we measure its complexity and discard it if $K_{LZ}(f)> K_M$. For the functions that are kept, we measure  the training error $\epsilon_S$ on the training set of $m$ input-output pairs and the generalization error $\epsilon_G$ on the test set made up of $n-m$ input-output pairs,  and record in the figures,    the mean and the standard deviations as a function of the complexity $K$.

For the $n=5$ systems there are  only $2^{32}\sim 4 \times 10^9$ strings so full sampling is relatively straightforward.    We (randomly) select a fixed training set $S$ of size $m$, and sampled over all functions $f$ for several target functions $ft$, namely \\
 (a) `0011' $\times$ 8 (a simple function); \\ 
    (b) `01001111000111111110101111110100' (a complex function); \\
    (c) `1001' $\times$ 8 (a lowK lowP function); \\
    (d) the parity function.\\
The simple functions (a) and (b) are chosen to contrast simple and complex targets.   (c) is a simple function which the DNN has difficulty producing; it was not observed in $10^8$ random samples. In other words, even though it has low complexity, the DNN is not biased towards it.  Such lowK lowP functions are also seen in other input-output maps~\cite{dingle2020generic}.  A bound can be derived that relates the distance that  $\log P(f)$ is below the AIT-inspired upper bound from~\cite{dingle2018input} to the  randomness deficit in the set of inputs. In other words, lowK lowP functions have inputs that are simpler than the average inputs.        One consequence of this connection is that the total probability of such lowK lowP functions is small, because the vast majority of inputs are complex, and so only a relatively small fraction are available to create such lowK lowP outputs (Note that the arguments in ~\cite{dingle2020generic} hold for discrete inputs, and here the weights are continuous; nevertheless the arguments can be extended if an arbitrary discretization is applied to the weights).   Finally,   the parity function is chosen because it is thought that DNNs have a hard time learning it, and so this provides an example of a function that the DNN may be biased against.

For the $n=7$ system, exhaustive enumeration is harder because the total number of possible functions is so large. Therefore  we take a relatively large training set $m=100$, so that  we can exhaustively enumerate over all $2^{28}$ strings that exactly fit the training data. When $\mathcal{H}_{\leq K_M}$ is not sufficiently expressive to reach zero training error, we sample  $10^6$ further strings  by perturbing a uniformly sampled integer $k$ in range $0<k<10$) labels using the a) true function b) all 0's or all 1's function and c) a randomly generated function with high symmetry, as bases for the perturbation. This generates a distribution over low-complexity functions with low train error.  This process produced results that looked qualitatively similar to those seen for the $n=5$ system (which is exact). 

Experiments on these uniform learners are shown in \cref{subfig:main_bv,app:fig:bv_n7,app:fig:bv_n5}. They illustrate some aspects of the bias-variance tradeoff for classification of the different possible target functions.  Some basic patterns are discussed below. 

Firstly, we note that these plots do not show double-descent behaviour as in~\cite{belkin2018reconciling} because we are changing the inductive bias of the model by increasing $K_{M}$.  We are not fixing the model and changing the ratio of parameters to data. 

For $K_M$ much less than  the complexity of the target $K_t=K_{LZ}(f_t)$, it will normally be the case that the learner cannot achieve zero training error, and most of the error will be due to bias. 
As $K_M$ is increased, there will be a $K'\leq K_t$ for which a function in $\mathcal{H}_{\leq K'}$ will first fit $S$ with zero error.  For large training sets, $K'$ will roughly equal $K_{LZ}(f_t)$. For very small $|S|$, $K'$ can be quite small, for example, if a function is highly class-imbalanced towards $0$'s, then for smaller training sets the probability  to find an i.i.d.\ training set with all $0's$ which could be fit by a trivial function $f=0000.....$ with minimal complexity may not be negligible. 

The smallest $\mathcal{H}_{\leq K'}$ with $\epsilon_{S}=0$ will have the smallest PAC-Bound on generalization error. In this case, the error due to bias may be small if the main prediction is close to the true complexity; this typically occurs when the true complexity of the function is close to this $K'$, and the error due to variance will also be small (as there are few functions in the posterior).

For $K=K_{max}$, $|\mathcal{H}_{\leq K_{max}}|=2^{2^n}$.  All possible functions on the test set are present and equiprobable, so the mean generalization error will be exactly 0.5. The error due to variance here will be the largest, as there is a huge range of functions to choose from;  depending on the target function the error due to bias can also increase as the learner is no longer biased towards low-complexity functions.

\subsection{Defining priors and posteriors for the unbiased learner} 

For an unbiased learner $U_{\leq K_M}$, the prior,
$P_{\leq K}(f)$, and likelihood rule, $P_{\leq K}(S\mid f)$, can be defined as follows:
\begin{align*}
    P_{\leq K}(f) = \mathbbm{1} [f \in \mathcal{H}_{K_m}] / \vert{\mathcal{H}_{K_m}}\vert \\
    P_{\leq K}(S\mid f)=\mathbbm{1}\left[\epsilon_S(f)==\min\{\epsilon_S(f'): f'\in \mathcal{H}_{K_m}\}\right]
\end{align*}
where $\epsilon_S(f)$ denotes the error of a function $f$ on the training set $S$. Explicitly, we have a uniform prior over functions with $K_{LZ}\leq K_M$, and a posterior distribution that is uniform over functions with $K_{LZ}\leq K_M$ that achieve the minimum possible training error.

\subsection{Neural networks with a maximum complexity cutoff $K_M$}
If we restrict hypotheses to 
 $\mathcal{H}_{\leq K}$,  then the posterior for DNNs trained on a training set $S$ can be defined as:
$$
P_{\leq K_M}(f\mid S) = \frac{P_{SGD}(f\mid S\mathbbm{1}(f\in \mathcal{H}_{\leq K})}{\sum_{f'} P_{SGD}(f'\mid S\mathbbm{1}(f'\in \mathcal{H}_{\leq K})}.
$$
The posterior $P_{\leq K_M}(f\mid S)$ is approximated by training 5000 DNNs to 100\% training accuracy (with AdvSGD, batch size 16, and networks of width 40), and rejecting all functions not found in each $\mathcal{H}_{\leq K}$, as with the unbiased learner.  For the $n=7$ system, we used a training set of size $100$ when restricting the complexity.  For each function that the DNN converges to, that is  within the restricted hypothesis class, we calculate the complexity, as well as $\epsilon_G(f)$ and $\epsilon_S(f)$. We  record their averages and standard deviations in the plots \cref{subfig:main_bv,app:fig:bv_n7,app:fig:bv_n5}.

The ReLU-based DNN and the $\tanh$ based DNN with $\sigma_w=1$  have a strong inductive bias towards simple functions, meaning in general the probability to obtain a highly complex $f$ is much smaller than the probability to obtain a  simple $f$. The strong suppression of these complex functions means that removing them with $\mathcal{H}_{\leq K}$ typically has a  relatively small impact on generalization error. By contrast,  the chaotic neural networks do not suppress these complex functions strongly enough to counteract the exponential growth in $\mathcal{H}_{\leq K}$ with complexity, and thus errors due to variance can be as  large as for the unbiased learner.

\section{PAC-Bayesian Bounds}\label{app:PAC-Bayes}

\subsection{Background on non-uniform PAC and PAC-Bayes bounds} 
The influential concept of PAC-Bayes bounds was introduced by 
McAllester in the late 1990s.  In~\cite{mcallester1998some} he first introduced  ``preliminary bound'' that is easy to use and understand.  In the realisable case 
 (where 0 training error can be achieved),  
 for all hypotheses $f$ consistent with a  training set $S$ of size $|S|=m$, the generalization error $\epsilon(f)$ can be bounded by
\begin{equation}
    \forall D, P_{S \sim D^m}\left[\epsilon_G(f)\leq\frac{\ln\frac{1}{P(f)}+\ln\frac{1}{\delta}}{m}\right]\geq 1-\delta\label{app:eqn:pac_main}
\end{equation}
where $0 < \delta < 1$, and $P(f)$ is a probability measure or prior on all functions (here we assume the set of all functions $f \in \mathcal{H}$ is finite).  We will call this bound the non-uniform PAC bound  because it can be derived by standard union-bound techniques that are also used for  similar textbook uniform PAC-like bounds (see also~\cite{shalev2014understanding}). In fact, in the simplest case where the prior is uniform so that $P(f)=1/|\mathcal{H}|$, the bound above reduces to the standard textbook uniform PAC bound~\cite{shalev2014understanding}:
\begin{align}\label{app:eqn:PACC}
\forall D, P_{S \sim D^m}\left[\epsilon_G\leq\frac{\ln(|\mathcal{H}|)+\ln\frac{1}{\delta}}{m}\right]\geq 1-\delta.
\end{align}.

The non-uniform PAC bound captures the simple intuition that if a learner has an inductive bias $P(h)$ that in some way aligns with the data so that the learner can obtain zero training error on $S$ with a high probability hypothesis, then we expect the generalization error to be smaller than in the case where there is no good inductive bias.   Of course, we can also see that in the opposite case of an (anti)  inductive bias $P(f)$ that can only fit the data with hypotheses for which $P(f) < 1/|\mathcal{H}|$, then the error will be greater than a uniform learner.  Another simple way to understand the effect of inductive bias is to consider the simple case  when prior knowledge allows you to exclude certain hypotheses. In that limit,  $P(h)$ would be highly concentrated and uniform  on some subset of $\mathcal{H}$. Then \cref{app:eqn:pac_main}  approaches the standard uniform PAC bound~\cref{app:eqn:PACC}, but for  a reduced hypothesis class, allowing a tighter bound. 

McAllester's PAC-Bayes bounds are more sophisticated than the simple bounds above, but they essentially also capture the effect of an inductive bias. We will use  the marginal-likelihood version of the PAC-Bayes bound for binary classification defined in\citep{valle2018deep,valle2020generalization}, which is closely linked to the original  bounds derived in~\citep{mcallester1999pac}. For a given training set $S$, with set of consistent hypotheses $U(S)$,
the expected generalization error (over runs of the training algorithm) is bounded by
\begin{align}
\label{app:eqn:guillermo_pac_bayes_1}
\forall D, P_{S \sim D^m}\left[\epsilon_g(Q^*)\leq 1-\exp\left(-\frac{\ln\frac{1}{P(U(S))}+\ln\frac{2m}{\delta}}{m-1}\right)\right] >1-\delta,
\end{align}
each hypothesis $h$ is chosen with the prior probability $Q^*(f)=\frac{P(f)} {\sum_{f\in U({S})}P(h)}$ so that $\epsilon(Q^*)$ is the expected generalization error, and  $P(U({S}))=\sum_{f\in U({S})}P(f)$ is just the marginal likelihood.

This marginal-likelihood bound has been shown to work remarkably well for DNNs, capturing the effect of architecture, model complexity, and data set size~\cite{valle2020generalization}.  Its good performance suggests that the marginal likelihood $P(U(S))$  captures something important about the effective capacity of the DNNs.  For the case that $P(f)$ is the probability that a DNN expresses a function (or hypothesis) $f$ upon random sampling of parameters, then $P(U(S))$ can be interpreted as the probability that upon this random sampling of parameters, the DNN  obtains zero classification error on $S$.  Since a larger $P(U({S}))$ implies that the sum over the probability of hypotheses that obtain zero classification error on $S$ is larger, this implies that $P(U({S}))$ provides a measure of the inductive bias of the DNN on the data (at least at initialization).   If, as argued earlier (see e.g.~\cite{mingard2021sgd}) the Bayesian posterior probability $P(f|S) = P(f)/P(S)$ for $f \in U({S})$ strongly correlates with the posterior probability that an SGD trained DNN converges on the function $f$, then this inductive bias carries through. 

\subsection{Bound calculations for Boolean classification}\label{app:subsec:our_pac_calc}

For the unbiased learner $\mathcal{U}_{\leq K}$, target function $f_t$ with training set $S$ of size $m$ (see e.g.\ \cref{subfig:main_bv,app:fig:bv_n5,app:fig:bv_n7}),  $P(U(S))=\frac{|\{f:K_{LZ}(f))\leq K, f|_{S}\}|}{|\{f:K_{LZ}(f))\leq K\}|}$. The notation $|_S$ denotes evaluating $f$ on only the training data in the training set $S$ (as opposed to evaluating on all $2^n$ datapoints).
Plugging this into \cref{app:eqn:guillermo_pac_bayes_1} leads to:
\begin{align}\label{app:eqn:pac_bayes}
    \epsilon(Q^*)\leq 1-\exp\left(-\frac{\ln\frac{|\{f:K_{LZ}(f))\leq k\}|}{|\{f:K_{LZ}(f))\leq k, f|_{s=S}\}|}+\ln\frac{2m}{\delta}}{m-1}\right),
\end{align}
the implementation of the PAC-Bayes bound for $\mathcal{U}_{\leq K}$.

If at least one function achieves zero training error for some $U_{\leq K_M}$, we can use the realisable PAC Bound adapted from (\cref{app:eqn:PACC}) (where we don't make the customary approximate step that $(1 - \epsilon)^m \approx e^{-m \epsilon}$ which only holds for small $m \epsilon^2$: 
\begin{align}
\epsilon_G\leq 1-\exp\left(-\frac{\ln{|\mathcal{H}_k|}+\ln\frac{1}{\delta}}{m}\right)\sim 1-\exp\left(-\frac{k\ln{2}+\ln\frac{1}{\delta}}{m}\right).
\end{align}
where the second relation holds if we assume that $|\mathcal{H}_K| \sim 2^k$.  In practice, we take the cumulative measure  $|\mathcal{H}_K| = \int_{K_{min}}^K P(K') dK'$ using our LZ complexity measure to calculate $P(K)$. This explains why the PAC bound flattens off at larger $K$ 

\subsection{Practical estimates for $n=5,7$}
The PAC and PAC-Bayes bounds (see \cref{app:eqn:pac_bayes}) can both be enumerated exactly (as $P_{K_{LZ}}(f)\mid S$ can be calculated exactly for all $f$) for the $n=5$ system, and for the $n=7$ system we use the approximation in \cref{app:fig:P(K)_n7} to calculate $P(K)$, and we can exactly enumerate the numerator in \cref{app:eqn:pac_bayes} by sampling. The PAC-Bayes bounds for the neural networks (for the n=5 system) use $10^8$ samples to approximate $P(f)$, and $P(S)$ is extracted from this distribution. In the event that no function found in $10^8$ samples fits the training set, the bound is plotted with $P(S)=10^{-8}$, to show the minimum value of the PAC-Bayes bound.

\section{Experimental details}

\subsection{Complexity measure}\label{app:complexitymeasures} 

In this paper, we mainly use a variation on the famous Lempel-Ziv (LZ) algorithm introduced in 1976 by Lempel and Ziv~\cite{lempel1976complexity} to calculate the complexity of binary strings. This compression-based measure has been a popular approximator for Kolmogorov complexity. It is thought to work better than other lossless compression-based measures for shorter strings~\cite{amigo2004estimating,lesne2009entropy}.  The essence of the Lempel-Ziv algorithm is to read through a string (of any finite alphabet size) and to create a dictionary of sub-patterns. 
If the number of words (distinct patterns) in the dictionary is $N_w(x)$, then, as in ~\cite{dingle2018input},  we define the following complexity measure  \begin{equation}
K(x) =\begin{cases}
     \log_2(n), &  \hspace*{-0.3cm}  \text{$x=0^n$ or $1^n$}\\
    \log_2(n) [N_w(x_1...x_n) + N_{w}(x_n...x_1)]/2, & \hspace*{-0.2cm} \text{otherwise}
\end{cases}\label{eq:CLZ}
\end{equation}
For further details on the performance of this measure, see the supplementary information of~\cite{dingle2018input}. In particular, it was shown that this measure correlates with other complexity measures for binary strings.   One weakness is that it tends to stretch out the complexity for large complexity strings, compared to theoretical expectations (see e.g.~the right panel in e.g.~\cref{app:fig:P(K)_n7}(b)).  However, we are not typically interested in complexities at the top of the range, so this weakness of the LZ measure is unlikely to be important for our main qualitative results.  In \cref{app:fig:K_measures} we compare both to a more naive measure (entropy) as well as to a more sophisticated (approximate) measure of Boolean complexity.  The main qualitative differences between the normal and the chaotic DNNs remain. 

\subsection{Prior over LZ complexity, P(K)}\label{app:exp_details:P(K)} The prior distribution over complexities $K$ is defined as
\begin{align}
P(K)=\sum_{f:C_{LZ}(f)=K}P(f)=\int_{\theta}d\theta P_{\rm par}(\theta)\mathbbm{1}(C_{LZ}(f)=K)
\end{align}
where $C_{LZ}(f)$ is the LZ complexity of the string representation of $f$. $P(K)$ is approximated by randomly initializing parameters $10^8$ times, recording the frequencies $\rho_f$ of observed $f$, so $P(K) \approx \sum_{f:C_{LZ}(f)=K}\rho_f/10^8$.

\subsection{Obtaining functions with a range of LZ complexity}\label{app:exp_details:gen_funcs_range} 

This algorithm is used in generating targets in e.g.\ \cref{subfig:trained_DNN_bias}
Low-complexity strings are rare.
One might think that one could use  a DNN to achieve produce them since it is biased towards simplicity.  However, there may be other biases in the DNN as well. So instead we use the algorithm below:

\begin{algorithm}[H]
\caption{Generating strings over a range of complexities}\label{alg:random_complexity}
\begin{algorithmic}[0]\label{alg:random_complexity_2}
\STATE \textbf{input:} Length of binary string, $2^n$ (typically $n=7$). $F = []$.
\FOR{m in $0,1,\dots,2^{n-1},2^n$}
    \FOR{\_ in range $10^6$}
        \STATE Generate a string $s$ with the first $m$ characters $1$ and all others 0. Generate a random permutation $S_{2^n}$, and use it to permute $s$.
        \STATE Append the tuple $(s,LZ(s))$ to $F$ if $s\notin F$.
    \ENDFOR
\ENDFOR
\STATE group functions in $F$ by $K$.
\STATE Return $F$
\end{algorithmic}
\end{algorithm}
This method heavily biases the sample towards $f$ with low LZ and low entropy. Experiments in \cref{app:bias_variance} therefore used strings obtained by generating a random string of length $2^l$ (for $l<n$) and repeating the algorithm $2^{n-l}$ times; thus generating a low LZ but high entropy string.

\subsection{generalization error v.s.\ LZ complexity}\label{app:exp_details:eg_vs_LZ} 

This algorithm produces results in e.g.\ \cref{subfig:trained_DNN_bias}
This method is designed to determine the expected generalization error for DNNs with $L$ layers, tanh activations, and initialization scheme $(\sigma_w,\sigma_b)$, for some training set size $m$ and for a range of target function complexities.

\begin{algorithm}[H]\caption{generalization error v.s.\ LZ Complexity}\label{alg:eg_vs_lz}
\begin{algorithmic}
\STATE \textbf{input:} DNN $N$ with $L$ layers tanh activations, initialization variances $\sigma_w$ ($\sigma_b=0$), optimiser $O$, batch size $b$. Dataset $\mathcal{B}(\cdot)_7$, training data size $0<m<2^7$ (typically $m=64$). Set $F$ of functions generated by \cref{alg:random_complexity}. Stochastic optimiser $O$, Loss function $L$.
\STATE Let $F_{out}={}$
\FOR{unique $K$ in $F$}
    \STATE Sample one function $f_K$
    \STATE $C_K=[]$
    \FOR{\_ in range(1000)}
        \STATE Randomly sample $m$ integers without replacement from $0,\dots,2^7$. These are the indices of $f_K$ used as training data $S$. The other part of $B(\cdot)_7$ is the test set $E$
        \STATE Train DNN with optimiser $O$ with loss function $L$ on $S$ until 100\% training accuracy is achieved. Record the error on $E$, $\epsilon$
        \STATE Append $\epsilon$ to $C_K$
    \ENDFOR
    \STATE Calculate mean $\mu_K$ and std $\rho_K$ of $C_K$. Append $(K,\mu_K,\rho_K)$ to $F_{out}$
\ENDFOR
\STATE Return $F_{out}$.
\end{algorithmic}
\end{algorithm}

For experiments with CE loss, the AdvSGD optimiser was used (see Appendix A in \citep{valle2018deep} for details) with batch size 16. For experiments with MSE loss, the Adam optimiser was used, with a learning rate of $10^{-5}$ and batch size $16$.

\subsection{generalization v.s.\ output function LZ complexity}\label{app:exp_details:eg_vs_LZ_single_func} 

These experiments aim to capture the distribution of functions found when training a DNN to some target $f$ with training set size $m$. See \cref{subfig:TF31} for an example.

\begin{algorithm}[H]\caption{$\epsilon(f_N)$ v.s.\ $C_{LZ}(f_N)$ for target function $f$}\label{alg:eg_vs_lz_scatter}
\begin{algorithmic}
\STATE \textbf{input:} DNN $N$ with $L$ layers Tanh activations, initialization variances $\sigma_w$ ($\sigma_b=0$), optimiser $O$, batch size $b$. Dataset $\mathcal{B}(\cdot)_7$, training data size $0<m<2^7$ (typically $m=64$). Set $F$ of functions generated by \cref{alg:random_complexity}. Stochastic optimiser $O$, Loss function $L$. Target complexity $K$.
\STATE Let $F_{out}=\{\}$
\STATE Sample one function $f_K$ from $F$
\FOR{\_ in range(1000)}
    \STATE Randomly sample $m$ integers without replacement from $0,\dots,2^7$. These are the indices of $f_K$ used as training data $S$. The other part of $B(\cdot)_7$ is the test set $E$
    \STATE Train DNN with optimiser $O$ with loss function $L$ on $S$ until 100\% training accuracy is achieved. Record the error on $E$, $\epsilon$, and $C_{LZ}(f_N(\theta)$.
    \STATE Append $(K,\epsilon,C_{LZ}(f_N(\theta))$ to $F_{out}$
\ENDFOR
\STATE Return $F_{out}$.
\end{algorithmic}
\end{algorithm}

\subsection{Likelihoods and  posteriors averaged over training sets}\label{app:exp_details:approx_bayes}

\subsubsection{Basic Bayes theorem for classification in terms of functions} 

For supervised learning, the posterior probability $P(f|S)$ is the conditional probability that the learning agent obtains the  function $f$ (or equivalently a hypothesis $h$ if you prefer),  conditioned on the training set data ${S}$ taken i.i.d.\ from a data distribution $\mathcal{D}$, can be calculated by Bayes' rule:
 \begin{equation} 
 P(S|f)  = \frac{ P(S |f)P(f)}{P(S)},
 \end{equation} 
in terms of the prior probability $P(f)$ of obtaining $f$ without training,  the  likelihood $P(S|f)$, and the marginal likelihood $P(S)$.
The simplest case is perhaps  where we condition on zero training error.  Then the likelihood is the conditional probability that the learning agent correctly produces all the outputs  in $S$, given that it is producing function $f$. This can be expressed in a  simple mathematical form:
\begin{equation}
P(S|f) =   \begin{cases} 1  \textrm{ if } \epsilon(f|S) =0\\
                            0\textrm{ otherwise }.
            \end{cases}
\end{equation} 
where $ \epsilon(f|S)$ is the error that function $f$ makes on the training set (to be defined more carefully later).
The posterior $P(f|S)$  is then given by
\begin{equation}\label{eq:PB1a}
P(f|S) =      \begin{cases}  \frac{P(f)}{P(S)}  \textrm{ if } \epsilon(f|S) =0\\
                            0\textrm{ otherwise }.
            \end{cases}
\end{equation} 
and  is directly proportional to the prior for functions that have zero error on $S$, which we will define as the set $U(S)$.
Similarly, the marginal likelihood takes a simple form  for discrete functions:
\begin{equation}
P(S) = \sum_f P(S|f) P(f)=\sum_{f\in U(S)} P(f), 
\end{equation} 
 The marginal likelihood  gives the (prior) probability that  upon random sampling of the parameters of the learning system, it correctly produces the labels in $S$, which is equivalent to saying that probability that it finds a function with zero error on $S$. One interpretation of this quantity  is that it is a measure of the inductive bias of the learning agent to produce $S$. Large $P(S)$ means there is a good inductive bias towards the data $S$, and vice versa.  The hope of course is that a large $P(S)$ also corresponds to a good inductive bias towards the full data $\mathcal{D}$. 

\subsubsection{Averaging over training sets} 

Further insight comes from  averaging over all training sets of size $m$, chosen i.i.d.\ from the data distribution $\mathcal{D}$. This averaging can also be interpreted as the expectation over $\mathcal{D}$ for a training set of $m$ instances.   The expected value of the likelihood is:
\begin{equation}
\langle P(f|S) \rangle_m = \left(1 - \epsilon(f)\right)^m
\end{equation} 
where $\epsilon(f)$ is the expected error (fraction of incorrect labels) of function $f$ on data in $\mathcal{D}$.  Similarly, 
the averaged marginal likelihood is then given by
\begin{equation}\label{eq:PD}
\langle P(S) \rangle_m =  \frac{1}{N_S(m)} \sum_{i=1}^{N_S(m)} \sum_{\forall f}  P(f) \langle P(f|S_i) \rangle_m  = \sum_{\forall f}  P(f) \left(1 - \epsilon(f)\right)^m \approx  \sum_{\forall f}  P(f) e^{ -m \epsilon(f)}
\end{equation}
where $N_S(m)$ is the number of training sets of size $m$.  The last step is a good approximation if $\epsilon(f)$ is small (more precisely, if  $m (\epsilon(f))^2  \ll 1$,  since $e^{-m \epsilon}/(1-\epsilon)^m = 1 + \frac12 m \epsilon^2 + \frac13 m \epsilon^3 + \ldots $. So how good the approximation is, depends in part on how large $m$ is). Note that $e^{-m \epsilon(f)} > (1-\epsilon(f))^m$, so the last approximation above is really an upper bound.    This averaged marginal likelihood has the simple interpretation of the (prior)  probability that for any training set of size $m$, chosen i.i.d. from $\mathcal{D}$, the learning agent obtains a function with zero training error upon random sampling of parameters.     One interpretation of $\langle P(S) \rangle_m$ is that it measures the match between the \textit{expected} inductive bias of the learning agent, and the structure of the data.

The expectation of the posterior probability upon averaging over training sets of size $m$ follows in the same vein:
\begin{equation}\label{eq:PBnew}
\langle P(f|S) \rangle_m =  P(f) \left \langle \frac{  P(S|f) }{P(S) }   \right \rangle_m \approx \frac{ P(f) \left(1 - \epsilon(f)\right)^m}{\langle P(S) \rangle_m} = t_F(f,m) P(f) \approx \frac{P(f) e^{ -m \epsilon(f)}}{\langle P(S) \rangle_m} 
\end{equation}
where the first approximate step  (average of the ratio is the ratio of the averages) should be good if the distribution over $P(S)$ is highly concentrated, which we expect it to be for larger $m$.  The second approximation needs $m (\epsilon(f))^2 \ll 1$, but it is always an upper bound.   We also define an average training factor, which can be rewritten as follows:
\begin{equation}\label{eq:TF}
t_F(f,m) = \frac{(1-\epsilon(f))^m}{\langle P(S) \rangle} = \frac{(1-\epsilon_G(m))^m}{\langle P(S) \rangle} \left(1-\frac{\epsilon(f)-\epsilon_G(m)}{1 - \epsilon_G(m)}\right)^m \approx  \frac{e^{ -m \epsilon(f)}}{\langle P(S) \rangle} =  \frac{e^{ -m \epsilon_G}}{\langle P(S) \rangle} e^{-m \left(\epsilon(f) - \epsilon_G\right)}
\end{equation}
where 
\begin{equation}
\epsilon_G (m) = \sum_f   \left< P(f|S)  \right>_m\epsilon(f) \approx  \frac{\sum_f P(f) (1-\epsilon(f))^m \epsilon(f)}{\langle P(S)\rangle_m} \approx  \frac{\sum_f P(f) e^{-m \epsilon(f)} \epsilon(f)}{\langle P(S)\rangle_m}
\end{equation} 
is the mean expected generalization error on test sets given that the system is trained on an i.i.d.\ chosen training set $S \in \mathcal{D}$ of size $m$. 
The training factor captures how training affects the value of the posterior beyond just the prior. The contribution of functions with $\epsilon(f) < \epsilon_G(m)$ to the posterior is exponentially enhanced by the training factor, and by contrast, functions with $\epsilon(f) > \epsilon_G(m)$ are exponentially suppressed by the training factor.  As $m$ increases we expect $\epsilon_G(m)$ to decrease. In other words, with increasing training set size m, fewer functions get enhanced and more functions get exponentially suppressed, and those that were already suppressed at lower $m$ get suppressed by even more as $m$ increases. This is the essence of the Bayesian approach to training.

For Boolean functions specifically, the number of training sets of size $m$ is given by 
\begin{equation}
 N_S(m) = {2^n \choose m}   
\end{equation}
 and the averaged posterior from \cref{eq:PB} becomes
\begin{align}
    \langle P(f| S) \rangle_{m} &\approx \frac{1}{\langle P(S)\rangle_{m} N_S(m)}\sum_{i=1}^{N_S(m)}P(S| f)P(f)\\
    &= \frac{1}{\langle P(S)\rangle_{m} }{2^n \choose m}^{-1}{2^n(1-\epsilon(f)) \choose m}P(f)
\end{align}
because there are  ${2^n(1-\epsilon(f)) \choose m}$ possible training sets per error, and ${2^n(1-\epsilon_g(f)) \choose m}=0$ if $m>2^n(1-\epsilon_g(f))$. 
Note that ${2^n \choose m}^{-1}{2^n(1-\epsilon_g(f)) \choose m}$ is closely approximated by $(1-\epsilon_f(g))^m$ for small $\epsilon_f(g)$ and large $m$.

In the main text, we also make a decoupling approximation for the averaged posterior over complexity: 
\begin{equation} \label{eq:PBapproxapp} 
\langle P(K'|S) \rangle_m = \hspace*{-0.5cm} 
  \sum_{K_{LZ}(f)=K'} \langle P(f|S) \rangle_S
\propto P(K') \langle \left(1-\epsilon(K')\right)^m \rangle_{l},
\end{equation}
where the decoupled likelihood is averaged over the $l$ largest error functions with complexity $K$.  These functions are found by exhaustively searching through all functions up to 5 bits different from the target function, and then sampling for larger differences.  While this approximation is rather crude, it works remarkably well across model complexity and for different training set sizes $m$, and is also quite robust to changes in the number of functions $l$ included -- this can all  be seen in \cref{fig:Cube_plots_2} in the main text, and in Figures S9-S12 in the Appendix. 

Finally, we note that   posterior distribution with 0-1 likelihood
is not the exact posterior distribution that DNNs may approximate. Despite halting training at 0 training error, we find that the DNNs often also arrive at very low training loss, which would be close to this loss though.  Similarly, Gaussian Processes with MSE use a mean-square error likelihood $P(S\mid f)=\exp\left(-\sum_{i\in S}(f(x_i;\theta)-g(x_i))^2/2\alpha^2\right)$, where $\alpha^2$ is typically very small. 

\subsection{The Gaussian Processes/Linearised Approximation}\label{app:exp_details:GP}

This approximation is used in \cref{fig:app:gp_nn_2}.
Instead of directly calculating the Gaussain Process (GP), we  approximate it using a very wide neural network and train only the last layer with MSE loss until low loss for a linear model closely approximating a GP with mean square error likelihood -- see \citep{mingard2021sgd,matthews2017sample} for an explanation of GPs with MSE likelihood and how they are equivalent to these linear models respectively. For the $n=7$ Boolean systems, 
we use DNNs with width 16384, freeze all but the final classification layer, and train this layer with Adam (without weight decay) and MSE loss until the loss reaches $10^{-5}$.   We tested that results are not sensitive to changes in the final loss, or to the width.

\subsection{Experimental details for MNIST and CIFAR10 datasets \& CSR complexity measure }\label{app:exp_details:image_dataset}

We also use MNIST dataset\footnote{\url{http://yann.lecun.com/exdb/mnist/}} and CIFAR10 dataset\footnote{\url{https://www.cs.toronto.edu/~kriz/cifar.html}}.  For both MNIST and CIFAR10, we use DNNs with hidden layer width 200, input width 784 and 3072 respectively and output dimension 10. Because functions on the datasets do not have an easily calculable intrinsic measure (not dependent on the neural network) of complexity, we use the critical sample ratio (CSR) as a measure of the complexity of a function. CSR was introduced in \citep{krueger2017deep}. It is defined with respect to a sample of inputs as the fraction of those samples which are critical samples, defined to be an input such that there is another input within a box of side $2r$ centred around the input, producing a different output (for discrete outputs).
Following~\citep{valle2018deep}, we use CSR as a rough estimate of the complexity of functions found in the posterior (conditioning on $S$), and the prior (i.e.\ functions on $S$). In our experiments, we used binarised MNIST with a training set size of $10000$ and a test set of size $100$ (analogously to the majority of our other experiments).

\section{Extended figures of priors $\bf P(K)$ and $\bf P(f)$ for DNNs}\label{app:priors}

\begin{figure}[H]
    \centering
    
    \begin{subfigure}[b]{0.99\columnwidth}
        \includegraphics[width = \textwidth]{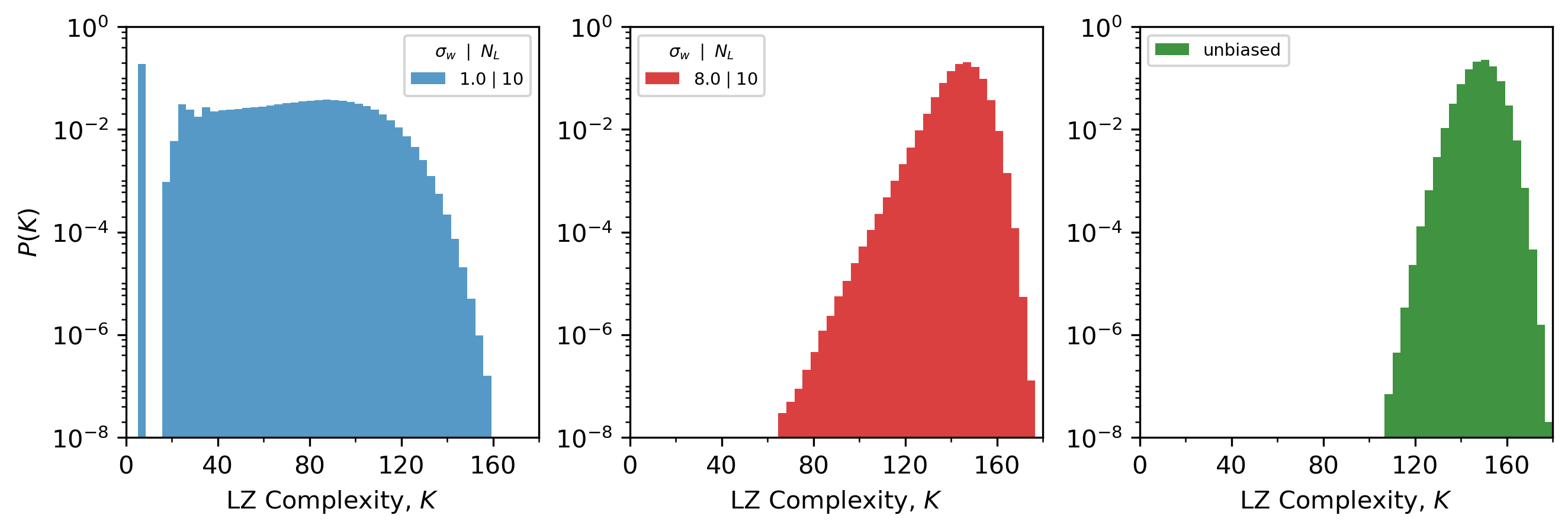}
        \caption{P(K) distributions for LZ complexity (raw data only)}
    \end{subfigure}
    
    \begin{subfigure}[b]{0.99\columnwidth}
        \includegraphics[width = \textwidth]{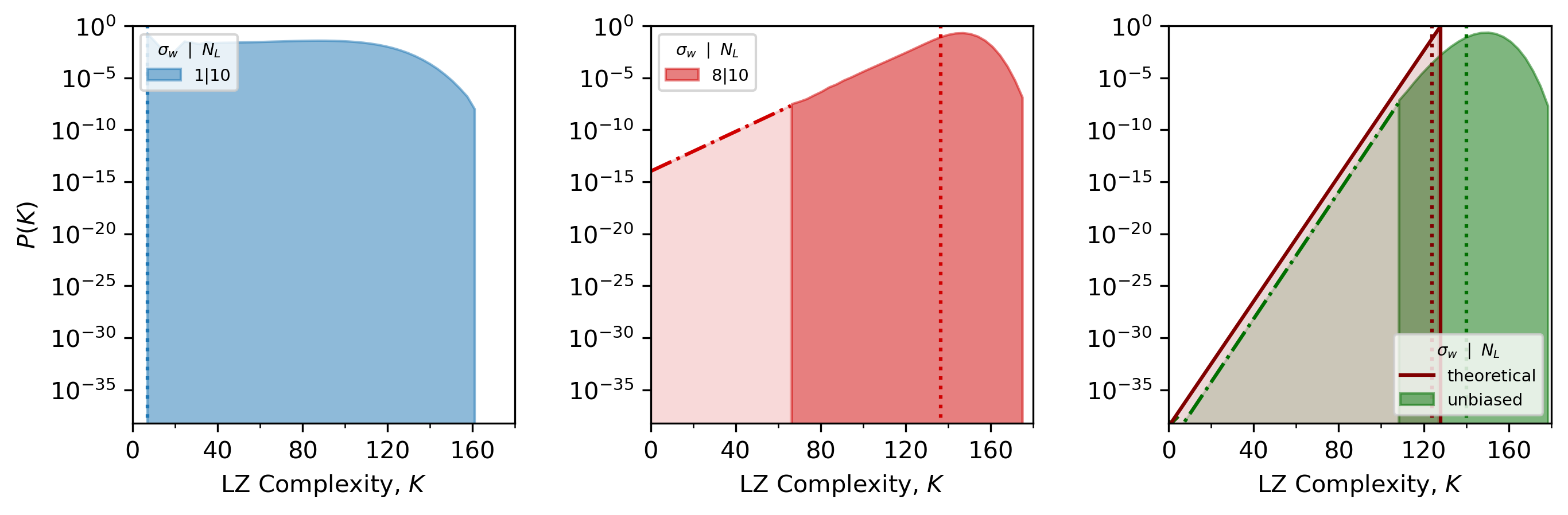}
        \caption{P(K) distributions for LZ complexity extended data}
    \end{subfigure}
    
    \caption{\textbf{$\bf P(K)$ v.s.\ LZ complexity $K$ for the $n=7$ Boolean system.}  Left and centre panes show DNNs with 10 layers,  $\sigma_b=0.2\sigma_w$  and $\sigma_w=1,8$ respectively, with tanh activations and hidden layer width 40. The right-hand pane shows $P(K)$ when strings of length 128 are uniformly sampled, for $10^8$ random samples. 
    (a) Raw values for LZ complexity (without extrapolation) and (b) with log-linear extrapolation (inspired by plots for other lengths in~\cite{dingle2018input})  here the $y$-axis is cut off at $2\times2^{-128}$ -- the exact value of $P(K)$ for the simplest functions when randomly sampling. The blue, red and green areas show the same data as found in \cref{subfig:order_bar_plot,subfig:P(K|S)_theoretical} respectively. 90\% of the probability mass lies to the right of the vertical dotted lines; the dash-dot lines are lines of best fit for the left-hand two panes; for the right-hand most pane, one point is fixed at $(7,2\times2^{-128})$ (the theoretical value for $P(K=7)=2\times2^{-128}$, as $K=7$ corresponds to the strings of all 0s and 1s only), and connected to the end of the sampled area. Max observed LZ complexity is  $K=179$ (in the right-hand plot), x-axis cut at $K=180$.  The red theoretical curve in the right plot is for an ideal compressor for which $P(K) = 2^K$. See \cref{app:exp_details:P(K)} for further experimental details
    }\label{app:fig:P(K)_n7}
\end{figure}

In this section, we plot in more detail than the main text some results for the priors of DNNs upon random sampling of parameters.  In \cref{app:fig:P(K)_n7} we show priors over complexity $P(K)$ for an ordered and chaotic FCN with 10 layers, as well as the $P(K)$ for randomly sampled strings of length $128$.

\begin{figure}[H]
    \centering
    \begin{subfigure}[ht]{0.22\linewidth}
        \includegraphics[width = \textwidth]{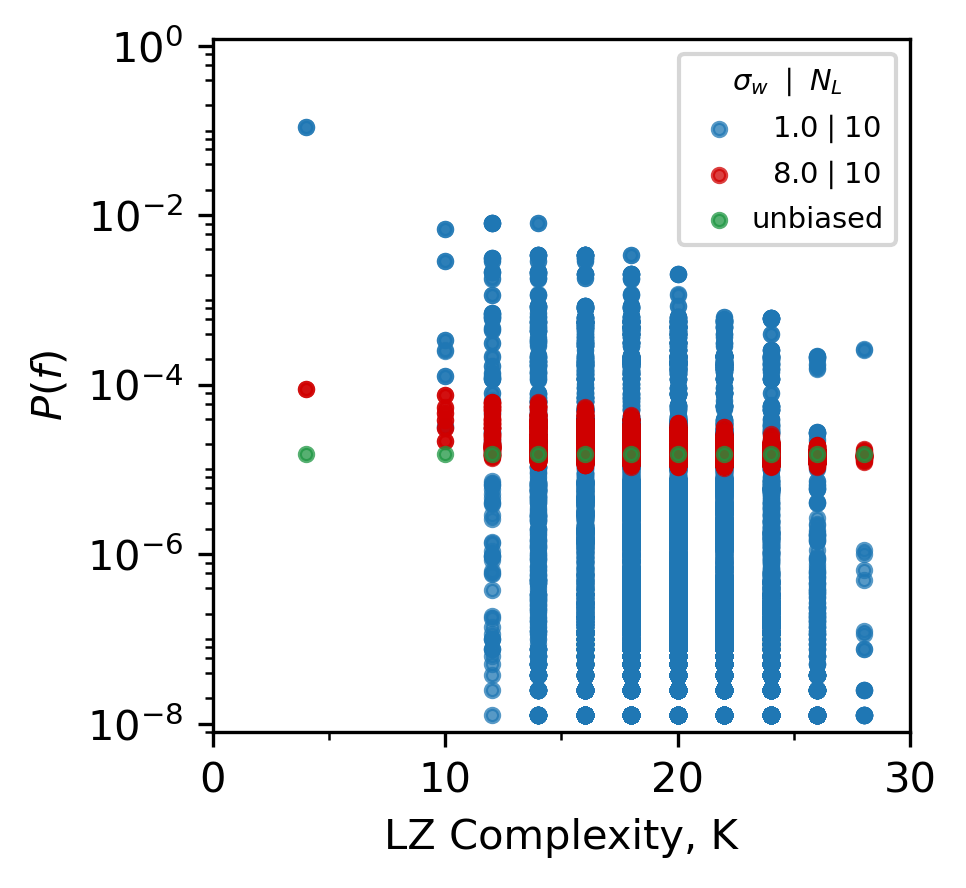}\caption{$P(f)$ v.s.\ $K_{LZ}(f)$, n=4 }\label{app:fig:n=5prior:n4}
    \end{subfigure}
    \begin{subfigure}[ht]{0.22\linewidth}
        \includegraphics[width = \textwidth]{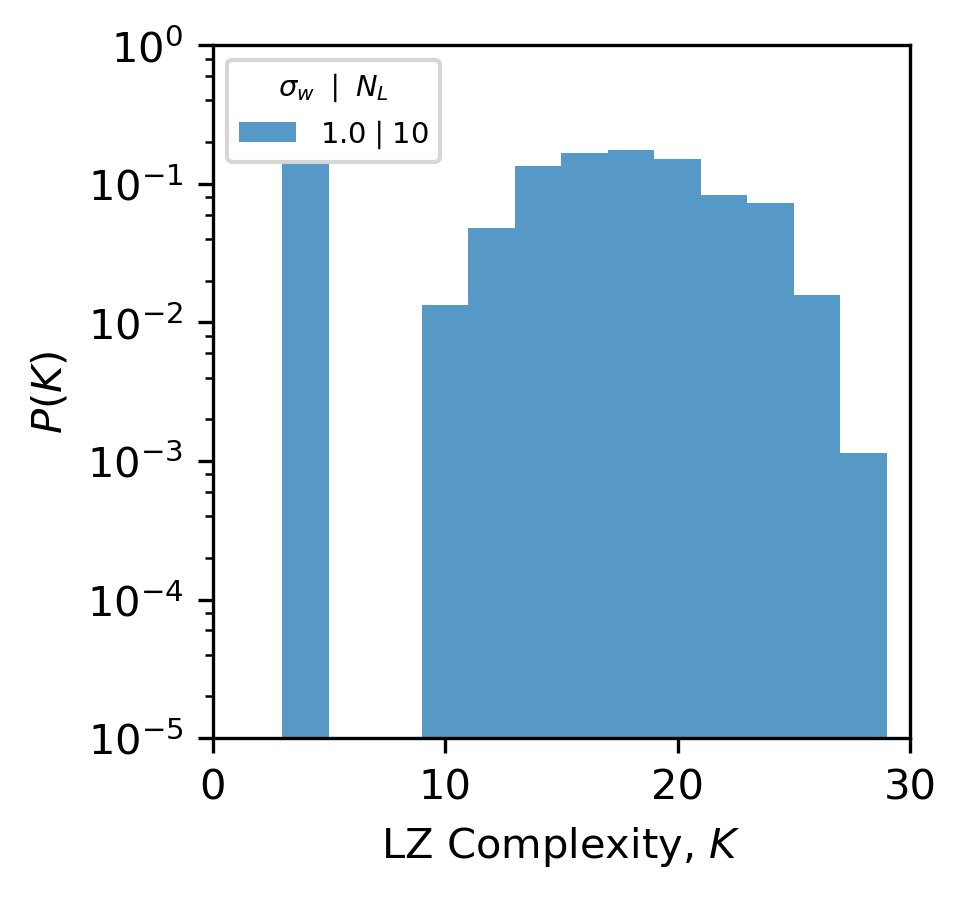}\caption{n=4, $\sigma_w=1$}
    \end{subfigure}
    \begin{subfigure}[ht]{0.22\linewidth}
        \includegraphics[width = \textwidth]{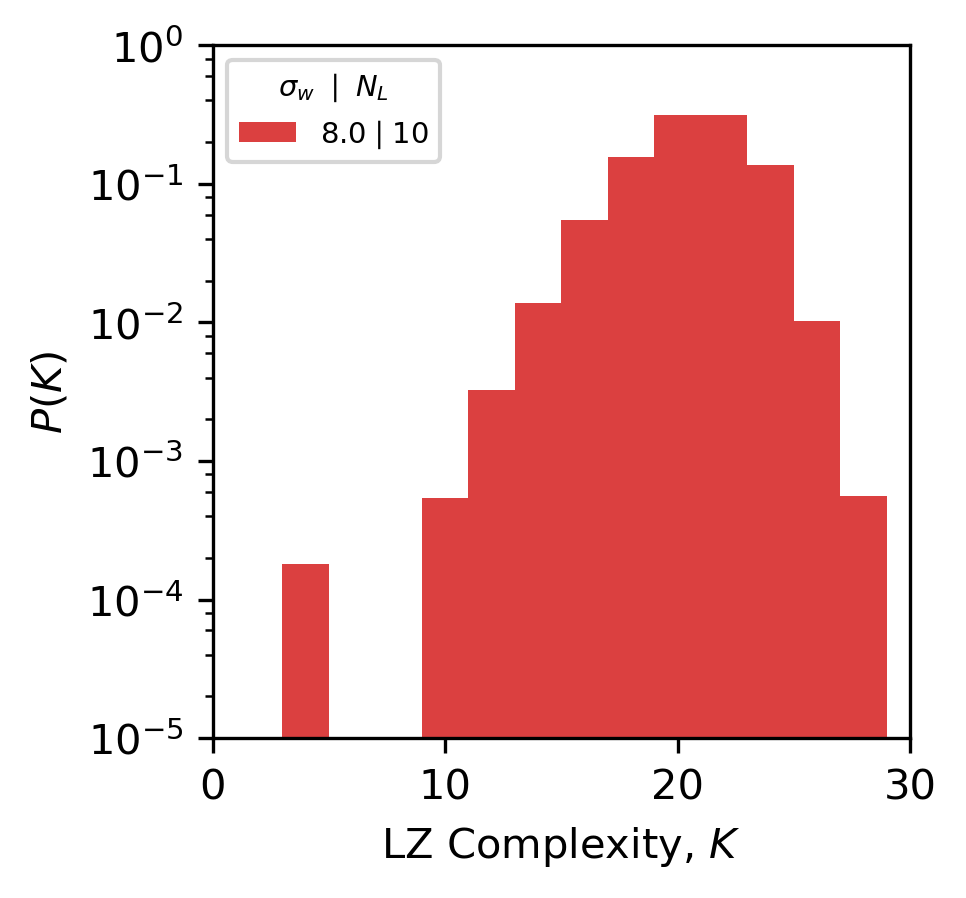}\caption{n=4, $\sigma_w=8$}
    \end{subfigure}
    \begin{subfigure}[ht]{0.22\linewidth}
        \includegraphics[width = \textwidth]{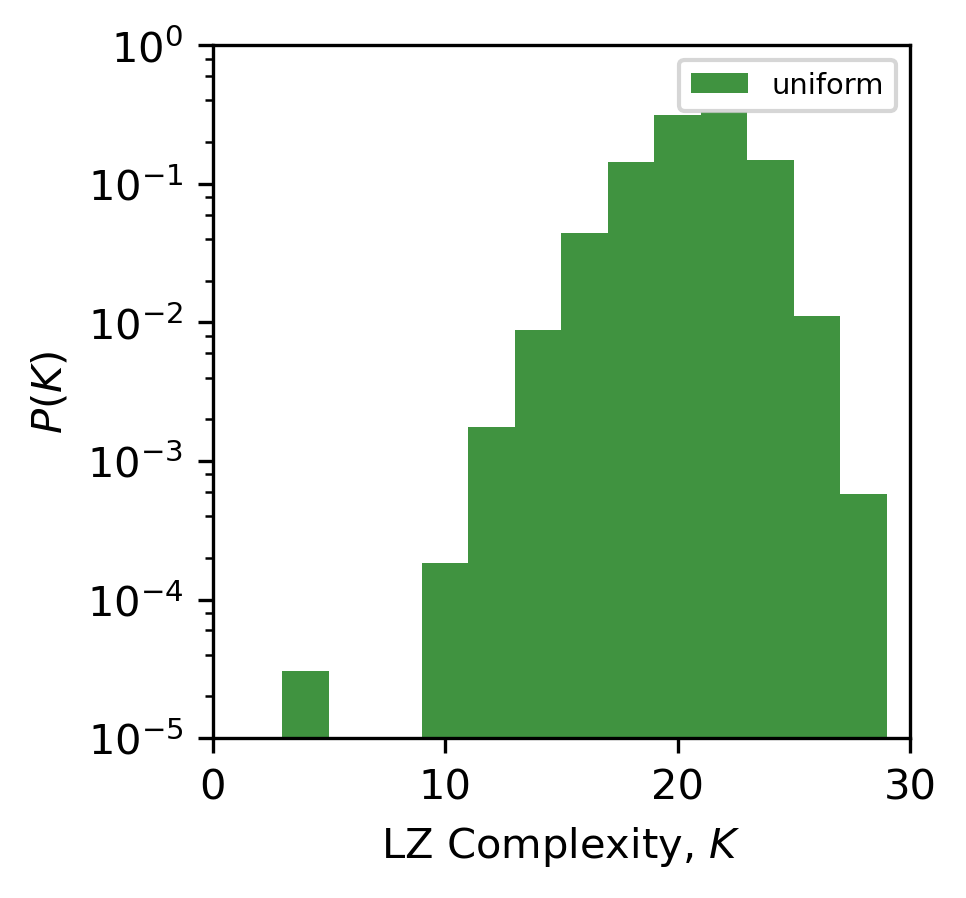}\caption{n=4, uniform (exact)}
    \end{subfigure}
    
    \begin{subfigure}[ht]{0.22\linewidth}
        \includegraphics[width = \textwidth]{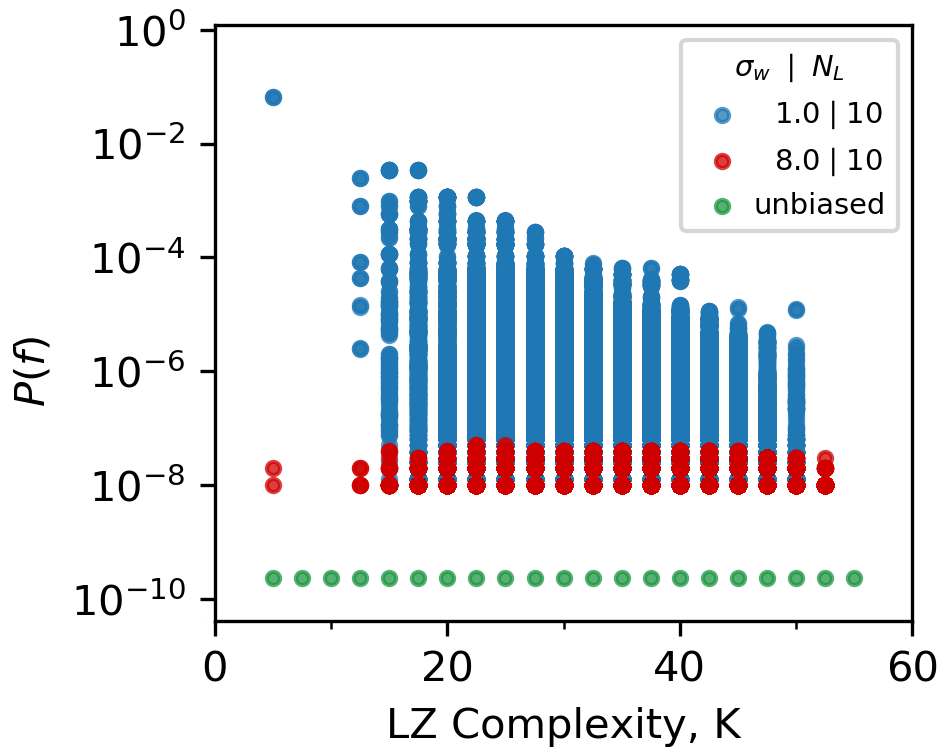}
        \caption{$P(f)$ v.s.\ $K_{LZ}(f)$, n=5}\label{app:fig:n=5prior:n5}
    \end{subfigure}
    \begin{subfigure}[ht]{0.22\linewidth}
        \includegraphics[width = \textwidth]{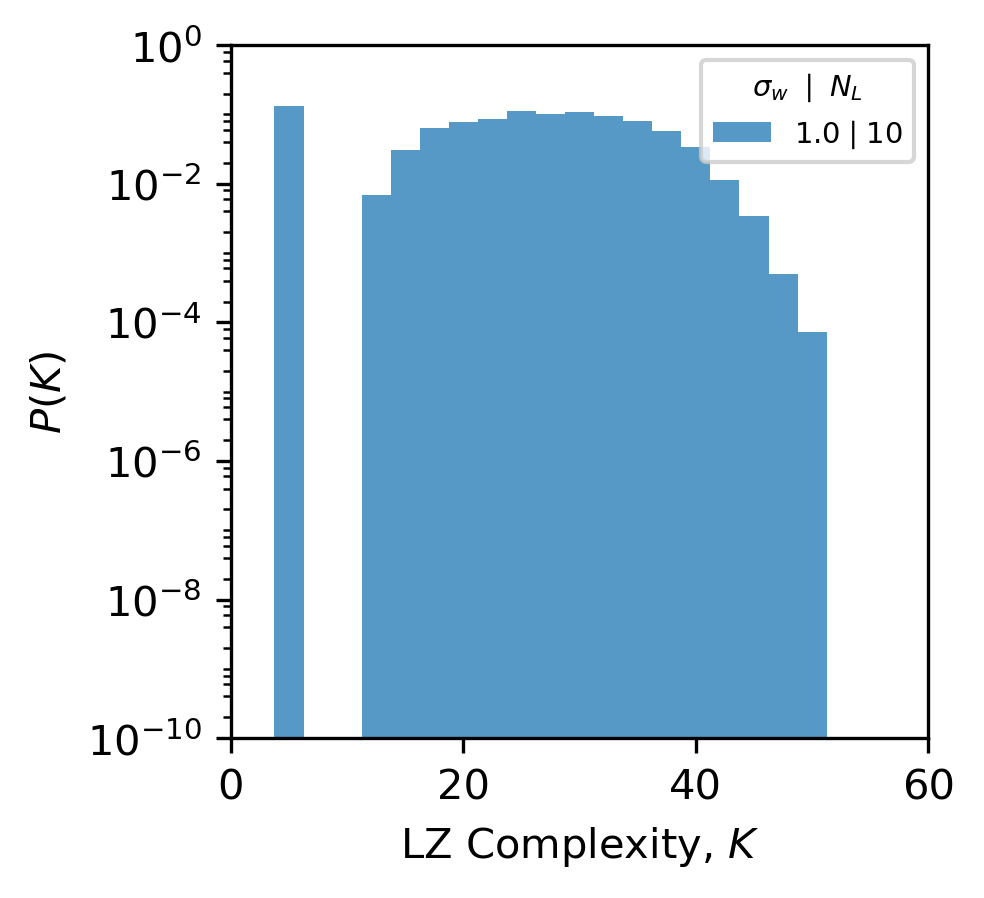}\caption{n=5, $\sigma_w=1$}
    \end{subfigure}
    \begin{subfigure}[ht]{0.22\linewidth}
        \includegraphics[width = \textwidth]{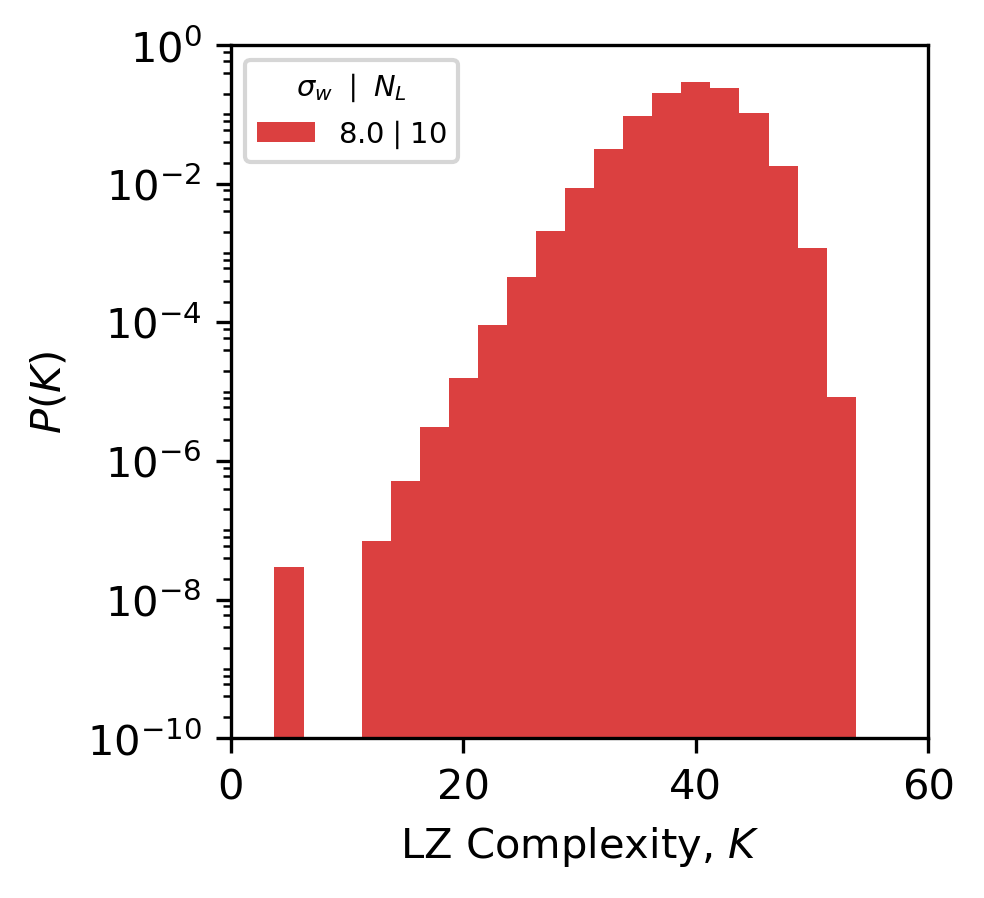}\caption{n=5, $\sigma_w=8$}
    \end{subfigure}
    \begin{subfigure}[ht]{0.22\linewidth}
        \includegraphics[width = \textwidth]{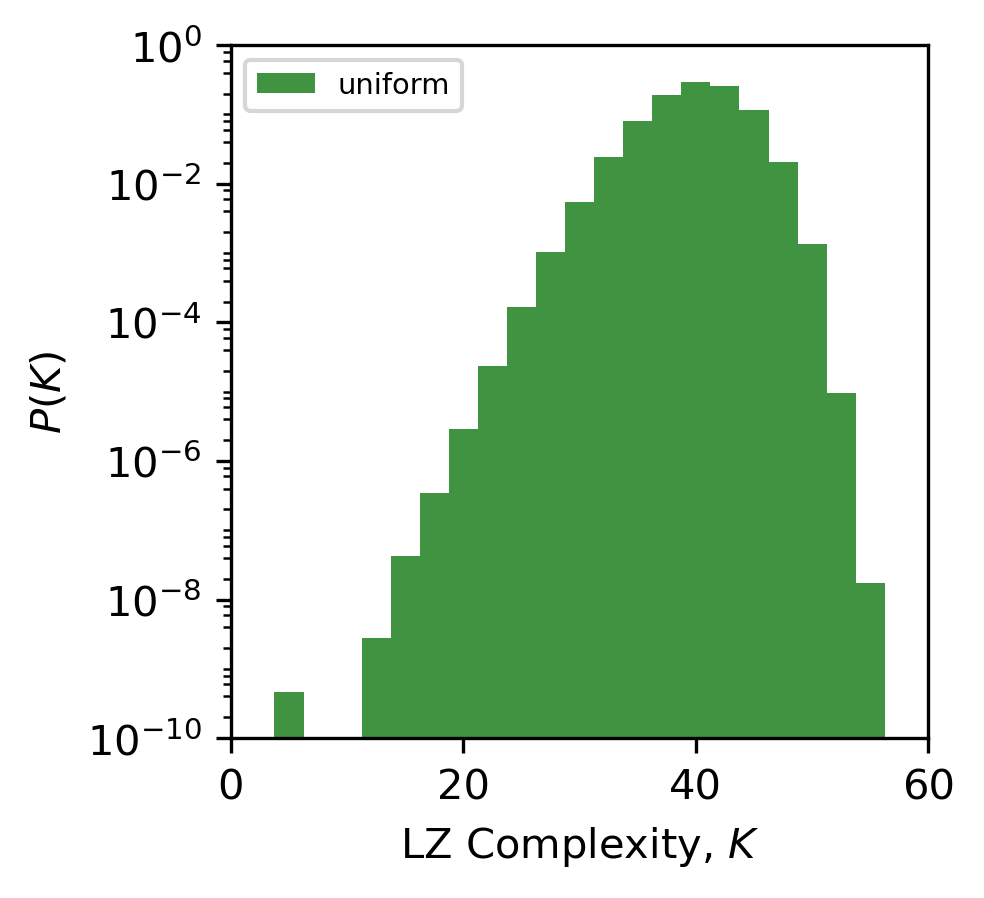}\caption{n=5 uniform (exact)}
    \end{subfigure}
    
    \caption{{\bf Priors upon random sampling of parameters v.s.\ LZ complexity $\bf K$ for the $\bf n=5$ and $\bf n=4$  Boolean systems.} Prior $P(f)$ on $\{0,1\}^5$ for 10 layered DNNs with $\sigma_w=1,8$ from $10^8$ samples. The green points show the theoretical $P(f)$ with a uniform prior over functions. The $n=4$ system is so small that $P(f)$ for the uniform learner is $2^{-16}$, and we can take enough samples to show the functions with $P(f)$ much lower than this for the ordered prior $\sigma_w=1$, for which generalization would be worse than random, as consistent with no free lunch theorems. The $n=5$ system is fractionally too large to see these functions, but we are able to get enough samples to see the trivial functions in the $\sigma_w=8$ prior, and clearly, $P(K)$ grows exponentially in $\sigma_w=8$, suggesting the approximation in \cref{app:fig:P(K)_n7} is sensible.
    }\label{app:fig:n=5prior}
\end{figure}

In \cref{app:fig:n=5prior} we show similar priors (and also $P(f)$) for smaller systems with $n=5$ and $n=4$, where the number of functions are $2^{2^5}=4,294,967,296$ and $2^{2^4}=65,536$ respectively, both of which can be fully sampled.   The downside of these very simple systems is that the finite-size issues with the LZ approximator (see \cref{app:complexitymeasures})  become more pronounced since the strings are only length $32$ and $16$, instead of $128$ as in the $n=7$ system.  Nevertheless, it is clear that the same qualitative differences between the two DNNs are observed in the $n=7$ system.

\begin{figure}[H]
    \centering
    \begin{subfigure}[ht]{0.35\linewidth}
        \includegraphics[width = \textwidth]{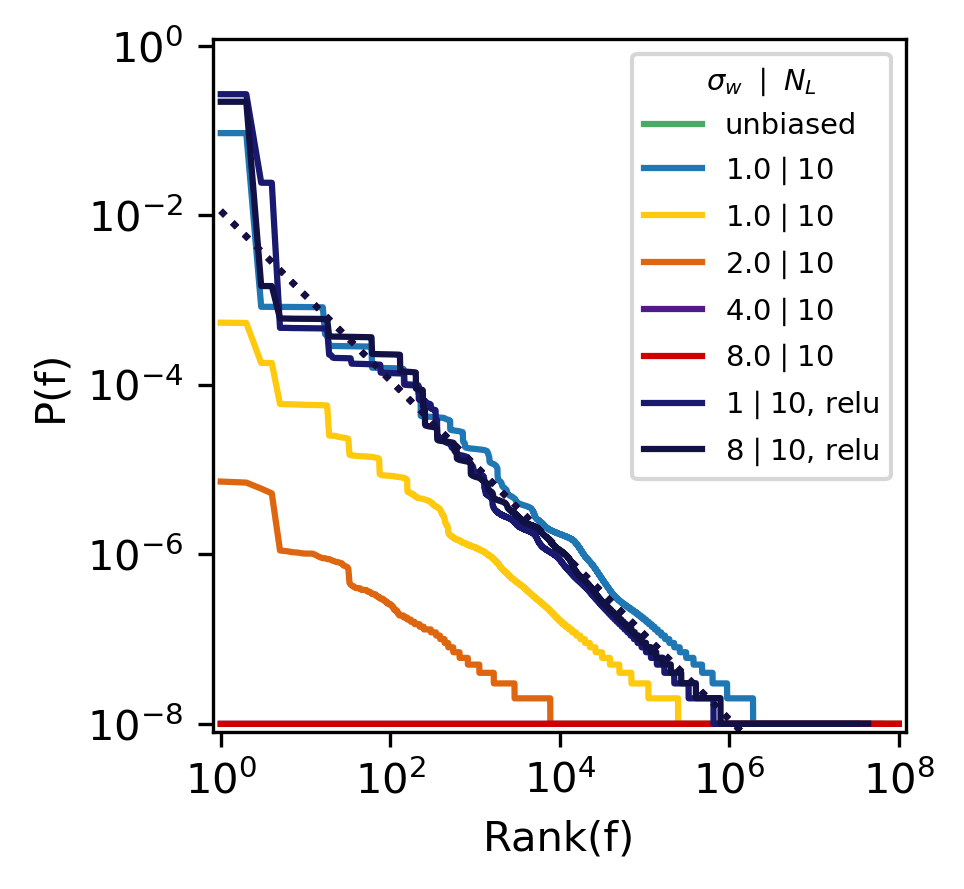}
        \caption{P(f) v.s.\ LZ Complexity}
    \end{subfigure}
    \begin{subfigure}[ht]{0.35\linewidth}
        \includegraphics[width = \textwidth]{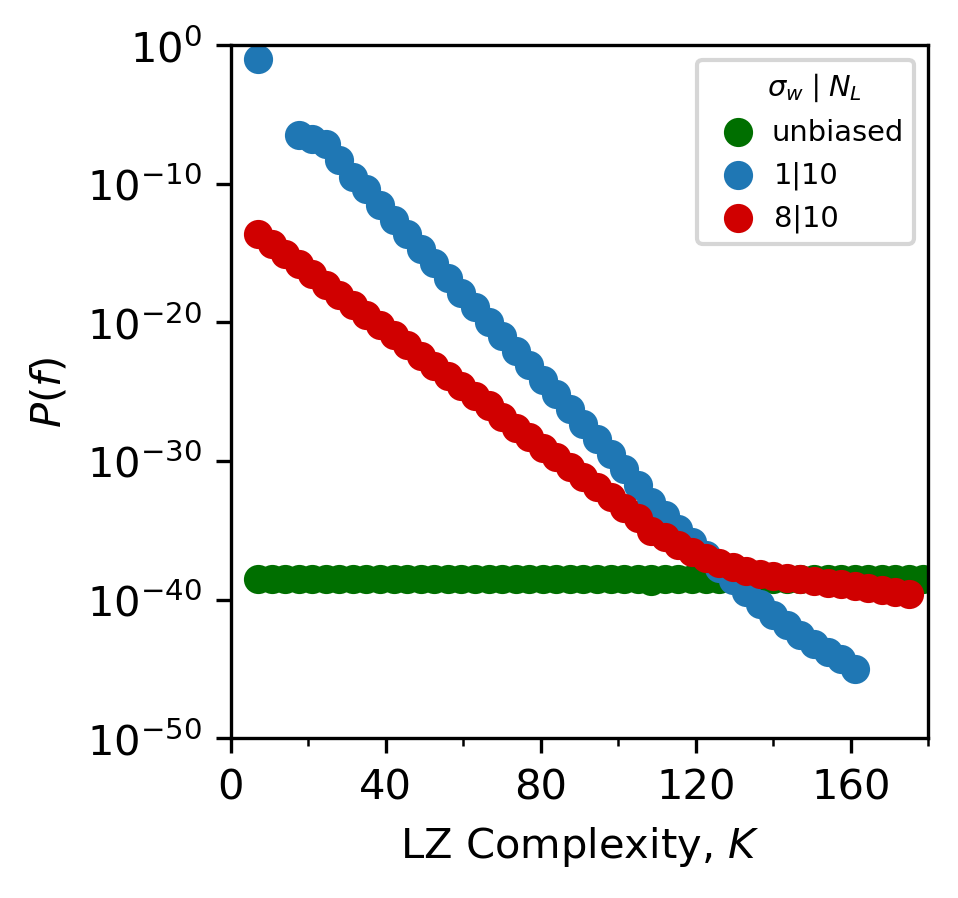}\caption{P(f), n=7, (approx)}
    \end{subfigure}
    \caption{\small {\bf Comparing priors over functions $\bf P(f)$} (a) Prior $P(f)$ for 10-layer FCNs with tanh and ReLU activations, on $n=7$ Boolean functions, ranked by probability of individual functions, generated from  $10^8$ random samples of parameters $\Theta\sim \mathcal{N}(0,\sigma_w)$.  (b) To get a sense of what  $P(f)$ v.s.\ LZ complexity $K$ looks like on a much larger scale,   $P(K)$ data from \cref{app:fig:P(K)_n7} was used to approximate $\langle P(f\mid K)\rangle \approx P(K)/|\{f:C_{LZ}(f)=K\}|$. Note that these points lie below many of the samples in \cref{subfig:LZ_order_chaos}; this is to be expected as the calculation returns the average value for $P(f\mid K)$, which is clearly much lower than $\max P(f\mid K)$ (see e.g.\ \cref{app:fig:n=5prior:n4}).
    Both the chaotic and the normal DNN are strongly simplicity biased. Nevertheless, as shown previously, the lower simplicity bias of the chaotic prior is not strong enough to overcome the growth of the number of functions with increasing complexity, so the prior over complexity is dominated by large-complexity functions for the chaotic regime. 
    }\label{app:fig:sw8relu}
\end{figure}

In \cref{app:fig:sw8relu} (a) we compare the priors $P(f)$ for DNNs with the more popular ReLU activations functions  to that of the tanh-based priors with different $\sigma_w$'s.  Our "standard" $\sigma_w=1$ prior is very close to the ReLU prior (this also holds for $P(K)$).  Moreover, both priors closely follow the idealised Zipfian curve
\begin{equation}
P(f)_{Zipf} = \frac{1}{\ln(N_f) ({\rm rank}(f))^\alpha}
\end{equation}
where the number of functions $N_f=2^{128}$ and with exponent $\alpha=1$.  
For larger $\sigma_w$, as the system enters the chaotic regime the prior begins to noticeably deviate from the ideal Zipfian curve, which is not surprising given that the fully chaotic fixed-point has a uniform prior over functions as shown in~\cref{app:sec:infinite}.

To get a sense of the full range probabilities  of $P(f)$ v.s.\ $K(f)$ we make a simple 
approximation $P(f)\approx P(K)/|\{f:K_{LZ}(f)=K\}|$, where the numerator comes from the measured $P(K)$ from the first two panels of \cref{app:fig:P(K)_n7}(b), and the denominator comes from the third panel in \cref{app:fig:P(K)_n7}(b).  
As shown \cref{app:fig:sw8relu}(b), this produces an approximate $P(f)$ versus $K$ over a wide range of values.  While this  approximation is rather crude, it does allow us to get a sense of the scale of $P(f)$ over the full range of K. 

This larger scale helps us see that the chaotic prior with $\sigma_w=8$ is still quite strongly simplicity biased compared to uniform sampling over functions.   Nevertheless, as can be seen in \cref{app:fig:P(K)_n7}, this simplicity bias is not nearly enough to counteract the $2^K$ growth in the number of functions with increasing $K$.  In other words, an Occam's razor-like simplicity bias is not enough for good generalization on simpler functions.  It has to be the right strength or else  the DNNs will still suffer from a tendency to converge on functions that are too complex, much as one expects from classical learning theory.

\begin{figure}[H]
    \centering
    \begin{subfigure}[b]{0.48\columnwidth}
        \includegraphics[width = \textwidth]{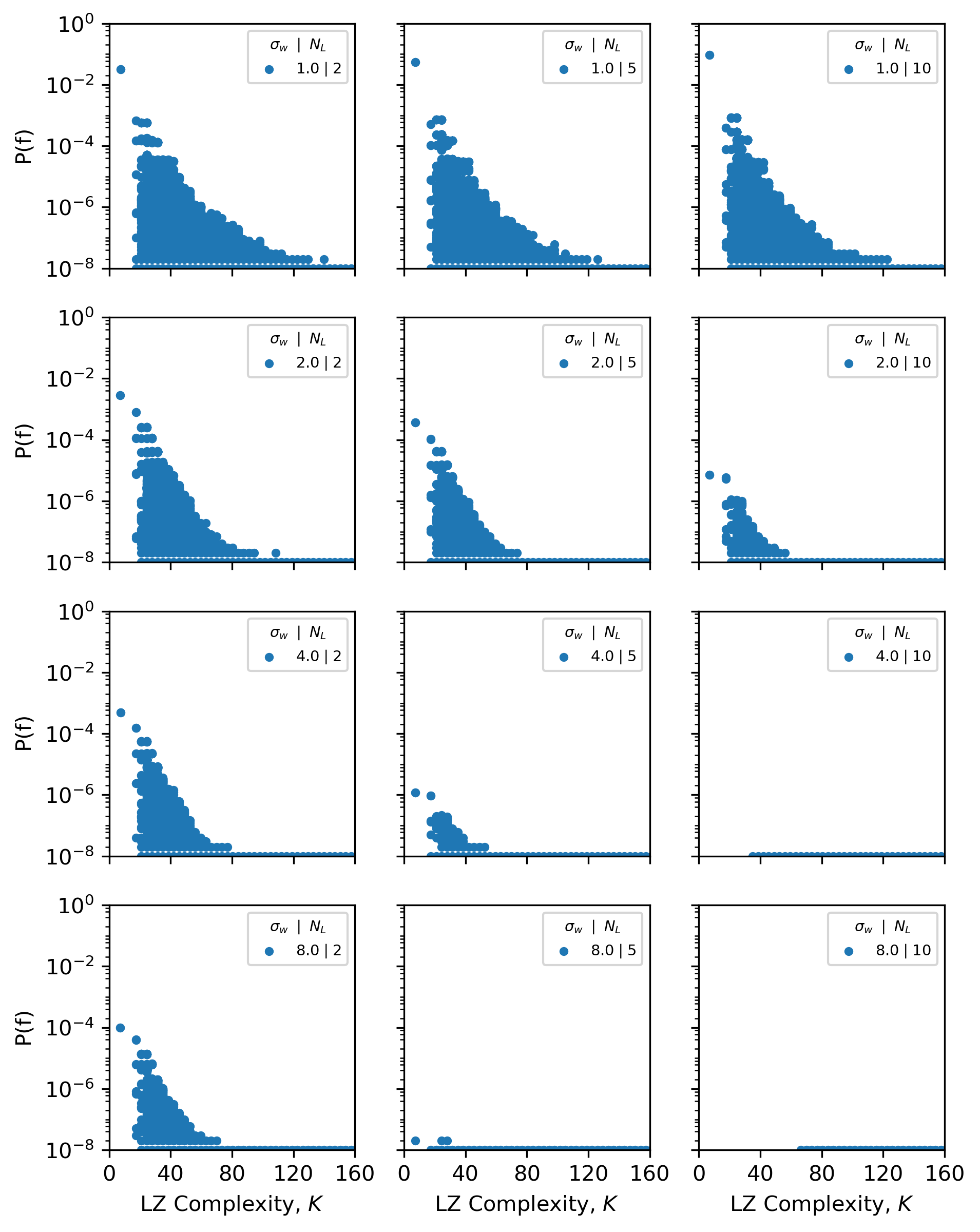}
        \caption{P(f) v.s.\ LZ Complexity}
    \end{subfigure}
    \begin{subfigure}[b]{0.48\columnwidth}
        \includegraphics[width = \textwidth]{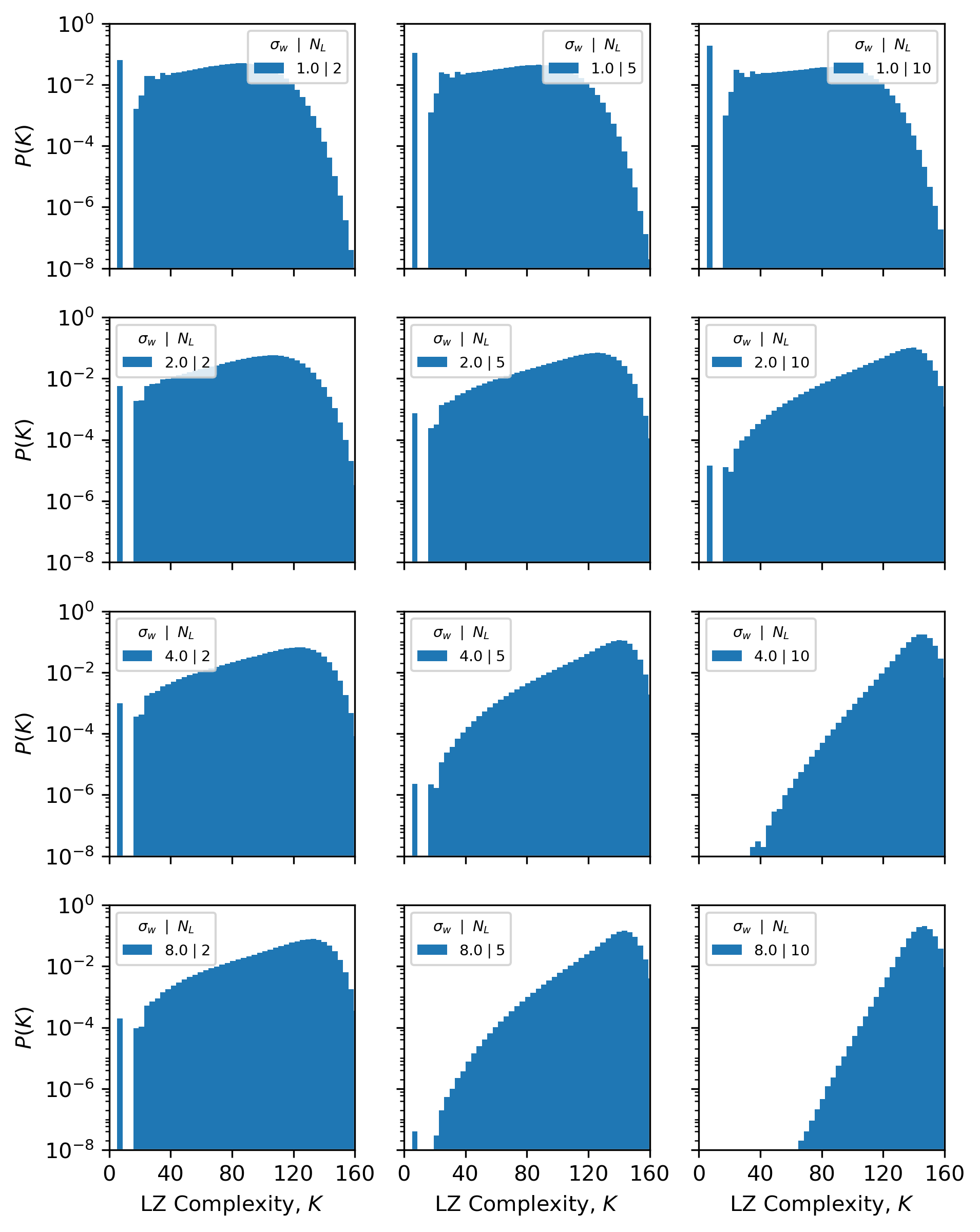}
        \caption{P(K) v.s.\ LZ Complexity}
    \end{subfigure}
    \caption{\textbf{DNN priors for different $\bf \sigma_w$ and depth $\bf N_l$:} (a) Empirical probability $P(f)$ of individual functions versus their LZ complexity for networks initialized with various $\sigma_w$ and number of layers $N_l$, with hidden layer width 40 and $\tanh$ activations for $10^8$ random samples. Points with a probability of $10^{-8}$ are not removed  (even though sampling errors for these low probability points are large) because in plots ($\sigma_w = 4,\; N_l =10$) and ($\sigma_w = 8,\; N_l =10$) only points of this type are found.  (b) \emph{A-priori} probability $P(K)=\sum_{f:K_{LZ}(f)=K} P(f)$ versus LZ Complexity based on $10^8$ samples of a neural network with $N_l$ hidden layers, hidden layer width 40 and tanh activations, initalized i.i.d with weight standard deviation $\sigma_w$. $\sigma_b=0.2\sigma_w$. Further examples for \cref{subfig:rank_plot}.
    See \cref{app:exp_details:P(K)} for further experimental details
    }\label{app:fig:priors}
\end{figure}

In \cref{app:fig:priors} we depict the priors for a range of $\sigma_w$ and depths $N_l$.  For increasing $\sigma_w$ and increasing $N_l$, which corresponds to entering more deeply into the chaotic regime,  the amount of simplicity bias decreases, and the $P(K)$ plots show a sharper rise towards complex functions.

\begin{figure}[H]
    \centering
    \begin{subfigure}[ht]{0.3\linewidth}
        \includegraphics[width = \textwidth]{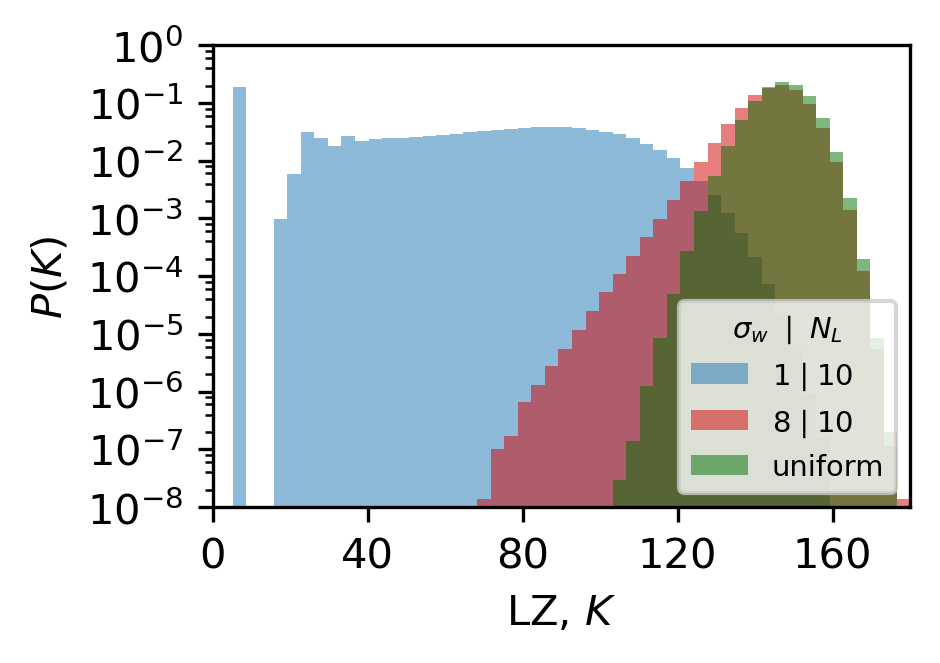}
        \caption{LZ Complexity}
    \end{subfigure}
    \begin{subfigure}[ht]{0.3\linewidth}
        \includegraphics[width = \textwidth]{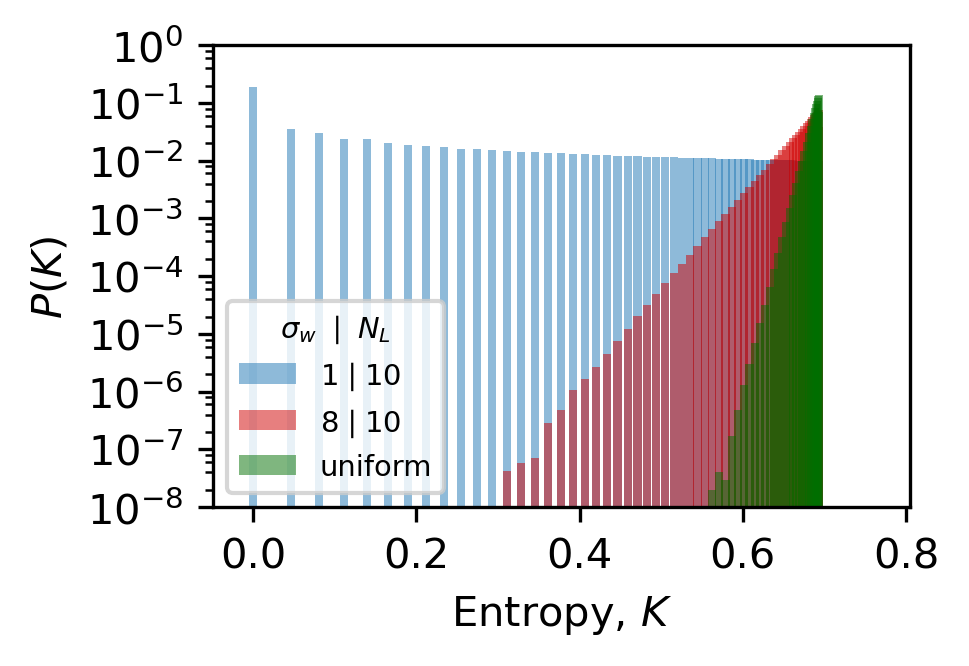}\caption{Entropy}
    \end{subfigure}
    \begin{subfigure}[ht]{0.3\linewidth}
        \includegraphics[width = \textwidth]{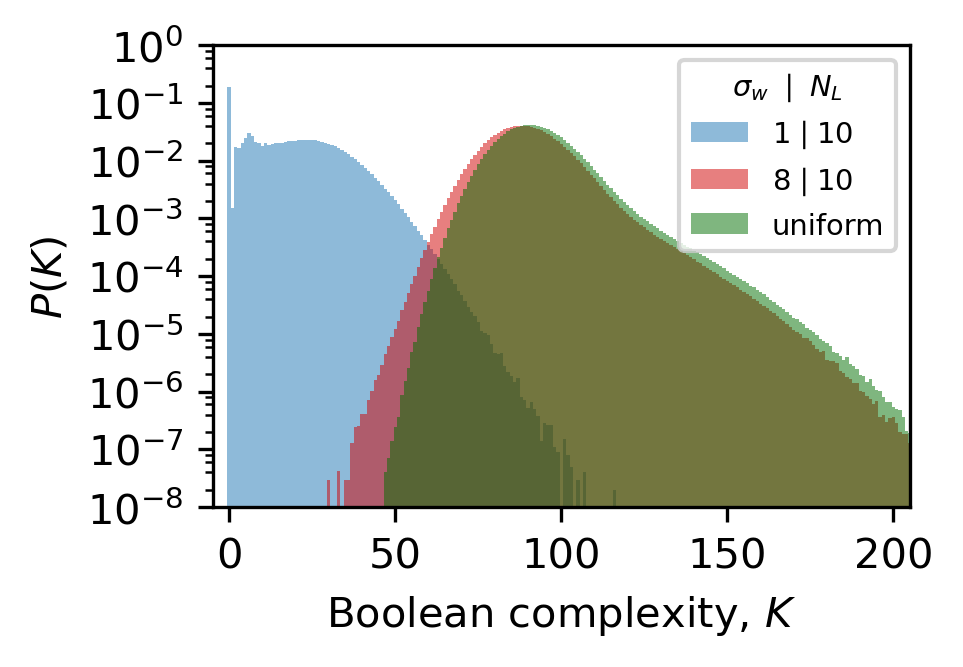}\caption{Boolean complexity}
    \end{subfigure}
    \caption{\small {\bf $\bf P(K)$ for different complexity measures.} Prior $P(K)$ for 10-layer FCNs with tanh activations, on $n=7$ Boolean functions, ranked by probability of individual functions, generated from  $10^8$ random samples of parameters $\Theta\sim \mathcal{N}(0,\sigma_w)$. Uniform sampling is included for reference. (a) LZ Complexity (b) Entropy (c) Boolean complexity. (b) the entropy here is that of the string representation of the function, and (c) is the same Boolean complexity measure as in \citep{valle2018deep} (the minimum of the Disjunctive/Conjunctive Normal Form, calculated with SOPform and POSform from Scipy).
    }\label{app:fig:K_measures}
\end{figure}

Finally, in \cref{app:fig:K_measures} we show the priors $P(K)$ for three different complexity measures.  The first is the LZ measure used before. The second is the
simple binary entropy $K_S=-\log_2 p -\log_2 (1-p)$ where $p$ is the fraction of $0$'s (or equivalently of $1$'s.   Simple strings such as all 0's or all 1s also have low LZ complexity, but strings such as "01010101..." have high binary entropy, but low LZ complexity. The third measure is a classic measure of Boolean complexity, the minimum of the disjunctive/conjunctive normal form.  While all three $P(K)$ plots differ in detail, as expected, they also all present the same qualitative behaviour in that the chaotic DNN prior is much closer to that of random strings than the order DNN prior is.

\break

\section{Extended figures for generalization error, posteriors, and likelihoods}\label{app:ce_advsgd}

In this section, we present extended data for the generalization error with data complexity, scatter plots of error v.s.\ complexity, as well as the approximate average likelihoods and posteriors for the decoupling approximation, compared to posteriors from SGD.  These figures complement main text Figures~\ref{fig:Cube_plots_1}~$\&$~\ref{fig:Cube_plots_2}.

\begin{figure}[H]
    \centering
    \begin{subfigure}[ht]{0.22\linewidth}
        \includegraphics[width = \textwidth]{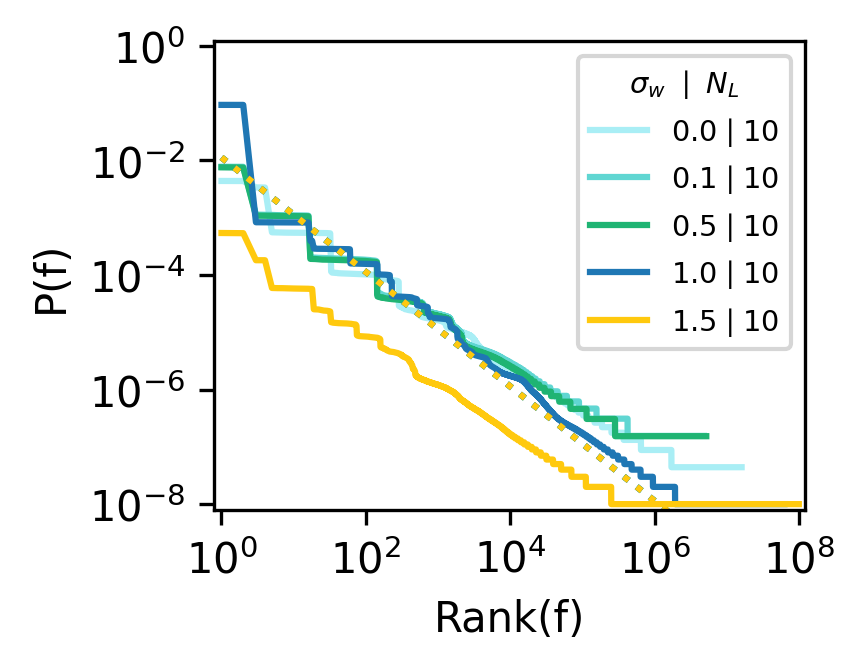}
        \caption{}
    \end{subfigure}
    \begin{subfigure}[ht]{0.22\linewidth}
        \includegraphics[width = \textwidth]{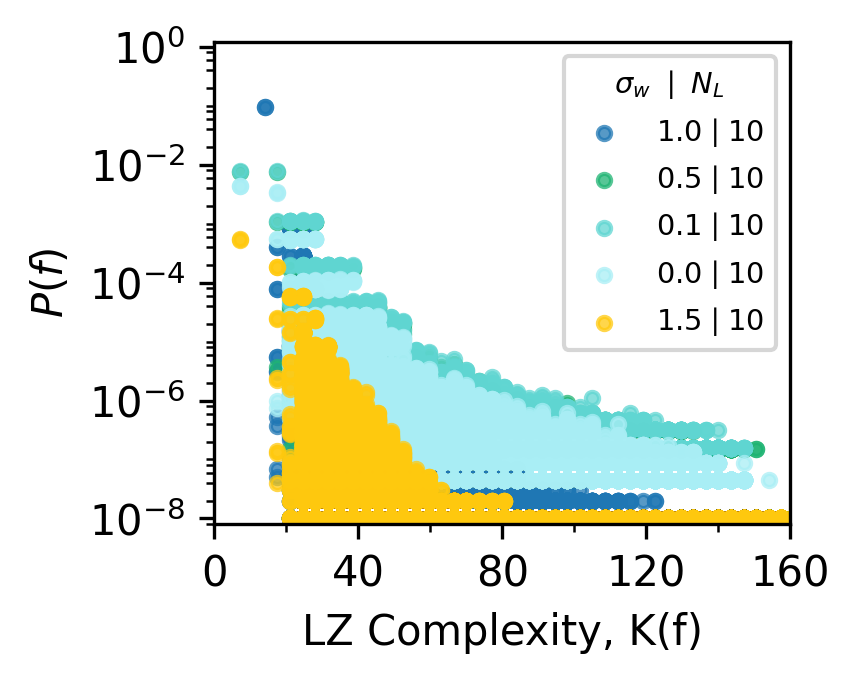}\caption{}
    \end{subfigure}
    \begin{subfigure}[ht]{0.22\linewidth}
        \includegraphics[width = \textwidth]{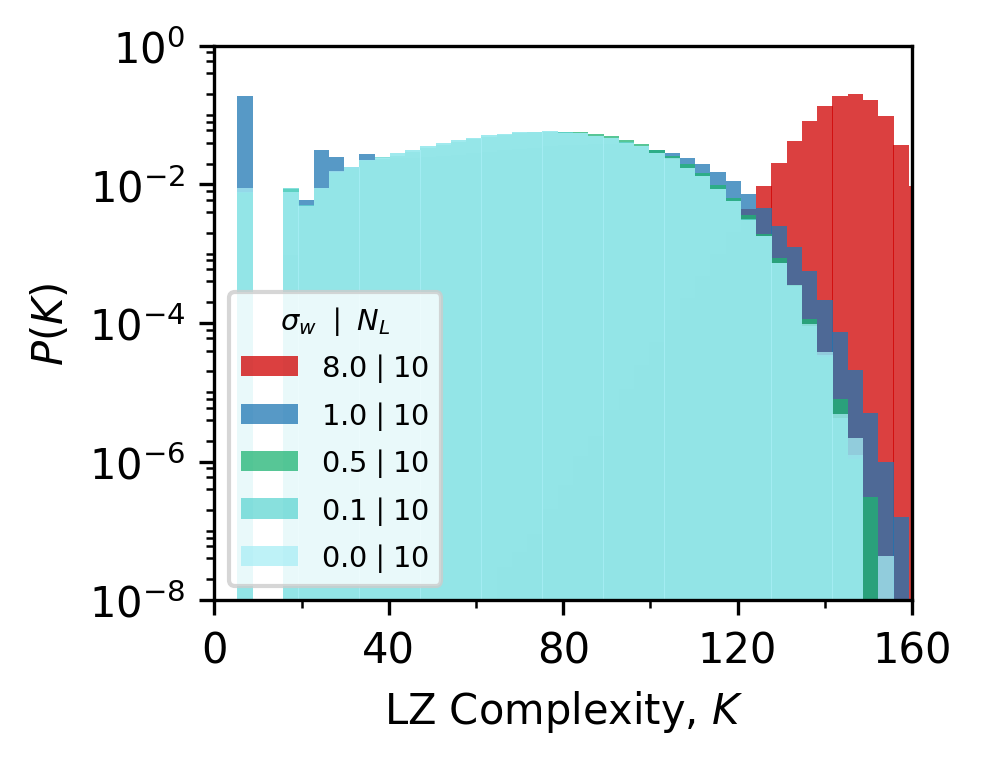}\caption{}
    \end{subfigure}
    \begin{subfigure}[ht]{0.22\linewidth}
        \includegraphics[width = \textwidth]{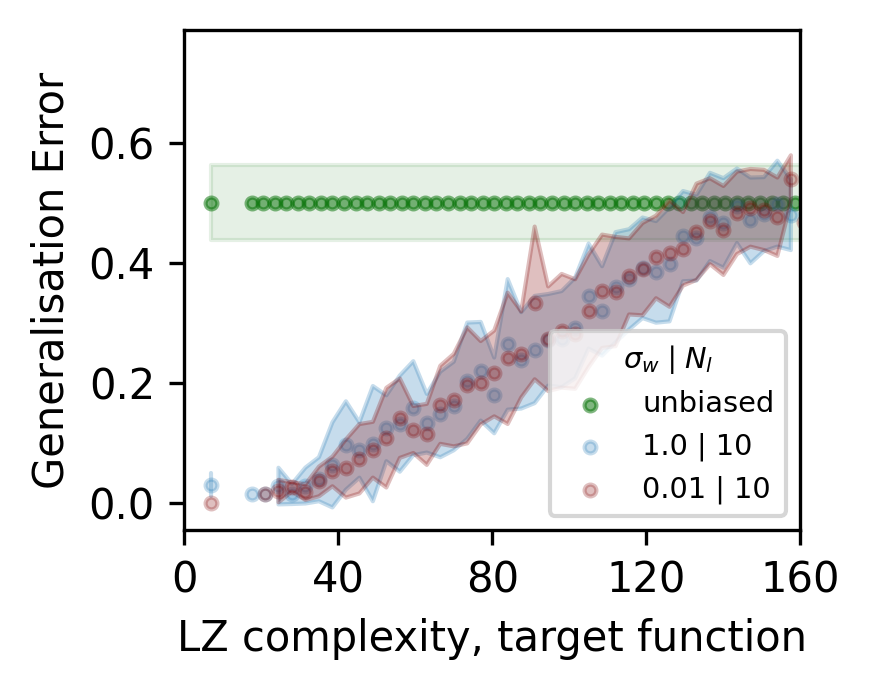}\caption{}
    \end{subfigure}
    
    \begin{subfigure}[ht]{0.3\linewidth}
        \includegraphics[width = \textwidth]{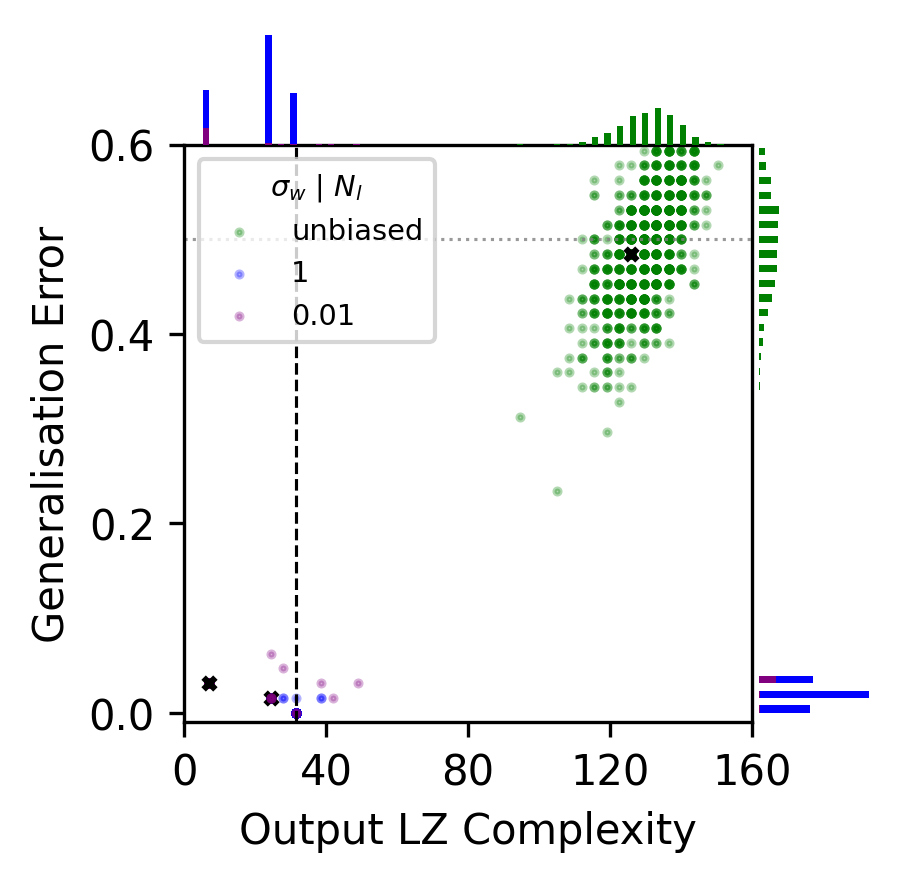}
        \caption{}
    \end{subfigure}
    \begin{subfigure}[ht]{0.3\linewidth}
        \includegraphics[width = \textwidth]{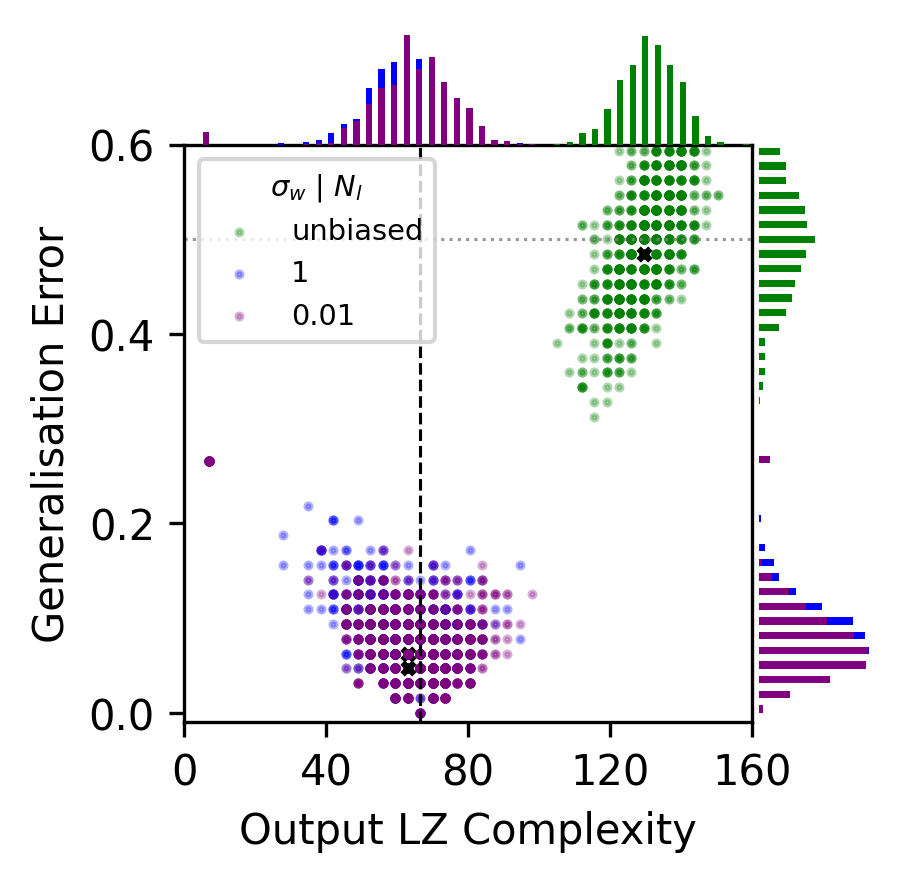}\caption{}
    \end{subfigure}
    \begin{subfigure}[ht]{0.3\linewidth}
        \includegraphics[width = \textwidth]{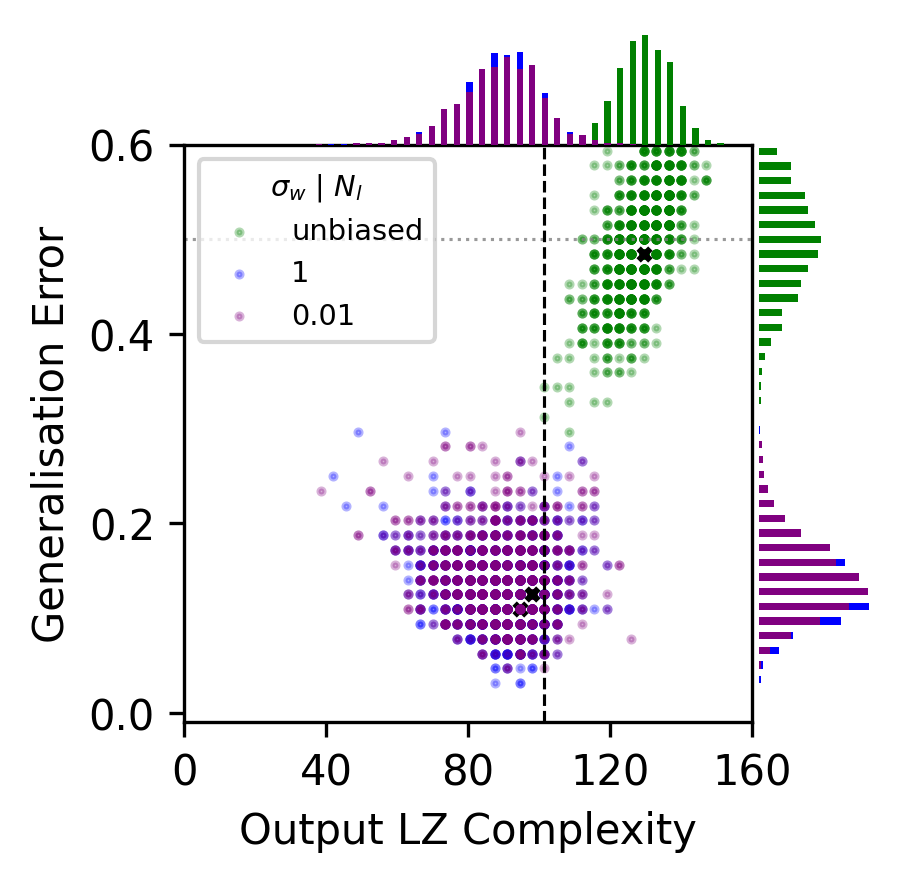}\caption{}
    \end{subfigure}
    
    \caption{\small {\bf Priors and posteriors in the ordered regime} 
    This figure extends results in \cref{fig:Cube_plots_1} to the ordered regime, where $\sigma_w<1$ ($\sigma_b=0$ for all $\sigma_w<1$). In (a) and (b), $\sigma_w=0$ denotes the limit of very small $\sigma_w$, where the tanh activations are essentially linear. This linear neural network regime has effectively the same expressivity as a perceptron. As observed in \citep{mingard2019neural}, the simplicity bias is slightly weaker for perceptrons than DNNs with non-linear activations. This can be seen in (a) by slightly shallower $P(f)$ v.s. $Rank(f)$ curves in the ordered regime. 
    (c) shows the $P(K)$ plots for (b). $\sigma_w=0.5,0.1,0.0$ are very close,  although the slight decrease in simplicity bias can be picked up at larger complexities. 
    In (d) we observe that these small differences in the priors do not appreciably  impact the generalisation performance of the SGD trained neural networks.
    This similarity is also observed in the scatterplots of the posteriors (e-g), which use target functions of the same complexity as those in \cref{fig:Cube_plots_1}(e-g), the ordered prior with $\sigma_w=0.1$ exhibits similar behaviour to an FCN with $\sigma_w = 1$, and both differ markedly from an unbiased learner.
    See \cref{fig:app:1c_extras,fig:app:1d_extras} for the behaviour of these networks when trained with other optimisers, activation functions, and loss-functions. 
    }\label{app:fig:ordered_plots}
\end{figure}

\begin{figure}[H]
    \centering
    \makebox[\columnwidth][c]{\includegraphics[width=0.75\columnwidth]{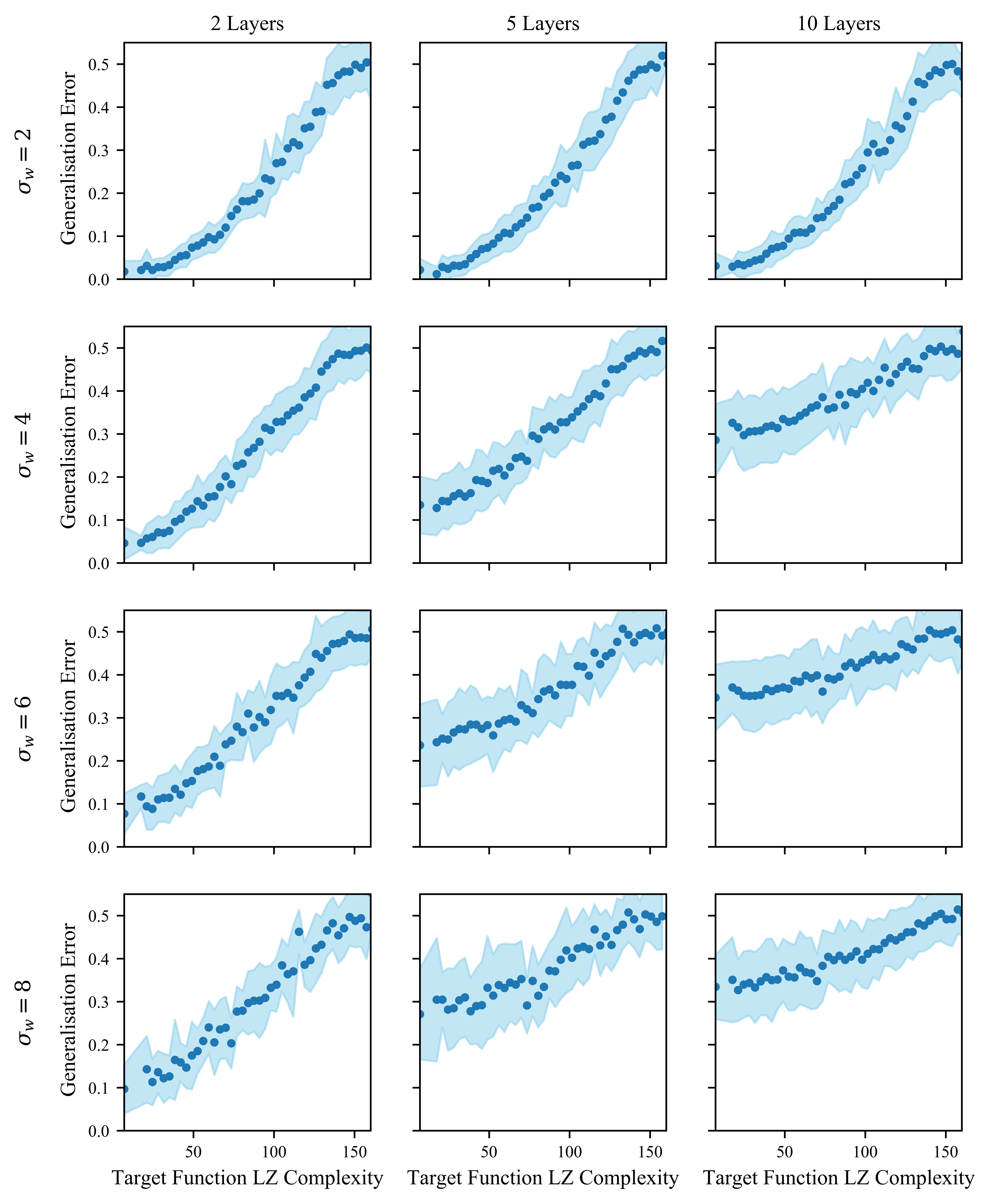}}%
    \caption{\textbf{Generalization error  versus LZ complexity with AdvSGD} for networks of varying numbers of layers $N_l$ and weight standard deviation $\sigma_w$. Architecture details are identical to those in \cref{app:fig:priors}, and were trained with AdvSGD. Error bars are one standard deviation.
    Moving further into the chaotic regime leads to poorer generalization performance.
    These examples complement the data shown in the main text in  \cref{subfig:trained_DNN_bias}. See \cref{app:exp_details:eg_vs_LZ} for further experimental details.
    }\label{app:fig:e_vs_lz}
\end{figure}

In \cref{app:fig:ordered_plots} we show extensions to plots in \cref{fig:Cube_plots_1} in the ordered regime, i.e.\ $\sigma_w<1$. In \cref{app:fig:e_vs_lz} we display the generalization error for DNNs with different $\sigma_w$ and $N_l$. This figure extends \cref{fig:Cube_plots_1}(c) in the main text.  By comparing to \cref{app:fig:priors}, one can see how   the change in bias strongly affects the generalization error.

\begin{figure}[H]
    \begin{subfigure}[b]{0.24\columnwidth}
        \includegraphics[width = \textwidth]{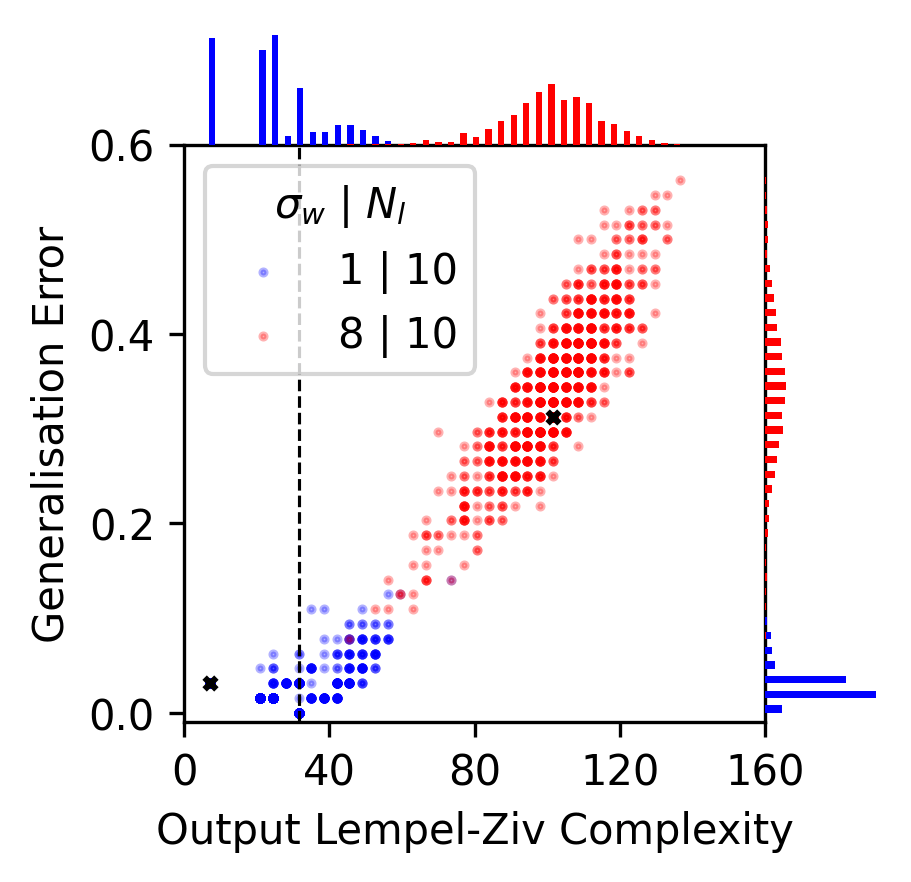}
        \caption{$C_{LZ}(f_t)=$31.5}
    \end{subfigure}
    \begin{subfigure}[b]{0.24\columnwidth}
        \includegraphics[width = \textwidth]{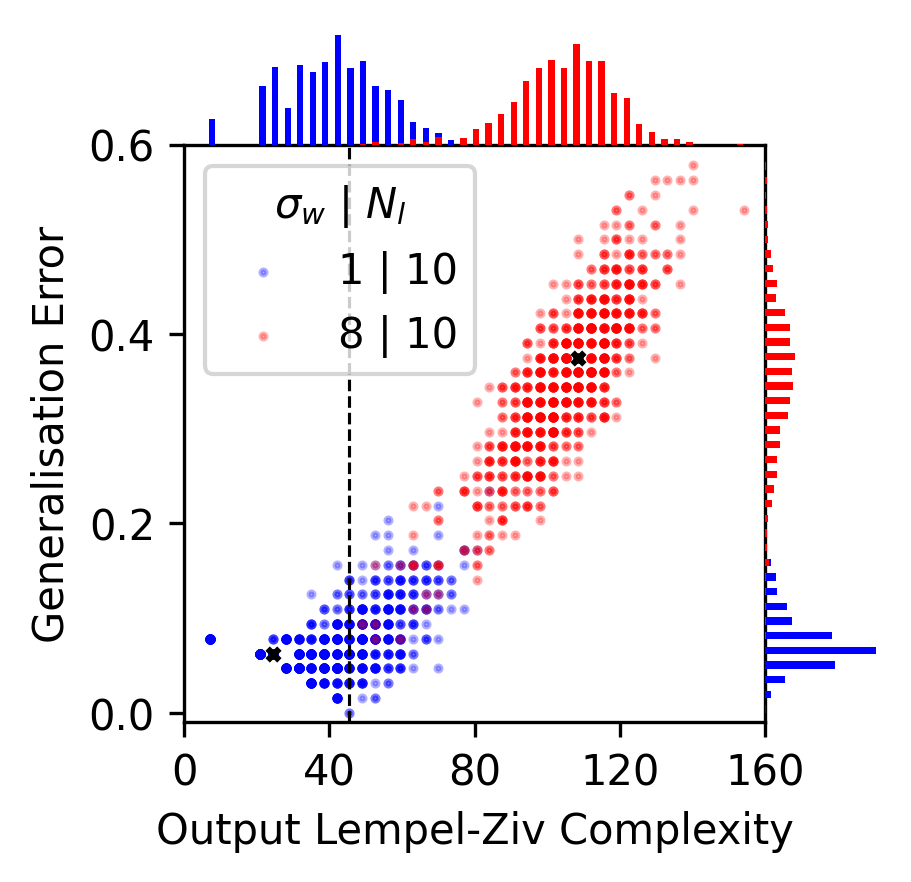}
        \caption{$C_{LZ}(f_t)=$45.5}
    \end{subfigure}
    \begin{subfigure}[b]{0.24\columnwidth}
        \includegraphics[width = \textwidth]{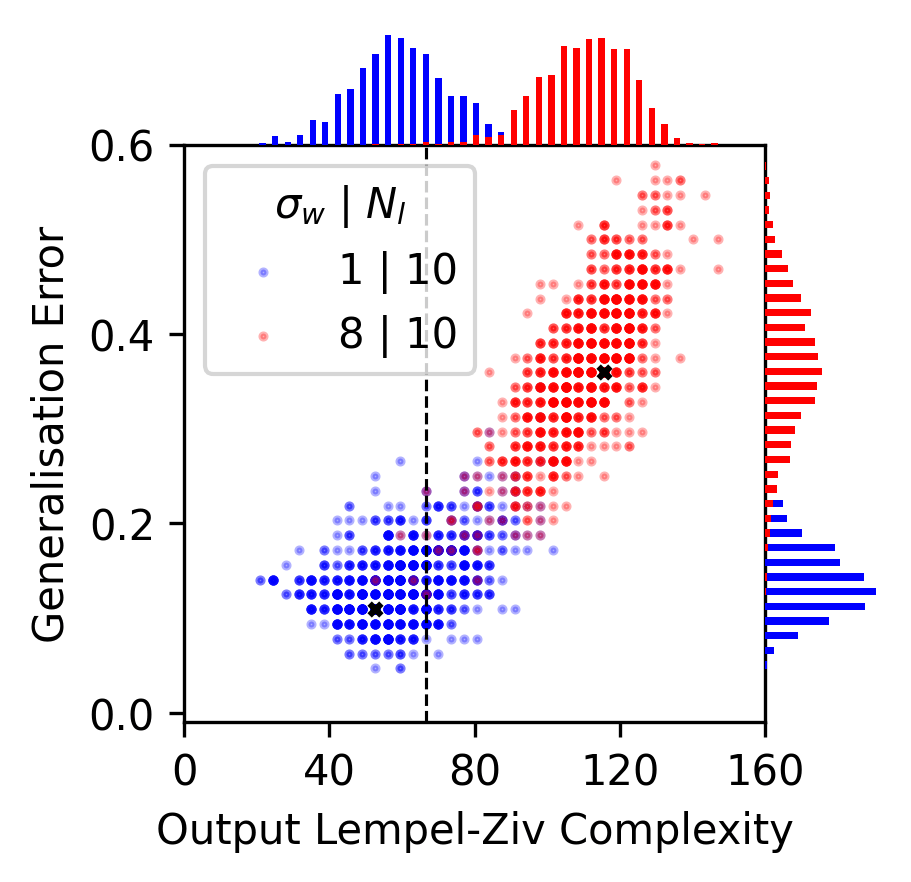}
        \caption{$C_{LZ}(f_t)=$66.5}
    \end{subfigure}
    \begin{subfigure}[b]{0.24\columnwidth}
        \includegraphics[width = \textwidth]{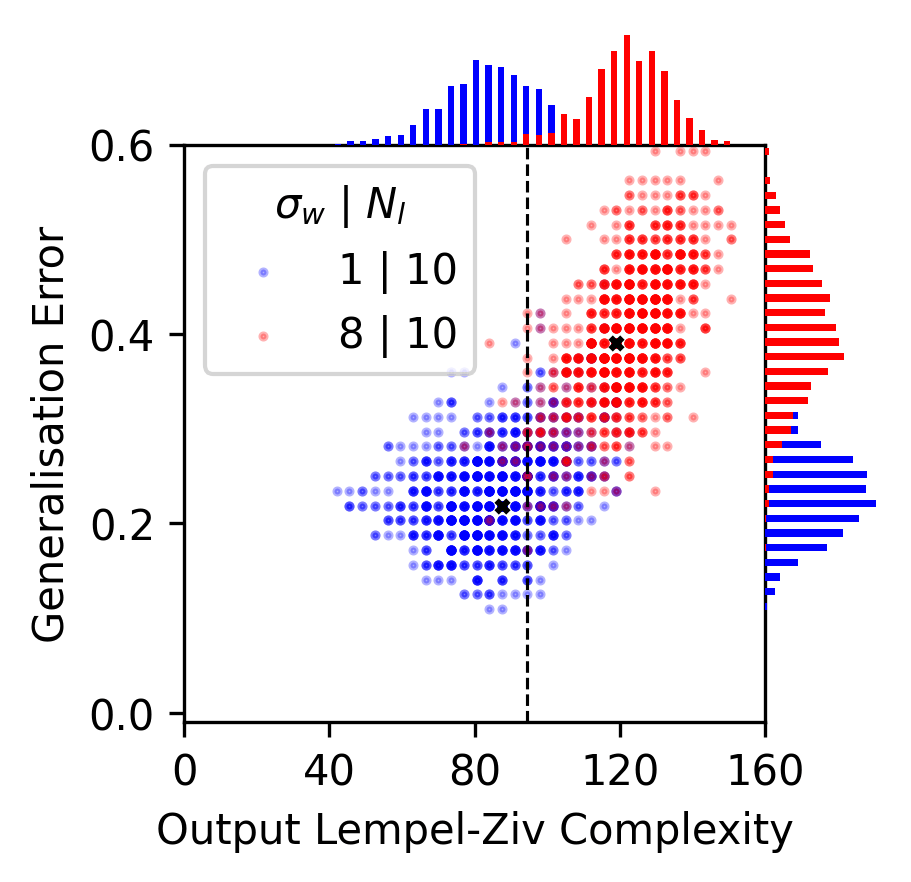}
        \caption{$C_{LZ}(f_t)=$94.5}
    \end{subfigure}
    
    \begin{subfigure}[b]{0.24\columnwidth}
        \includegraphics[width = \textwidth]{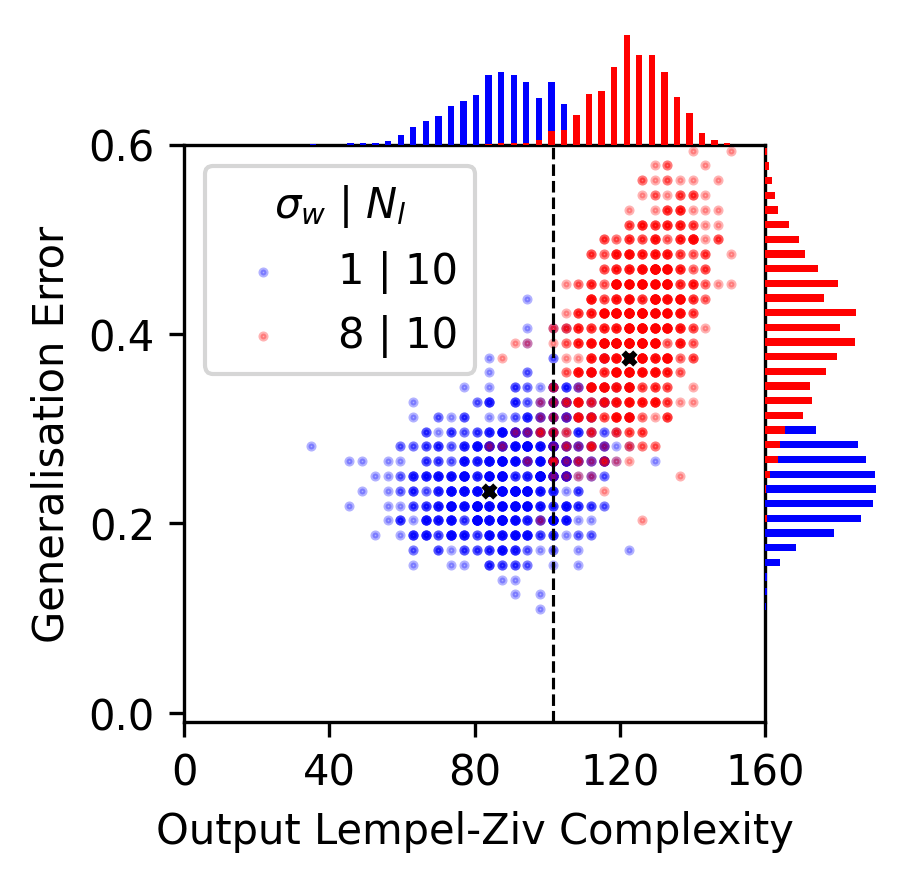}
        \caption{$C_{LZ}(f_t)=$101.5}
    \end{subfigure}
    \begin{subfigure}[b]{0.24\columnwidth}
        \includegraphics[width = \textwidth]{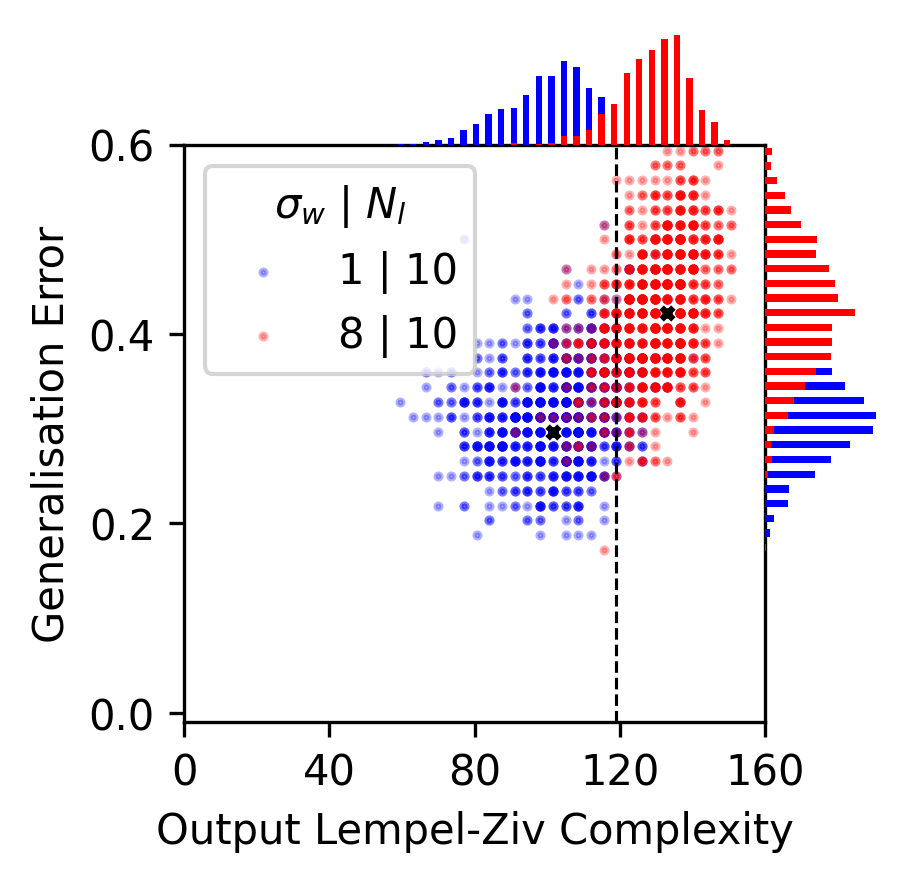}
        \caption{$C_{LZ}(f_t)=$119.0}
    \end{subfigure}
    \begin{subfigure}[b]{0.24\columnwidth}
        \includegraphics[width = \textwidth]{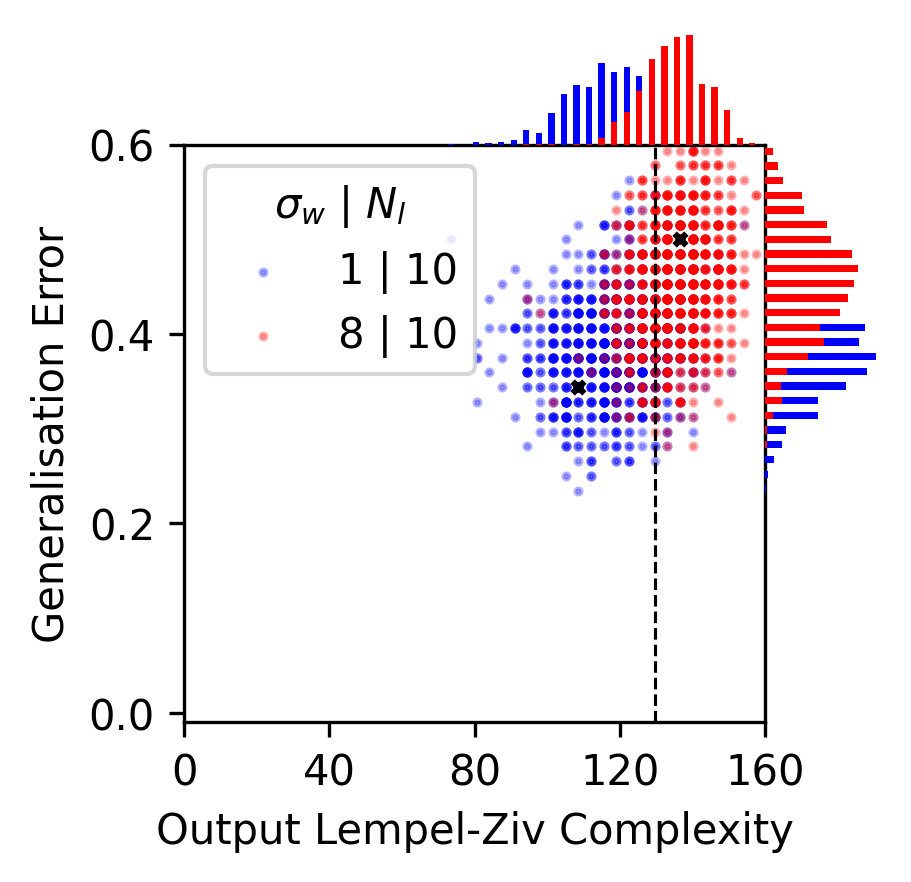}
        \caption{$C_{LZ}(f_t)=$129.5}
    \end{subfigure}
    \begin{subfigure}[b]{0.24\columnwidth}
        \includegraphics[width = \textwidth]{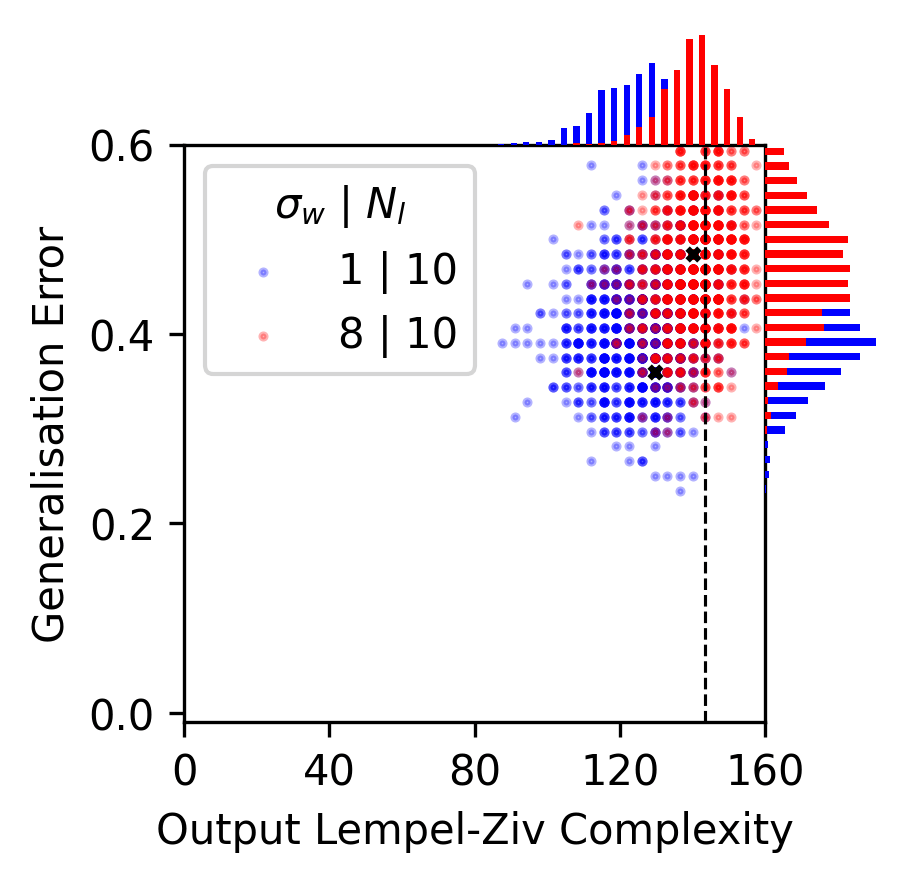}
        \caption{$C_{LZ}(f_t)=$143.5}
    \end{subfigure}
    \caption{{\bf Generalization error versus learned function LZ complexity scatter plots with AdvSGD}.  These complement the main text \cref{subfig:TF31,subfig:TF66,subfig:TF101}.  Each point depicts a generalization error/complexity pair obtained for 1000 random initializations of the $\sigma_w =1$ (blue) and $\sigma_w=8$ (red) DNNs.  The networks were trained on target functions of various LZ complexity and a training set size of 64 chosen i.i.d.\ for each run from the full set of 128 inputs. generalization error is measured on the other $64$ inputs.  The histograms on the top/side of the plots show the distribution of complexity/generalization errors. The black cross represents the most commonly learned function in each case.} \label{app:fig:scatter1} 
\end{figure}

\begin{figure}[H]
\centering

    \begin{subfigure}[ht]{0.3\linewidth}
        \includegraphics[width=\textwidth]{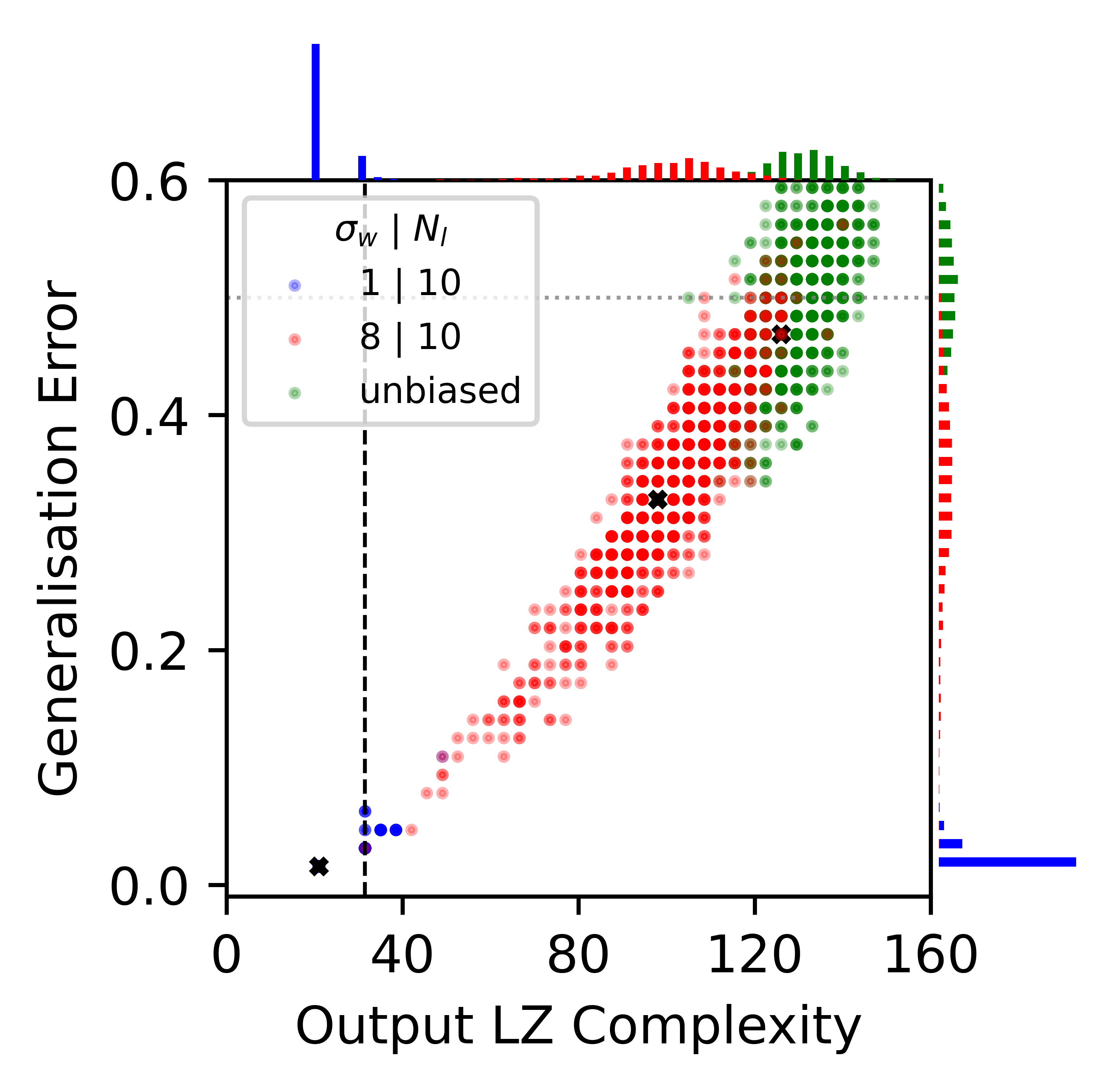}\caption{LZ=31.5}
    \end{subfigure}
    \begin{subfigure}[ht]{0.3\linewidth}
        \includegraphics[width=\textwidth]{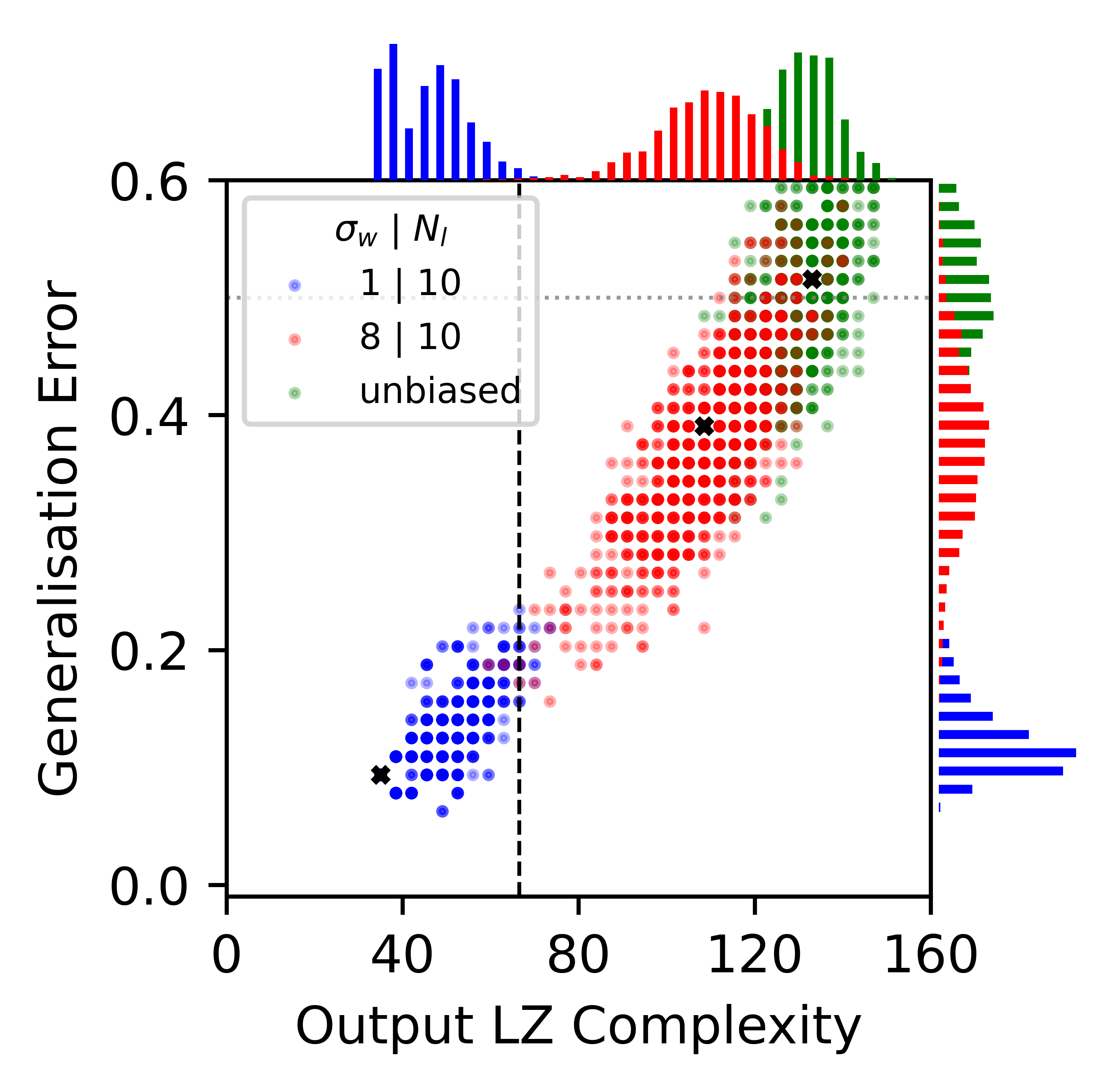}\caption{LZ=66.5}
    \end{subfigure}
    \begin{subfigure}[ht]{0.3\linewidth}
        \includegraphics[width=\textwidth]{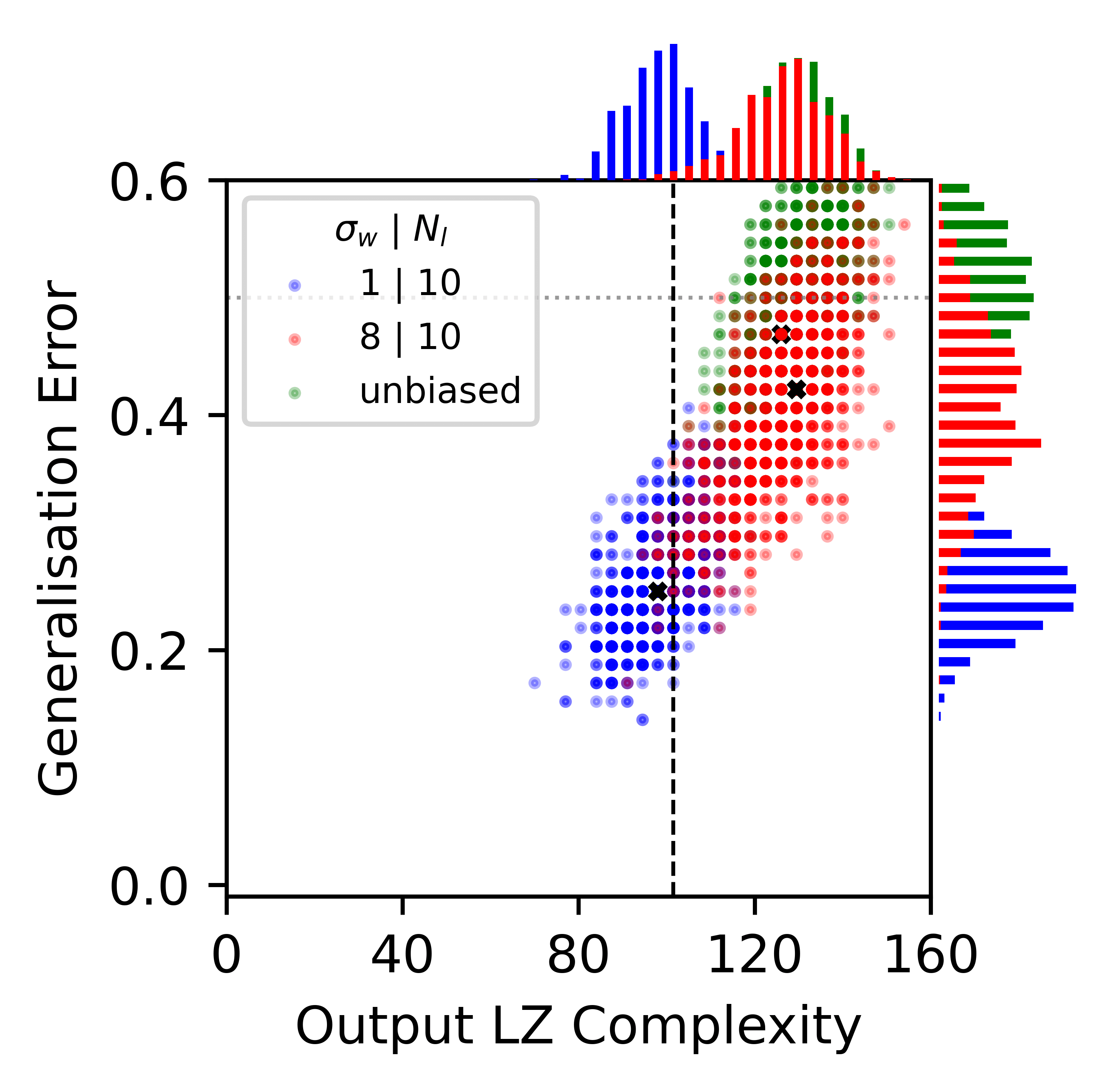}\caption{LZ=101.5}
    \end{subfigure} 
    
    \caption{  \small {\bf \small Scatter plots with on a single representative training set from scatter plots in \cref{fig:Cube_plots_1} with AdvSGD.
    }. Model and optimiser identical to \cref{fig:Cube_plots_1}. Here, as opposed to previous figures, the training set consists of 64 inputs that are held  constant across all 1000 runs. As expected, the results are qualitatively similar to \cref{fig:Cube_plots_1}, where a different random training set is used per training run.}\label{app:fig:single_train}
\end{figure}

In \cref{app:fig:scatter1} and \cref{app:fig:single_train}, we show scatter plots that complement the main text scatter plots \cref{subfig:TF31,subfig:TF66,subfig:TF101}.

In \cref{app:fig:error_func} we show the approximations to the posterior via the decoupling approximation \cref{eq:PBapprox} for a wider range of complexities than shown in \cref{fig:Cube_plots_2}. Similarly, in \cref{app:fig:k=1_approx_bayes} and \cref{app:fig:k=50_approx_bayes} we complement \cref{fig:Cube_plots_2} by showing the results of the decoupling approximation for $l=1$ and $l=50$ functions (respectively), chosen to calculate the mean likelihood.  Note that while the decoupling approximation provides an estimate that should  be proportional to the posterior, it likely will not lead to a correct normalisation on its own.  In the plots, we normalise the posteriors separately.  As can be seen in these plots, despite the rather strong approximations inherent in \cref{eq:PBapprox}, this approach turns out to be robust to changes in the number of functions used, and works for a range of data complexities and training set sizes. See also \cref{fig:app:gp_nn_2} for similar calculations, but with MSE loss, and including a near-exact Bayesian approximation using a Gaussian Process (GP) approach.   In all these cases the decoupling approximation captures key qualitative features of the posterior, providing an a-posteriori justification for its use for this system.   It would be interesting to find an a-prior justification for this approximation as well.

\begin{figure}[H]
    \centering
    \begin{subfigure}[b]{0.23\columnwidth}
        \includegraphics[width = \textwidth]{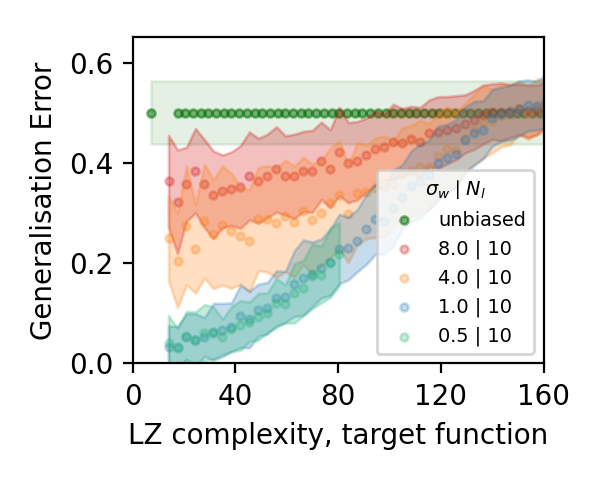}
        \caption{m=32, ce, tanh}
    \end{subfigure}
    \begin{subfigure}[b]{0.23\columnwidth}
        \includegraphics[width = \textwidth]{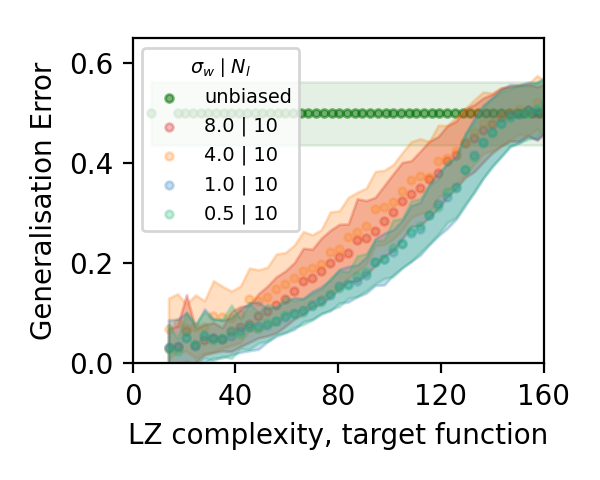}
        \caption{m=32, ce, relu}
    \end{subfigure}
    \begin{subfigure}[b]{0.23\columnwidth}
        \includegraphics[width = \textwidth]{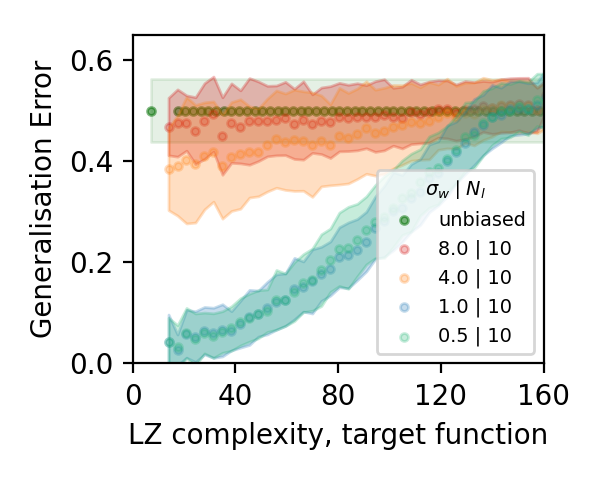}
        \caption{m=32, mse, tanh}
    \end{subfigure}
    \begin{subfigure}[b]{0.23\columnwidth}
        \includegraphics[width = \textwidth]{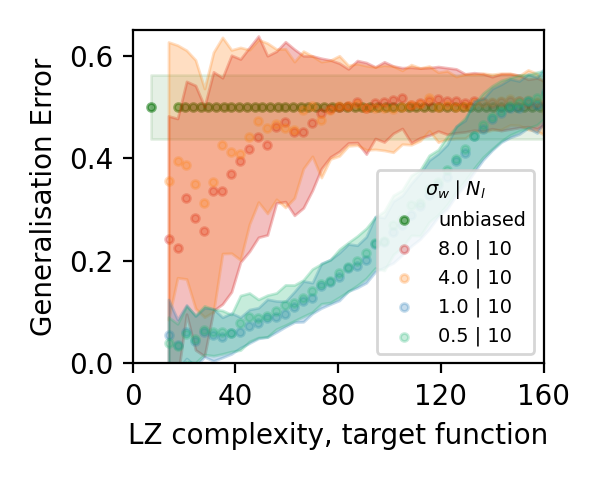}
        \caption{m=32, mse, relu}
    \end{subfigure}
    
    \begin{subfigure}[b]{0.23\columnwidth}
        \includegraphics[width = \textwidth]{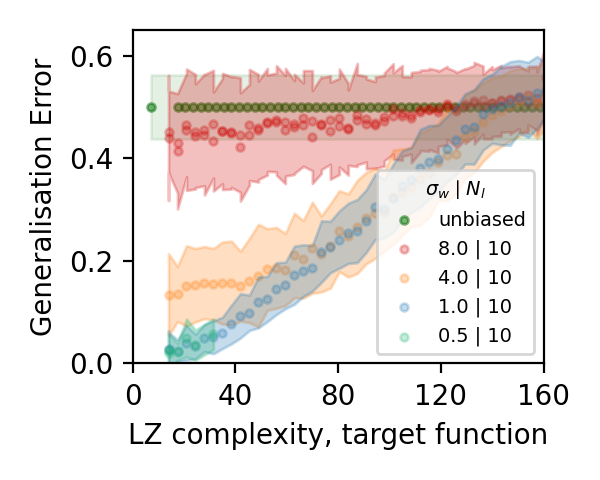}
        \caption{m=64, ce, tanh}
    \end{subfigure}
    \begin{subfigure}[b]{0.23\columnwidth}
        \includegraphics[width = \textwidth]{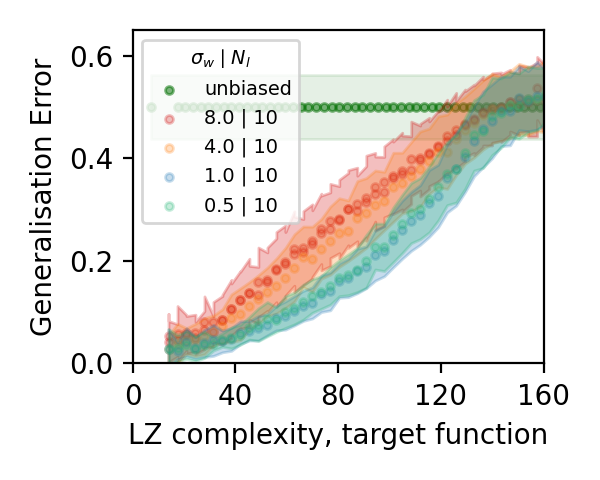}
        \caption{m=64, ce, relu}
    \end{subfigure}
    \begin{subfigure}[b]{0.23\columnwidth}
        \includegraphics[width = \textwidth]{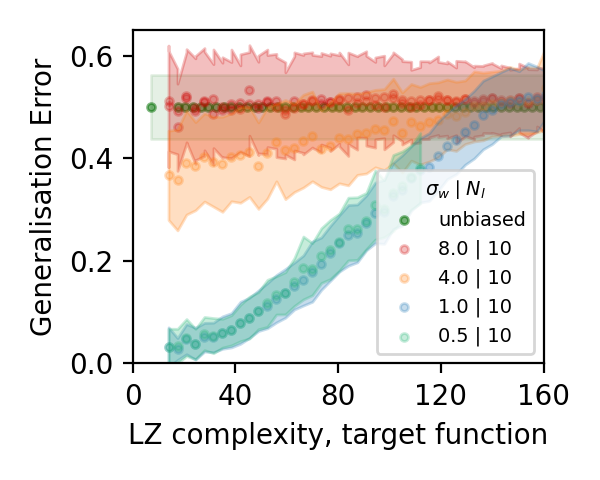}
        \caption{m=64, mse, tanh}
    \end{subfigure}
    \begin{subfigure}[b]{0.23\columnwidth}
        \includegraphics[width = \textwidth]{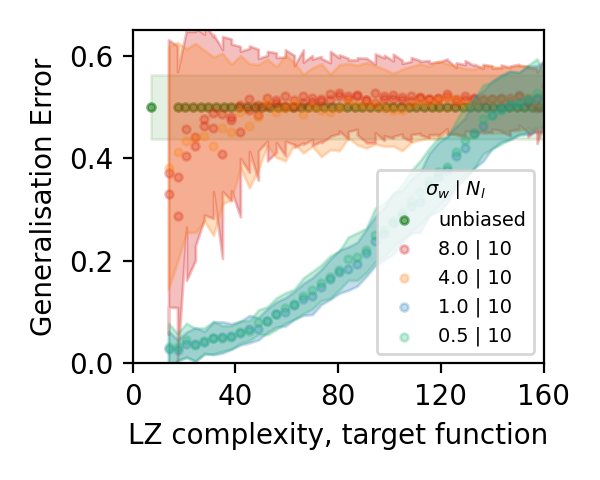}
        \caption{m=64, mse, relu}
    \end{subfigure}
    
    \begin{subfigure}[b]{0.23\columnwidth}
        \includegraphics[width = \textwidth]{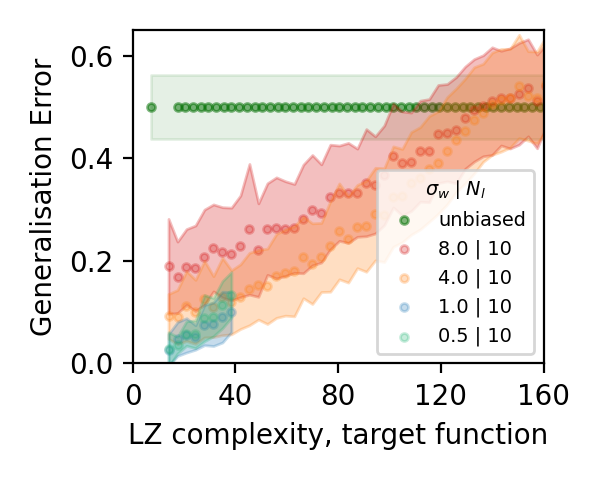}
        \caption{m=100, ce, tanh}
    \end{subfigure}
    \begin{subfigure}[b]{0.23\columnwidth}
        \includegraphics[width = \textwidth]{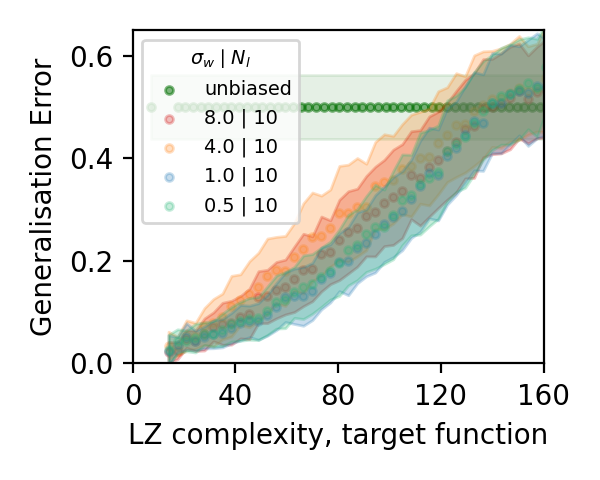}
        \caption{m=100, ce, relu}
    \end{subfigure}
    \begin{subfigure}[b]{0.23\columnwidth}
        \includegraphics[width = \textwidth]{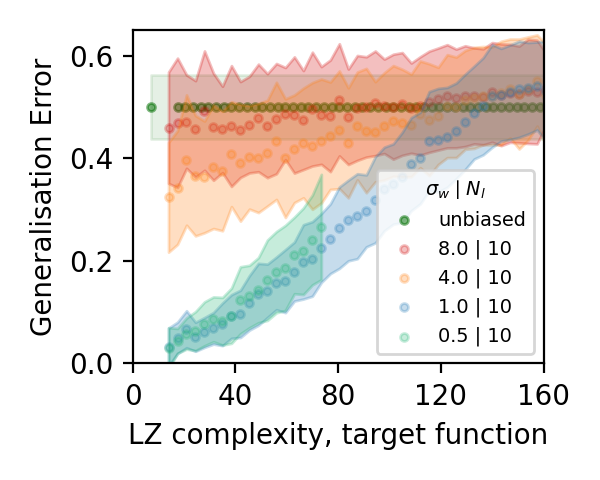}
        \caption{m=100, mse, tanh}
    \end{subfigure}
    \begin{subfigure}[b]{0.23\columnwidth}
        \includegraphics[width = \textwidth]{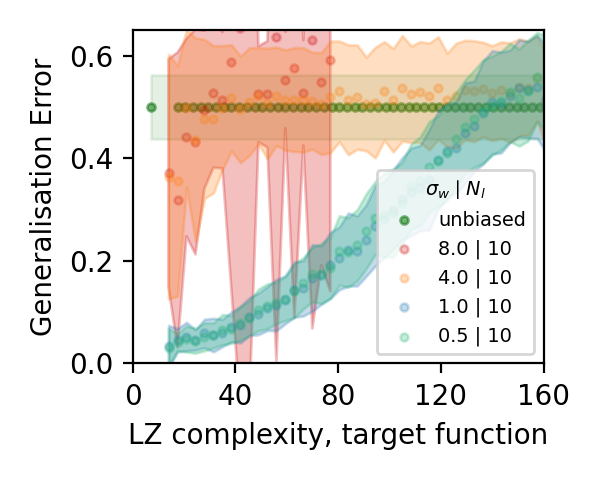}
        \caption{m=100, mse, relu}
    \end{subfigure}
    
    \caption{\small {\bf generalization error v.s. complexity for different training set sizes, activation functions and loss functions}
    The models used are all 10-layer FCNs on the $n=7$ boolean system with either relu or tanh activations, and initialised with different values of $\sigma_w$. Models are trained with Adam (unlike \cref{fig:Cube_plots_1}(c) which use AdvSGD).
    Training set size $m$ increases vertically. 
    The first column shows cross-entropy loss with tanh activations (identical networks to \cref{fig:Cube_plots_1}(c)). The simplicity bias is strongly dependent on $\sigma_W$. 
    The second column shows ReLU actiavtions and cross-entropy loss. The ReLU activations mean that the priors $P(f)$ are independent of $\sigma_w$, and we see simplicity bias at all values of $\sigma_w$. 
    The final two columns show tanh and relu activated FCNs with mse loss. Because the mse loss introduces a stronger sense of scale (as the targets are at +1 and -1 rather than + and - infinity). Thus, smaller $\sigma_w$ solutions find smaller margin solutions which generalise better with ReLU activations \citep{farhang2022investigating}. Note that $\sigma_w=0.5$ was very slow to train, so those curves are sometimes incomplete. However, where present they match up very closely with $\sigma_w=1$.
}\label{fig:app:1c_extras}
\end{figure}

\begin{figure}[H]
    \centering
    \begin{subfigure}[b]{0.23\columnwidth}
        \includegraphics[width = \textwidth]{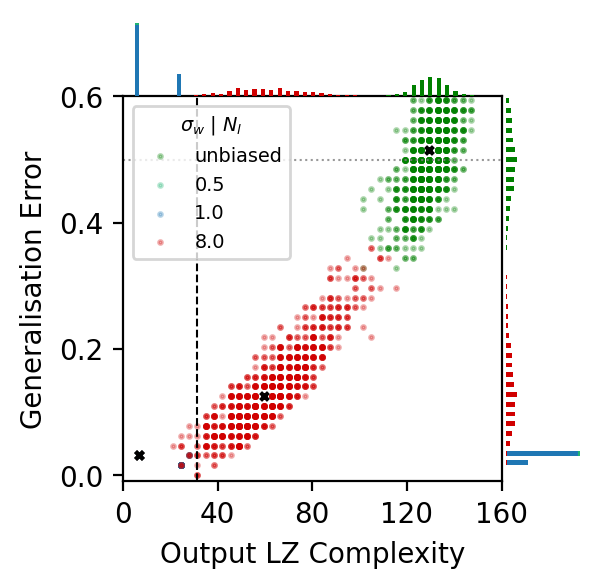}
        \caption{ce, tanh}
    \end{subfigure}
    \begin{subfigure}[b]{0.23\columnwidth}
        \includegraphics[width = \textwidth]{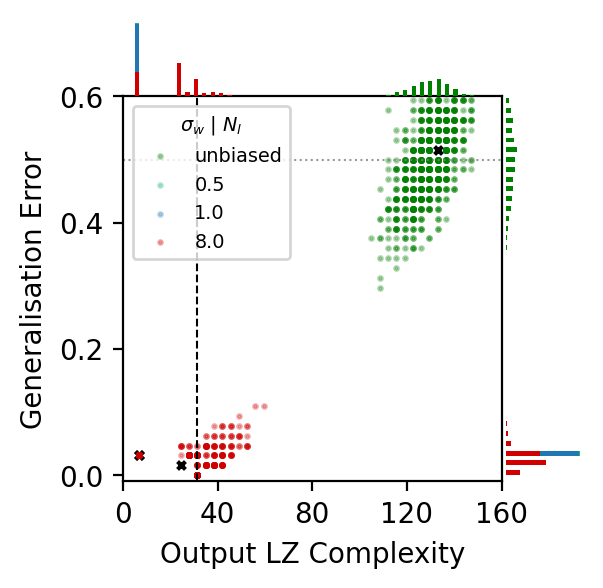}
        \caption{ce, relu}
    \end{subfigure}
    \begin{subfigure}[b]{0.23\columnwidth}
        \includegraphics[width = \textwidth]{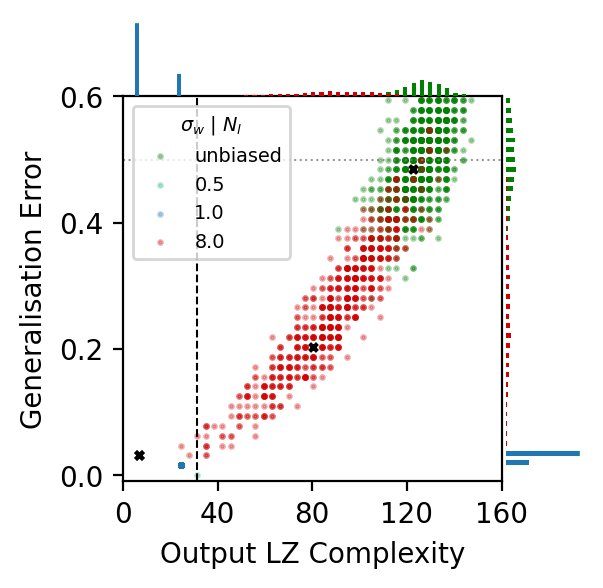}
        \caption{mse, tanh}
    \end{subfigure}
    \begin{subfigure}[b]{0.23\columnwidth}
        \includegraphics[width = \textwidth]{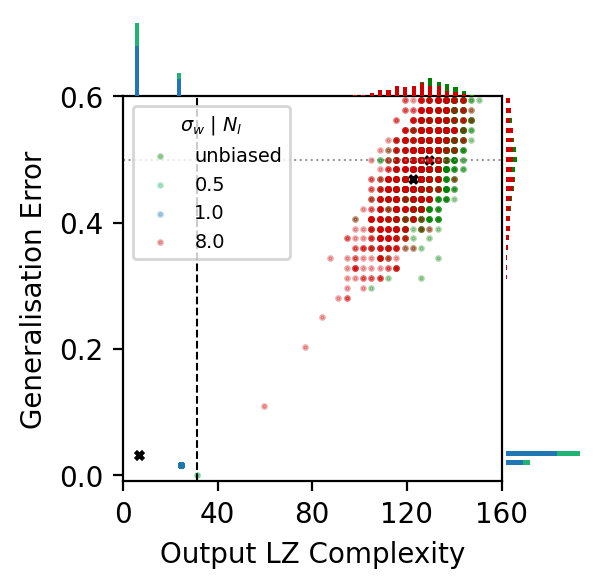}
        \caption{mse, relu}
    \end{subfigure}
    
    \begin{subfigure}[b]{0.23\columnwidth}
        \includegraphics[width = \textwidth]{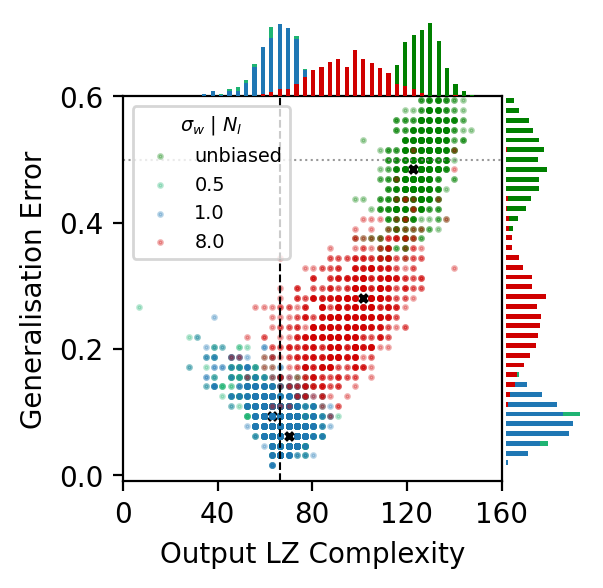}
        \caption{ce, tanh}
    \end{subfigure}
    \begin{subfigure}[b]{0.23\columnwidth}
        \includegraphics[width = \textwidth]{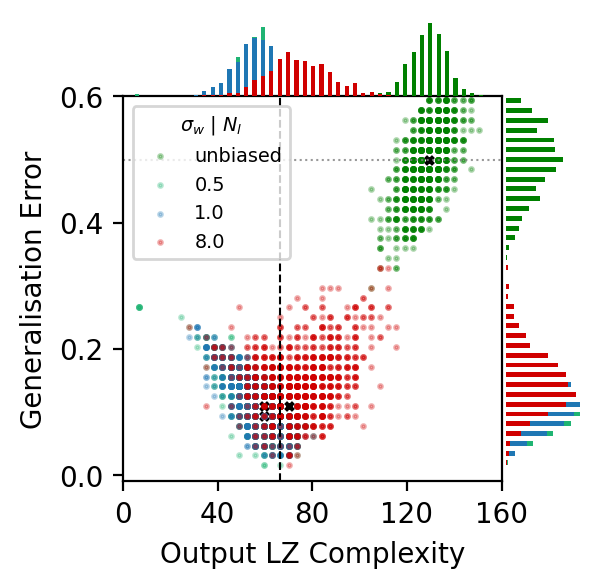}
        \caption{ce, relu}
    \end{subfigure}
    \begin{subfigure}[b]{0.23\columnwidth}
        \includegraphics[width = \textwidth]{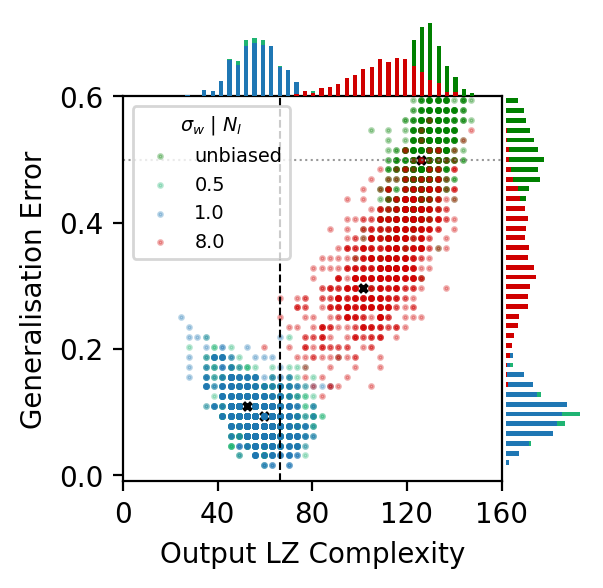}
        \caption{mse, tanh}
    \end{subfigure}
    \begin{subfigure}[b]{0.23\columnwidth}
        \includegraphics[width = \textwidth]{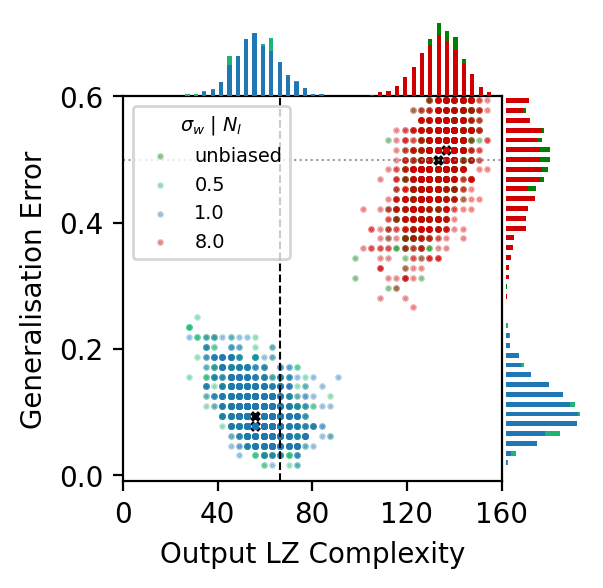}
        \caption{mse, relu}
    \end{subfigure}
    
    \begin{subfigure}[b]{0.23\columnwidth}
        \includegraphics[width = \textwidth]{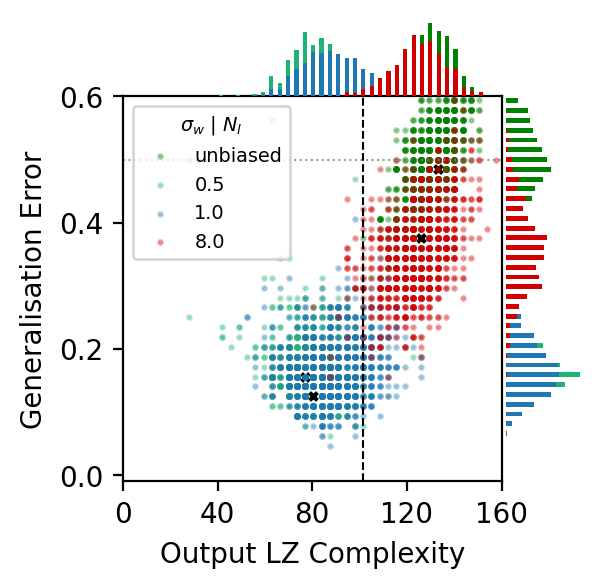}
        \caption{ce, tanh}
    \end{subfigure}
    \begin{subfigure}[b]{0.23\columnwidth}
        \includegraphics[width = \textwidth]{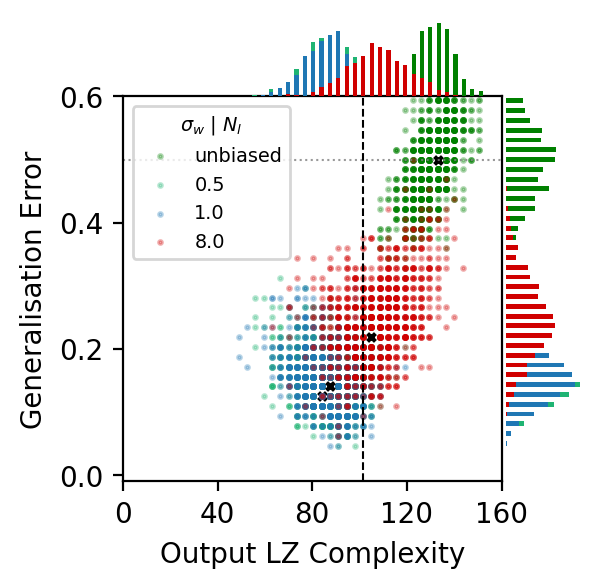}
        \caption{ce, relu}
    \end{subfigure}
    \begin{subfigure}[b]{0.23\columnwidth}
        \includegraphics[width = \textwidth]{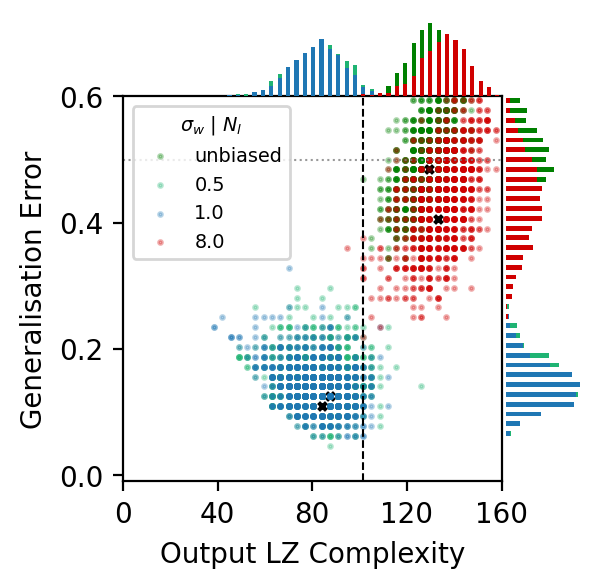}
        \caption{mse, tanh}
    \end{subfigure}
    \begin{subfigure}[b]{0.23\columnwidth}
        \includegraphics[width = \textwidth]{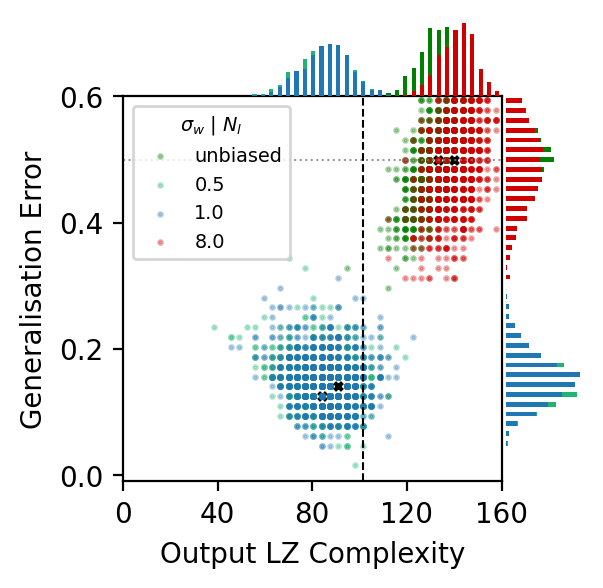}
        \caption{mse, relu}
    \end{subfigure}
    
    \caption{\small {\bf generalization error v.s. complexity scatter plots for different activation functions, loss functions and target function complexities}
    The models used are all 10-layer FCNs on the $n=7$ boolean system with either relu or tanh activations, and initialised with different values of $\sigma_w$. Models are trained with Adam (unlike \cref{fig:Cube_plots_1}(d-f) which use AdvSGD). Training set size $m=64$.
    As in \cref{fig:app:1c_extras}, tanh activations with cross-entropy loss generalise poorly with increasing $\sigma_w$, as their priors are not simplicity biased (the first column). The prior $P(f)$ is not dependent on $\sigma_w$ for ReLU activated networks, and as cross-entropy loss is close to scale-invariant (in that the pre-thresholded targets are at infinity), the scatter plots in the second column are not strongly affected by $\sigma_w$. The third and fourth columns are trained with mse loss, and as discussed in \cref{fig:app:1c_extras}, large $\sigma_w$ will lead to low margin solutions and thus worse generalisation, so both ReLU and tanh activations perform badly.
}\label{fig:app:1d_extras}
\end{figure}

\begin{figure}[H]
\centering

    \begin{subfigure}[b]{0.24\columnwidth}
        \includegraphics[width = \textwidth]{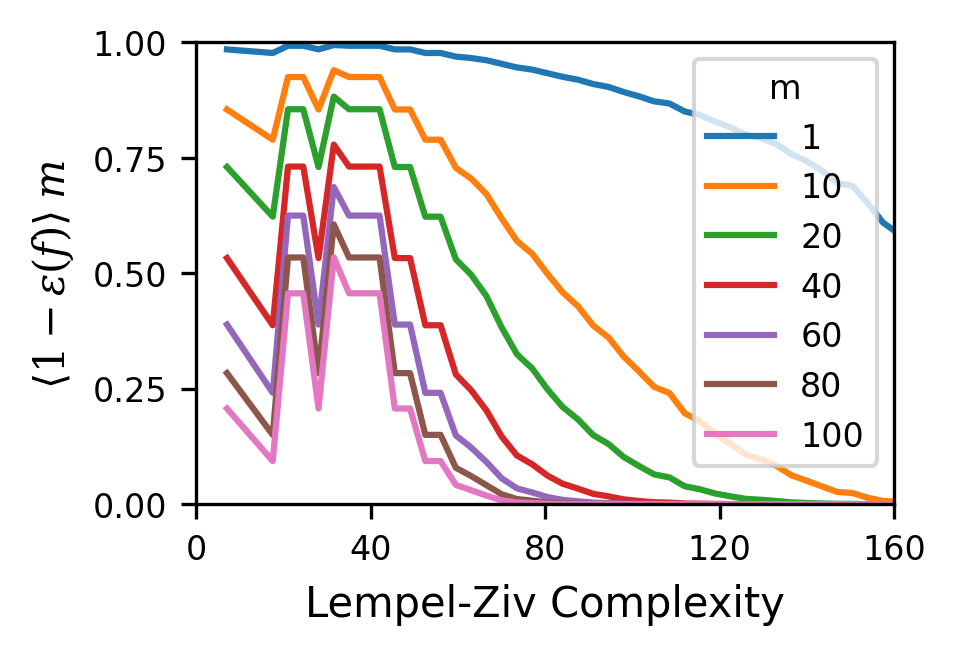}
    \end{subfigure}
    \begin{subfigure}[b]{0.24\columnwidth}
        \includegraphics[width = \textwidth]{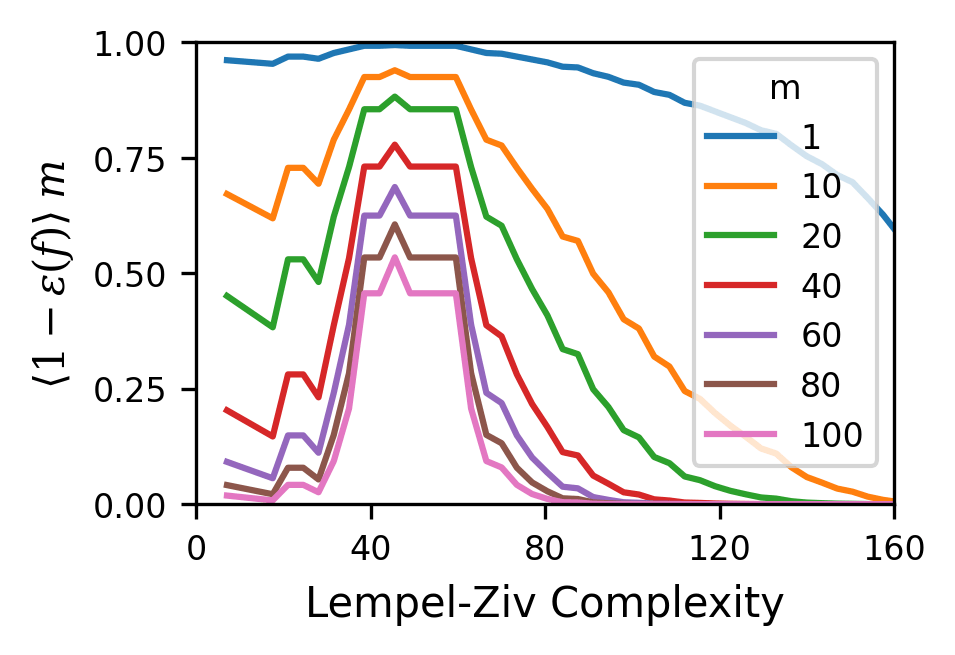}
    \end{subfigure}
    \begin{subfigure}[b]{0.24\columnwidth}
        \includegraphics[width = \textwidth]{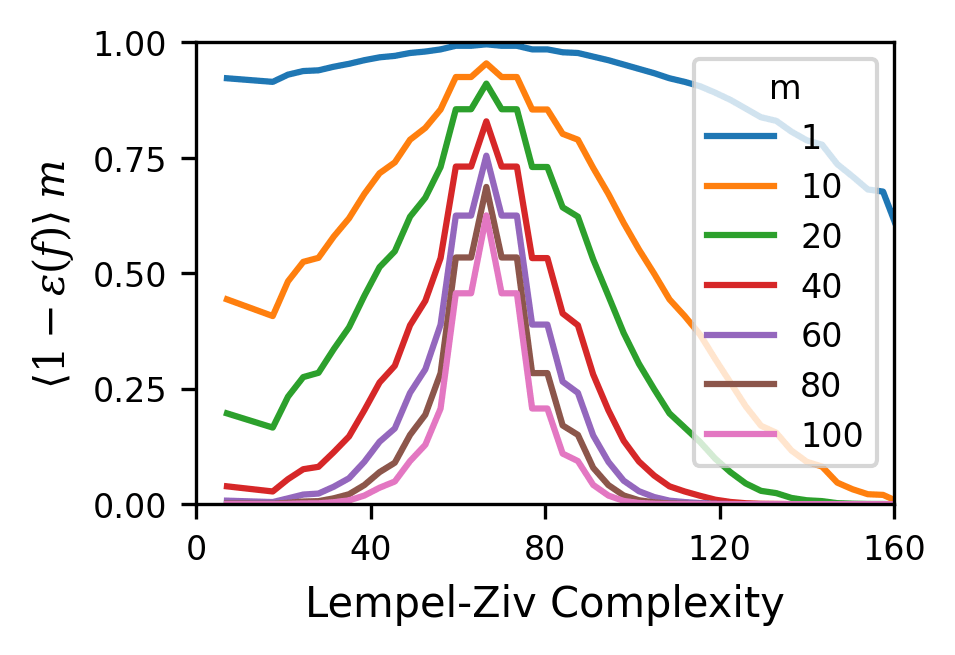}
    \end{subfigure}
    \begin{subfigure}[b]{0.24\columnwidth}
        \includegraphics[width = \textwidth]{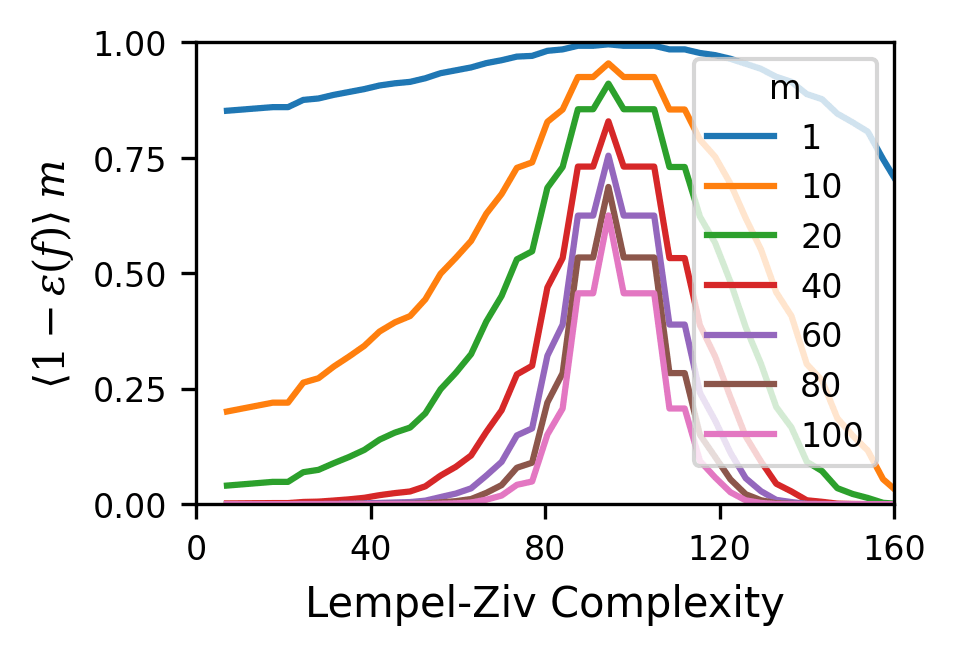}
    \end{subfigure}

    \begin{subfigure}[b]{0.24\columnwidth}
        \includegraphics[width = \textwidth]{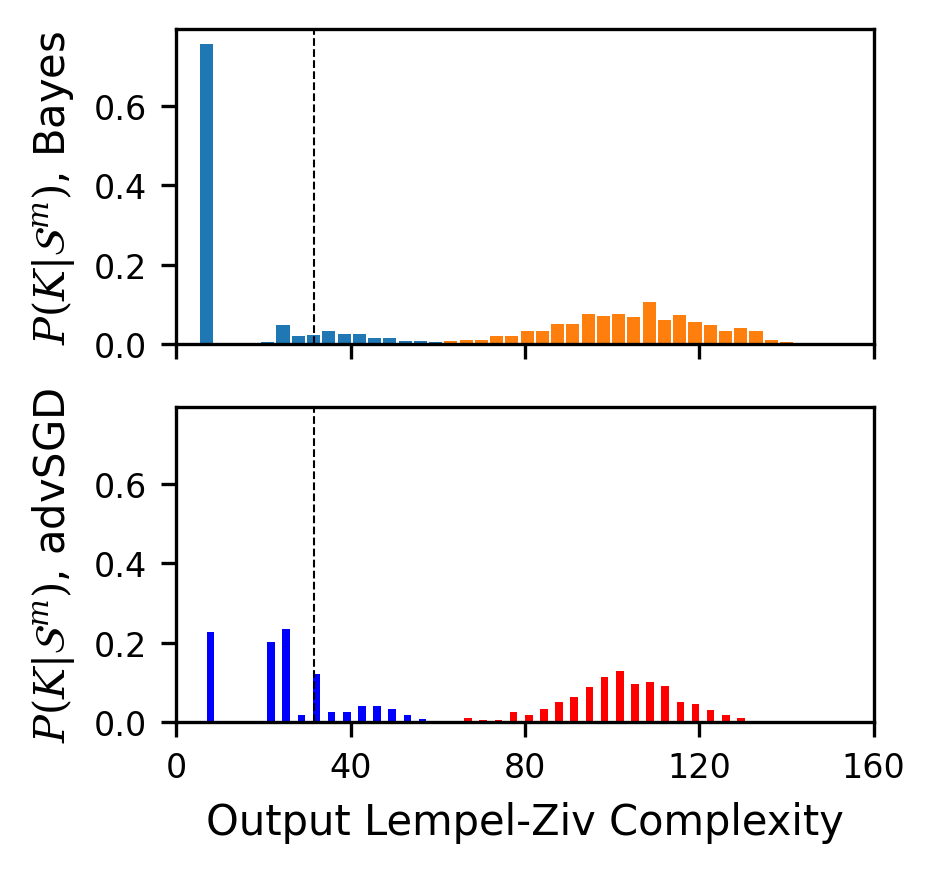}
        \caption{$C_{LZ}(f_t)=$31.5}
    \end{subfigure}
    \begin{subfigure}[b]{0.24\columnwidth}
        \includegraphics[width = \textwidth]{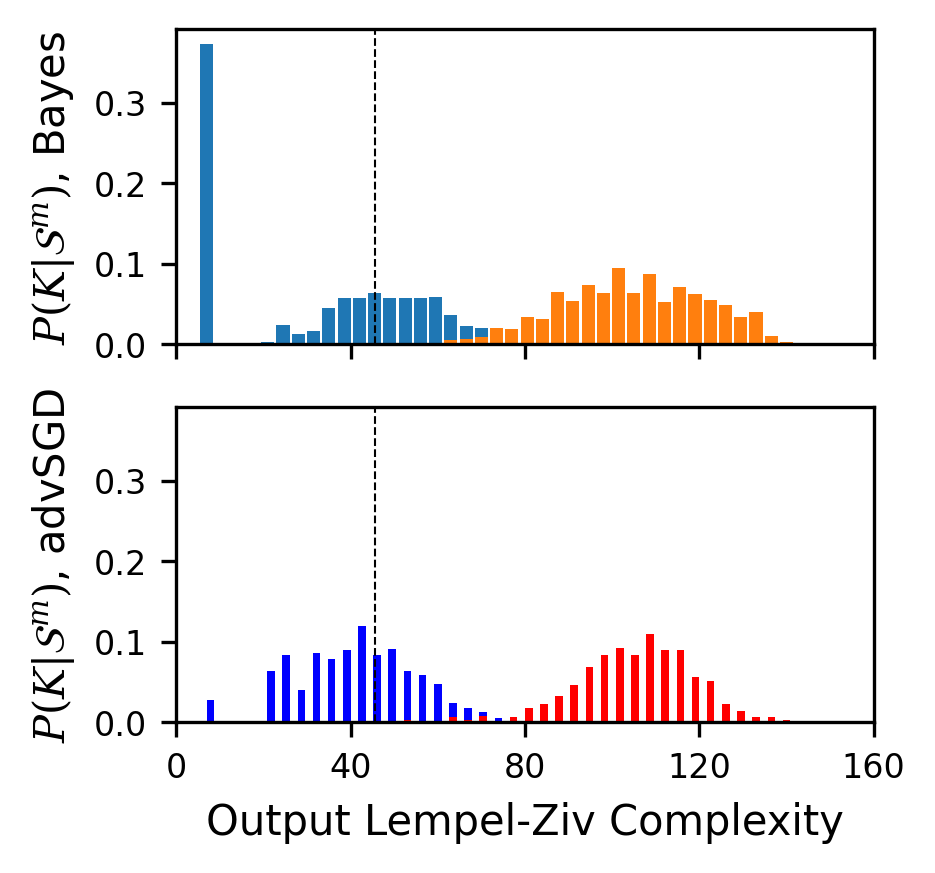}
        \caption{$C_{LZ}(f_t)=$45.5}
    \end{subfigure}
    \begin{subfigure}[b]{0.24\columnwidth}
        \includegraphics[width = \textwidth]{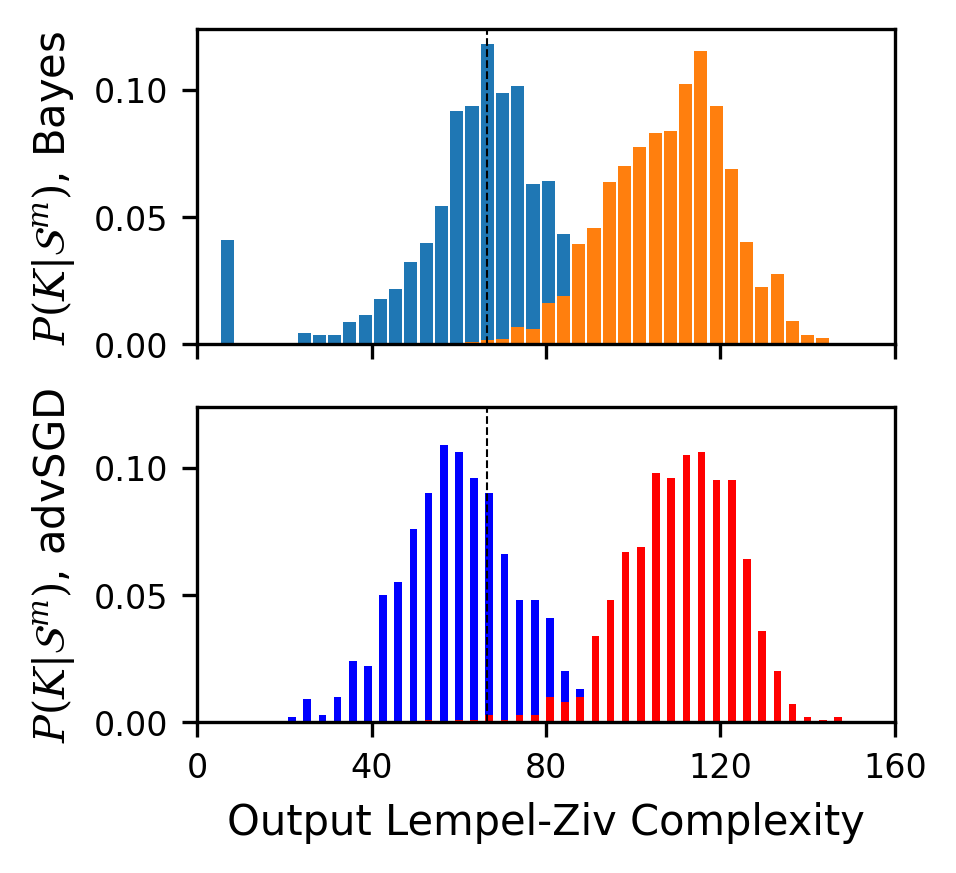}
        \caption{$C_{LZ}(f_t)=$66.5}
    \end{subfigure}
    \begin{subfigure}[b]{0.24\columnwidth}
        \includegraphics[width = \textwidth]{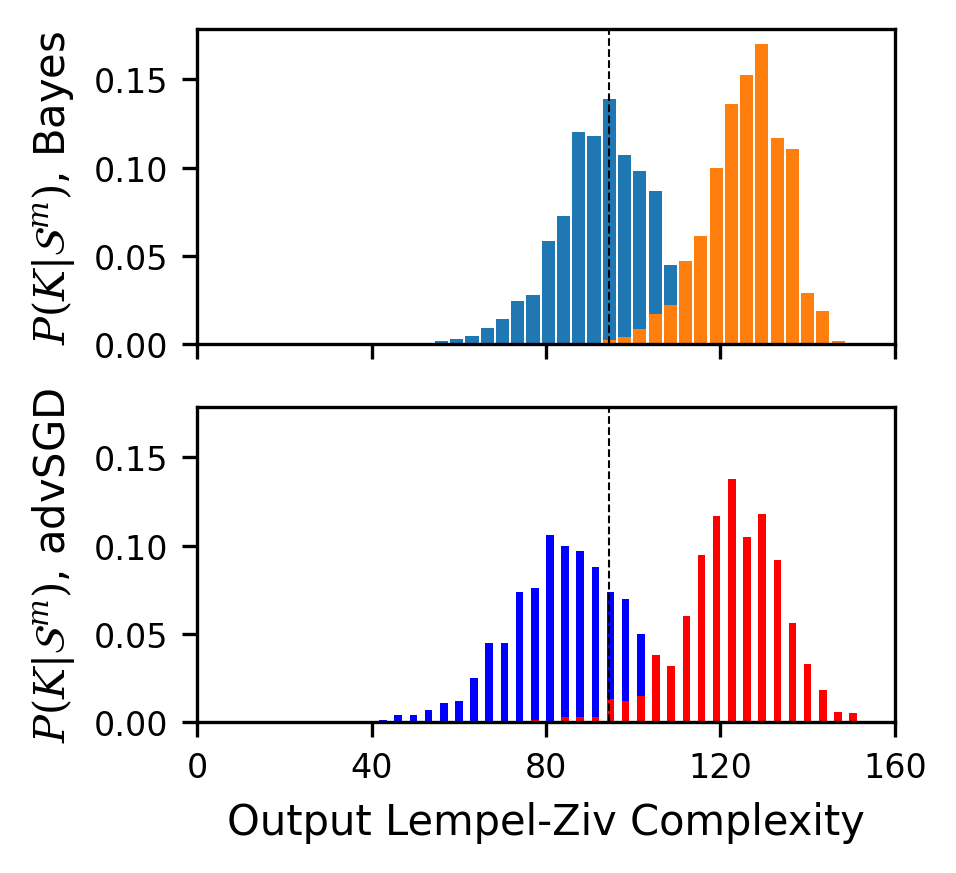}
        \caption{$C_{LZ}(f_t)=$94.5}
    \end{subfigure}
    
    \begin{subfigure}[b]{0.24\columnwidth}
        \includegraphics[width = \textwidth]{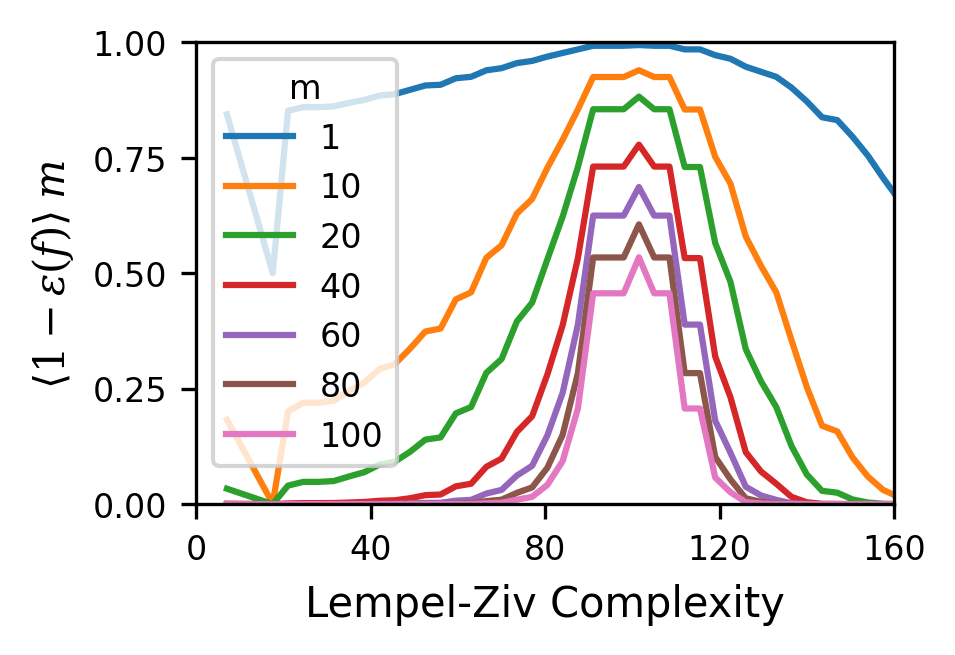}
    \end{subfigure}
    \begin{subfigure}[b]{0.24\columnwidth}
        \includegraphics[width = \textwidth]{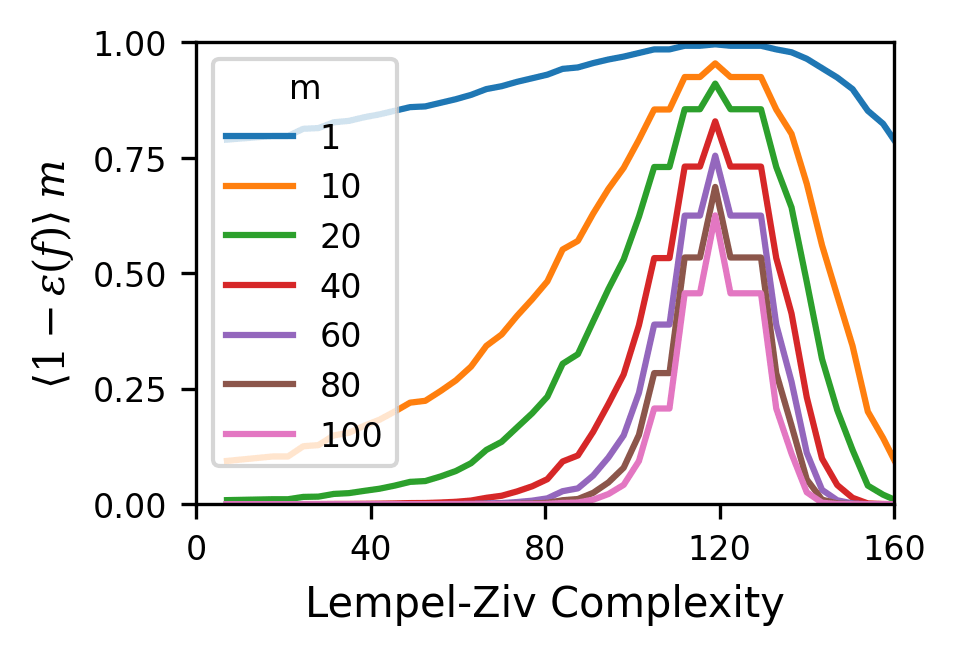}
    \end{subfigure}
    \begin{subfigure}[b]{0.24\columnwidth}
        \includegraphics[width = \textwidth]{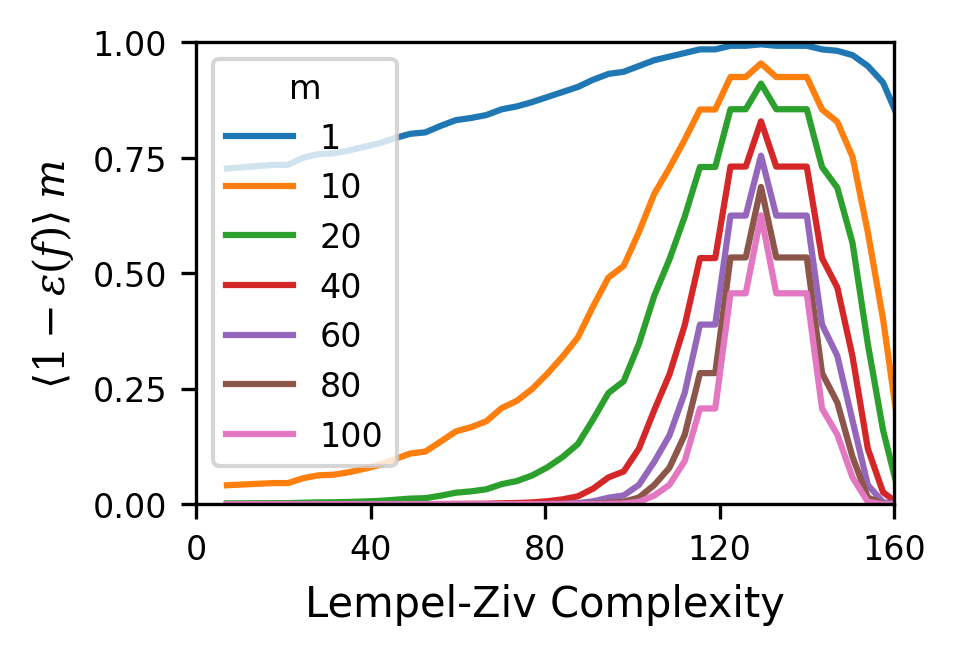}
    \end{subfigure}
    \begin{subfigure}[b]{0.24\columnwidth}
        \includegraphics[width = \textwidth]{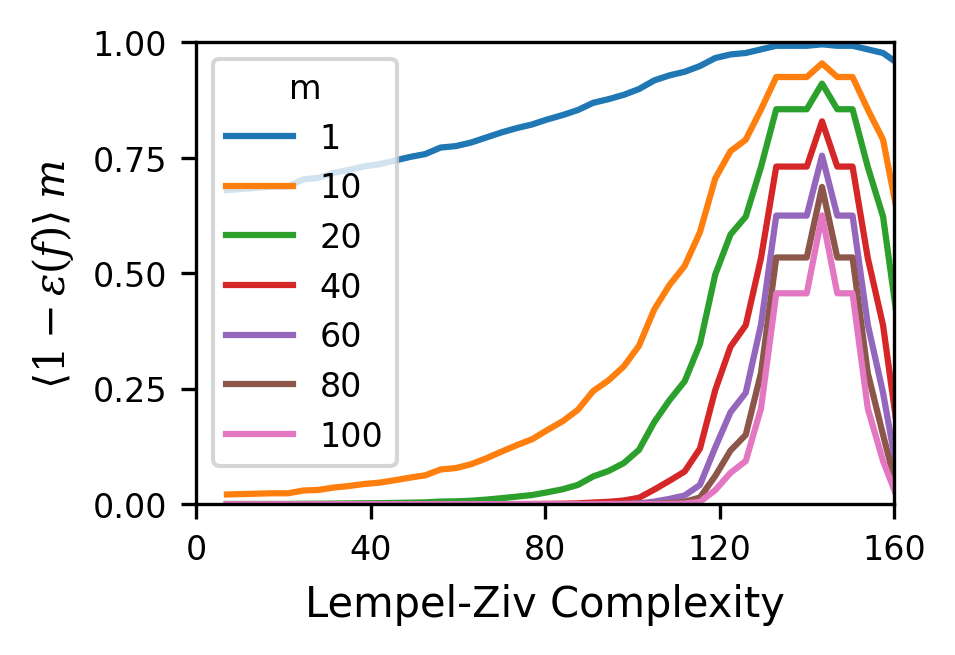}
    \end{subfigure}
    
    \begin{subfigure}[b]{0.24\columnwidth}
        \includegraphics[width = \textwidth]{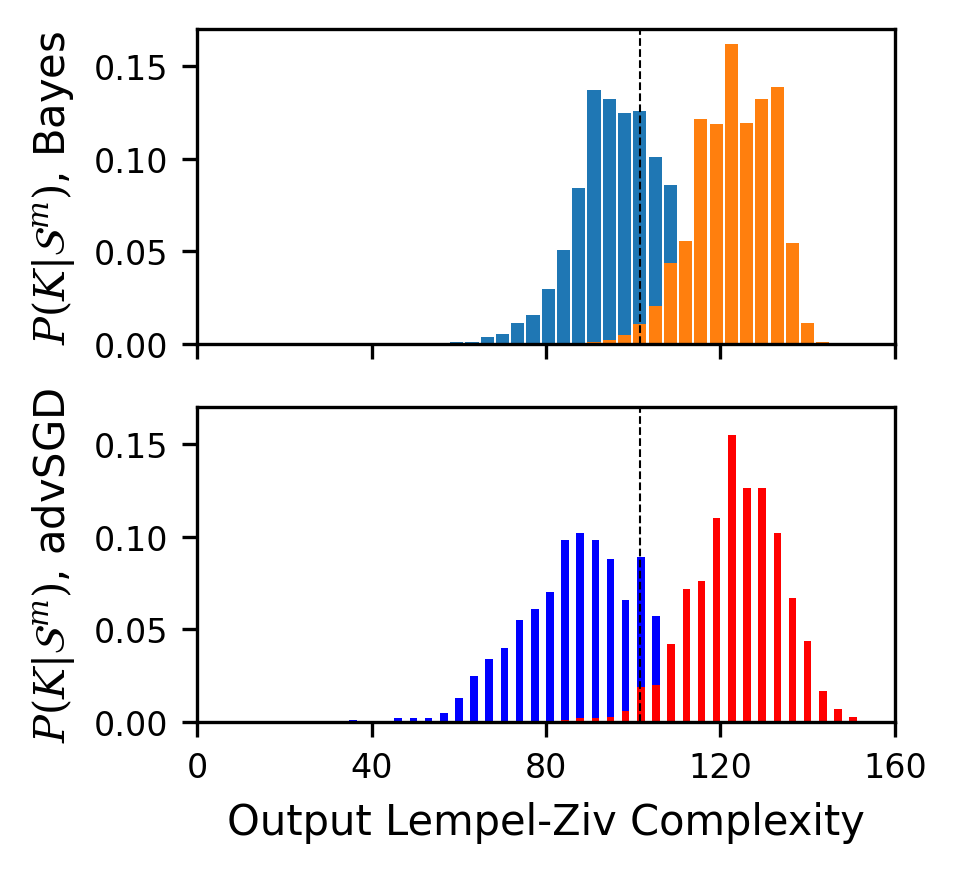}
        \caption{$C_{LZ}(f_t)=$101.5}
    \end{subfigure}
    \begin{subfigure}[b]{0.24\columnwidth}
        \includegraphics[width = \textwidth]{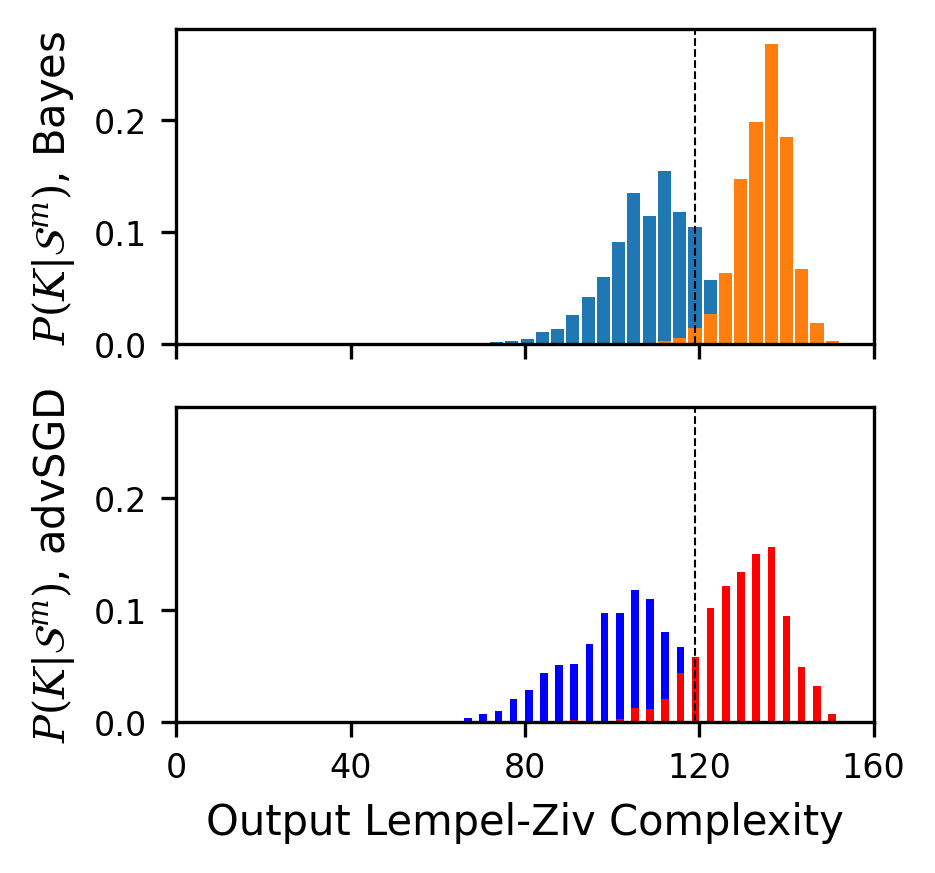}
        \caption{$C_{LZ}(f_t)=$119.0}
    \end{subfigure}
    \begin{subfigure}[b]{0.24\columnwidth}
        \includegraphics[width = \textwidth]{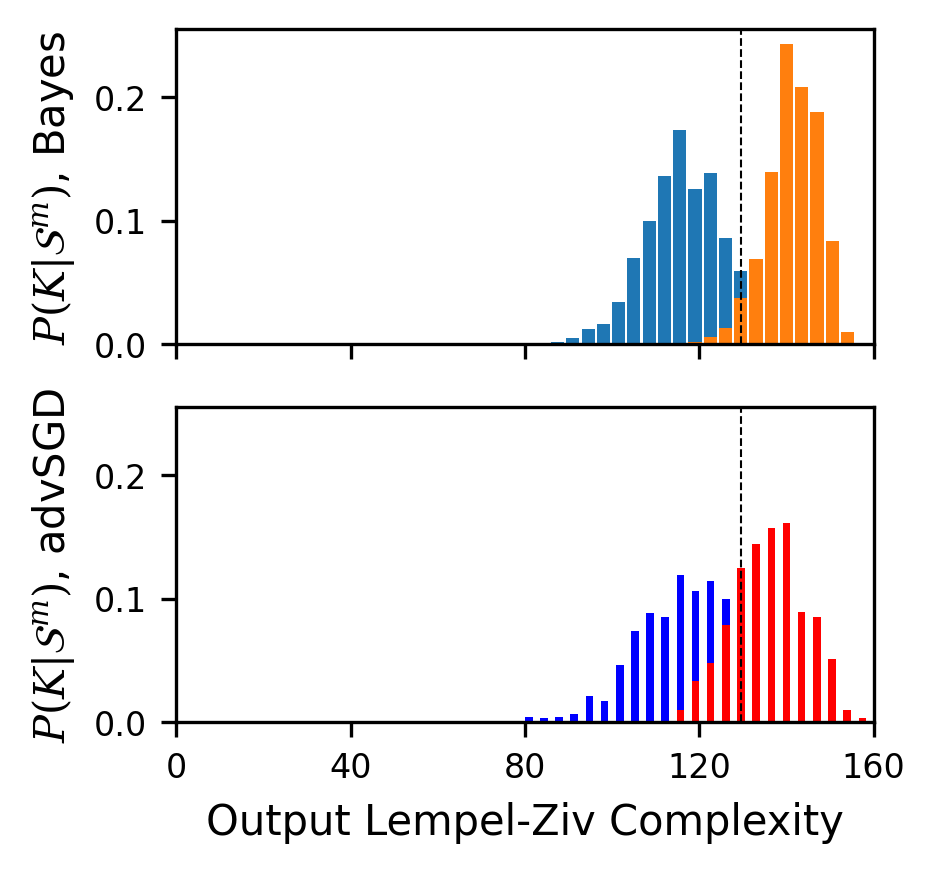}
        \caption{$C_{LZ}(f_t)=$129.5}
    \end{subfigure}
    \begin{subfigure}[b]{0.24\columnwidth}
        \includegraphics[width = \textwidth]{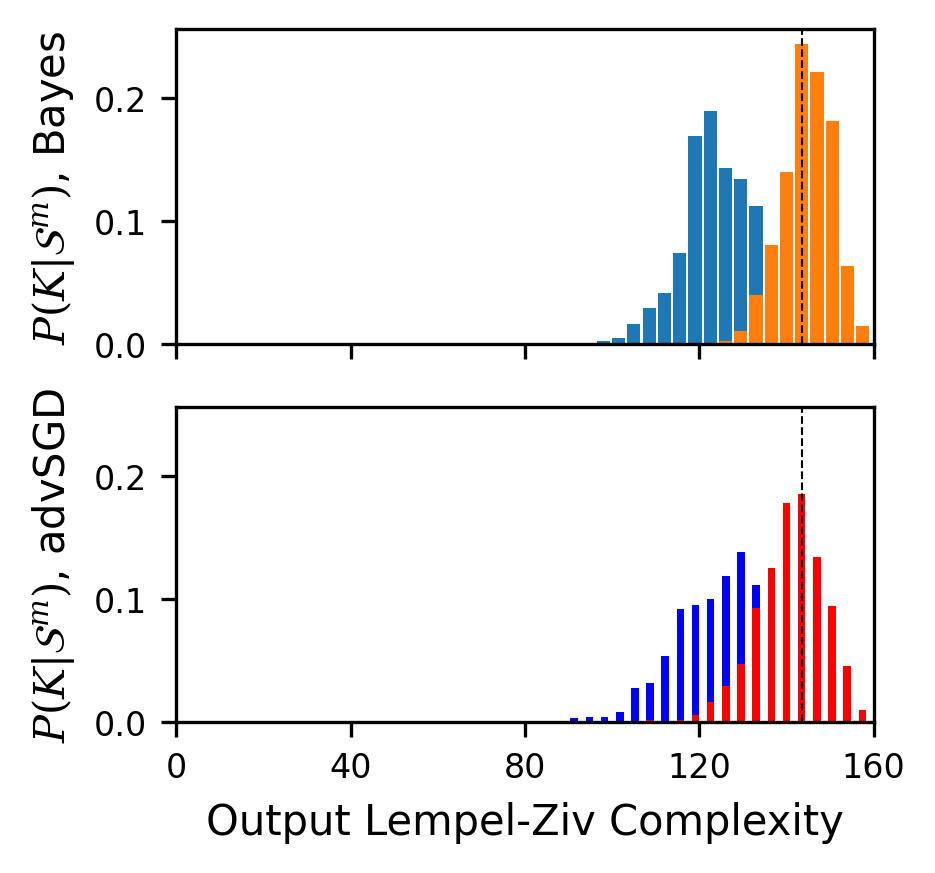}
        \caption{$C_{LZ}(f_t)=$143.5}
    \end{subfigure}
    
    \caption{\small {\bf Comparing  posteriors calculated with the decoupling approximation of  \cref{eq:PBapprox} to  directly measured SGD posteriors for a range of complexities.}
This figure complements the main text \cref{fig:Cube_plots_2} by showing a wider range of complexities. 
\textbf{Top row}: The mean likelihood   $\langle(1 - \epsilon(f))^m\rangle_5$ from \cref{eq:PBapprox}, averaged over training sets, and also over the 5 lowest error functions at each LZ complexity $K$ for four different target function complexities in (a)($K=31.5$) -- (d)($K 94.5$). This property is independent of the architecture. It typically peaks around the target function complexity.
\textbf{Middle two rows} 
The top and bottom histograms  compare the posterior probability of obtaining functions of complexity $K$ from  \cref{eq:PBapprox} (lighter blue and orange) to the posteriors calculated from independent initializations of SGD (darker blue and red) averaged over training sets of size $m=64$.  As can be seen, the predictions and the measurements agree remarkably well.
\textbf{Bottom 3 rows} These are the same plots as the top 3 rows, but now for complexities $K=101.5$ - $K=143.5$ ((a) to (d)).
   } \label{app:fig:error_func}
\end{figure}

\begin{figure}[H]
\centering
    \begin{subfigure}[ht]{0.3\linewidth}
        \includegraphics[width=\textwidth]{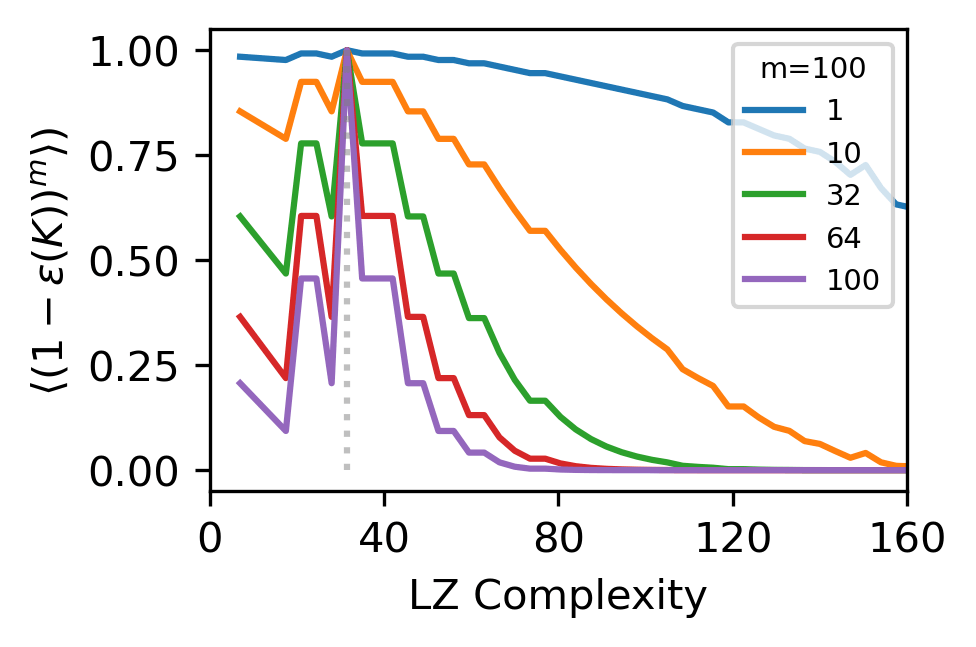}
    \end{subfigure}
    \begin{subfigure}[ht]{0.3\linewidth}
        \includegraphics[width=\textwidth]{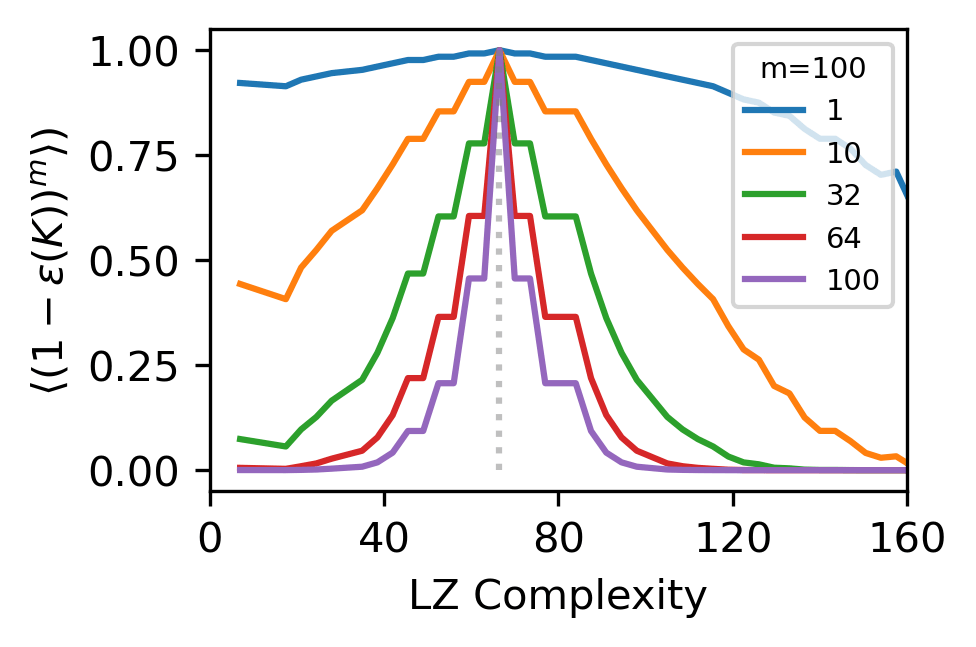}
    \end{subfigure}
    \begin{subfigure}[ht]{0.3\linewidth}
        \includegraphics[width=\textwidth]{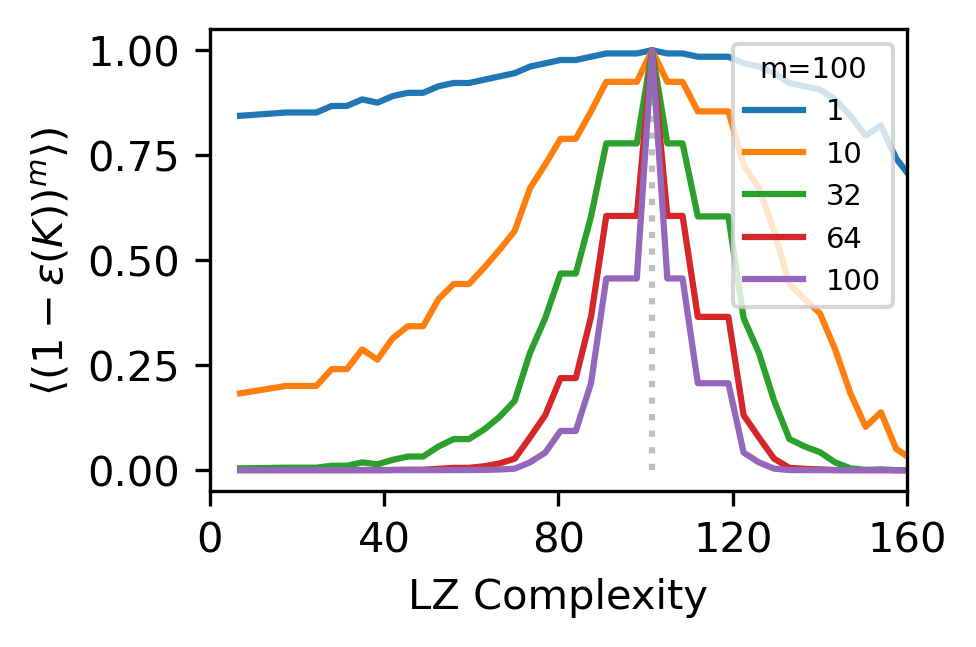}
    \end{subfigure}
    
    \begin{subfigure}[ht]{0.3\linewidth}
        \includegraphics[width=\textwidth]{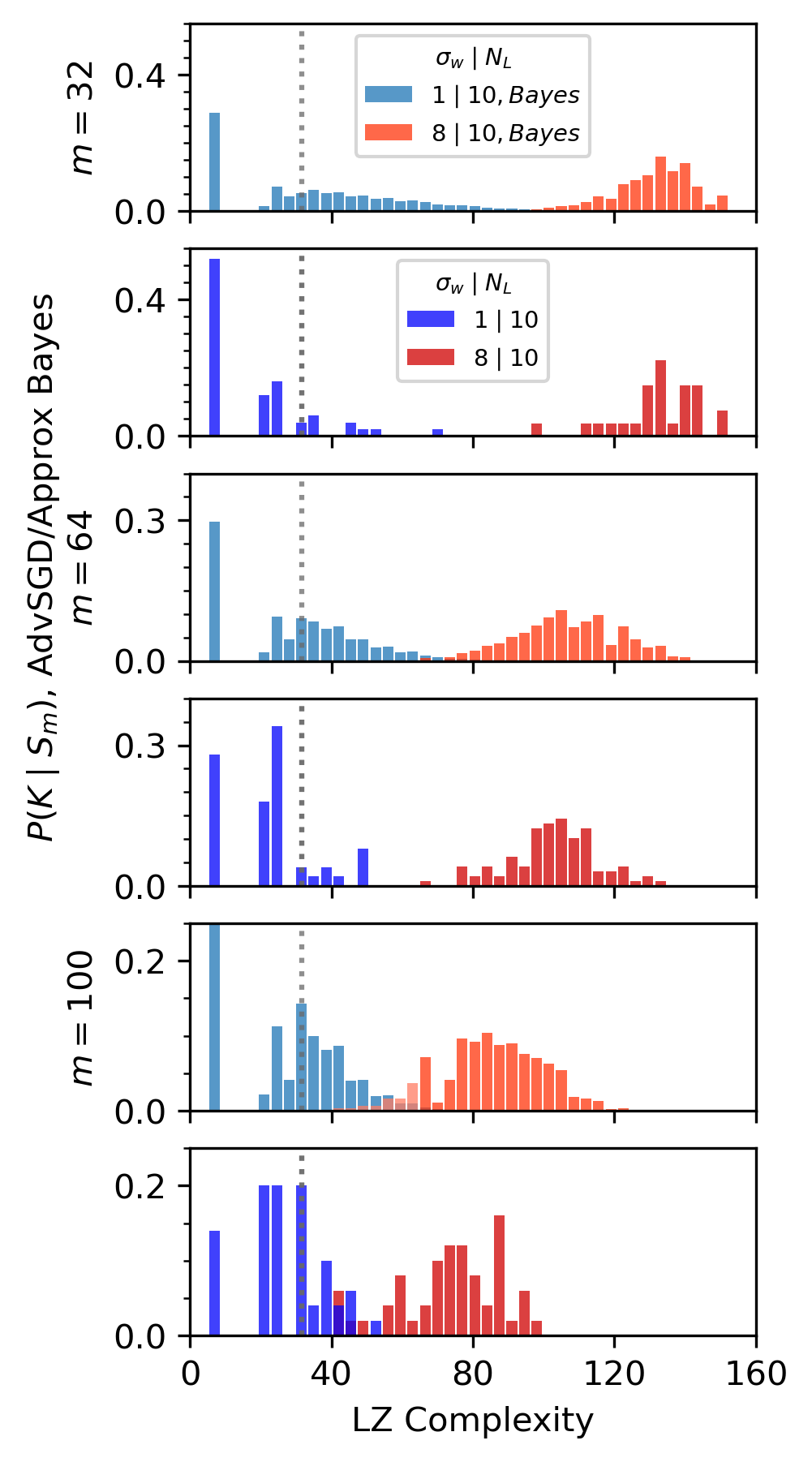}\caption{k=1, LZ=31.5}
    \end{subfigure}
    \begin{subfigure}[ht]{0.3\linewidth}
        \includegraphics[width=\textwidth]{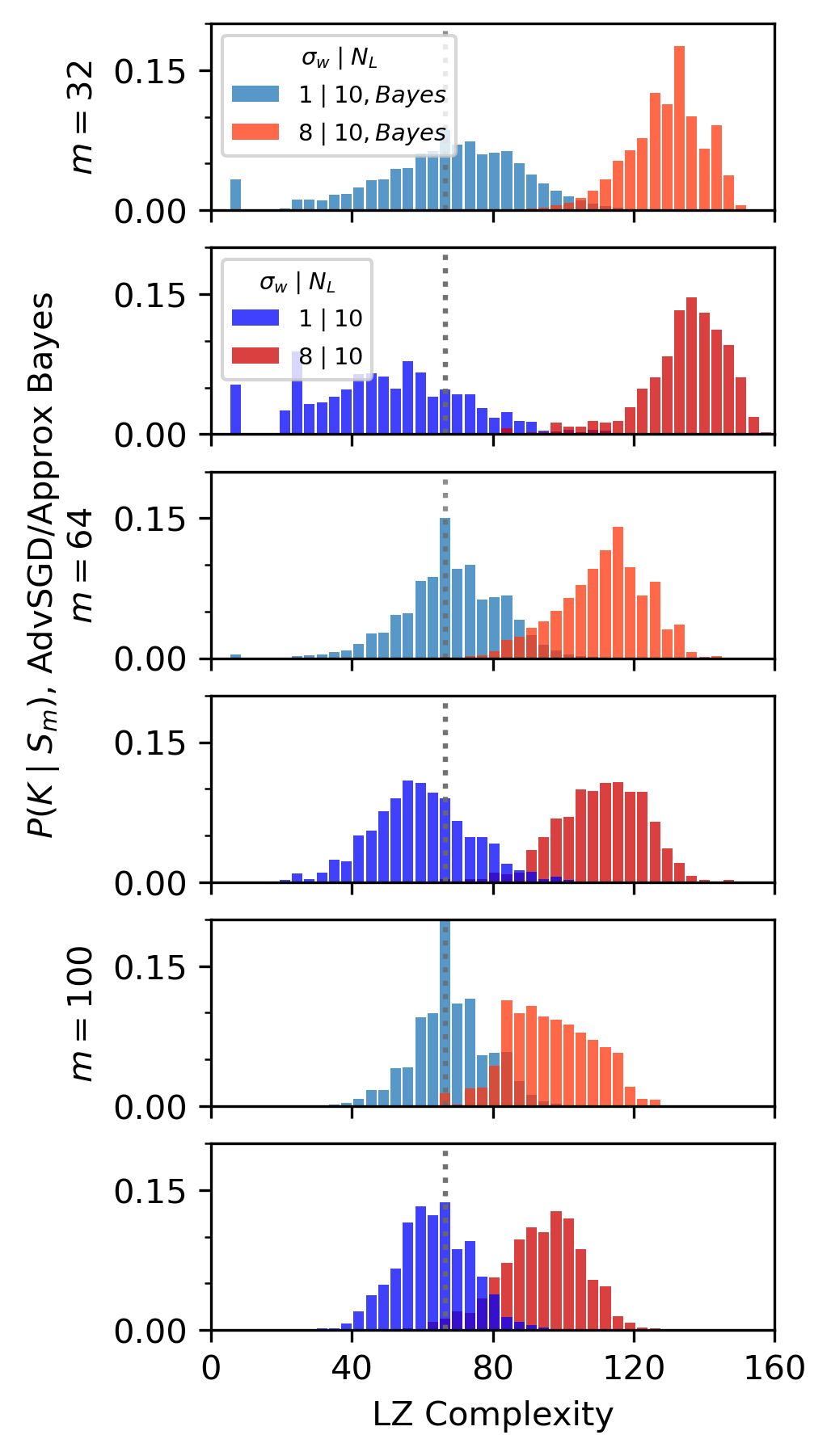}\caption{k=1, LZ=66.5}
    \end{subfigure}
    \begin{subfigure}[ht]{0.3\linewidth}
        \includegraphics[width=\textwidth]{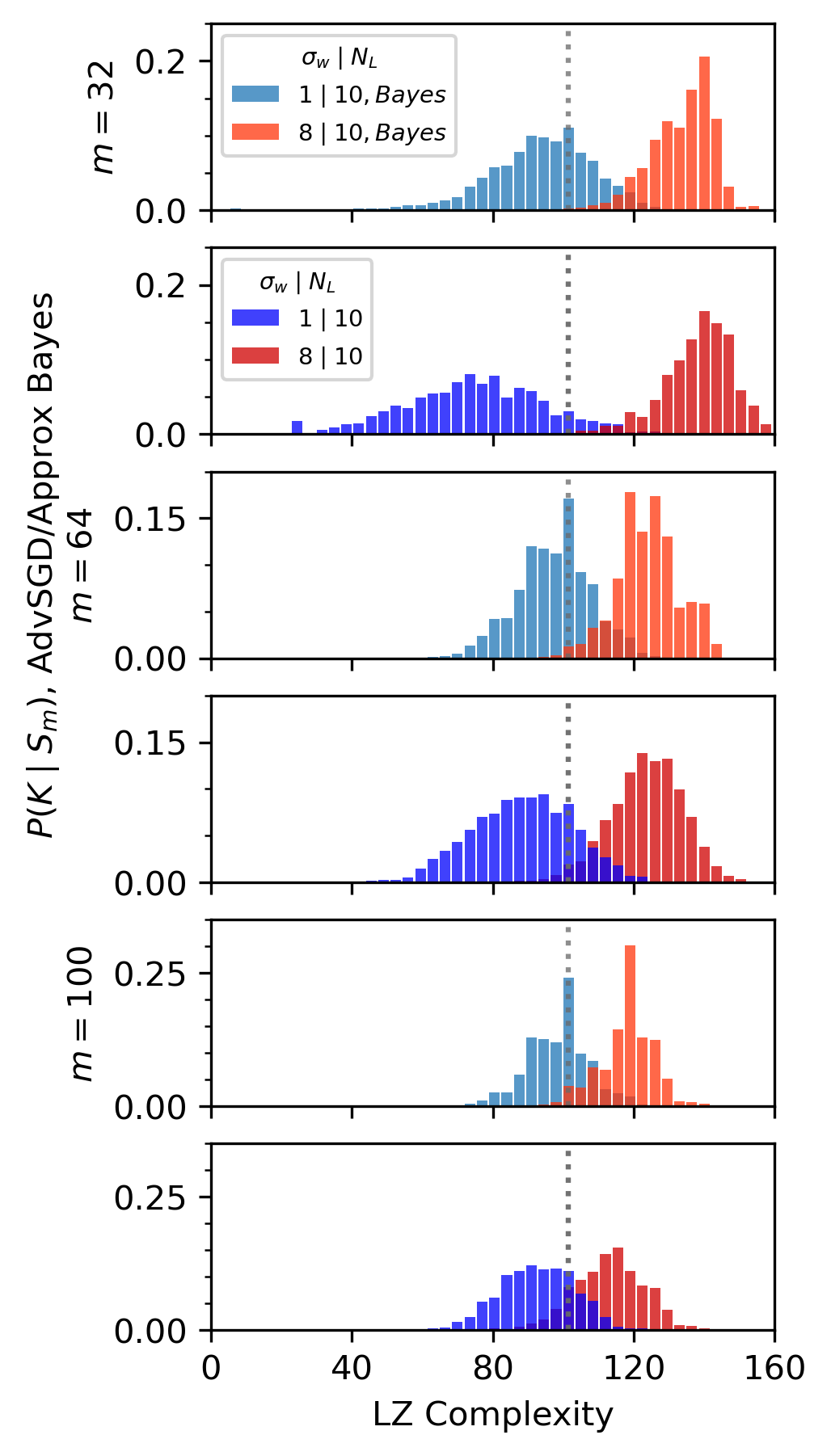}\caption{k=1, LZ=101.5}
    \end{subfigure} 
    
    \caption{  \small {\bf \small   Comparing  posteriors calculated with the decoupling approximation of  \cref{eq:PBapprox} to  directly measured SGD posteriors, but for $l=1$ functions.
    }The same experiment as \cref{fig:Cube_plots_2} but taking only the most probable function per LZ complexity to calculate the likelihood (l=1 in \cref{app:exp_details:approx_bayes}). 
As can be seen by comparing to \cref{fig:Cube_plots_2},  \cref{app:fig:error_func}, and \cref{app:fig:k=50_approx_bayes}, the choice of the number of top functions to include in the likelihood estimation does not have a strong qualitative effect on the predicted posteriors. }\label{app:fig:k=1_approx_bayes}
\end{figure} 

\begin{figure}[H]
    \centering
        
    \begin{subfigure}[ht]{0.3\linewidth}
        \includegraphics[width=\textwidth]{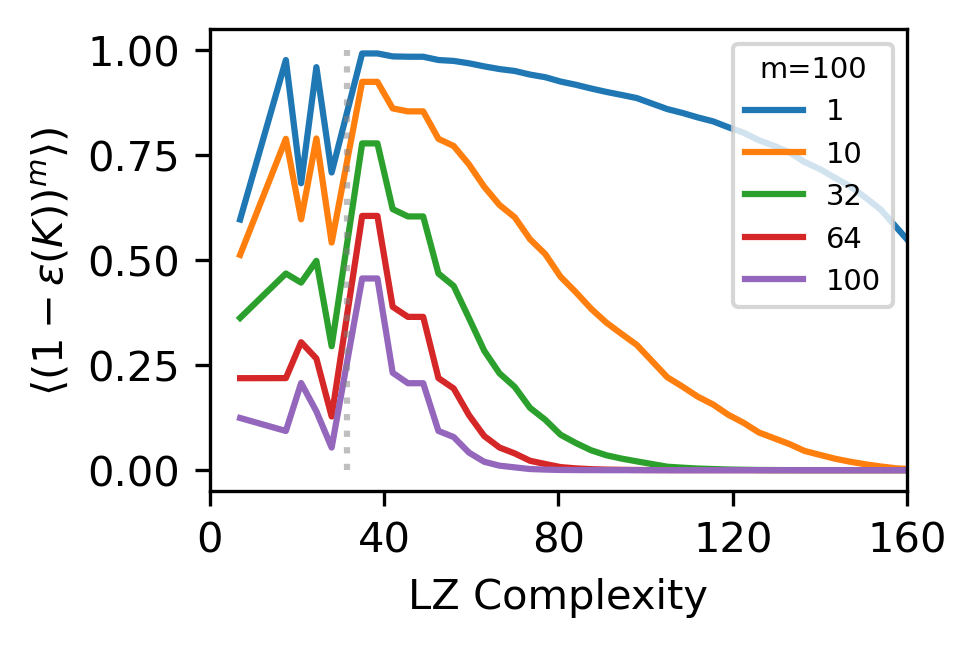}
    \end{subfigure}
    \begin{subfigure}[ht]{0.3\linewidth}
        \includegraphics[width=\textwidth]{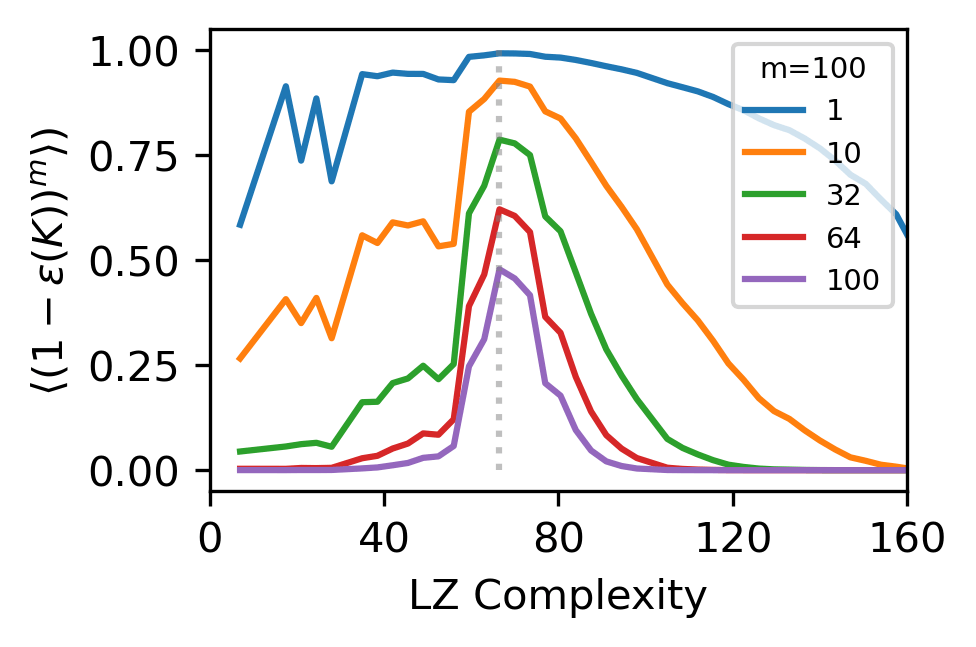}
    \end{subfigure}
    \begin{subfigure}[ht]{0.3\linewidth}
        \includegraphics[width=\textwidth]{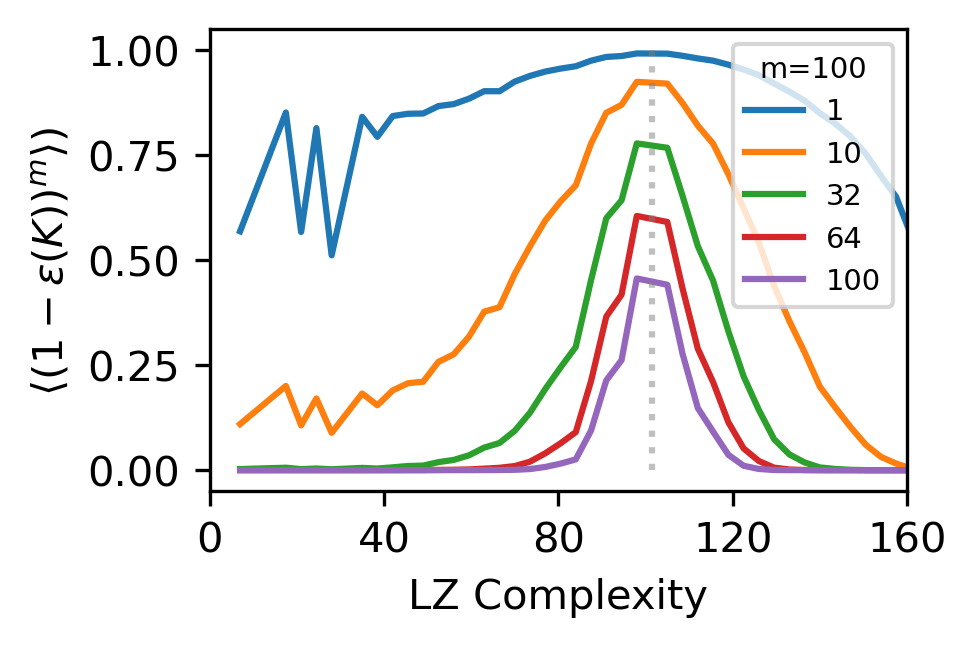}
    \end{subfigure} 
    
        \begin{subfigure}[ht]{0.3\linewidth}
        \includegraphics[width=\textwidth]{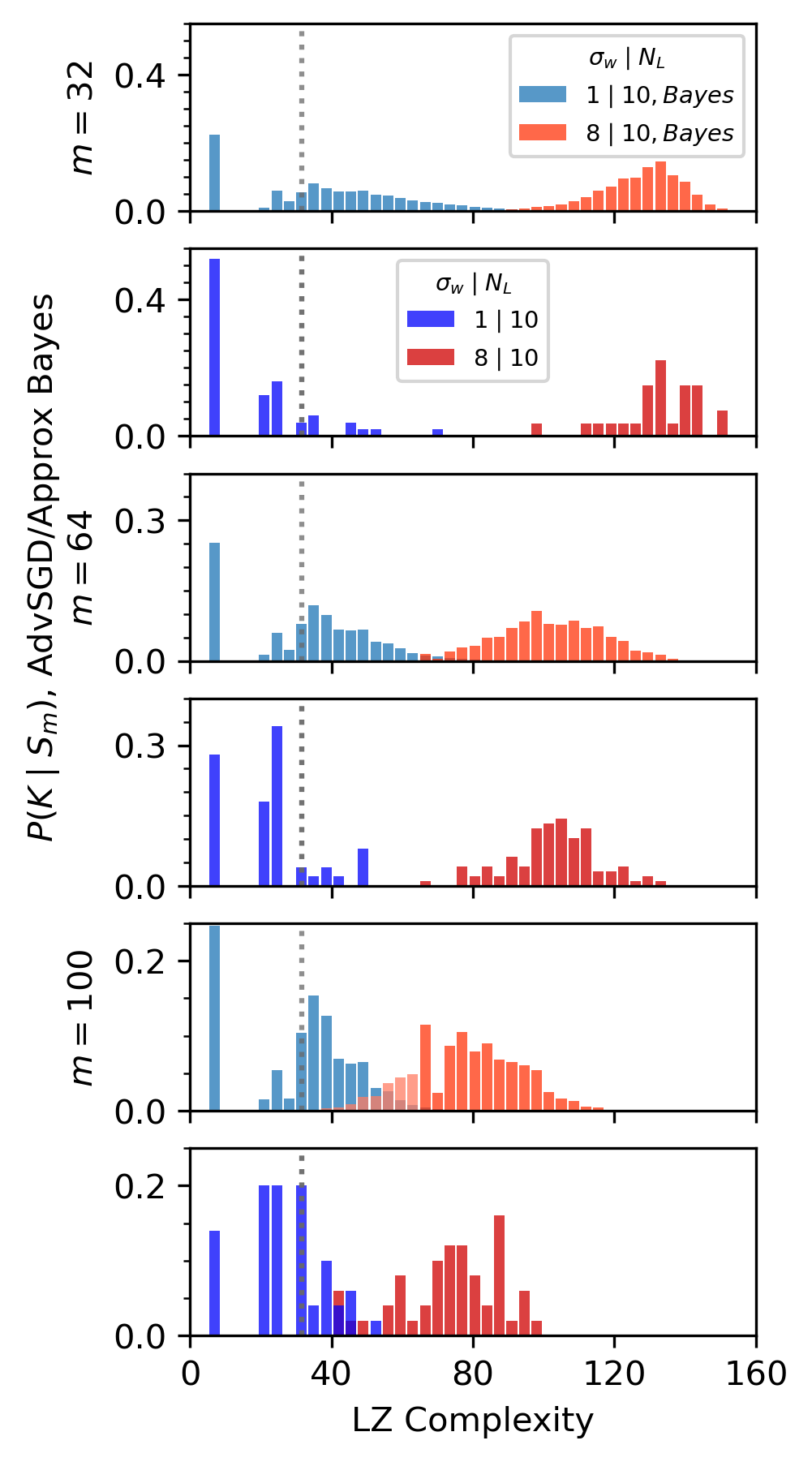}\caption{k=50, LZ=31.5}
    \end{subfigure}
    \begin{subfigure}[ht]{0.3\linewidth}
        \includegraphics[width=\textwidth]{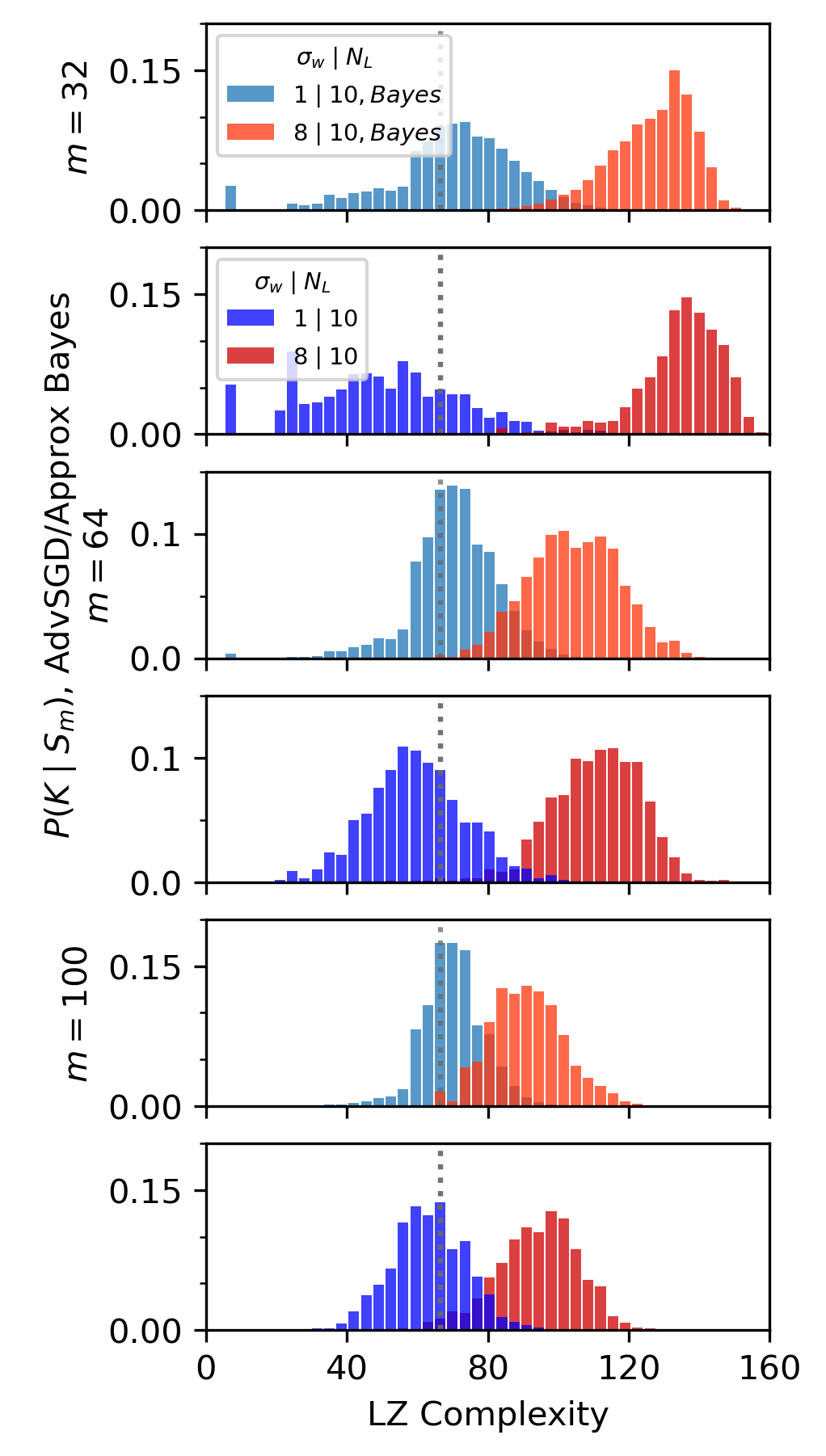}\caption{k=50, LZ=66.5}
    \end{subfigure}
    \begin{subfigure}[ht]{0.3\linewidth}
        \includegraphics[width=\textwidth]{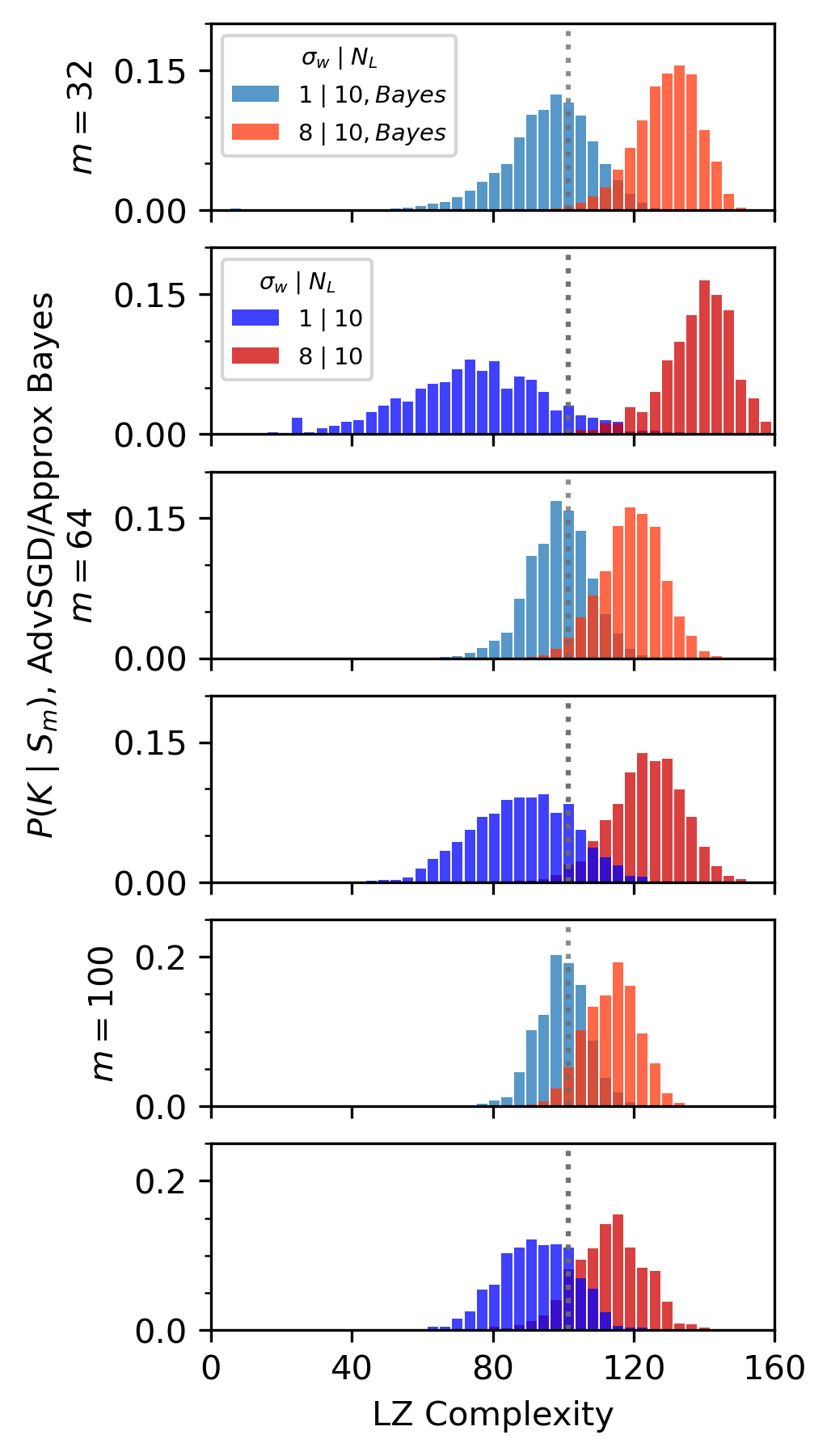}\caption{k=50, LZ=101.5}
    \end{subfigure} 

    \caption{  \small {\bf \small  Comparing  posteriors calculated with the decoupling approximation of  \cref{eq:PBapprox} to  directly measured SGD posteriors, but for $l=50$ functions.
    } The same experiment as \cref{fig:Cube_plots_2} in the main text,  but taking the 50 most probable function per LZ complexity to calculate the likelihood (l=50 in \cref{app:exp_details:approx_bayes}). 
 As can be seen by comparing to \cref{fig:Cube_plots_2},  \cref{app:fig:error_func}, and \cref{app:fig:k=1_approx_bayes}, the choice of the number of top functions to include in the likelihood estimation does not have a strong qualitative effect on the predicted posteriors}\label{app:fig:k=50_approx_bayes}
\end{figure}

\begin{figure}[H]
    \centering
    \makebox[\columnwidth][c]{\includegraphics[width=1\columnwidth]{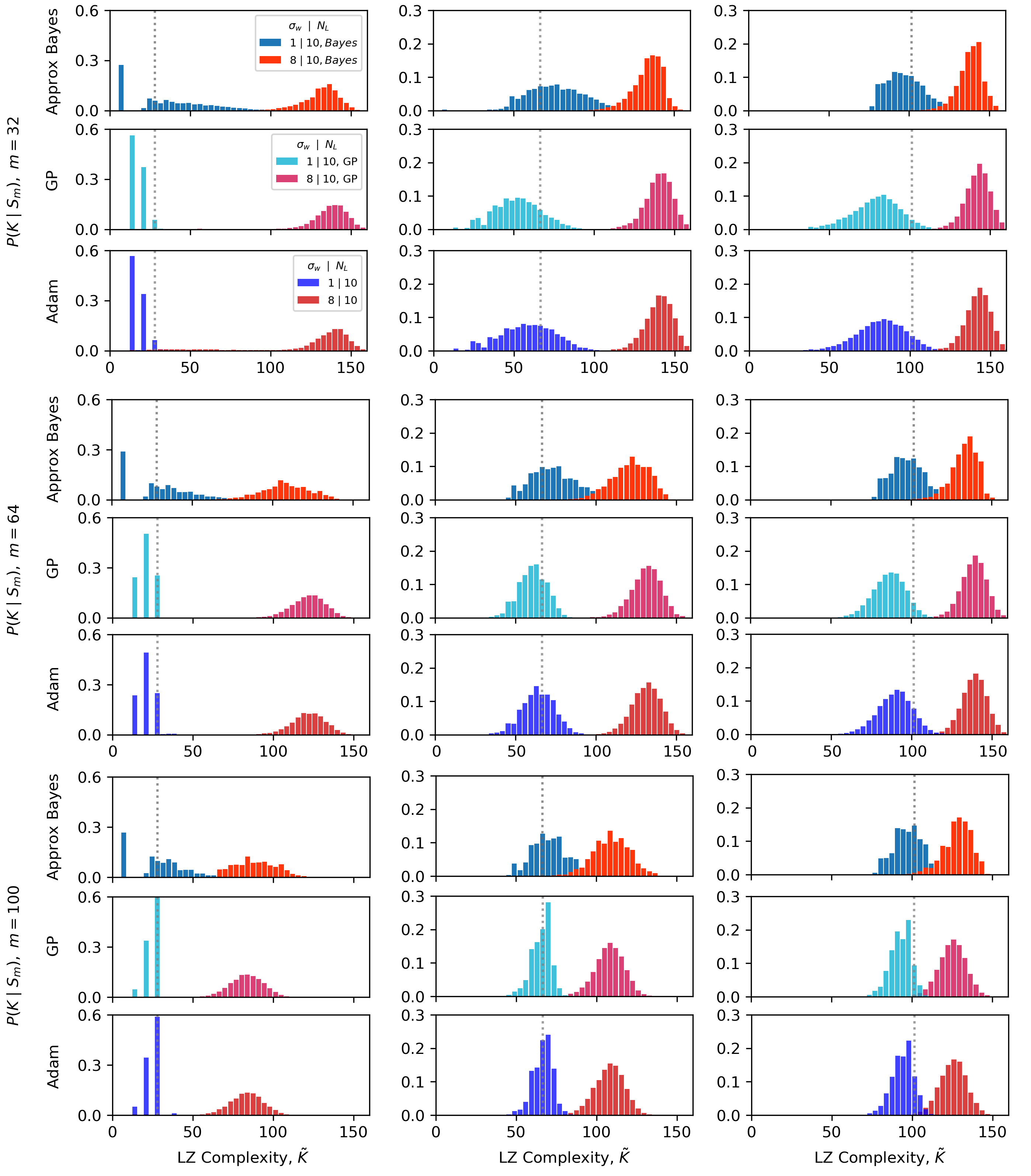}}
    \caption{\small {\bf \small
    Comparing posteriors calculated with the decoupling approximation of  \cref{eq:PBapprox} to  directly measured SGD posteriors and to Gaussian Process (GP) posteriors, but with MSE loss instead of CE loss.}
Left to right are target function complexities $LZ=31.5, LZ=66.5$ and $LZ=101.5$ respectively.  Top to bottom are for training set sizes $m=32, m=64$ and $m=100$ respectively. In each set, the prediction from \cref{eq:PBapprox} is at the top, the Bayesian prediction from the GP is in the middle, and the prediction from SGD  on the bottom row.  In this case, SGD means Adam with learning rate $5\times10^{-4}$ and batch size $32$ on an FCN with 10 layers and layer widths 64 was used.  The GP approximation exploits the correspondence between very wide-layer DNNs and GPs. A DNN with all frozen layers (bar the final classification layer) and width 16384 was trained with SGD with an MSE objective function (until zero loss, where the correspondence with GPs becomes exact).
As the plots show, using MSE loss, or the GP approximation, does not significantly change  the agreement with  \cref{eq:PBapprox}.
    }\label{fig:app:gp_nn_2}
\end{figure}

\begin{figure}[H]
    \centering
    \begin{subfigure}[b]{0.23\columnwidth}
        \includegraphics[width = \textwidth]{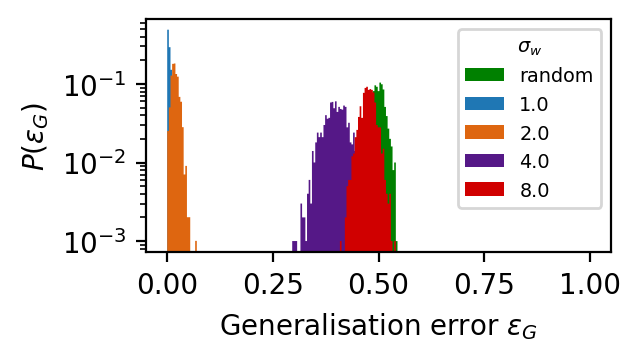}
        \caption{$m=1024$, $f=10\dots$}
    \end{subfigure}
    \begin{subfigure}[b]{0.23\columnwidth}
        \includegraphics[width = \textwidth]{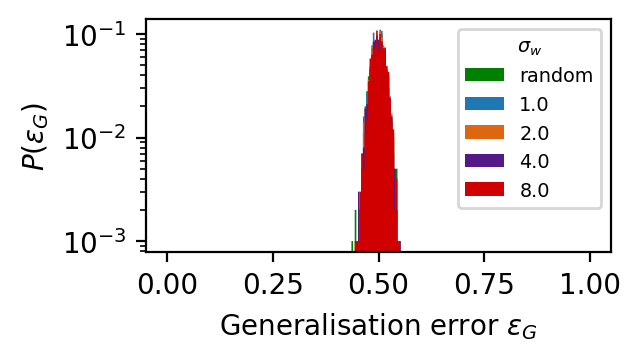}
        \caption{$m=1024$, $f=3-$parity}
    \end{subfigure}
    \begin{subfigure}[b]{0.23\columnwidth}
        \includegraphics[width = \textwidth]{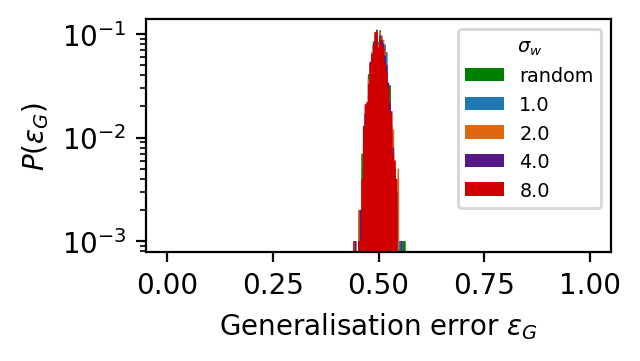}
        \caption{$m=1024$, $f=$ parity}
    \end{subfigure}
    \begin{subfigure}[b]{0.23\columnwidth}
        \includegraphics[width = \textwidth]{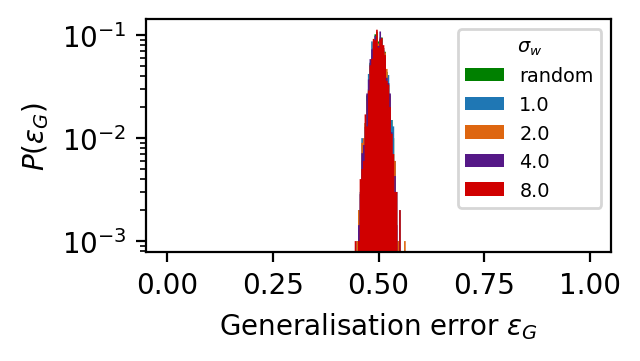}
        \caption{$m=1024$, $f=$ random}
    \end{subfigure}
    
    \begin{subfigure}[b]{0.23\columnwidth}
        \includegraphics[width = \textwidth]{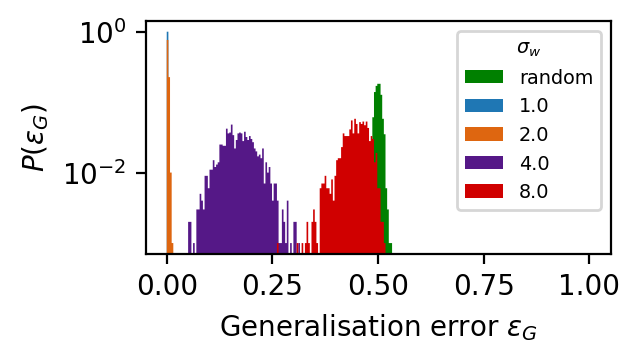}
        \caption{$m=4096$, $f=10\dots$}
    \end{subfigure}
    \begin{subfigure}[b]{0.23\columnwidth}
        \includegraphics[width = \textwidth]{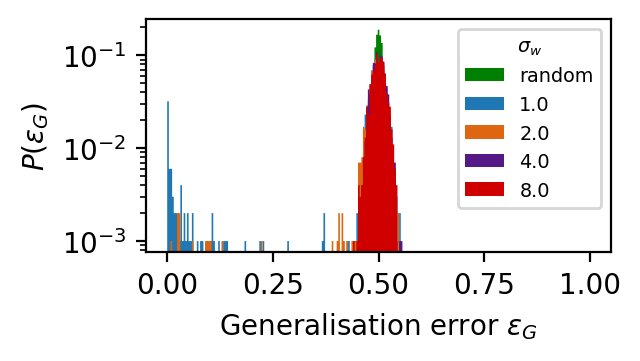}
        \caption{$m=4096$, $f=3-$parity}
    \end{subfigure}
    \begin{subfigure}[b]{0.23\columnwidth}
        \includegraphics[width = \textwidth]{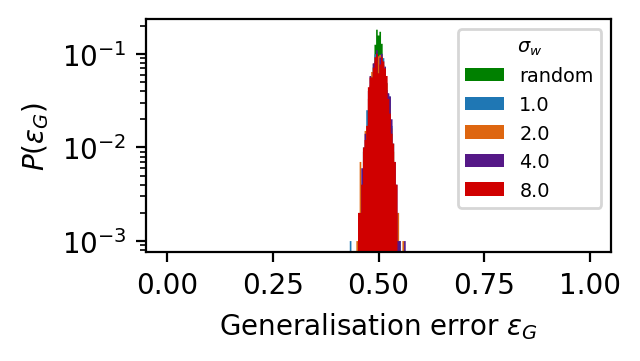}
        \caption{$m=4096$, $f=$ parity}
    \end{subfigure}
    \begin{subfigure}[b]{0.23\columnwidth}
        \includegraphics[width = \textwidth]{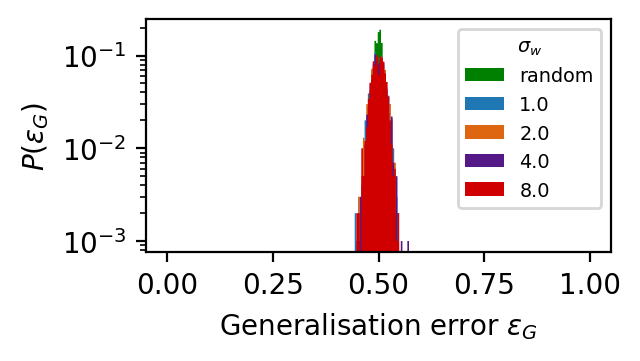}
        \caption{$m=4096$, $f=$ random}
    \end{subfigure}
    
    \begin{subfigure}[b]{0.23\columnwidth}
        \includegraphics[width = \textwidth]{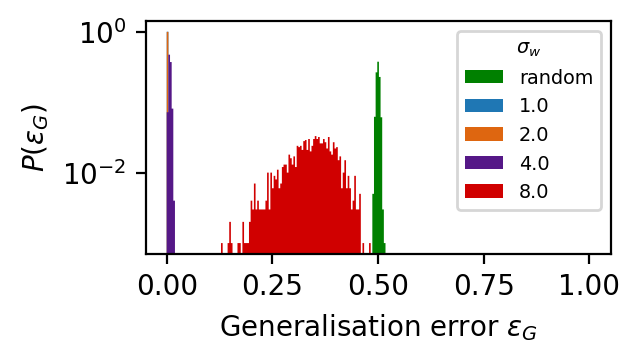}
        \caption{$m=16384$, $f=10\dots$}
    \end{subfigure}
    \begin{subfigure}[b]{0.23\columnwidth}
        \includegraphics[width = \textwidth]{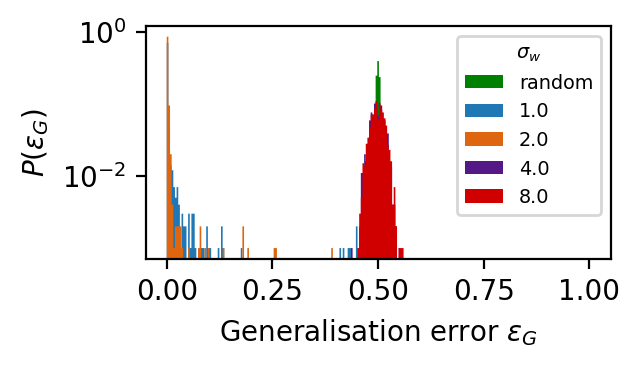}
        \caption{$m=16384$, $f=3-$parity}
    \end{subfigure}
    \begin{subfigure}[b]{0.23\columnwidth}
        \includegraphics[width = \textwidth]{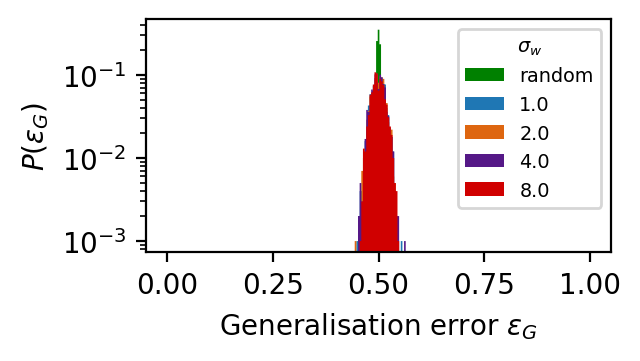}
        \caption{$m=16384$, $f=$ parity}
    \end{subfigure}
    \begin{subfigure}[b]{0.23\columnwidth}
        \includegraphics[width = \textwidth]{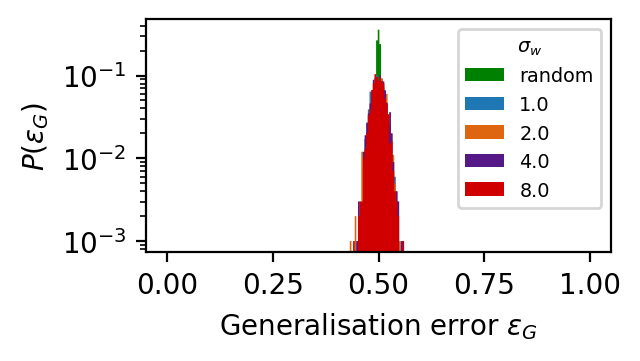}
        \caption{$m=16384$, $f=$ random}
    \end{subfigure}
    
    \caption{\small {\bf Boolean data with $\mathbf{n=128}$}
    In this figure, we trained FCNs with 10 layers, input dimension 128, hidden width 256 and tanh activations on randomly sampled training sets from the $n=128$ boolean dataset (with size $m=1024,4096,16384$). The training and test sets had to be randomly sampled as there are $2^{128}$ possible datapoints, far too many to be trained on. Training uses cross-entropy loss. we tested on 1024 random sampled test datapoints each time. As there are $2^{2^{128}}$ possible functions, we chose them by their properties on the overall data. 
    The first column uses target function which is 1 if the first input bit is 1 else 0. The second column uses the parity function on the first 3 input bits. The third column uses the parity function all input bits, and the final column uses a randomly generated target. Similarly to the small systems in e.g.\ \cref{fig:Cube_plots_1,fig:app:1d_extras}, the easiest function can be learned by $\sigma_w=1,2$ (and given enough data, even the large $\sigma_w$ can begin to learn them), and as functions grow more complex, they become impossible to learn for the amount of data we are using.
}\label{fig:app:large_bool}
\end{figure}

\begin{figure}[H]
    \centering

    \includegraphics[width = 0.5\textwidth]{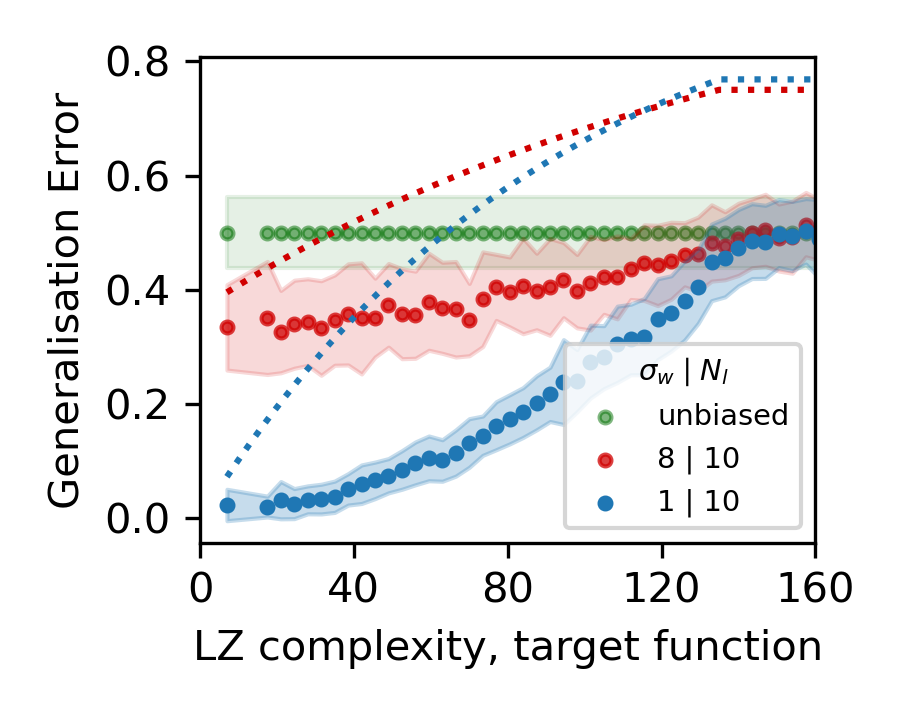}
    
    \caption{\small {\bf \cref{fig:Cube_plots_1}(c) with PAC-Bayesian bounds}
    We used extrapolated $P(K)$ data from \cref{app:fig:P(K)_n7} to calculate PAC-Bayesian bounds for the task shown in \cref{fig:Cube_plots_1}(c).
    We assumed that each function in each complexity class has the same probability, meaning the marginal likelihood for a function with complexity $K$, $P(S)_K\approx P(K) / N_K$, where $N_K$ is the number of functions with complexity $K$. $N_K$ is estimated from the rightmost plot in \cref{app:fig:P(K)_n7}(b), and $P(K)$ from the middle and leftmost plots in the same figure. Note that the PAC-Bayes bound for $\sigma_w=8$ is lower than that of $\sigma_w=1$ for the larger complexities -- which is to be expected, as careful examination of \cref{app:fig:P(K)_n7} shows a greater prior probability assigned to the more complex functions for $\sigma_w=8$ than $\sigma_w=1$.
}\label{fig:app:pb_1c}
\end{figure}

\newpage

\section{Extended data for MNIST and CIFAR datasets and CSR complexity}

In this section, we show extended data for \cref{fig3}, where we plot generalization errors, priors, and scatterplots for MNIST and CIFAR10 datasets.

\begin{figure}[H]
    \centering
    \makebox[\columnwidth][c]{\includegraphics[width=0.8\columnwidth]{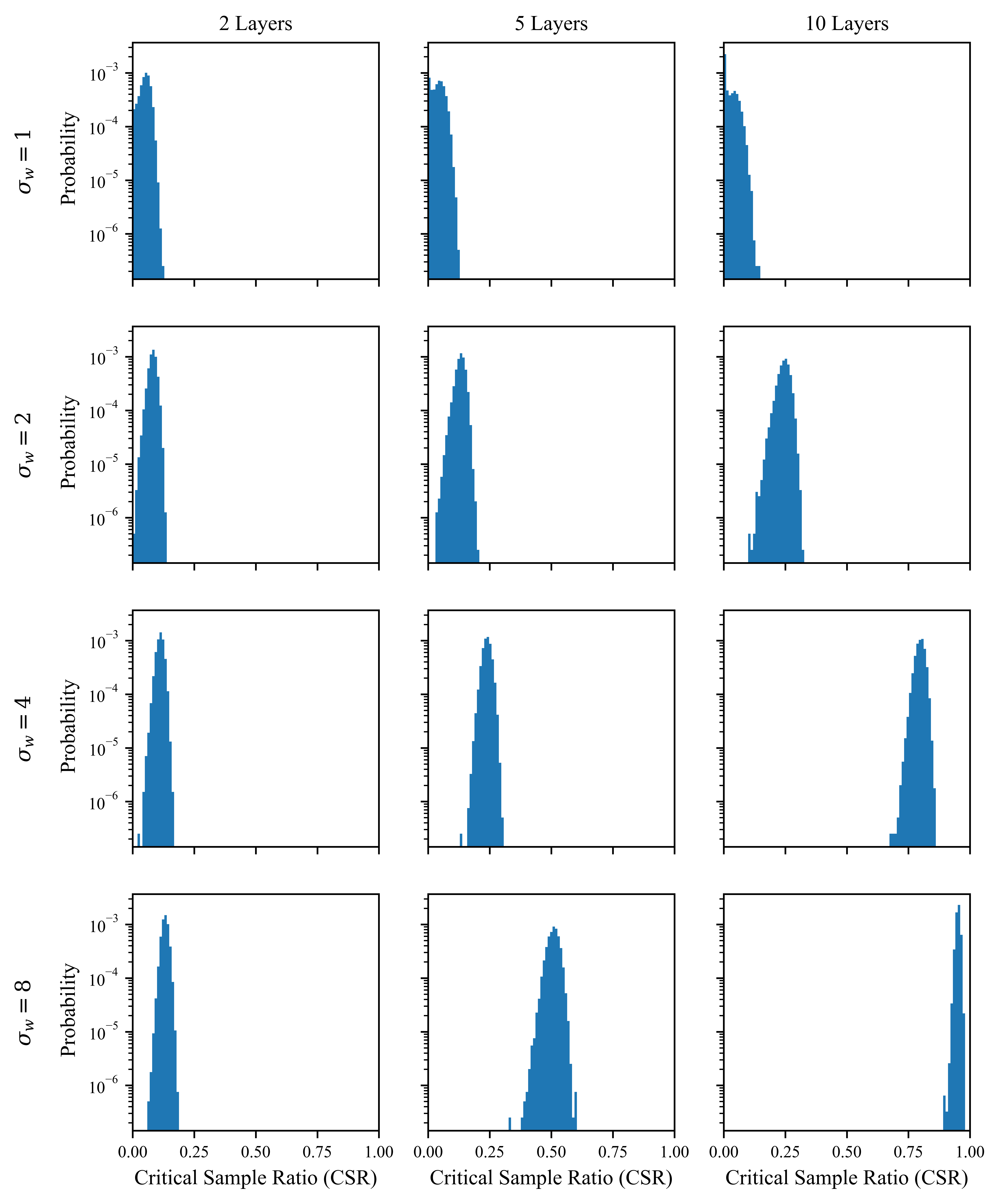}}
    \caption{{\bf Prior probability versus CSR complexity for MNIST}. Extended data for main text~\cref{fig:3_c}.  Probabilities were calculated  1000 MNIST images for randomly initalized FCN networks of 2, 5 and 10 layers with 200 neurons and weight standard deviations $\sigma_w =1,2,4,8$. Probabilities are estimated from a sample of $2 \times 10^4$ parameters. }
\end{figure}

\begin{figure}[H]
    \begin{subfigure}[b]{0.24\columnwidth}
        \includegraphics[width = \textwidth]{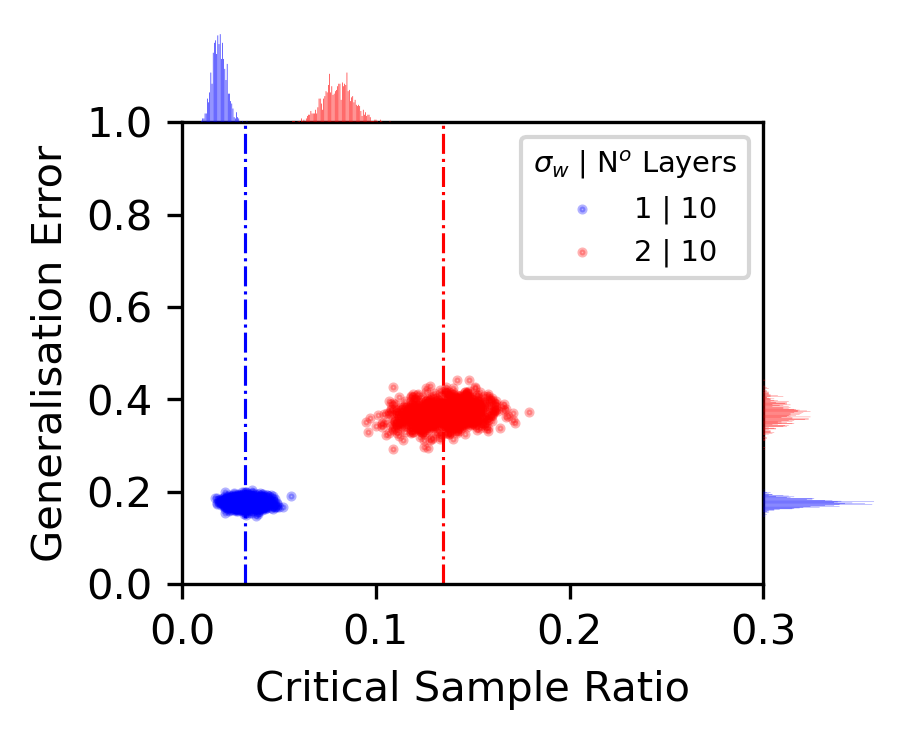}
        \caption{Uncorrupted}
    \end{subfigure}
    \begin{subfigure}[b]{0.24\columnwidth}
        \includegraphics[width = \textwidth]{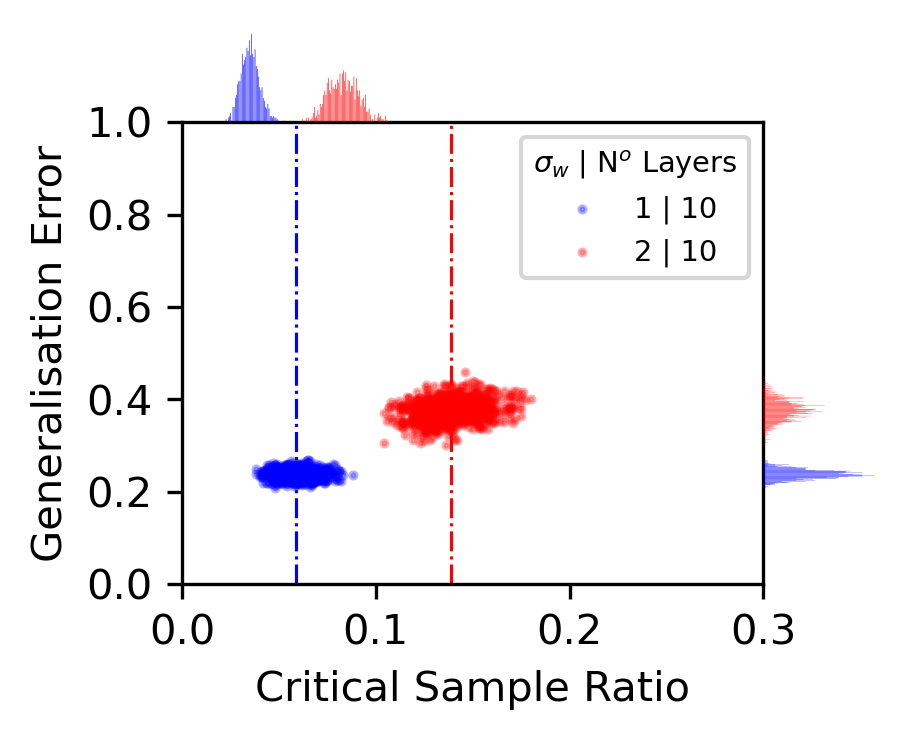}
        \caption{5\% corruption}
    \end{subfigure}
    \begin{subfigure}[b]{0.24\columnwidth}
        \includegraphics[width = \textwidth]{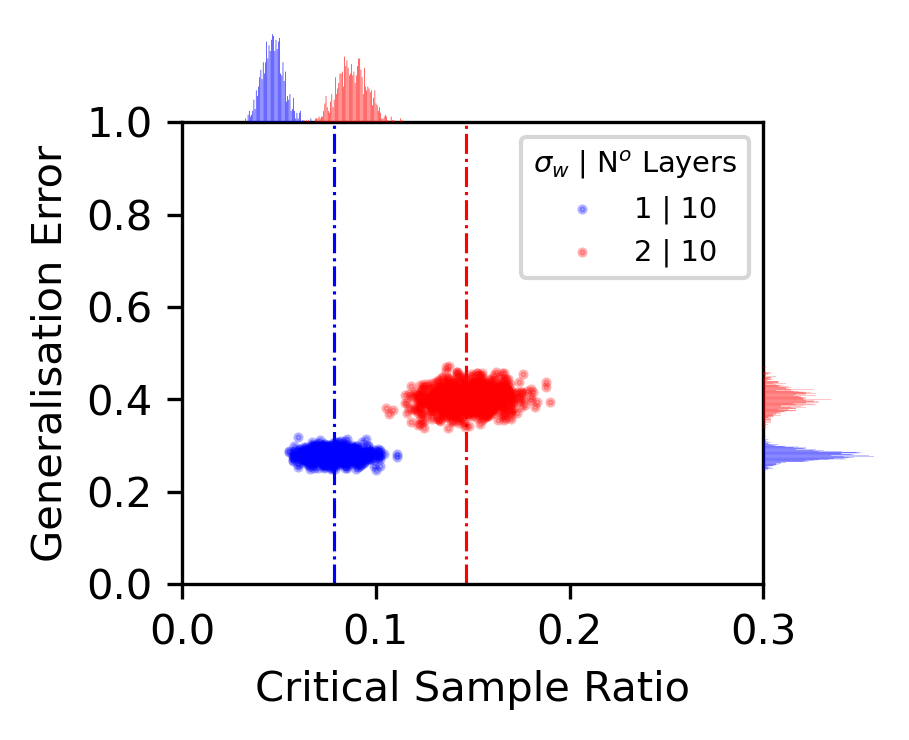}
        \caption{10\% corruption}
    \end{subfigure}
    \begin{subfigure}[b]{0.24\columnwidth}
        \includegraphics[width = \textwidth]{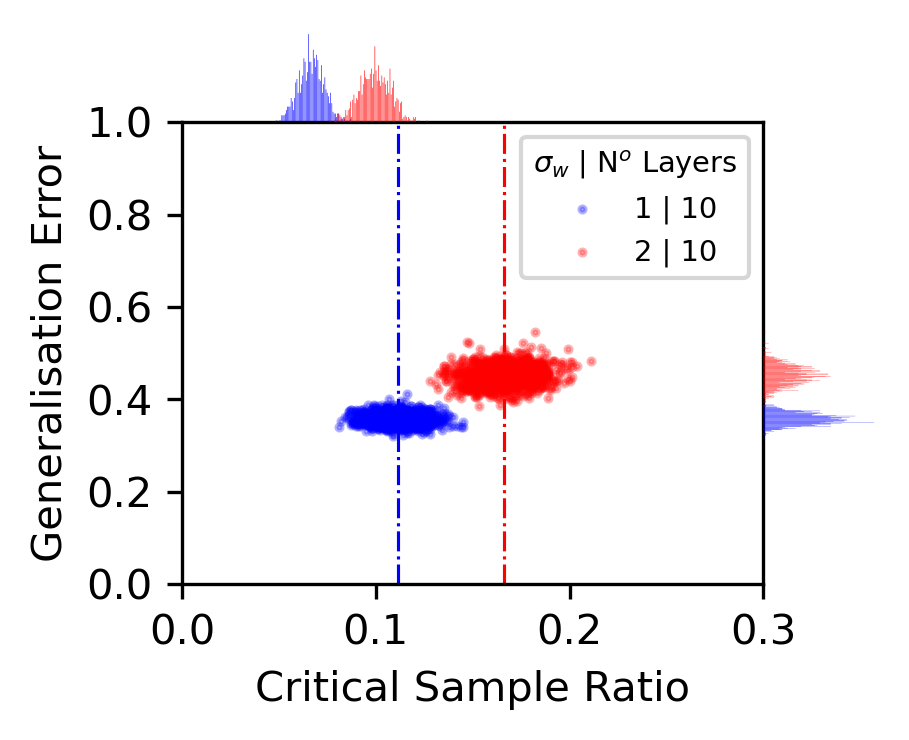}
        \caption{20\% corruption}
    \end{subfigure}

    \begin{subfigure}[b]{0.24\columnwidth}
        \includegraphics[width = \textwidth]{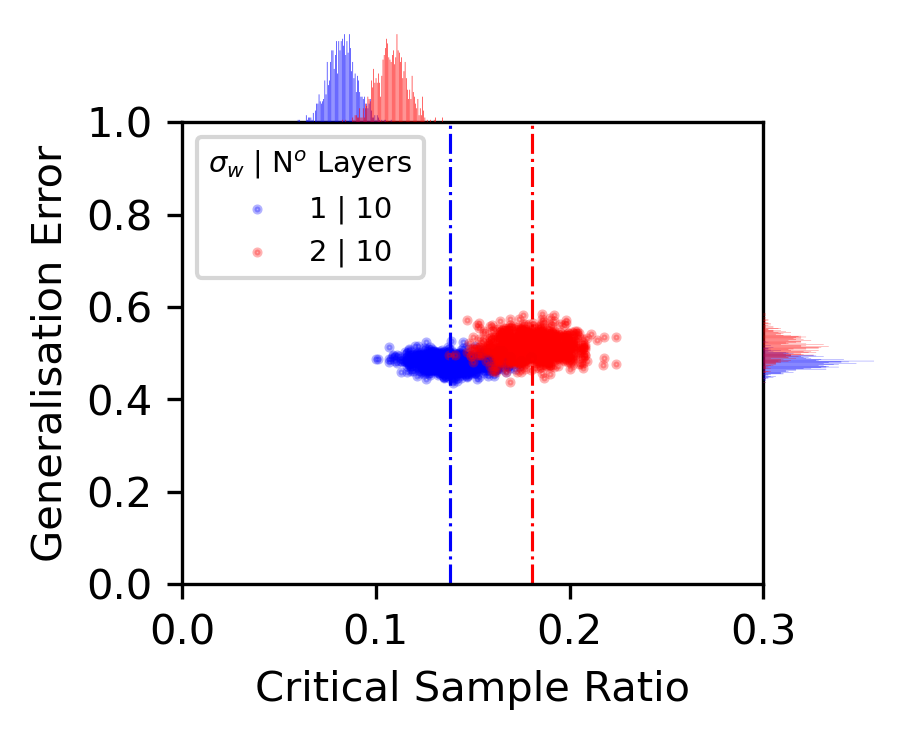}
        \caption{30\% corruption}
    \end{subfigure}
    \begin{subfigure}[b]{0.24\columnwidth}
        \includegraphics[width = \textwidth]{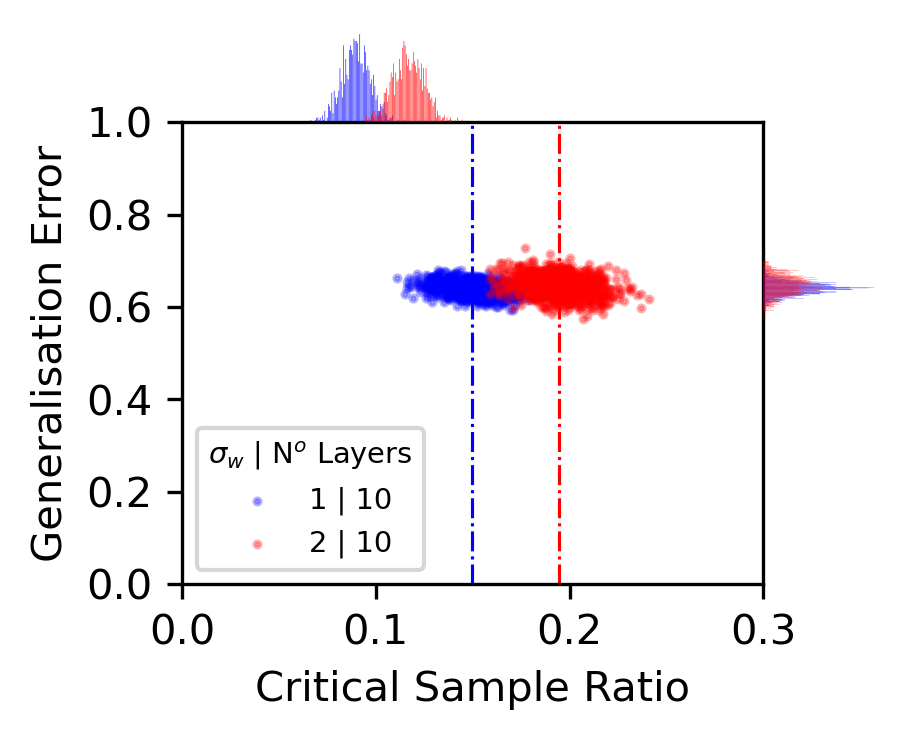}
        \caption{50\% corruption}
    \end{subfigure}
    \begin{subfigure}[b]{0.24\columnwidth}
        \includegraphics[width = \textwidth]{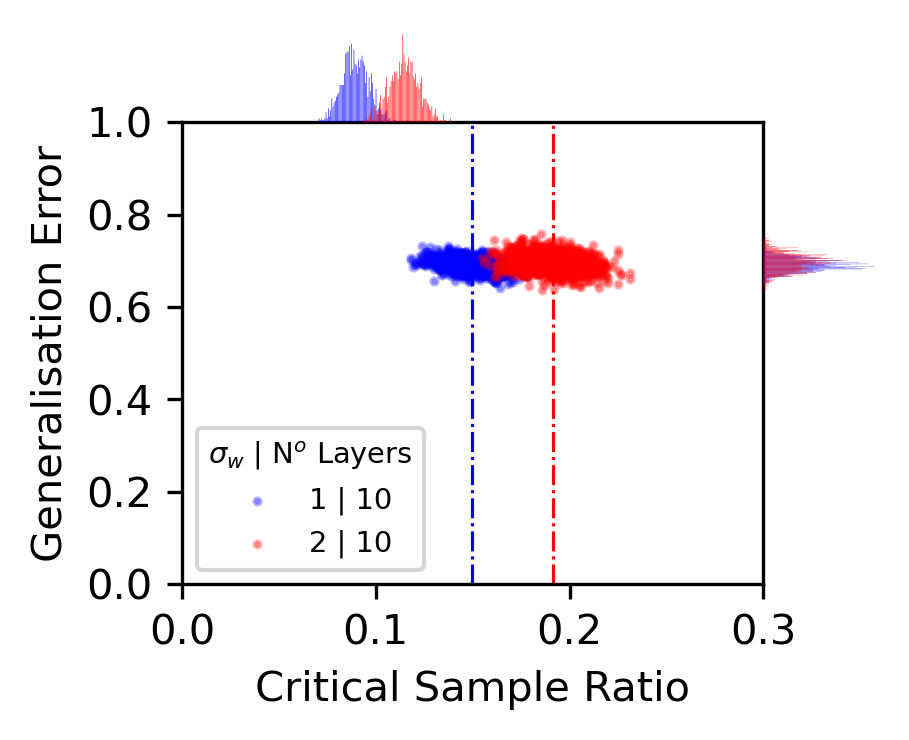}
        \caption{60\% corruption}
    \end{subfigure}
    \begin{subfigure}[b]{0.24\columnwidth}
        \includegraphics[width = \textwidth]{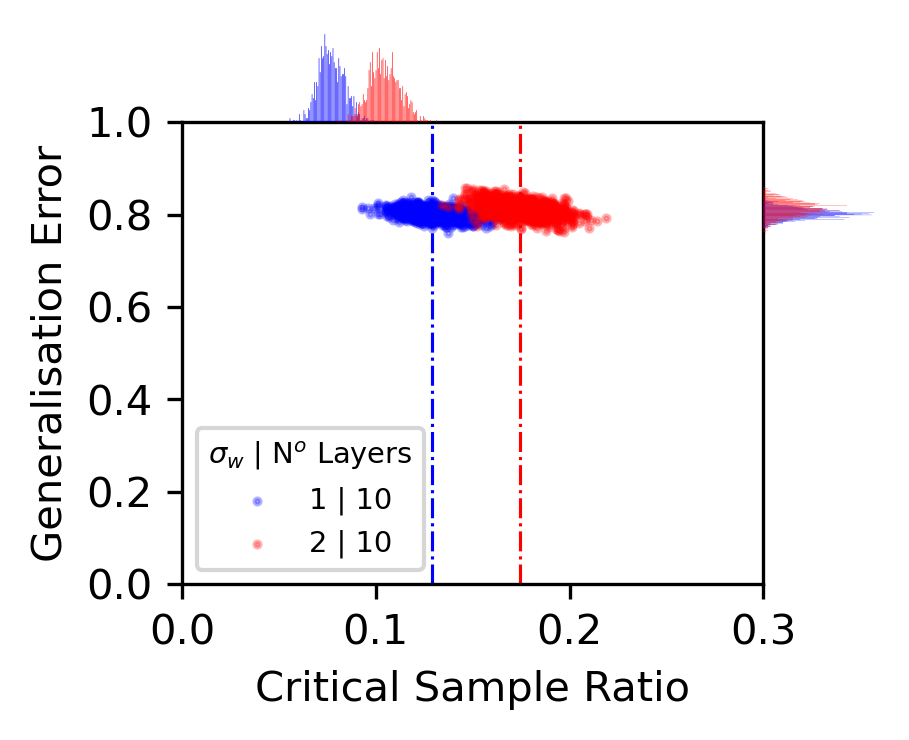}
        \caption{80\% corruption}
    \end{subfigure}
    \caption{{\bf generalization error v.s.\ CSR complexity scatter plots.} Extended data for main text \cref{fig:3_CSR_0,fig:3_CSR_25,fig:3_CSR_50}: 
    generalization error versus the critical sample ratio of the test set for 1000 networks trained to 100\% accuracy on 1000 MNIST images and tested on another 1000 images. The training examples in each plot have been corrupted to varying degrees by randomising the labels.}
\end{figure}

\section{Extended data for K-learner with PAC and PAC-Bayes bounds}\label{app:bias_variance}

\begin{figure}[H]
    \centering
    \begin{subfigure}[b]{0.34\columnwidth}
        \includegraphics[width = \textwidth]{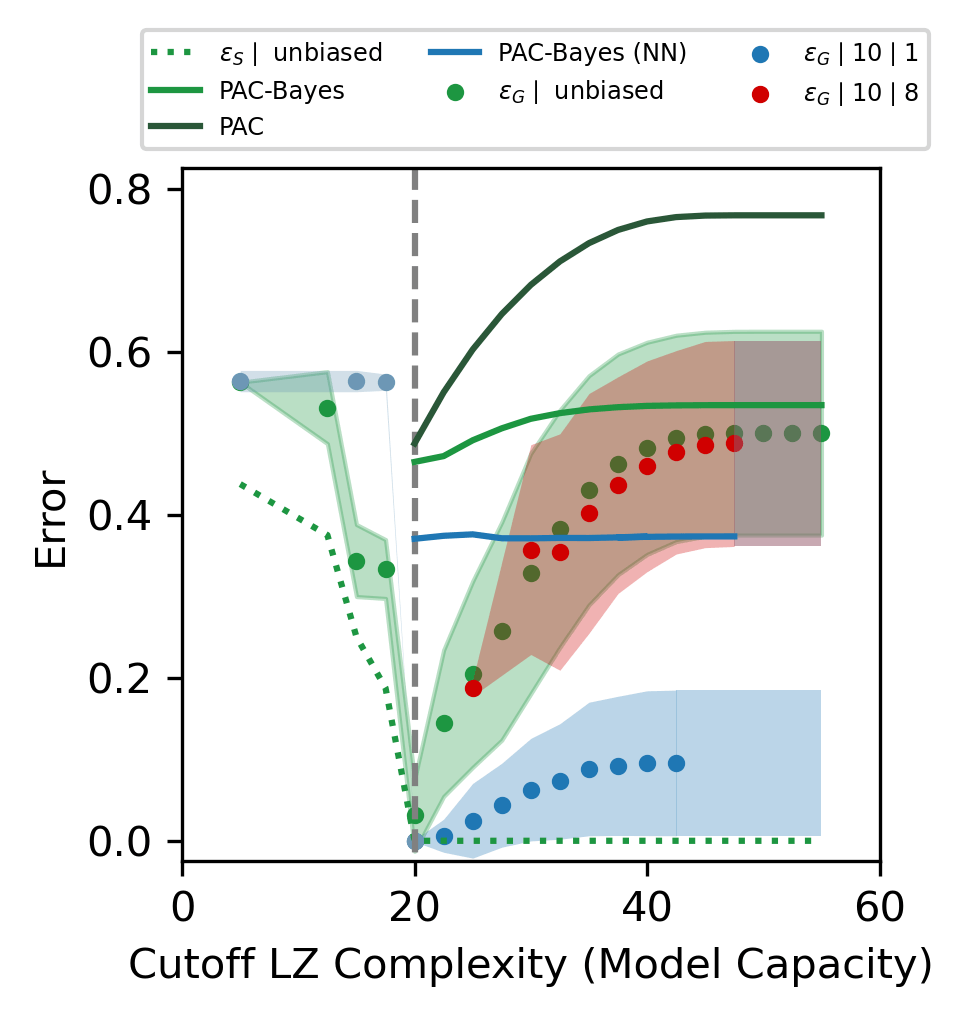}
        \caption{Simple function}
    \end{subfigure}
    \begin{subfigure}[b]{0.34\columnwidth}
        \includegraphics[width = \textwidth]{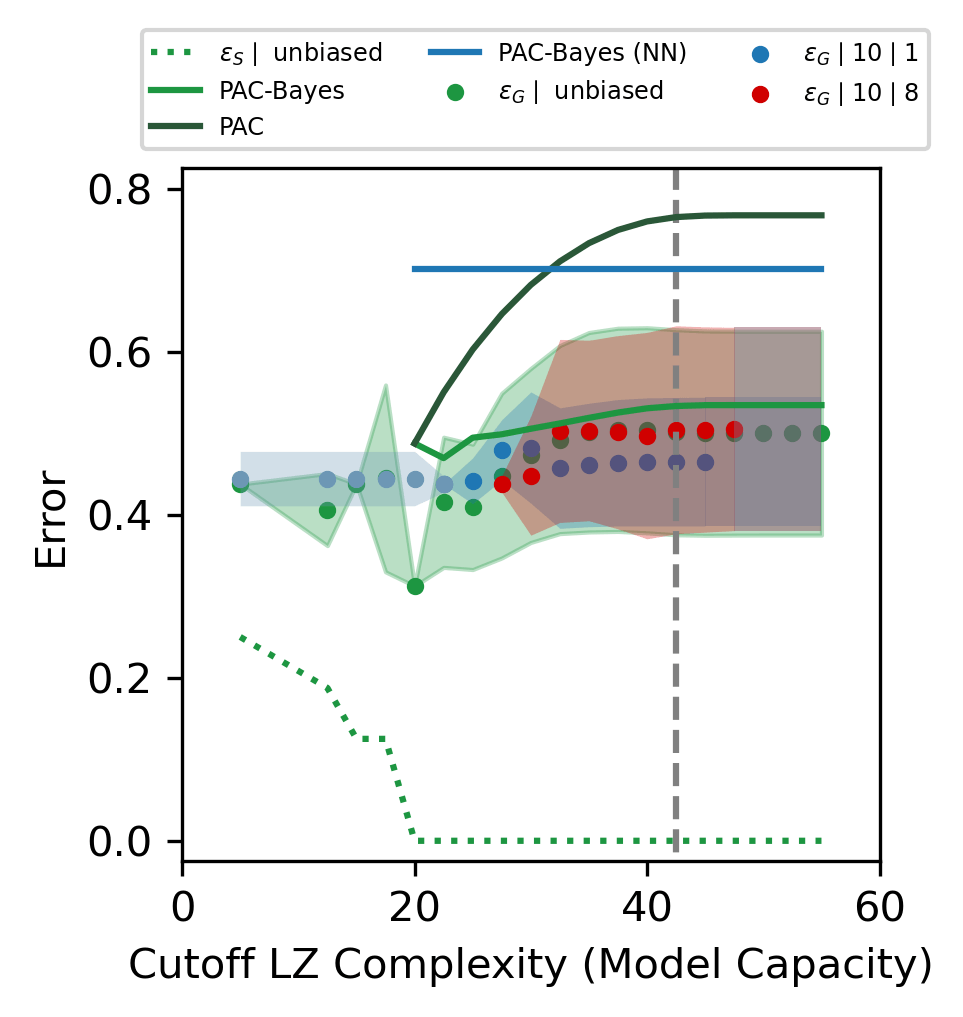}
        \caption{Complex function}
    \end{subfigure}
    
    \begin{subfigure}[b]{0.34\columnwidth}
        \includegraphics[width = \textwidth]{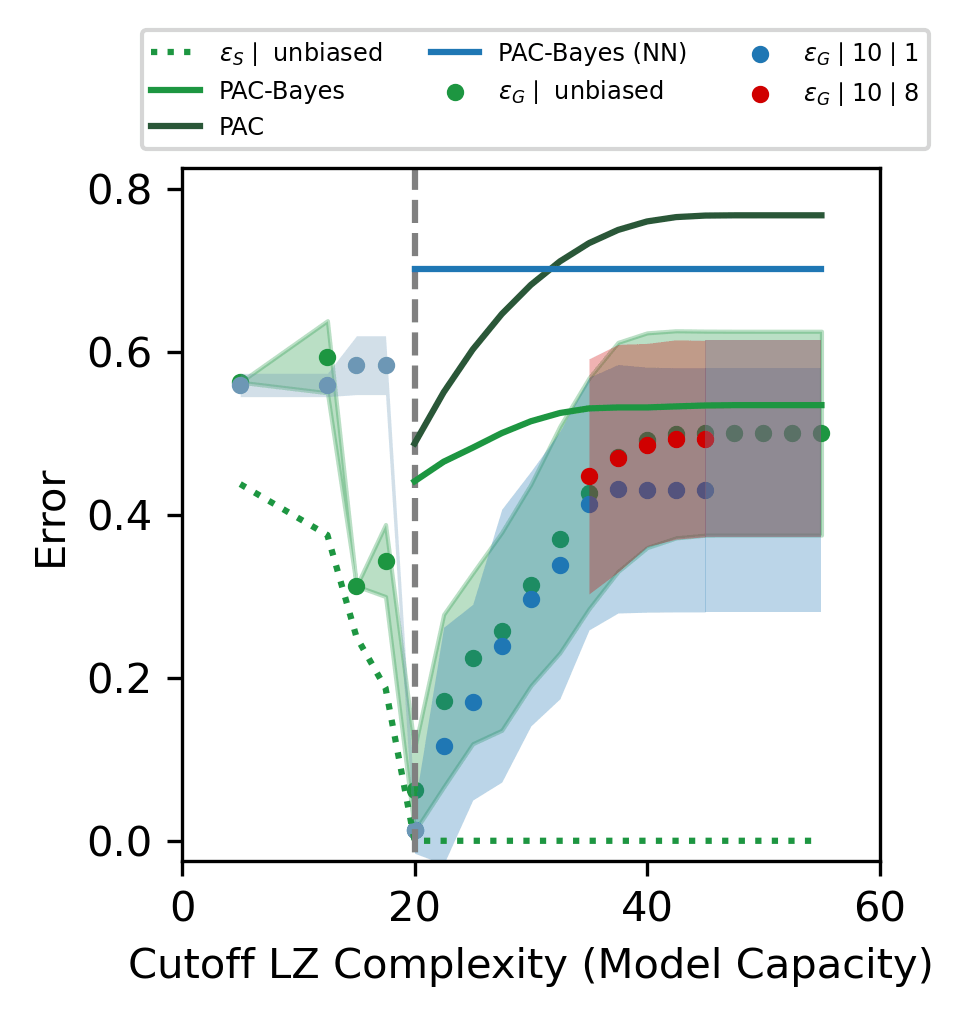}
        \caption{low-K low-P}\label{app:subfig:bv_trans}
    \end{subfigure}
    \begin{subfigure}[b]{0.34\columnwidth}
        \includegraphics[width = \textwidth]{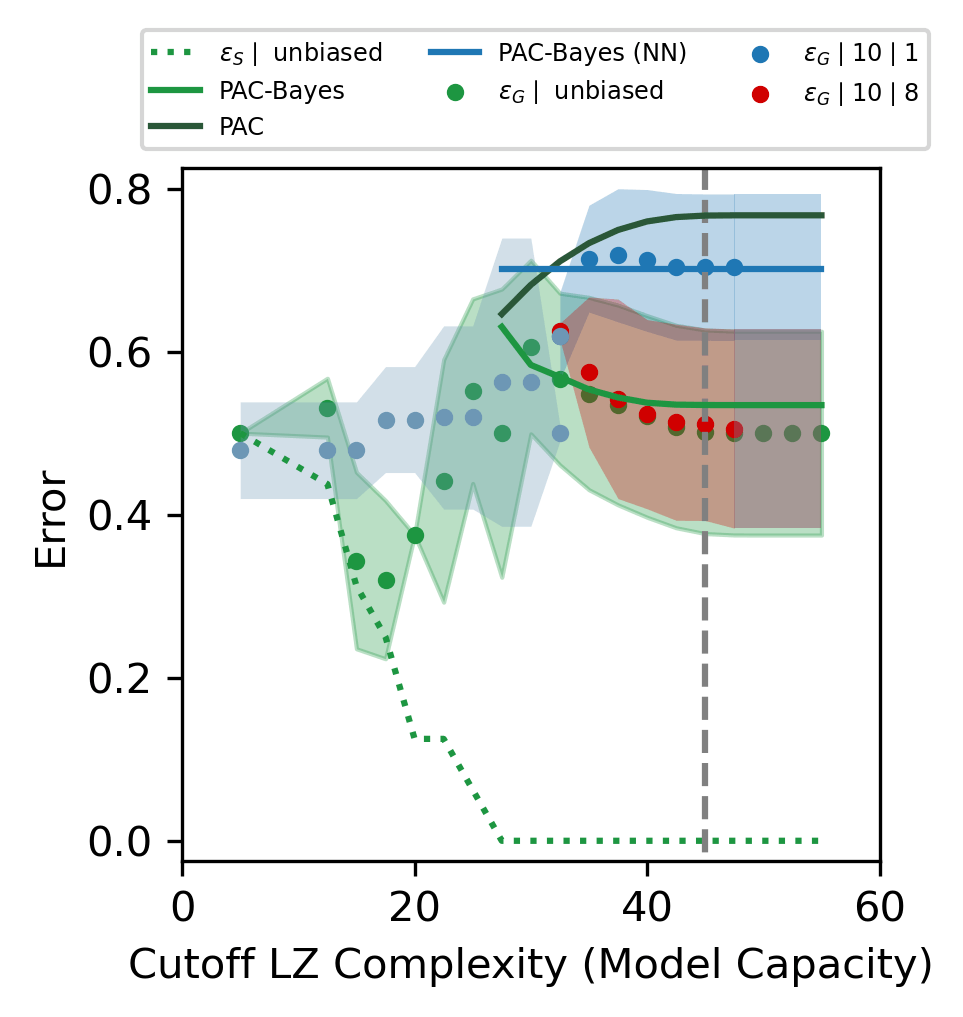}
        \caption{Parity function}
    \end{subfigure}
    
    \caption{\small {\bf \small K-learner for $\bf n=5$ Boolean system} Training set size $m=16$. Refer to \cref{app:bias_variance_main} for full details on these experiments.
    $\epsilon_T \mid \textrm{, unbiased}$ denotes the minimum training error achievable with functions $f$ of $C_{LZ}(f)\leq K$. $\epsilon_G \mid \textrm{, unbiased}$ denotes the test error of those functions, assuming each is equally likely.
    $\epsilon_G,10\mid \sigma_w$ denotes the generalization error of a DNN with 10 layers and weight initialization $\sigma_w$, upon training to 100\% training accuracy. The lighter blue datapoints are for $\epsilon_G,10\mid 1$, with function predictions sampled from the DNN each time training accuracy decreased. The chaotic DNN initializes at high K functions, so low K functions are not reached as part of training (in $10^4$ samples). The PAC-Bayes (NN) bound is for the DNN with $\sigma_w=1$, uses $\delta = 0.01$, and the marginal likelihood is calculated with 0-1 loss using $10^8$ samples from the prior of the $\sigma_w=1$ DNN (divide the number of functions that fit the training dataset $S$ found during sampling by $10^8$). See \cref{app:PAC-Bayes} for details about the bound. The PAC-Bayes bound is for the unbiased learner (where the marginal likelihood is calculated exactly with 0-1 loss); and the PAC bound is given in \cref{app:eqn:PACC} Both use $\delta=0.01$.\\
    (a) `0011' $\times$ 8\\
    (b) `01001111000111111110101111110100'\\
    (c) `1001' $\times$ 8\\
    (d) Parity function\\
    A simple function (a) is learnable by the uniform learner with a hypothesis space restricted to the complexity of the target function, or the DNN with $\sigma_w=1$.
    A highly complex function (b) should not be learnable no matter the chosen training set and hypothesis restriction.
    (c) A simple function (with respect to LZ complexity) has very low prior probability (was not found during sampling the prior, so $P(f)<10^{-8}$) is  a low-K low-P function. Therefore, the DNN with $\sigma_w=1$ does not generalise that well either.
    (d) The parity function is clearly very hard for the DNN to learn.  Interestingly it seems biased against it, as it generalises worse than a random learner.  
    However, the parity function is learnable for the  $n=7$ dataset with
 $m=100$. (see \cref{app:fig:bv_n7}).
    }\label{app:fig:bv_n5}
\end{figure}

\begin{figure}[H]
    \centering
    \begin{subfigure}[b]{0.34\columnwidth}
        \includegraphics[width = \textwidth]{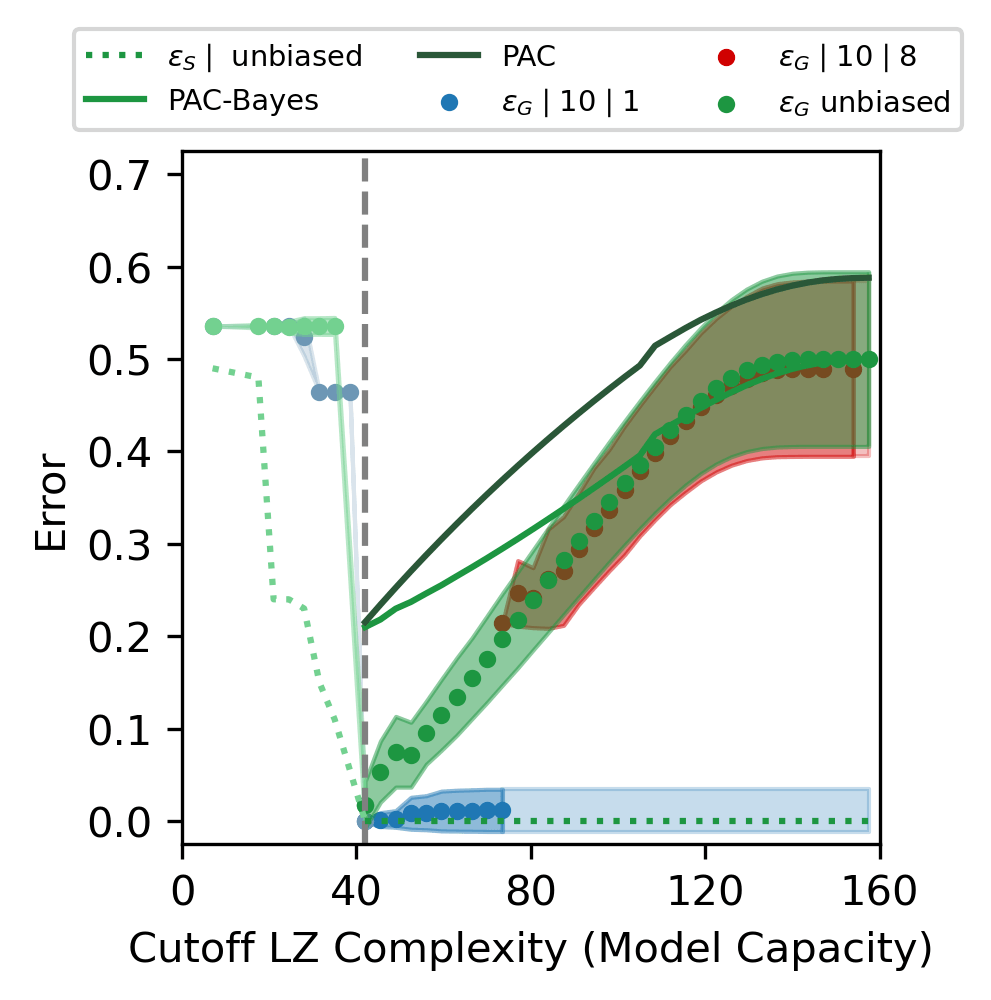}
        \caption{Simple function}
    \end{subfigure}
    \begin{subfigure}[b]{0.34\columnwidth}
        \includegraphics[width = \textwidth]{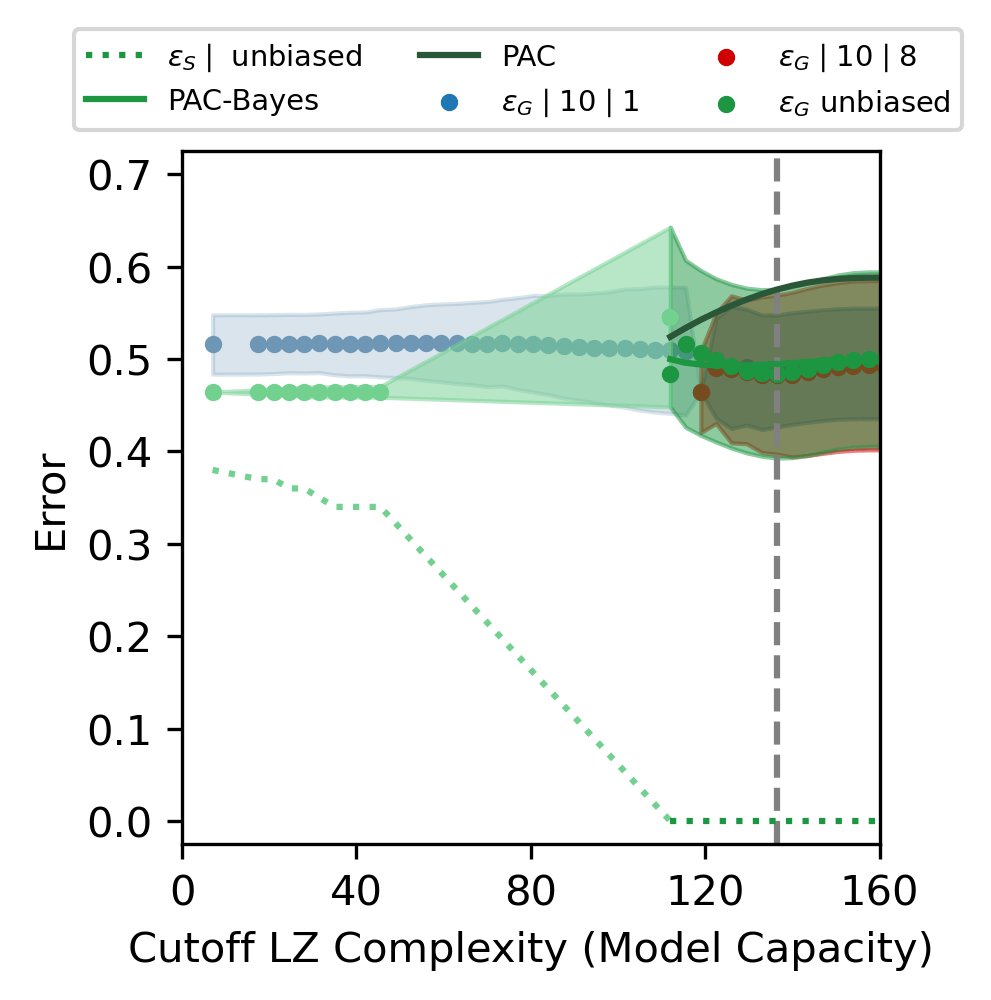}
        \caption{Complex function}
    \end{subfigure}

    \begin{subfigure}[b]{0.34\columnwidth}
        \includegraphics[width = \textwidth]{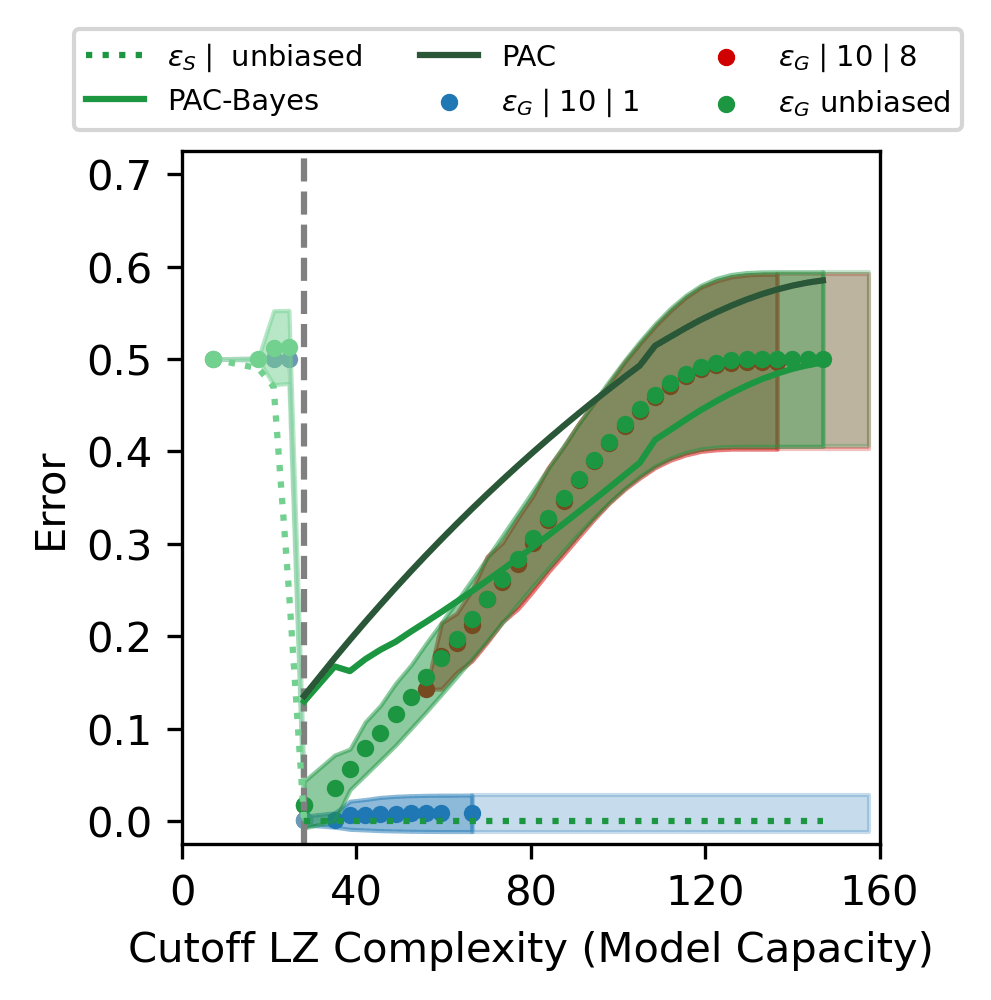}
        \caption{LowK lowP function at $n=5$}
    \end{subfigure}
    \begin{subfigure}[b]{0.34\columnwidth}
        \includegraphics[width = \textwidth]{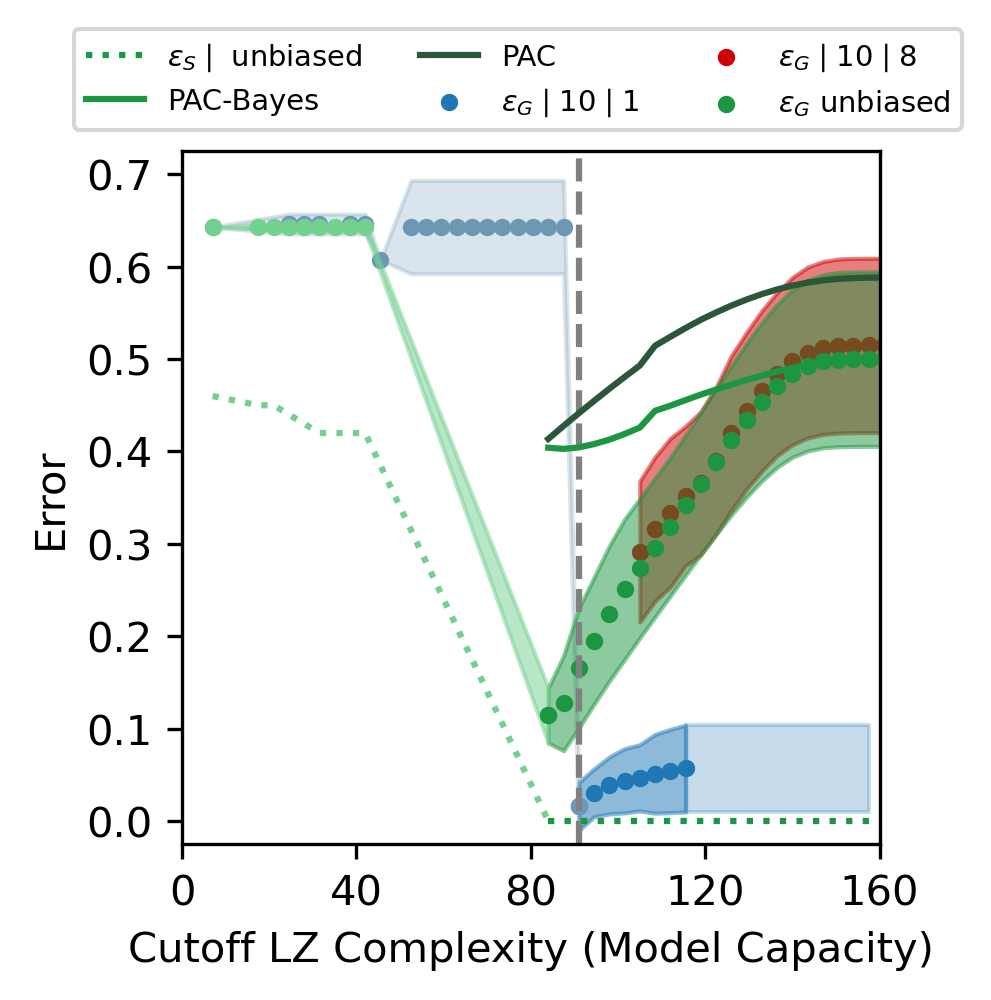}
        \caption{Parity function}
    \end{subfigure}
    
    \caption{\small {\bf \small K-learner for $\bf n=7$ Boolean system}
    See \cref{app:fig:bv_n5} for notation. Training set size $m=100$. For details of the implementation of the different learners, as well as  the realisable PAC and PAC-Bayes bounds (which use $\delta=0.01$), see \cref{app:bias_variance_main} for full details.\\
    (a) simple function `10011010' $\times$ 16\\
    (b) complex function `0011011110100111010010111001011100011100101110111011101011101111\\0000110001100011111111100111011001111010010101011101111001110001'\\
    (c)lowK lowP function at $n=5$ `1001' $\times$ 32\\
    (d) Parity function\\
    Unlike the $n=5$ case, good generalization is achieved for the lowK lowP function in (c) and  parity function in (d).  This may be due in part to the relatively large training set ($m=100$).}\label{app:fig:bv_n7}
\end{figure}

In this section, we expand on \cref{fig:Cube_plots_1} (i) for different target functions, as well as for the smaller $n=5$ system.  As can be seen in \cref{app:fig:bv_n5} for $n=5$ and in \cref{app:fig:bv_n7} for $n=7$, there are clear differences in performance on simple rather than complex functions.  We also include a lowK lowP function that is simple, but has low probability (and so is far from the bound) (see \cite{dingle2020generic} for a longer discussion of such functions). We would expect (and indeed observe for $n=5$) that the DNN struggles more with functions for which it has a smaller $P(f)$, and therefore a smaller inductive bias at initialization.  Finally, we study the parity function, for which $f=(1,0)$ if the number of ones in the input is even (odd).  It, therefore, has the highest input sensitivity of any function, because any change of input changes the function output.  There are well-known questions in the field about whether or not a DNN can learn the parity function~\cite{nye2018efficient,bhattamishra2022simplicity}.

\FloatBarrier

\section{Analysis of histograms in \cref{fig:Cube_plots_2}}

\begin{figure}[H]
    \centering
    \begin{subfigure}[t]{0.8\columnwidth}
        \includegraphics[width = \textwidth]{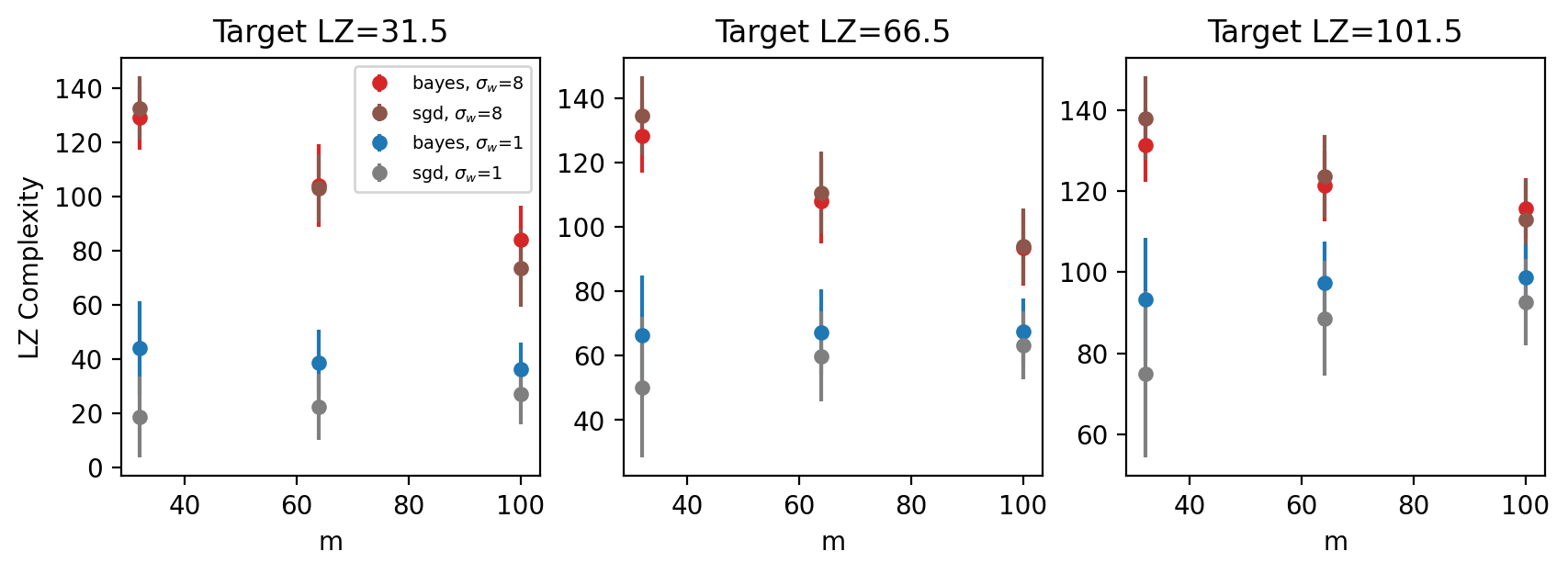}
        \caption{Means and standard deviations}
    \end{subfigure}
    
    \begin{subfigure}[t]{0.8\columnwidth}
        \includegraphics[width = \textwidth]{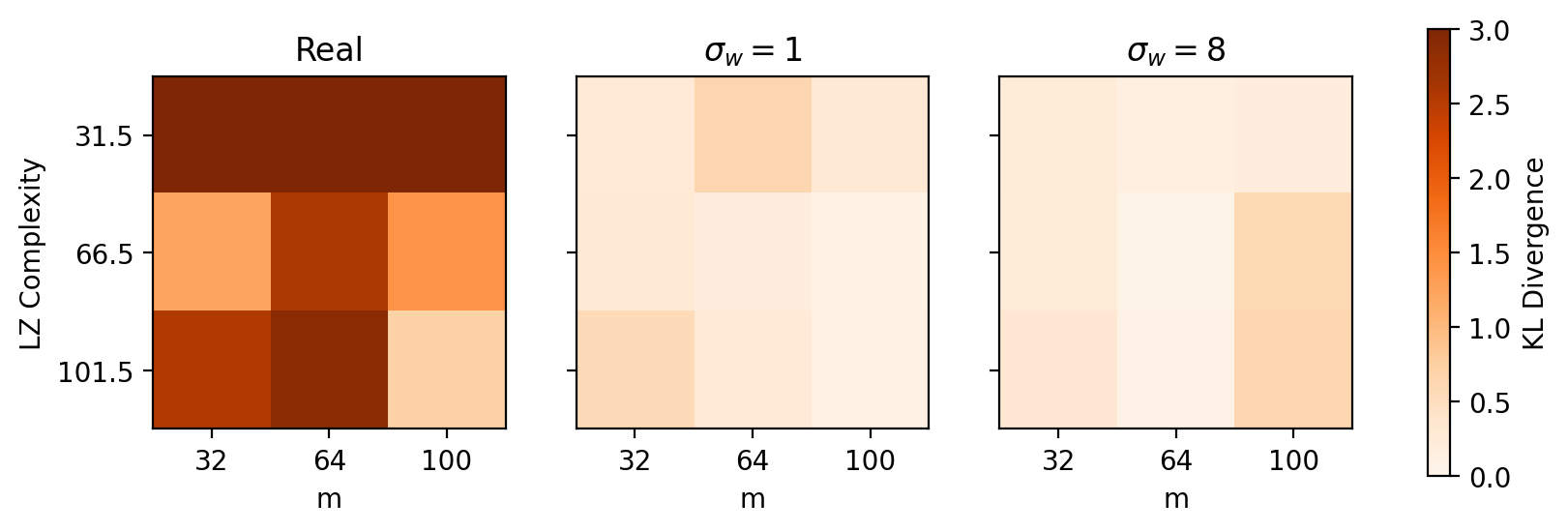}
        \caption{KL divergences}
    \end{subfigure}
    
    \caption{\small {\bf \small Analysis of histograms in \cref{fig:Cube_plots_2}}
    (a) shows the means and standard deviations for histograms of $P(f\mid S)$ in \cref{fig:Cube_plots_2}, for each of the three functions analysed, as a function of training data $m$.  The left-most panel in (b) shows the KL divergences between the $\sigma_w=1$ and $\sigma_w$=8 histograms from the real data (the scale goes up to 3 as a maximum, but the divergence is clearly very large when the histograms do not overlap). This gives a sense of scale for the divergences show in the next two panels: calculated between the base model and the Bayesian approximation. The average KL divergence in the three panels is 2.26, 0.27 and 0.28 respectively. Two-tailed t-tests on the histograms clearly show that the underlying distributions are very unlikely to have the same mean, with t-values greater than 3 in all but 2 cases ($m=64$, $LZ=31.5$, $\sigma_w=8$ and $m=100$, $LZ=66.5$, $\sigma_w=8$).}\label{app:Cube_plots_analysis}
\end{figure}

\newpage

\end{document}